\documentclass[12pt]{article} % For LaTeX2e

\usepackage{url,fullpage,color}
\usepackage{graphicx}
\usepackage[font=small]{caption}
\usepackage[font=small]{subcaption}
\usepackage{amsmath, amsthm, amssymb, bm}
\usepackage{algpseudocode}
\usepackage{algorithm}
\usepackage{comment}
\RequirePackage[round]{natbib}
\RequirePackage[colorlinks,citecolor=blue,urlcolor=blue]{hyperref}

\usepackage{setspace}

\usepackage{multirow}

\let\hat\widehat

\theoremstyle{remark}
\newtheorem{remark}{Remark}

\newtheorem{thm}{Theorem}
\newtheorem*{thm*}{Theorem}

\theoremstyle{definition}

\newtheorem*{defn*}{Definition}
\theoremstyle{remark}

\newcommand{\norm}[1]{\left\Vert#1\right\Vert}
\newcommand{\abs}[1]{\left\vert#1\right\vert}
\newcommand{\set}[1]{\left\{#1\right\}}

\newcommand{\eps}{\varepsilon}
\newcommand{\al}{\alpha}

\newcommand{\sig}{\sigma}

\newcommand{\CC}{{\mathbb C}}

\newcommand{\EE}{{\mathbb E}}

\newcommand{\LL}{{\mathbb L}}

\newcommand{\PP}{{\mathbb P}}

\newcommand{\RR}{{\mathbb R}}

\newcommand{\XX}{{\mathbb X}}

\newcommand{\pDisk}{{\sf pDisk}}
\newcommand{\Disk}{{\sf Disk}}

\newcommand{\calC}{{\mathcal C}}

\newcommand{\calL}{{\mathcal L}}

\newcommand{\calS}{{\mathcal S}}

\newcommand{\calX}{{\mathcal X}}

\usepackage{etoolbox}
\newcommand{\zerodisplayskips}{%
  \setlength{\abovedisplayskip}{2pt}%
  \setlength{\belowdisplayskip}{2pt}%
  \setlength{\abovedisplayshortskip}{2pt}%
  \setlength{\belowdisplayshortskip}{2pt}}
\appto{\normalsize}{\zerodisplayskips}
\appto{\small}{\zerodisplayskips}
\appto{\footnotesize}{\zerodisplayskips}

\catcode`@=11
\newskip\beforeproofvskip
\newskip\afterproofvskip
\beforeproofvskip=\medskipamount
\afterproofvskip=\bigskipamount

\def\prooftag{Proof}
\def\proofskip{\enspace}

\def\proof{\@ifnextchar[{\@@proof}{\@proof}}  %] for emacs matching
\def\@startproof{\par\vskip\beforeproofvskip\leavevmode}
\def\@proof{\@startproof{\scshape\prooftag.}\proofskip}
\def\@@proof[#1]{\@startproof {\scshape\prooftag #1.}\proofskip}

\catcode`@=12

\usepackage{enumitem}
\setlist{nosep}

\begin{document}

\def\spacingset#1{\renewcommand{\baselinestretch}%
{#1}\small\normalsize} \spacingset{1}

  \title{\bf Skeleton Clustering: Dimension-Free Density- {Aided} Clustering}
  \author{ Zeyu Wei \hspace{.2cm}\\
    Department of Statistics, University of Washington\\
    and \\
    Yen-Chi Chen \\
    Department of Statistics, University of Washington}
  \maketitle

\begin{abstract}
\noindent

We introduce a density-aided clustering method called Skeleton Clustering that can detect clusters in multivariate and even high-dimensional data with irregular shapes. To bypass the curse of dimensionality, we propose surrogate density measures that are less dependent on the dimension but have intuitive geometric interpretations. The clustering framework constructs a concise representation of the given data as an intermediate step and can be thought of as a combination of prototype methods, density-based clustering, and hierarchical clustering. We show by theoretical analysis and empirical studies that the skeleton clustering leads to reliable clusters in multivariate and high-dimensional scenarios.

%We develop a new framework for clustering, the Skeleton Clustering framework, that can detect clusters in multivariate and even high-dimensional data with irregular shapes. In this framework, data is first represented by some knots, and the connections between the knots are weighted by some similarity measures. The produced skeleton structure is then segmented into clusters and the data points are clustered with respect to the knots. Novel similarity measures between knots are proposed, and the estimations of those measures are not dependent on the dimension of the data, hence avoiding the curse of dimensionality. Some theoretical properties and empirical results justify those advantages of skeleton clustering.
\end{abstract}

\noindent%
\vfill

\newpage
\spacingset{1.9} % DON'T change the spacing!

\section{Introduction}

~~~~Density-based clustering \citep{azzalini2007clustering,menardi2014advancement,  Chacon2015} 
is a popular framework for grouping observations into clusters defined based on the underlying probability density function (PDF).
In practice, when the PDF is usually unknown, it is estimated via the random sample and the estimated PDF is then used to obtain the resulting clusters.
Many clustering methods have been proposed within the framework of density-based clustering.
The mode clustering \citep{li2007nonparametric, chacon2013data, Chen2016} finds clusters via the local modes of the underlying PDF.
When the kernel density estimator (KDE) is used for density estimation,
the mode clustering can be done easily via the mean-shift algorithm \citep{fukunaga1975estimation, cheng1995mean, carreira2015review}.
Another famous density-based clustering approach is the level-set clustering \citep{cuevas2000estimating, Cuevas2001, mason2009asymptotic, rinaldo2012stability}, which creates clusters as the connected components of high-density regions. 
The well-known DBSCAN (Density-Based Spatial Clustering of Applications with Noise) method \citep{EsterKriegel1996} 
is also a special case of level-set clustering. 
Moreover, the cluster tree \citep{Wernergsl, Chaudhuri,Chaudhuri2014, Eldridge2015, Kim} 
% \citep{klemela2009smoothing}
is a density-based clustering approach combining information from both modes and level sets. 
This method creates a tree structure with each leaf representing a mode and the tree describes the evolution of level-set clusters at different density levels.

Compared to the classical k-means clustering \citep{Lloyd1982, Hartigan1979, Pollard1982} and the model-based clustering methods \citep{Fraley2002},
a density-based clustering approach is capable of finding clusters with irregular shapes
and gives an intuitive interpretation based on the underlying PDF. 
Furthermore, defining clusters based on the density function makes it possible to view the clustering problem as an estimation problem: the clusters
from the true PDF are the parameters of interest
and the estimated clusters are sample quantities utilized for approximation.

Although density-based clustering enjoys many advantages, it has a fundamental limitation: the curse of dimensionality.
Because a density-based clustering method often involves a density estimation step, it does not scale well with the dimension.
Specifically, the convergence rate of a density estimator is $O_P(n^{-\frac{2}{4+d}})$ under usual smoothness conditions \citep{scott2015multivariate,Wasserman2006}, which is slow when $d$ is large.
To overcome the curse of dimensionality and apply density-based clustering to high-dimensional data, 
we follow the idea of merging a large number of clusters \citep{peterson2018merging, synClus2020, FredJain2005,Maitra2009, Shin2019},  
% In particular, \cite{peterson2018merging} propose to first construct k-Means clusters and then merge the clusters guided by the overlapping probabilities between clusters as estimated through fitting Gaussian distributions. 
to explicitly construct a graph representation of the data based on the initial protoclusters and propose density-aided similarity measures suitable for high-dimensional settings.

 {The idea of merging prototypes has also attracted great attention from model-based clustering  to overcome the limitations of parametric assumptions. 
In particular, there are several methods for merging Gaussian-mixture models \citep{Hennig2010}
such as
Dip test approach \citep{Hartigan1985},
ridgeline elevation \citep{RayLindsay2005}, 
misclassification method \citep{Tibshirani2005},
multi-layer approach \citep{Li2005},
entropy-based method \citep{Baudry2010},
level set-based method \citep{SCRUCCA2016},
and modal clustering \citep{Chacon2019}.
%\cite{Hennig2010} proposed several merging strategies for Gaussian-mixture models. 
%\citep{RayLindsay2005} proposed to use the ridgeline elevation
%to merge clusters. 
%\citep{Hartigan1985} proposed a Dip test approach.
%\citep{Tibshirani2005} proposed a 
%a prediction-based method for merging Gaussian mixture models.
%%misclassification-based approaches focus on different estimates of misclassification probabilities or the prediction strength \citep{Tibshirani2005} when merging Gaussian components.
%Moreover, \cite{Li2005} uses two layers of Gaussian mixtures as a merging process, and \cite{Baudry2010} discusses entropy-based merging. 
%Additionally, \cite{SCRUCCA2016} starts with Gaussian mixture density estimates and identifies the connected components of high density regions, 
%while \cite{Chacon2019} employs the Gaussian mixture density estimation for modal clustering with mean-shift algorithm. 
% More recently, \cite{Shin2019} use the conformal prediction idea to turn the k-means clusters as balls and merge them into connected components based on Lebesgue measure of such union of balls.
The work by \cite{Aragam2020} reconstructs a nonparametric mixture model by 
fitting the data with a large number of general nonparametric mixture components and then partitions them into a small number of final clusters.
}

%Specifically, we propose 
%a multistage clustering method
%that uses ideas from prototype clustering and hierarchical clustering and combines them with density-based clustering.
Our idea can be summarized as follows.
We first find a large set of protoclusters (called \textit{knots}) by running $k$-means clustering. Nearby knots are then connected by edges to form a graph that we call the \textit{skeleton}.
The similarities between connected knots are measured by density-aided criteria that are estimable even in high dimensions.
Finally, we merge knots according to a linkage criterion 
to create the final clusters. 
Because the construction involves creating a \textit{skeleton} representation of the data,
we call this method \emph{Skeleton Clustering}.

\begin{figure}[ht]
\centering
\includegraphics[width=4cm]{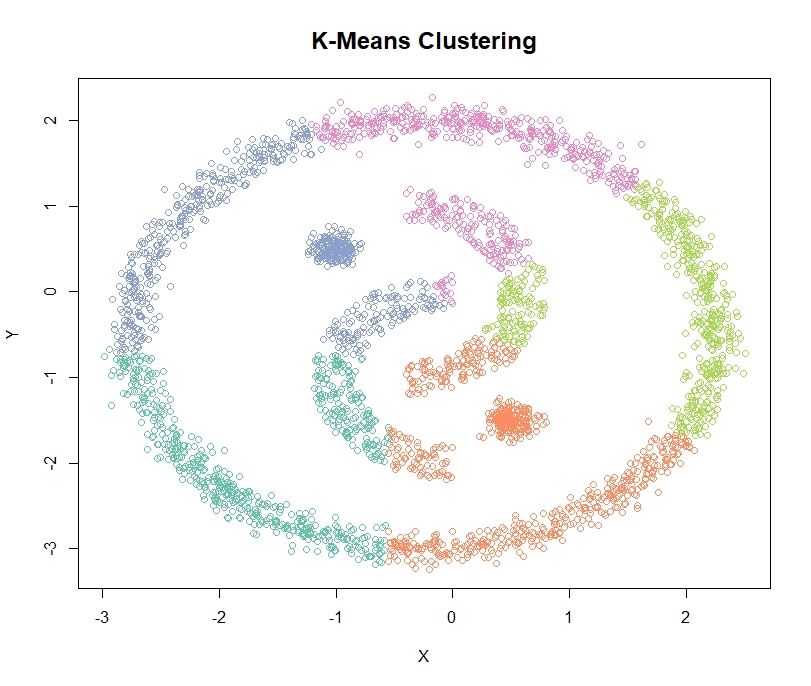}
\includegraphics[width=4cm]{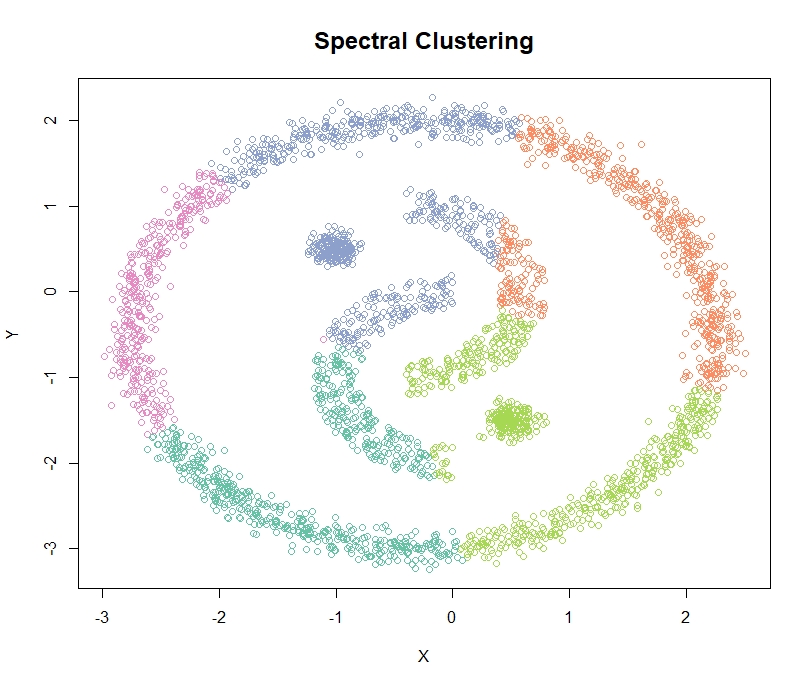}\\
\includegraphics[width=4cm]{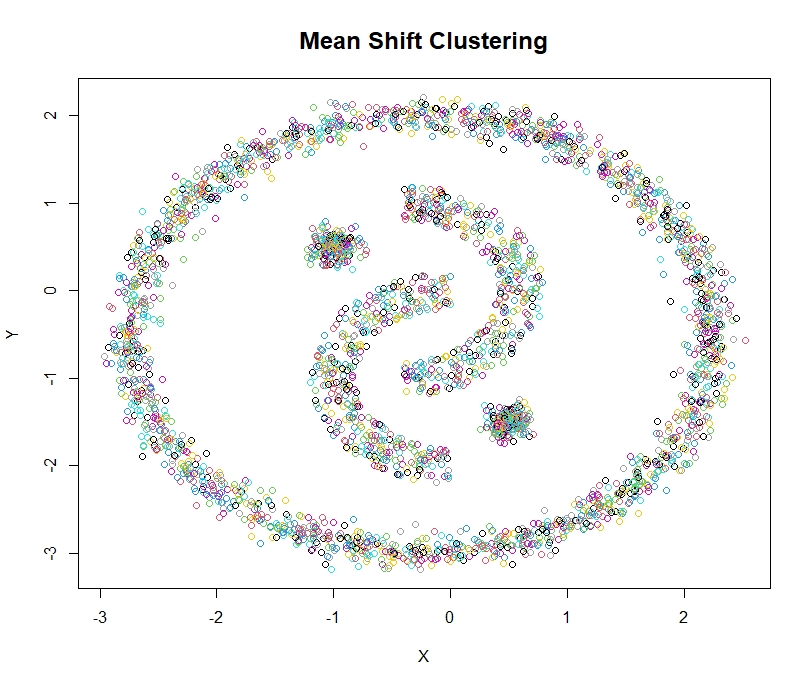}
\includegraphics[width=4cm]{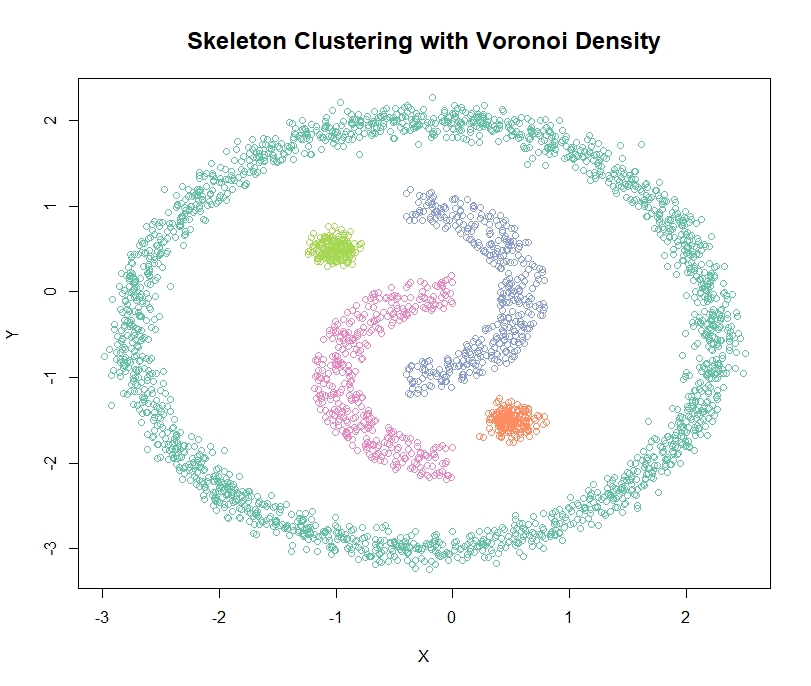}
\caption{Yinyang Data with dimension 200. On the bottom-right is the clustering result of the skeleton clustering with the proposed Voronoi density similarity measure.}
\label{fig::ex01}
\end{figure}

To illustrate the limitation of the classical approaches and to highlight the effectiveness of skeleton clustering, 
we conduct a simple simulation in 
Figure \ref{fig::ex01}.
It is a $d=200$ dimensional data consisting of five components with non-spherical shapes.
The actual structure is in $2$-dimensional space as illustrated in Figure \ref{fig::ex01}.
We add Gaussian noises in other dimensions to make it a $d=200$ dimensional data (see Section~\ref{sec::simulation}
for more details). 
Traditional $k$-means and spectral clustering fail to find the five components and the mean shift algorithm cannot form clusters due to the high dimensionality of the data.
However, our proposed method (bottom-right panel) can successfully recover the underlying five components.

\emph{Outline.} 
In section \ref{sec::clustering}, we describe the skeleton clustering framework. 
In section \ref{sec::construction}, we introduce similarity measures that can be utilized in the skeleton clustering framework. 
In section \ref{sec::theory}, we provide some consistency results of the sample similarity measures and the clustering performance guarantee.  
In section \ref{sec::simulation}, we present simulation results to demonstrate the effectiveness of skeleton clustering in dealing with different data scenarios and to guide some choices in the framework for applications.
In section \ref{sec::real}, we test the performance of skeleton clustering on real datasets.
In section \ref{sec::conclusion}, we conclude the paper and point out some directions for future research.

\section{Skeleton Clustering Framework}	\label{sec::clustering}

\vspace{-1em}
\begin{algorithm}
\caption{Skeleton clustering}
\label{alg::SC}
\begin{algorithmic}
\State \textbf{Input:} 
Observations $X_1,\cdots, X_n$, final number of clusters $S$.
%Grid points $x_1,\cdots,x_N$ and the functional evaluations $f(x_1),\cdots,f(x_N)$.
\State 1. {\bf Knot construction.} Perform $k$-means clustering with a large number of $k$; the centers are the knots (Section \ref{sec::knots}). 
% Generally, we choose $k = [\sqrt{n}]$.
\State 2. {\bf Edge construction.} 
Apply approximate Delaunay triangulation to the knots (Section \ref{sec::edge}).
\State 3. {\bf Edge weights construction.}
Add weights to each edge using 
either Voronoi density, Face density, or Tube density similarity measure (Section~\ref{sec::construction}). 
\State 4. {\bf Knots segmentation.}
Use linkage criterion to segment knots into $S$ groups based on the edge weights (Section \ref{sec::segmenting}).
\State 5. {\bf Assignment of labels.}
Assign a cluster label to each observation based on which knot group the nearest knot belongs to (Section \ref{sec::assignment}). 

%\State \textbf{Output:} 
\end{algorithmic}
\end{algorithm}

%flow chart of the skeleton clustering procedure
% \begin{figure}
% \centering
% \includegraphics[width=13cm]{Skeleton_flowchart2.png}
% \caption{Skeleton Clustering Flow Chart. {\color{magenta}YC: Given that we have algorithm \ref{alg::SC}, we probably do not 
% need this flow chart--it does not provide any additional information.}}
% \label{fig::flow}
% \end{figure}

~~~~~In this section, we formally introduce the skeleton clustering framework. Let $\XX= \{X_1, \dots, X_n\}$ be a random sample from an unknown distribution with density $p$ supported on a compact set $\calX \in \RR^d$. 
The goal of clustering is to partition $\XX$ into clusters $\XX_1, \dots \XX_S$, where $S$ is the final number of clusters. 

\begin{figure}
\captionsetup{skip=1pt}
\centering
    \begin{subfigure}[t]{0.18\textwidth}
        \centering
        \includegraphics[width=\linewidth]{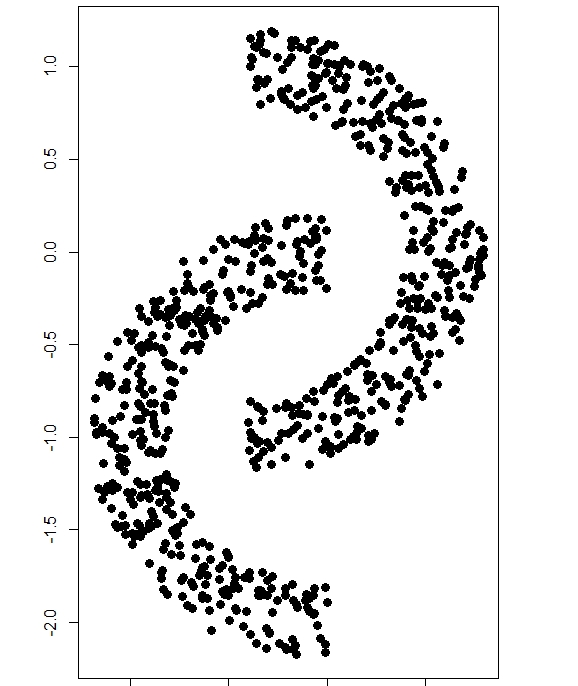} 
        \caption{Data}
    \end{subfigure}
    \begin{subfigure}[t]{0.18\textwidth}
        \centering
        \includegraphics[width=\linewidth]{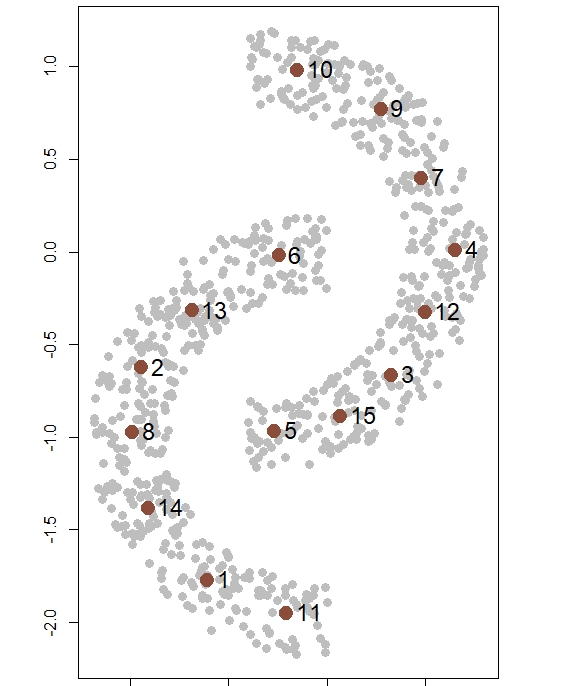}
        \caption{Knots}
    \end{subfigure}
    \begin{subfigure}[t]{0.18\textwidth}
        \centering
        \includegraphics[width=\linewidth]{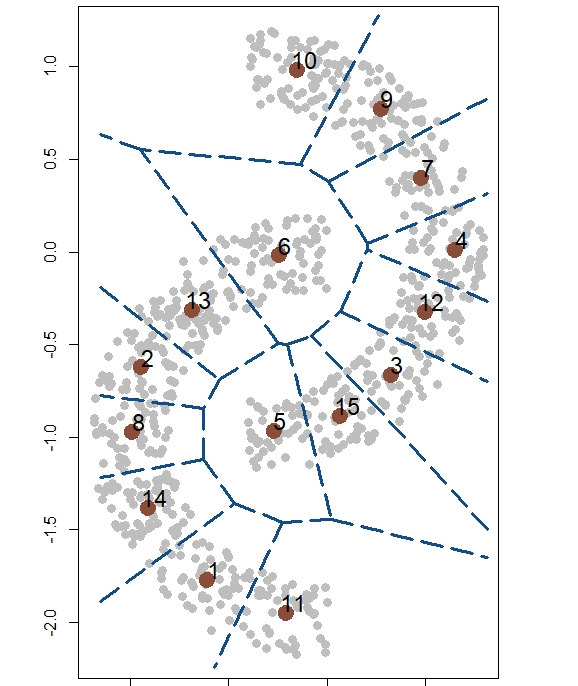}
        \caption{Voronoi Cells}
    \end{subfigure}
    \begin{subfigure}[t]{0.18\textwidth}
        \centering
        \includegraphics[width=\linewidth]{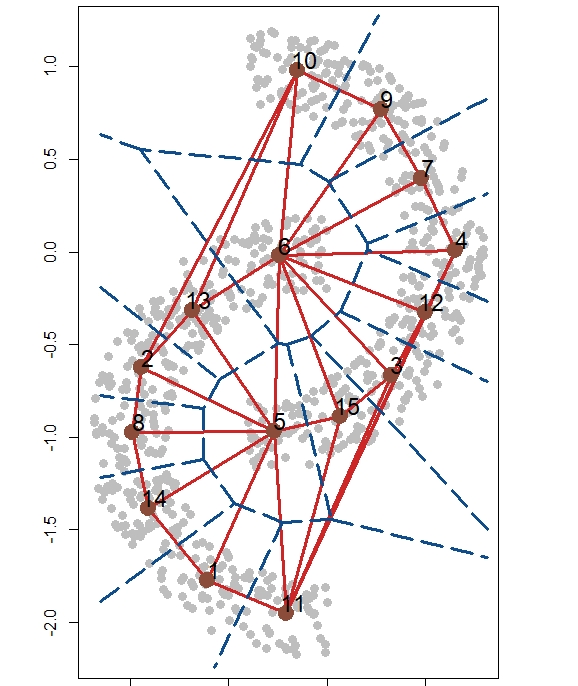} 
        \caption{Skeleton}
    \end{subfigure}\\
    \begin{subfigure}[t]{0.18\textwidth}
        \centering
        \includegraphics[width=\linewidth]{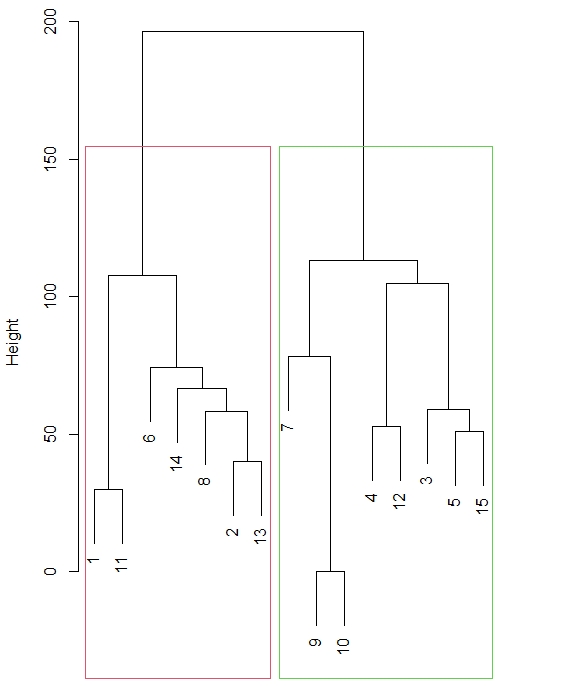} 
        \caption{Dendrogram}
    \end{subfigure} 
    \begin{subfigure}[t]{0.18\textwidth}
        \centering
        \includegraphics[width=\linewidth]{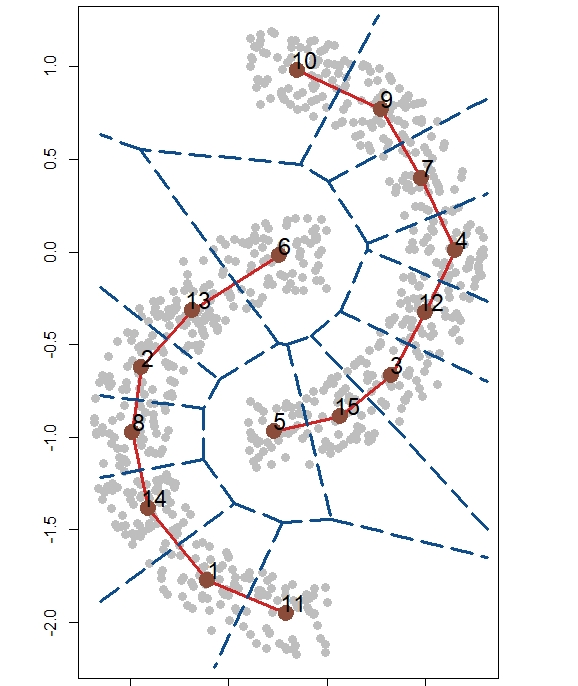} 
        \caption{Segmentation}
    \end{subfigure}  
    \begin{subfigure}[t]{0.17\textwidth}
        \centering
        \includegraphics[width=\linewidth]{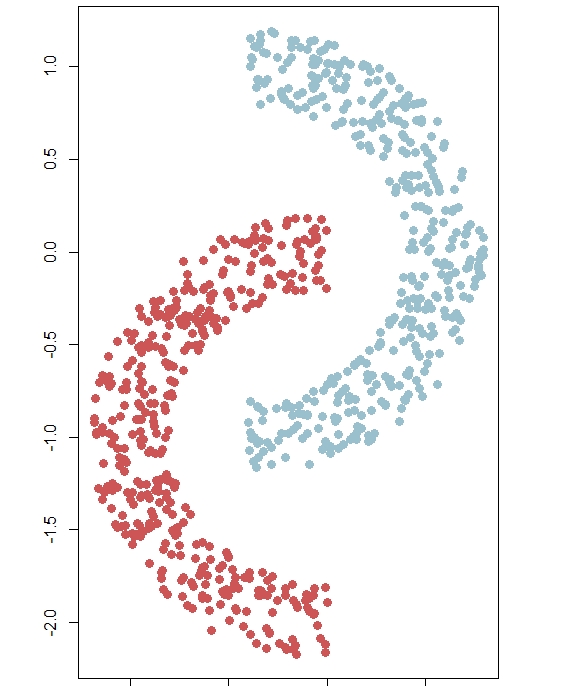} 
        \caption{Clustering}
    \end{subfigure} 
\caption{Skeleton Clustering illustrated by Two Moon Data (d=2).}
\label{fig::ex02}
\end{figure}

A summary of the skeleton clustering framework is provided in Algorithm~\ref{alg::SC}.
%We illustrate the procedure with simulated Two Moon data in Figure \ref{fig::ex02}. 
Figure \ref{fig::ex02} illustrates the overall procedure of the skeleton clustering method.
Starting with a collection of observations (panel (a)),
we first find knots, the representative points of the entire data (panel (b)). 
%In this work the knots are chosen by the $k$-means algorithm. 
Then we compute the corresponding Voronoi cells induced by the knots (panel (c))
and the edges connecting the nearby Voronoi cells (panel (d)). 
For each edge in the graph, we compute a density-aided similarity measure that quantifies
the closeness of each pair of knots.
For the next step, we segment knots into groups based on a linkage criterion (single linkage in this example), leading to the dendrogram in panel (e). 
Finally, we choose a threshold that 
cuts the dendrogram into $S = 2$ clusters (panel (f))
and assign a cluster label to each observation according to the knot-cluster that it belongs to (panel (g)).

%Panel (a) shows the scatter plot of the data. Panel (b) displays the knots  chosen by K-means and panel (c) gives the Voronoi cells induced by the knots. Panel (d) displays the constructed skeleton (knots and edges) of the data with bones as the red line segments. We then compute the weight of edges and segment knots using a linkage criterion. This leads to the dendrogram in panel (e), which clearly shows two major clusters.  We choose a threshold that cuts the dendrogram into $S = 2$ clusters, and the segmented skeleton structure is illustrated in panel (f). The final clustering result is given in panel (g).

In summary,
the skeleton clustering consists of the following five steps:
% \begin{itemize}
% \setlength\itemsep{-0.3em}
%     \item[1.] Knots construction. 
% %    Constructing knots from the data.
% %    Construct knots for the data. The number of knots $k$ is greater than the final number of clusters $S$.
%     \item[2.] Edges construction.
% %    Constructing edge between the knots.
%     \item[3.] Edge weights construction.
% %    Measuring the weights of each edges.
%     \item[4.] Knots segmentation.
% %    Segmenting skeleton structure based on the weight matrix.
%     \item[5.] 
% %    Assigning cluster labels according to the segmented skeleton structure.
% \end{itemize}
(1) Knots construction, (2) Edges construction, (3) Edge weights construction, (4) Knots segmentation, and (5) Assignment of labels.
In what follows in this section, we provide a detailed description of each step except Step 3.
Step 3 is the key step in our clustering framework where we incorporate the information from the underlying density for clustering in a less dimension-dependent way and we defer the detailed discussion of Step 3 to
%We provide a comprehensive discussion of this in 
Section~\ref{sec::construction}
%and theoretical analysis on the weights in 
and
Section~\ref{sec::theory}.
We include a short analysis of the computational complexity of our skeleton clustering framework in Appendix \ref{sec::complexity}.

%	Classifying each data point with respect to its nearest knot we get the final clustering in graph (d), and we see that this Skeleton clustering procedure successfully captures the data structure.

%With the overall framework introduced, we then discuss the main methods for each step in the procedure in detail. Some extensions about modifying the procedure for uncertainty clustering, anomaly detection, and dimension reduction are discussed in Appendix A, and the computation complexity of Skeleton Clustering is discussed in Appendix D.

\subsection{Knots Construction}	\label{sec::knots}
% \revise{ \citep{Aragam2020} reconstruct nonparametric mixing measures by fitting the data with $L \gg K$  general nonparametric mixture components and then partition into $K$ clusters.}

~~~~The construction of knots is a step aiming at finding representative points in the data that can help measure similarities between regions in the later stage. 
%to reduce the complexity of data.
The knots can be viewed as landmarks inside the data where we can shift our focus from the entire data to these local locations.
%We first detail the ways to construct representative knots from high-dimensional data. The main goal of this step is to reduce the high-dimensional large-size data to a reduced representation, on which classical low-dimensional methods can apply and computation complexity is reduced. To be representative, we want the chosen knots to capture the main data structure, and this naturally relates to the prototype methods for clustering mentioned in the Introduction section. In this work, we take K-means as the default method for knots construction and discuss it in detail below. 
%\subsubsection{Overfitting $k$-means}
A simple but reliable approach for constructing knots 
is the $k$-means algorithm.
We apply the $k$-means algorithm with a large number $k \gg S$  the desired number of final clusters, and this procedure behaves like overfitting the $k$-means.
Notably, we do not use the $k$-means procedure to obtain final clustering, but, instead, we use it as an intermediate step to find concise representations of the original data.
% In panel (a) of Figure~\ref{fig::ex02}, 
% the knots constructed by overfitting $k$-means form a concise representation of the original data.

%The choice of $k$  is often a challenging task in practice. 
The number of knots $k$ is a key parameter in the knots construction step. 
It controls the trade-off between the quality of the data representation and the reliability of each knot. 
More knots can give a better representation of the data, but, if we have too many knots,
the number of observations per knot will be small,
so the uncertainty in estimation in the later stage will be large.
We find that a simple reference rule for $k$ to be around $\sqrt{n}$ works well in our empirical studies (Section \ref{sec::KnotSize}).
%In practice, our skeleton clustering methods with the newly proposed edge measures 
%give reliable clustering performance as long as the number of knots is sufficiently large.
% While this reference rule seems to be reliable,
% we would recommend to examine the knot-size distribution (Figure~\ref{fig::ClusSize}, for more plots see Appendix ~\ref{sec::KnotSize}) 
% before making the final decision.
% The knot-size distribution is the empirical distribution
% size of each knot (number of observations of each knot) with a given choice of $k$.
 {In practice, it is also advisable to prune knots with a small number of corresponding observations because the density-aided weights (in Step 3, Section \ref{sec::construction}) are estimated locally by the data belonging to each pair of knots. Knots with a few data points can lead to unstable similarity measurements and unreliable final clustering.
Moreover, to take care of observations in the low-density areas that could cause problems for the $k$-means clustering,
one may first pre-process or denoise the data by removing observations in the low-density area and
then apply the $k$-means clustering to find out the knots.}

In this work, we use overfitting $k$-means as the default way for knots construction, but there are alternative approaches to find knots such as subsampling, the coreset construction methods \citep{bachem2017practical, coresetDensityEst}, and the Self-Organizing Maps (SOM) \citep{SOM2001}.
%However, they do not perform well in practice;
We show in Appendix \ref{sec::SOM} 
that the SOM can also be used to find knots
but requires more careful treatments such as removing knots with few or even no observations and the performance is slightly worse than that of the overfitting $k$-means.
 {The $k$-medians algorithm can be another alternative method
but it gave an unstable result when the dimension is large. Therefore, we choose to use the overfitting $k$-means algorithm in this work and recommend using it in practice.}

\begin{remark}
{\small
Since the $k$-means algorithm does not always find the global optimum, we repeat it many times with random initial points (generally $1,000$ times or more) and choose the one with the optimal objective function. This works well for all of our numerical analyses. Moreover, since we are only using $k$-means as a tool to find a useful representation, we do not need to find the actual global optimum. All we need is a set of knots forming a useful representation.}
%K-means is known for its tendency to converge to local minimums, and, with a large number of knots, the overfitting K-means step almost surely cannot find the global minimum, even with a large number of random starts. The uncertainty in this step of the algorithm can lead to changes in the final clustering performance. But based on empirical studies, local minimums of the K-means algorithm are sufficient for good clustering results.}
\end{remark}
% \begin{figure}
% \captionsetup{skip=1pt}
% \centering
% %    \begin{subfigure}[t]{0.28\textwidth}
% %        \centering
% %        \includegraphics[width=\linewidth]{Yinyang_MST.JPG} 
% %        \caption{}
% %    \end{subfigure}
% %    \begin{subfigure}[t]{0.34\textwidth}
% %        \centering
%         \includegraphics[width=6cm]{figures/Yinyang_clusdistr_k57}
% %        \caption{}
% %    \end{subfigure}
% \caption{The knot-size diagram: a histogram of cluster sizes given by $k$-means with $k= [\sqrt{n}]$ on Yinyang data with dimension $200$;
% see Section~\ref{sec::YY} for more details.}
% \label{fig::ClusSize}
% \end{figure}

% {\color{magenta}YC: I remove these two remarks since they are not directly answering any referee's comments
% and are too short. A remark has to be a thoughtful comment and generally consists of 2 or more sentences.}

%\revise{
%\begin{remark} \label{remark::denoise}
%{\small For data with noisy points it can be advised to first denoise the data before constructing the knots. A practical denoising method can be removing low density points based on some nearest neighbor based density estimators.}
%\end{remark}}
%
%\revise{
%\begin{remark}
%{\small 
%Note that
%the Approximate-Nearest-Neighbor finding algorithm from \citep{InvertedFiles} (call \textit{the inverted files}) 
%also suggest to partition/cluster the data  into Voronoi cells using k-means as the first step with $\sqrt{n}$ centers.
%}
%\end{remark}}

\vspace{-1em}
\subsection{Edges Construction}	\label{sec::edge}
\begin{figure}[ht]
\captionsetup{skip=-1em}
\centering
\includegraphics[width=4cm]{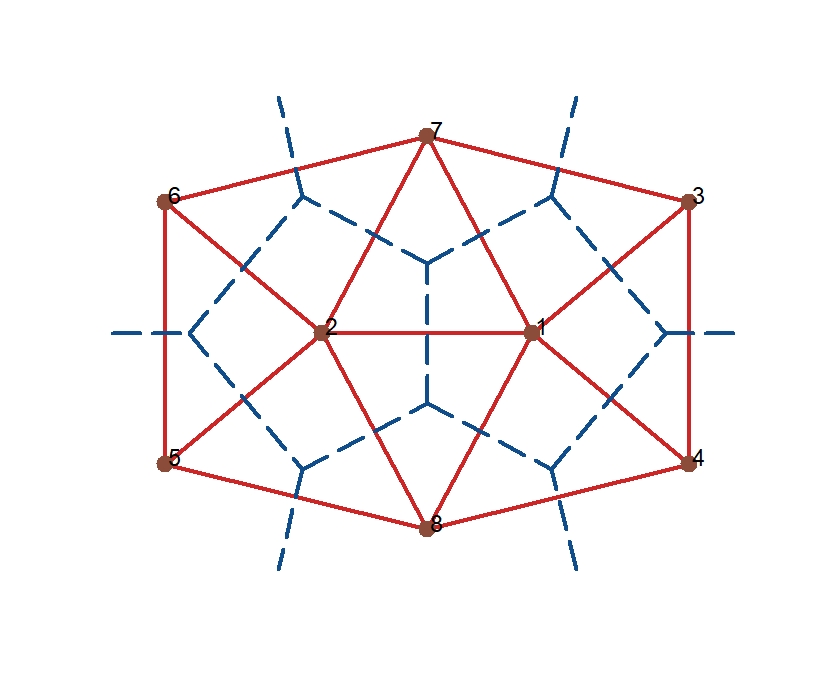}
\caption{Voronoi Tessellation as blue dashed lines and Delaunay Triangulation by red solid lines.}
\label{fig::DT}
\end{figure}

~~~~With the constructed knots, our next step is to find the edges connecting them. 
Let $c_1,\cdots, c_k$ be the given knots and we use $\calC  = \{c_1,\cdots, c_k\}$
to denote their collection of them.
We add an edge between a pair of knots if they are neighbors, with the neighboring condition being that the corresponding Voronoi cells \citep{voronoi1908recherches}
share a common boundary. 
The Voronoi cell, or Voronoi region, $\CC_{j}$, associated with a knot $c_{j}$ is the set of all points in $\calX$ whose distance to $c_{j}$ is the smallest
compared to other knots  (See Figure \ref{fig::DT}).
That is, 
\begin{align}
    \CC_j = \{x \in \calX: d(x, c_j) \leq d(x, c_\ell) \ \  \forall \ell \neq j\},
\end{align}
where $d(x,y)$ is the usual Euclidean distance.
Therefore, we add an edge between knots $(c_i,c_j)$ if $\CC_i\cap \CC_j\neq \emptyset$.
Such resulting graph is the Delaunay triangulation \citep{Delaunay1934}
of the set of knots $\calC$ and we denote it as $DT(\calC)$.
In a nutshell, the skeleton graph in our framework is given by the Delaunay triangulation of $\calC$.

\vspace{-0.5em}
The Delaunay triangulation graph is conceptually intuitive and appealing and is utilized by some clustering methods to identify connected components \citep{ azzalini2007clustering, SCRUCCA2016}, but empirically the computational complexity of the exact Delaunay triangulation algorithm has an exponential dependence on the ambient dimension $d$ \citep{DelaunayComplexity,Chazelle1993}. 
Given our multivariate and even high-dimensional data setting, exact Delaunay triangulation is empirically unfavorable due to its high computational cost \citep{VoronoiGraphTraversal}. 
Therefore, in practice, we approximate the exact Delaunay Triangulation with $\hat{DT}(\calC)$ by examining the 2-nearest knots of the sample data points.
%(the Voronoi Density regions in \ref{sec::VD} ). 
The key observation is that, if the Voronoi cells of two knots $c_i, c_j$  share a nontrivial boundary, there is likely to be a non-empty region of points whose 2-nearest knots are $c_i, c_j$. Consequently, for approximation, we query the two nearest knots for each data point and have an edge between $c_i, c_j$ if there is at least one data point whose two nearest neighbors are $c_i, c_j$. The complexity of the neighbor search depends linearly on the dimension $d$, which is desirable for high-dimensional setting \citep{search1998}, and this sample-based approximation to the Delaunay Triangulation has reliable empirical performance.

% Conceptually, a sample point in between serves as a bone connecting two knots, 

\subsection{Edge Weight Construction}

~~~~Given the constructed edges and knots,
we assign each edge a weight that represents the similarity between the pair of knots. 
In this work, we propose some novel density-aided quantities as the edge weights. 
Since the description of the similarity measures is more involved,
we defer the detailed discussion of the similarity measures to Section~\ref{sec::construction}. It is worth noting here that the similarity measures proposed in this work are estimated based on surrogates of the underlying density function (hence density-aided) and the estimation procedure has minimal dependence on the ambient dimension. Therefore, the estimations of the newly proposed similarity measures are reliable even under high-dimensional settings.

\subsection{Knots Segmentation}	\label{sec::segmenting}

%So far, we have already obtained the knots and edges and the corresponding weights (often represented by
%a weight matrix).
~~~~Given the weighted skeleton graph, the next step is to partition the knots into the desired number of final clusters, and we apply hierarchical clustering by converting the similarity measures into distances. 
Particularly, for given similarity measures $\{s_{ij}\}_{i\neq j}$ where only connected pairs can take nonzero entries and let $s_{\max} = \max_{i \neq j} s_{ij} $, we define the corresponding distances as
$d_{ij} = 0$ if $i = j$ and $d_{ij} = s_{\max} - s_{ij}$ otherwise.

The choice of linkage criterion for hierarchical clustering  may depend on the underlying geometric structure of the data. We analyze several linkage criteria under various simulation scenarios in Appendix \ref{sec::simLinkage}. Generally, single linkage gives reliable clustering results when the components are well-separated, but average linkage works better when there are overlapping clusters of approximately spherical shapes. Therefore, in practice, such a choice of linkage should be made based on some exploratory understanding of the data structure, and experimenting with different linkage methods is computationally tractable as only the knots need to be segmented.

The number of final clusters $S$ is an essential parameter for the hierarchical clustering procedure but can be unknown. 
The dendrograms given by hierarchical clustering can be a helpful tool in this situation, displaying the clustering structure at different resolutions. 
Consequently, analysts can experiment with different numbers of final clusters and choose a cut that preserves the meaningful structures based on the dendrograms, which takes little extra computation.  {However, it is worth pointing out that with the presence of noisy data points, the final number $S$ being larger than the true number of meaningful components may be needed to achieve better clustering results (see Appendix \ref{sec::simLinkage}).}

\begin{remark}
\small
Although the dendrogram for knots given by our method is not exactly the cluster trees, the pruning graph cluster tree procedure proposed in \cite{nugent2010clustering} with excess mass can be applied to help decide the final segmentation.
\cite{peterson2018merging} also presented similar ideas choosing the final number of clusters by looking at the lifetime of the clusters in the dendrogram. 
Additionally, the traditional ``elbow'' methods can be used to determine the number of clusters. An inferential choice can also be made using the gap statistics \citep{Tibshirani2001}. 
\end{remark}

\vspace{-1em}
\subsection{Assignment of Labels}	\label{sec::assignment}

~~~~In the previous step, we created $S$ groups of knots and each group has a cluster label.
To pass the cluster membership to each observation,
we assign a hard clustering label to each observation according to which group its nearest knot belongs.
For instance, if an observation $X_i$ is closest to knot $c_j$ and $c_j$ belongs to cluster $\ell$,
we assign cluster membership label $\ell$ to observation $X_i$. 
% This step assigns a hard clustering label to every observation.

 {
\begin{remark}
{\small There are other methods in clustering literature for assigning labels of observations based on  identified structures. \cite{azzalini2007clustering} and \cite{SCRUCCA2016} assign unlabelled data based on density ratios. DBSCAN and HDBSCAN  \citep{Campello2015, EsterKriegel1996} assign labels (and identify noisy points) based on k-nearest-neighbor considerations. One may use these alternatives to assign the cluster label to each observation.}
%The simple and direct approach in this framework works well in practice, but for future research, noise detection techniques can be incorporated at this step.}
\end{remark}
}

%gives the final clustering of all the data. 

%As the last step in the Skeleton Clustering framework, we assign individual labels to all data points based on the segmented skeleton. For notation, we assume the $k$ knots are already segmented into $S$ parts, and each knot has a cluster label. A natural way to assign a label to an individual data point is through the nearest knot calculation, where the label of a data point is assigned as the label of the knot closet to that point. The one nearest knot identification can be generalized to $m$-nearest knots \citep{Chen2016} and is discussed in Appendix A. By default the one nearest knot calculation is employed for this step of skeleton clustering if not specified otherwise.

\vspace{-1em}
\section{Density-Based Edge Weights Construction}	\label{sec::construction}

%To preserve the advantage of density-based clustering in our framework, 
~~~~To incorporate the information of density into clustering, we calculate the edge weights in the constructed skeleton based on the underlying density function.
However, the conventional notion of PDF is not feasible
in multivariate or even high-dimensional data due to the curse of dimensionality.
To resolve this issue,
we introduce three density-related quantities
that are estimable even when the dimension is high.

%
%In this section, we propose different ways to measure bone weights given the skeleton structure with knots $\{c_1, \dots, c_K\}$ and the corresponding Voronoi cells $\{C_1, \dots, C_K\}$. We follow the density-based clustering approach  \citep{EsterKriegel1996, Chacon2015} to construct the similarity measures using quantities that are related to the probability density. The density-based clustering approach has the appealing property that it is easy to define the population versions of the quantities used in clustering. This allows us to study the large sample theory and statistical consistency and the resulting clusters are easy to interpret.
%However, most density-based clustering approaches suffer from the curse of dimensionality because estimating the density function often has an error rate $O_p(n^{-\frac{2}{d+4}})$ \citep{Kim, Chen2016}, which is very slow when the dimension $d$ is greater than $2$. To handle this problem, we use some surrogates of density instead of the density function itself to measure the connection strength between knots. The use of density surrogates leads to a fast convergence rate even in high-dimensional settings, which is studied in Section 4.

\subsection{Voronoi Density}	\label{sec::VD}

% \revise{\citep{Shin2019} connects the spherical regions in their algorithm by looking at if there exists at least a single point in the intersection part.}

\begin{figure}[ht]
\captionsetup{skip=1pt}
\centering
\includegraphics[height=4.5cm]{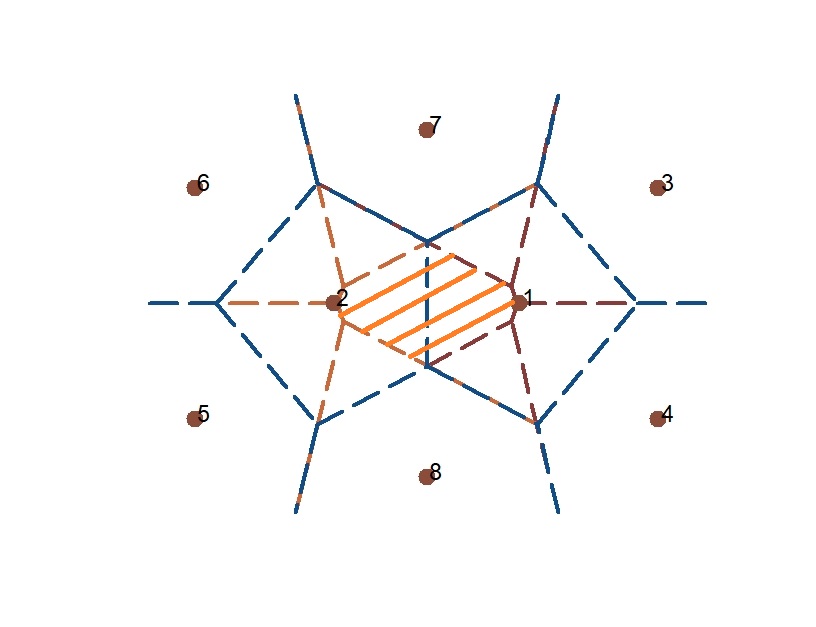}
\includegraphics[height=4cm]{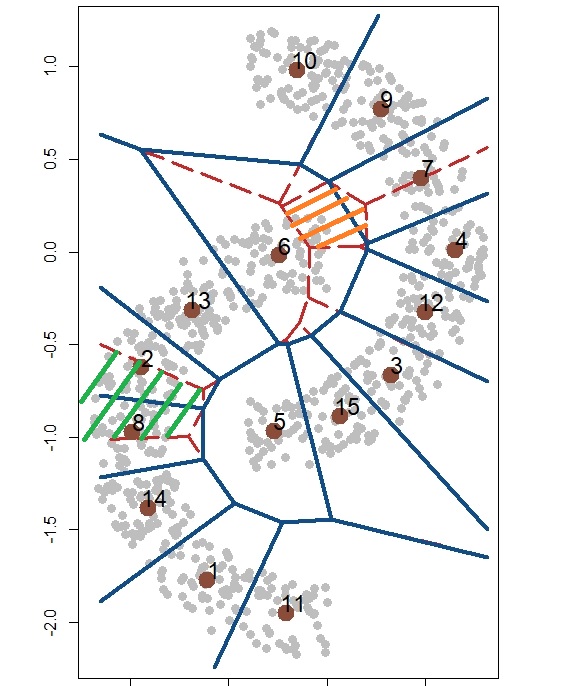}
\caption{{\bf Left:} Orange shaded area illustrates the 2-NN region of knots $1,2$.
{\bf Right:} Shaded areas illustrate the 2-NN region of knots $6,7$ and knots $2,8$.}
\label{fig::2nn}
\end{figure}

~~~~The \emph{Voronoi density (VD)} measures the similarity between a pair of knots $(c_j,c_\ell)$
based on the number of observations whose 2-nearest knots are $c_j$ and $c_\ell$. 
We start with defining the Voronoi density based on the underlying probability measure and then
introduce its sample analog. 
Given  a metric $d$ on $\RR^d$, the 2-Nearest-Neighbor (2-NN) region of  a pair of knots $(c_j,c_\ell)$ is defined as
\begin{align}
\label{eq::define2NN}
    A_{j\ell} = \{ x \in \calX: d(x, c_i) > \max\{ d(x,c_j), d(x, c_\ell) \}, \forall i \neq j, \ell \}.
\end{align}
In this work, we take $d(.,.)$ to be the usual Euclidean distance and use $||.||$ to denote the Euclidean norm. 
An example 2-NN region of a pair of knots is illustrated in Figure~\ref{fig::2nn}.

%
%For motivation of Voronoi density, we can think of each data point as a muscle fiber connecting the two nearest knots, and hence the strength of the connection between knots, the bone weights, can be assessed by the intensity of muscle fibers attached. With this idea, we say a data point $x \in \calX$ is attached to the edge $E_{j\ell}$ connecting $c_j,c_\ell$ if the knots $c_j, c_\ell$ are the two closest knots to point $x$, and the connection intensity between $c_j, c_\ell$ can be measured by the number of such $x$ attached to the edge $E_{j\ell}$. To frame in another way, for two knots $c_j, c_\ell$, we define the 2-Nearest-Neighbor region as 

%We then att
%and all the points in $A_{j\ell}$ are attached to the edge $E_{j\ell}$. 
%Also, we note that the strength of individual fiber is inversely proportional to its length. Therefore, we define the Voronoi density as
%We count the number of observations within $A_{j\ell}$
%and divide this number by
%the distance between the pair $(c_j,c_\ell)$, which leads to the formal definition of Voronoi density:

% and Figure~\ref{fig::VDex}.

Following the idea of density-based clustering, two knots $c_j, c_\ell$ belong to the same clusters if they are in a connected high-density region, and we would expect the 2-NN region of $c_j, c_\ell$ to have a high probability measure. Hence, the probability $\PP(A_{j\ell}) = P(X_1 \in A_{j\ell})$ can measure the association between $c_j$ and $c_\ell$ (see illustration in Figure~\ref{fig::2nn} right).
Based on this insight, the Voronoi density measures the edge weight of  $(c_j,c_\ell)$ with
\begin{align}
    S_{j\ell}^{VD} = \frac{\PP(A_{j\ell})}{\|c_j - c_\ell \|}.
\end{align}
Namely, we divide the probability of the in-between region by the mutual Euclidean distance.
The division of the distance adjusts for the fact that 2-NN regions have different sizes and provides more weights to edges between knots close in distance. 
However, such division makes the Voronoi density to be in the unit of $1/\norm{c_j - c_\ell}$ and hence can be scale-dependent.
%\begin{figure}[h]
%\captionsetup{skip=1pt}
%\centering
%\includegraphics[width=5cm]{delvor_2nnshaded.jpeg}
%\caption{Shaded areas illustrate the 2-NN region between knots $6,7$ and knots $2,8$.}
%\label{fig::VDex}
%\end{figure}
%We name it Voronoi density since its based on the 2-NN region and the division of distance make it like a density measure of the fiber connecting two knots.
%Note that even if we do not restrict ourselves to pairs 
%$(c_j,\c_\ell)$ in 
%
%\begin{lem}
%Let $DT(\calC)$ be the Delaunay triangulation of $\calC$ using Euclidean distance. If $c_j$ and $c_\ell$ has no edge in $DT(\calC)$, then $S^{VD}_{j\ell} = 0$
%\end{lem}
%{\color{magenta}YC: given that we only define weights for edges, there is no need for this lemma:
%
%\begin{lem}
%Let $DT(\calC)$ be the Delaunay triangulation of $\calC$ using Euclidean distance. If $c_j$ and $c_\ell$ has no edge in $DT(\calC)$, then $S^{VD}_{j\ell} = 0$
%\end{lem}
%}
%Let $\hat \calC \equiv \{\hat c_1, \dots, \hat c_k\}$ be the set of knots constructed based on sample data, and let $n$ be the sample size, the sample estimator of $S_{j\ell}^{VD}$ can be constructed as:
%The sample analogue of $ S_{j\ell}^{VD}$ is the quantity

In practice, we estimate $ S_{j\ell}^{VD}$ by a sample average. 
Specifically, the numerator $\PP(A_{j\ell}) $ is estimated by $\hat{P}_n (A_{j\ell}) = \frac{1}{n}\sum_{i=1}^n I(X_i\in A_{j\ell})$
and the final estimator for the VD is
\begin{align}
    \hat{S}_{j\ell}^{VD} &= \frac{\hat{P}_n ({A}_{j\ell})}{\|{c}_j - {c}_\ell\|}.
\end{align}
Note that here we are assuming that $c_1,\cdots, c_k$
as given beforehand. 
In the sample version, we replace them with the sample analog $\hat c_1,\cdots, \hat c_k$
and replace the region $A_{j\ell}$ by $\hat A_{j\ell}$.

The Voronoi density can be computed in a fast way. 
The numerator, which only depends on $2$-nearest-neighbors calculation, can be computed efficiently by the k-d tree algorithm \citep{kdtree}.  
For high-dimensional space, space partitioning search approaches like the k-d tree can be inefficient but a direct linear search still gives a short run-time \citep{search1998}, and with a large number of observations approximate nearest neighbor algorithms can be incorporated. 
The denominator requires distance calculation and can be burdensome in high-dimensional settings, but note that we only need to calculate the distance for edges present in $\hat{DT}(\calC)$, which is far less than $k(k-1)/2$, where $k$ is the number of knots. 
Hence, the calculation of VD can be carried out in a fast way even for high-dimensional data with a large sample size.

\subsection{Face Density}	\label{sec::FD}

% \revise{The ridgeline elevation function proposed by \citep{RayLindsay2005}, from a parametric mixture perspective, looks at the ``density ridge'' between two modes and share same ideology here. The ``ridge'' is in a sense the edge we are having here.}

~~~~Here we present another density-based quantity to measure the similarity between two knots.
Since the Voronoi cell of a knot describes the associated region,
a natural way to measure the similarity between two knots is to investigate the shared boundary of the corresponding Voronoi cells. 
If two knots are highly similar,
we would expect the boundary to lie in a high-density region and to be surrounded by many observations. 
% The intuition is to look at the boundary between the corresponding Voronoi cells. If the two knots belong to the same high-density region, then their face region should have high density also. 
Based on this idea, we define the \emph{Face Density (FD)} as the integrated PDF over the ``face'' (boundary) region. 
Note that, although the density is involved in FD, by integrating over the face region the problem reduces to a 1-dimensional density estimation task regardless of the dimension of the ambient space.
Formally,
let the face region between two knots $c_j, c_\ell $ be $F_{j\ell} = {\CC}_j \cap {\CC}_\ell$. 
At the population level, the FD is defined as 
\begin{align}
    S_{j\ell}^{FD} = \int_{F_{j\ell}} p(x) \mu_{d-1}(dx) = \int_{F_{j\ell}}  d \PP(x),
\end{align}
where $ \mu_{m}(dx)$ denotes the $m$-dimensional volume measure.

%{\color{magenta}YC: I don't think we need the following lemma since we only define all quantities for pairs
%that are neighboring to each other.
%
%\begin{lem}
%Let $DT(\calC)$ be the Delaunay triangulation of $\calC$ using Euclidean distance. If $c_j$ and $c_\ell$ has no edge in $DT(\calC)$, then $S^{FD}_{j\ell} = 0$
%\end{lem}
%
%}

To estimate the FD, we utilize the idea of kernel smoothing in combination with data projection.
%However, the density function $p$ is unknown in practice and we need to make an estimation.
% A naive way is to estimate the high-dimensional density function directly, but high-dimensional density estimation has a slow convergence rate \citep{Chen2017, Chacon2011, Wasserman2006}. 
By the construction of the Voronoi diagram, the boundary of two Voronoi cells is orthogonal to the line passing through the two corresponding knots (called the `central line') and intersects the central line at the middle point regardless of the dimension of the data (see Figure~\ref{fig::DT} for reference).
Therefore, we estimate the FD by first projecting the observations onto the central line and then using the $1$-dimensional kernel density estimator(KDE) to evaluate the density at the midpoint.
Specifically, 
%we project observation onto the $1$-dimensional line, perform $1$-dimensional kernel density estimation, and take the density in the middle as the estimator. 
fix two knots ${c}_j, {c}_\ell$, let ${  \CC_j}, { \CC_\ell}$ be the corresponding Voronoi cells, and denote $\Pi_{j\ell}(x)$ as the projection of $x \in \calX$ onto the central line passing through ${c}_j$ and  ${c}_\ell$, we define the estimator $\hat{S}^{FD}_{j\ell}$ to be
\begin{equation}
    \hat{S}_{j\ell}^{FD}  = \frac{1}{n h} \sum_{X_i \in {  \CC_j} \cup {  \CC_\ell} } K \bigg(\frac{\Pi_{j\ell}(X_i)  - ({c}_\ell + {c}_j)/2}{h} \bigg)
    \label{eq::AD::est}
\end{equation}
where $K$ is a smooth, symmetric kernel function (e.g. Gaussian kernel) and $h>0$ is the bandwidth
that controls the amount of smoothing. It is noteworthy that, while conventional kernel smoothing suffers
from the curse of dimensionality \citep{Chen2017,Chacon2011, Wasserman2006}, 
the kernel estimator in equation \eqref{eq::AD::est} 
bypasses it.

\subsection{Tube Density}	\label{sec::TD}
\begin{figure}
\captionsetup{skip=1pt}
\centering
\includegraphics[width=3cm]{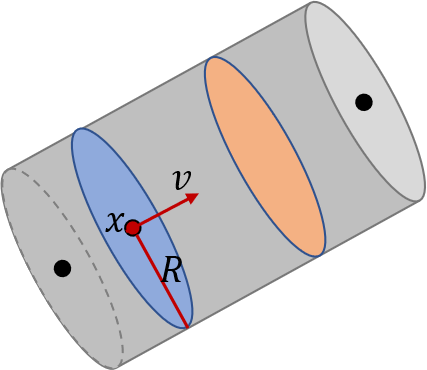}
\caption{The disk area centered at $x$ with a radius $R$ and a direction $\nu$.}
\label{fig::Disk}
\end{figure}

~~~~While FD is conceptually appealing, the characterization of the face between two Voronoi cells could be challenging
since the shapes of the boundaries can be irregular, and the characteristics of the boundaries can affect the estimation of the Face density from a theoretical perspective (Section \ref{sec::FDproof}).
%This leads to a slower convergence rate of estimating the FD (see Theorem~\ref{thm::FD}).
Here we propose a measure similar to the Face density measure but has a predefined regular shape. For a point $x$, we define 
the \textit{Disk Area} centered at $x$ with radius $R$ and normal direction $\nu$  (see Figure ~\ref{fig::Disk} for an illustration) as 
\begin{align}
    \Disk(x, R, \nu) = \{y : ||x-y|| \leq R, (x-y)^T \nu = 0\}
\end{align}
To measure the similarity between knots $c_j$ and $c_\ell $, we examine the integrated density within the disk areas along the central line. 
In more detail, the central line can be expressed as
%with normal vector $c_\ell - c_j$. 
%We parametrize the line segment by $t$ that 
$\{c_j + t(c_\ell - c_j): t \in [0,1]\}$, and any point on the central line can be written as $c_j + t(c_\ell - c_j)$
for some $t$.
For a point $c_j + t(c_\ell - c_j)$, 
we define the integrated density in the disk region (called \textit{Disk Density}) as
%Specifically, we look at the  disk density with given $c_j, c_\ell$, $R$ at each $t$  as
\begin{align}
    \pDisk_{j\ell,R}(t) = \PP \left(\Disk(c_j + t(c_\ell - c_j), R, c_\ell - c_j) \right)= \int_{\Disk(c_j + t(c_\ell - c_j), R, c_\ell - c_j)} p(x) dx.
\end{align}
The \emph{Tube Density (TD)} measures the similarity between $c_j$ and $c_\ell$ as the minimal disk density along the central line, i.e.,
%and the Tube density between $c_j$ and $c_\ell$ is defined as
\begin{align}
 S_{j\ell}^{TD} = \inf_{t \in [0,1]} \pDisk_{j\ell,R}(t)
\end{align}
In other words, with given $c_j, c_\ell$, we survey all Disk Density along the central line and retrieve the infimum as the similarity measure between two knots. 
% Since the regions being surveyed together form a tube, we call it the Tube density. 

%\begin{remark}
%The central line between two neighboring knots may not pass through the face region between the two knots (see Figure ~\ref{fig::ex02} panel (d)).
%\end{remark}
 {In this work, we set $R$ based on the root mean squared distances within each Voronoi cell.
Specifically, for knot $c_j$ and the corresponding Voronoi cell $\CC_j$, we calculate
\begin{align}
    R_j = \sqrt{\frac{1}{\abs{\CC_j}-1} \sum_{X_{\ell} \in \CC_j} \norm{X_\ell - c_j}^2}
\end{align}
where $\abs{\CC_j}$ denotes the size of set $\CC_j$. With the uniform radius paradigm where the radius is the same for all pairs of knots, we set $R = \frac{1}{k} \sum_{j=1}^k R_j$. 
Our empirical studies show that this rule leads to good clustering performances and theoretical analysis also shows that this reference rule for $R$ leads to the consistency of the sample analog of the TD.}

 {Note that the radius may also be chosen adaptively for each pair: we set the disk radius at $c_j$ to be $R_j$ for all knots and set the disk radius along the edge to be the linear interpolation of the radii at the two connected knots. The comparison between the uniform and adaptive $R$ is presented in Appendix \ref{sec::adaptiveTube}, and similar clustering performance is observed for the two approaches. Hence we use uniform $R$ by default for simplicity.} 
%For this work, $R$ is the same for all pairs of knots and is taken as the average variance within the Voronoi cells. The use of a disk with a fixed radius helps to make the estimated similarity measure between different regions comparable. 

Similar to the FD, we estimate the TD by a projected KDE. Let $\Pi_{j\ell}(x)$ be the projection of a point $x$ on the line through $c_j, c_\ell$.
%and $\XX_{j\ell} = \{x \in \XX: ||s - \Pi_{j\ell}(s)|| \leq R\}$.
We first estimate the $\pDisk$ via
\begin{align*}
%\begin{split}
    \widehat{\pDisk}_{j\ell,R}(t) &= \frac{1}{n h} \sum_{i = 1}^n K\bigg( \frac{\Pi_{j\ell}(X_i) - c_j - t (c_\ell - c_j)}{h} \bigg) I(||X_i - \Pi_{j\ell}(X_i)|| \leq R)  
%    \\&= \frac{1}{n h} \sum_{X_i \in \XX_{j\ell}} K\bigg( \frac{\Pi_{j\ell}(X_i) - c_j - t (c_\ell - c_j)}{h} \bigg) \\
%    \hat{S}_{j\ell}^{TD} &= \inf_{t \in [0,1]} \widehat{\pDisk}_{j\ell,R}(t) 
%\end{split}
\end{align*}
and then estimate the TD as
\begin{equation}
    \hat{S}_{j\ell}^{TD} = \inf_{t \in [0,1]} \widehat{\pDisk}_{j\ell,R}(t) .
    \label{eq::TD}
\end{equation}
where the infimum is approximated by grid search.

\begin{remark}
The estimations of the FD and the TD involve the use of the projected kernel density estimation, and we discuss the choices of kernel and the bandwidth selections for kernel density estimations in Appendix \ref{sec::bandwidth}.
By default, we use the Gaussian kernel with the normal scale bandwidth selector (NS) \citep{Chacon2011} for the best empirical results.

\end{remark}

\vspace{-1em}
\section{Asymptotic Theory of Edge Weight Estimation}	\label{sec::theory}

~~~~
% In our previous simulation results (Figure~\ref{fig::ex01}, \ref{sec::comp}), we saw that the skeleton clustering method based
% on the newly proposed density-aided similarity measures work well even when the dimension $d$ is large. 
In this section, we focus on the theoretical properties of the similarity measures to theoretically explain the effectiveness of the newly proposed density-aided similarity measures. 
We assume the set of knots $\calC = \{c_1, \dots, c_k\}$ is given and non-random  to simplify the analysis because (1) it is hard to quantify k-means uncertainty, and (2) with large $k$, it is extremely likely for k-means to stuck within the local minimum.
Note that this implies the corresponding Voronoi cells $\CC = \{\CC_1, \dots, \CC_k\}$ and the 2-NN regions $\{A_{j\ell}\}_{j,\ell = 1,\dots, k, j\neq \ell}$ (Equation \ref{eq::define2NN}) of all pairs of knots are fixed as well. 
We allow $k=k_n$ to grow with respect to the sample size $n$.
 {Theoretical results for Voronoi density are described in this section and theoretical properties for the Face density and Tube density are deferred to Appendix \ref{sec::FDtheory} and \ref{sec::TDtheory} respectively. In summary, the consistency of FD and TD are obtained based on the analysis of KDE with additional geometric considerations, resulting in rates similar to that of the 1-dimensional KDE under some regularity conditions. All proofs are included in Appendix \ref{sec::proof}.}

\subsection{Voronoi Density Consistency} \label{sec::VoronDensity}

~~~~We start with the convergence rate of the VD. 
We consider the following condition:
\begin{itemize}
\item[\textbf{(B1)}] There exists a constant $c_0$ such that the minimal knot size $\min_{(j,\ell)\in E} \mathbb{P}(A_{j\ell})\geq \frac{c_0}{k}$
and $\min_{(j,\ell)\in E}\|c_j-c_\ell\|\geq \frac{c_0}{k^{1/d}}$.
\end{itemize}
where $(j,\ell)\in E$ means that there is an edge between knots $c_j,c_\ell$ in the Delaunay Triangulation.
Condition (B1) is a condition requiring that 
no Voronoi cell $A_{j\ell}$ has a particularly small size and all edges have sufficient length.
This condition is mild because when the dimension of data $d$ is fixed, the total number of edges in the Delaunay triangulation of $k$ points scale at rate $O(k)$ \citep{CompGeometry}.
Because the volume shrinks at rate $O(k^{-1})$, the distance is expected to shrink at rate $O(k^{-1/d})$.

\begin{remark}
If we assume there exists constant $c_1, c_2 > 0$ that the density function $ f_\PP(x) > c_1 > 0$ for $\PP-a.e.$ and that $\min_{(j,\ell)} \abs{A_{j\ell}} \geq c_2\frac{\abs{\calX}}{k} $ where $\abs{\calX}$ is the volume of the support, then we have $\min_{(j,\ell)\in E} \mathbb{P}(A_{j\ell})\geq \frac{c_1 c_2}{k}$. So (B1), as implied by the above two common conditions, is a relatively weak assumption.
\end{remark}

%{\color{magenta}YC: I remove all the uniform bound;
%the formal derivation is standard but long. I don't think we need to include them here.}

\begin{thm}[Voronoi Density Convergence]
Assume (B1).
Then for any pair $j\neq\ell$ that shares an edge,
the similarity measure based on the Voronoi density satisfies
\begin{align}
\label{eq::VoronConv}
%    |\hat{S}_{j\ell}^{VD} - S_{j\ell}^{VD}|&= O_p\left(n^{-1/2}\right),\\
\left |\frac{\hat{S}_{j\ell}^{VD}}{S_{j\ell}^{VD}} -1 \right|&= O_p\left(\sqrt{\frac{k}{n}}\right),
%    \sup_{j,\ell}||\hat{S}_{j\ell}^{VD} - S_{j\ell}^{VD}|| 
%    \|\hat{S}^{VD} - S^{VD}\|_{\max} 
%\max_{(j,\ell)\in E}\left |\frac{\hat{S}_{j\ell}^{VD}}{S_{j\ell}^{VD}} -1 \right|
%    &= O_p\bigg( \sqrt{\frac{k\log k}{n}}\bigg)
\end{align}
\begin{align}
\label{eq::VoronUnifConv}
\max_{j,\ell}\left |\frac{\hat{S}_{j\ell}^{VD}}{S_{j\ell}^{VD}} -1 \right|&= O_p\left(\sqrt{\frac{k}{n}} \log k\right),
\end{align}
when $n\rightarrow\infty, k\rightarrow\infty, \frac{n}{k } \to \infty$.
\label{thm::VD}
\end{thm}

Theorem~\ref{thm::VD} provides the convergence rates of the sample-based Voronoi density to the population version of Voronoi density. 
This result is reasonable because when the knots $\calC$ are given,
the randomness in the sample-based Voronoi density 
is just the empirical proportion in each cell,
so it is a square-root-rate estimator based on the effective local sample size $n/k$.
%The $\log k$ term in the uniform bound is a natural outcome when we use the maximal deviation technique.
Consequentially, Theorem~\ref{thm::VD} suggests that  estimating the Voronoi density
is easy in the multivariate case when the knots are given--there is no dependency with respect to the ambient dimension. The extra $\log k$ factor in the uniform bound (Equation \ref{eq::VoronUnifConv}) comes from the Gaussian concentration bounds.

%\begin{remark}{\small
%Voronoi density does not involve any density estimation and hence the rates do not depend on the dimension of the ambient space.}
%\end{remark}
{
\subsection{Performance Guarantee for Voronoi Density}

% The adjusted Rand index is a correction of the Rand index that measures the similarity between two classifications of the same objects by the proportions of agreements between the two partitions. The correction is obtained by subtracting from the Rand index its expected value.
~~~~We provide below a performance guarantee for skeleton clustering with Voronoi density in terms of the adjusted Rand Index  \citep{Rand1971, Hubert1985}, which measures the agreements between two clusterings after adjusting for permutation chance.
To simplify the problem, we define the true clusters as the connected components of the skeleton graph with edges having true Voronoi density similarities $S_{j\ell}^{VD}$ over a known threshold $\tau > 0$. 
We show below that cutting the skeleton graph based on estimated edge similarities at the same threshold $\tau$ recovers the true clustering with a high probability.
Since the knots are fixed, the clustering error comes from partitioning knots into the wrong groups, so we will focus on the adjusted Rand Index of clustering the knots.
Let the true partition of the knots be $\calL^* = \{\calL_\ell^*\}_{\ell = 1, \dots, L}$, where $\calL_\ell^* $ contains all the knot indices belonging to the partition $\ell$.
Let the partition based on estimated edge similarities be $\hat \calL$. We assume that
\begin{itemize}
\item[\textbf{(P1)}] The true partition $\calL^*$ under the threshold $\tau$
remains the same when 
the thresholding level is within $(\tau(1 - \eps), \tau(1+ \eps) )$ for some $\eps > 0$.
\end{itemize}
This is a mild assumption because when we vary the threshold level $\tau$,
only a finite number of values will create a change in the partition. 
So (P1) holds under almost all values of $\tau$ except for a set of Lebesgue measure 0.
Let $ARI(\calL^*, \hat \calL)$ denotes the adjusted Rand Index of the estimated partition.

\begin{thm}[Adjusted Rand Index Guarantee]
Assume (B1) and (P1) and let $p_{min} = \min_{j,\ell} \PP(A_{j\ell})$, then
\begin{align}
    \PP\left\{ ARI(\calL^*, \hat \calL) < 1 \right\} \leq k(k-1)  \exp\left(-\frac{ \frac{1}{2} \eps^2p_{min} n}{ (1-p_{min})  + \frac{1}{3}\eps } \right) 
\end{align}
\label{thm::ARIVoron}
\end{thm}
\vspace{-1em}
Theorem~\ref{thm::ARIVoron} shows that
we have a good chance of recovering the ``true'' clusters defined by the 
actual Voronoi density. 
The above bound is derived from the uniform concentration bound of the Voronoi density.

%The above bound is derived from the uniform bounds of the concentration probability (so we have a factor of $k(k-1)$ in frou)  

%This result provides the concentration bound for the probability that the partition based on estimated Voronoi density similarities  differs from the partition based on the population quantity.

}

\vspace{-2em}
\section{Simulations}	\label{sec::simulation}

%In addition to the theoretical analysis of the  proposed skeleton clustering method,
%we conduct several Monte Carlo experiments
%to illustrate the effectiveness of the proposed method. 
~~~~
% The above theoretical analysis justifies the reliability of the density-aided weights proposed for skeleton clustering.
To study the effectiveness of skeleton clustering as a clustering method, we conduct several Monte Carlo experiments.
In this section, we present some empirical results to illustrate the performance of skeleton clustering in multivariate and high-dimensional settings (with additional data examples in Appendix \ref{sec::simDataAdd}).
Generally, our framework with the Voronoi density similarity measure is superior among all the compared clustering methods. 
 {In Appendix \ref{sec::simLinkage}, we use a systematic set of simulation studies to discuss the choice of linkage criteria within our clustering framework when dealing with different datasets and at the same time to demonstrate the robustness of the proposed framework to noisy data points and overlapping clusters.}
We include some additional simulations to support some choices within our framework in Appendix \ref{sec::additional}. 
The R implementation of the skeleton clustering methods along with some simulations can be founded at \url{https://github.com/JerryBubble/skeletonMethods}.

\subsection{High-dimensional Setting}	\label{sec::HD}
~~~~In this section, we demonstrate the performance of 
skeleton clustering on simulated datasets: the Yinyang data and the Mickey data. 
We also include a simulated dataset consisting of manifold structures of different dimensions, called the Manifold Mixture data, in Appendix \ref{sim::ManifoldMixture} and an additional simulation called the Ring data in Appendix \ref{sec::ring}.
%Additional simulation results are given in Appendix \ref{}
%xxxxx
%Yinyang data is composed of components of different un-spherical shapes, Mickey data has components that are unbalanced in sizes, and Manifold Mixture data have structures of different dimensions. Such datasets serve to demonstrate the advantages of our skeleton clustering framework in such challenging settings. See Appendix C.5 for additional example.
For the simulations within Section \ref{sec::HD} and Appendix \ref{sec::simDataAdd}, when using the skeleton clustering methods, the number of knots is set to be $k=[\sqrt{n}]$ and the knots are chosen by $k$-means with $1000$ random initialization. We select smoothing bandwidth by the normal scale bandwidth selector for the FD and TD, and the radius of TD is set to be the same for all edges with the value chosen as described in Section \ref{sec::TD}. We use single linkage hierarchical clustering when merging knots into final clusters  { with the true number of final clusters $S$ being provided.}

To highlight the importance of density-aided similarity measures, we include a similarity measure called the average distance (AD) for comparison. AD measures the similarity between $c_j$  and $c_\ell$ using the inverse of the average Euclidean distances between all pairs of observations in the two corresponding Voronoi cells.
All simulations are repeated 100 times to obtain the distribution of the empirical performances. 

\subsubsection{Yinyang Data}	\label{sec::YY}

~~~~The Yinyang dataset is an intrinsically $2$-dimensional data containing $5$ components: a big outer circle with $2000$ uniformly distributed data points, two inner semi-circles each with $200$ data points generated as 2D Gaussian with standard deviation $0.1$, and two clumps each with $200$ data points  {(generated with the \texttt{shapes.two.moon} function with default parameters in the \texttt{clusterSim} library in R \citep{clusterSim})}. The total sample size is $n=3200$ and, according to our reference rule, we choose $k=[\sqrt{3200}] = 57$ knots for the skeleton clustering procedure.
To make the data high-dimensional, we include additional variables from a Gaussian distribution with mean $0$ and standard deviation $0.1$, and we  increase the dimension of noise variables so that the total dimensions are $d=10, 100, 500, 1000$. We present results with larger standard deviations for the noisy variable in Appendix \ref{sec::highSigma}.

We empirically compare with the following clustering approaches: direct single-linkage hierarchical clustering (SL), direct $k$-means clustering (KM), spectral clustering (SC), and the merging K-means with hierarchical clustering method proposed by \cite{peterson2018merging}(KmH). 
We include the comparison with merging model-based clusters approaches in Appendix \ref{sec::mcomb}.

For skeleton clustering, we present the results with average distance density (AD), Voronoi density (Voron), Face density (Face), and Tube density (Tube).
%The number of knots $k=\sqrt{n}$ is chosen by the reference rule
%and we use single linkage clustering when merging knots into final clusters.
%We select smoothing bandwidth by Silverman's rule of thumb in FD and TD.
Since this is simulated data, we know that there are exactly $5$ clusters and we know which cluster an observation belongs to. 
 {The true number of clusters is provided to all the clustering algorithms.}
We use the adjusted Rand Index to measure the performance of each clustering method. 

%The studied methods are direct single-linkage hierarchical clustering (SL), spectral clustering (SC), skeleton clustering with average distance density (AD), skeleton clustering with Voronoi density (Voron), skeleton clustering with Face density (Face), and skeleton clustering with Tube density (Tube).

%The result 

%The first two dimensions of this simulated data depict the Yinyang structure, and higher dimensions are filled with random Gaussian noises with mean $0$ and standard deviation $0.1$. There are $n=3200$ sample points in the data. The chosen knots and the resulting dendrogram from single-linkage hierarchical clustering on dimension $1000$ data is depicted in Figure \ref{fig::yingyang1}.

\begin{figure}
\centering
\includegraphics[width=4.5cm]{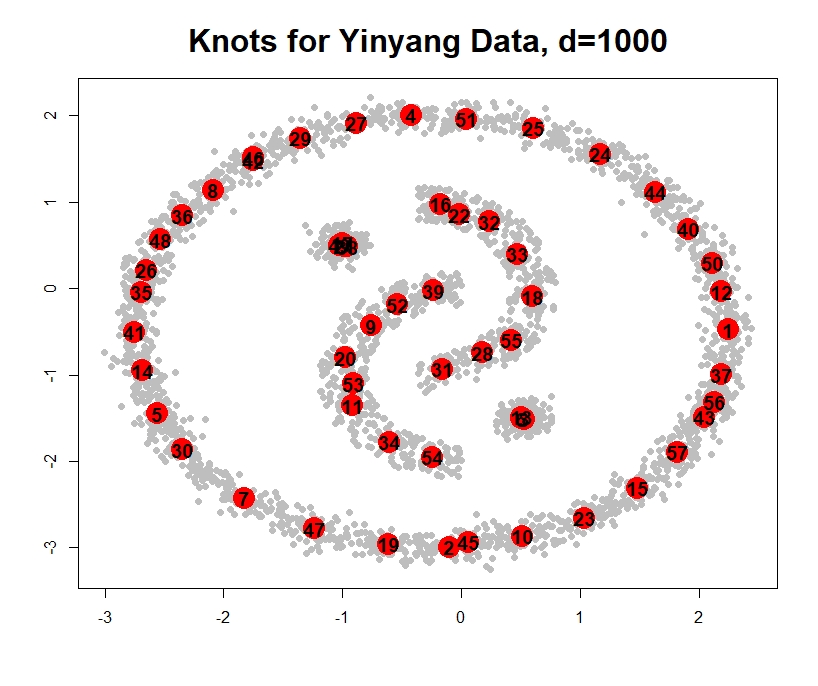}
\includegraphics[width=4.5cm]{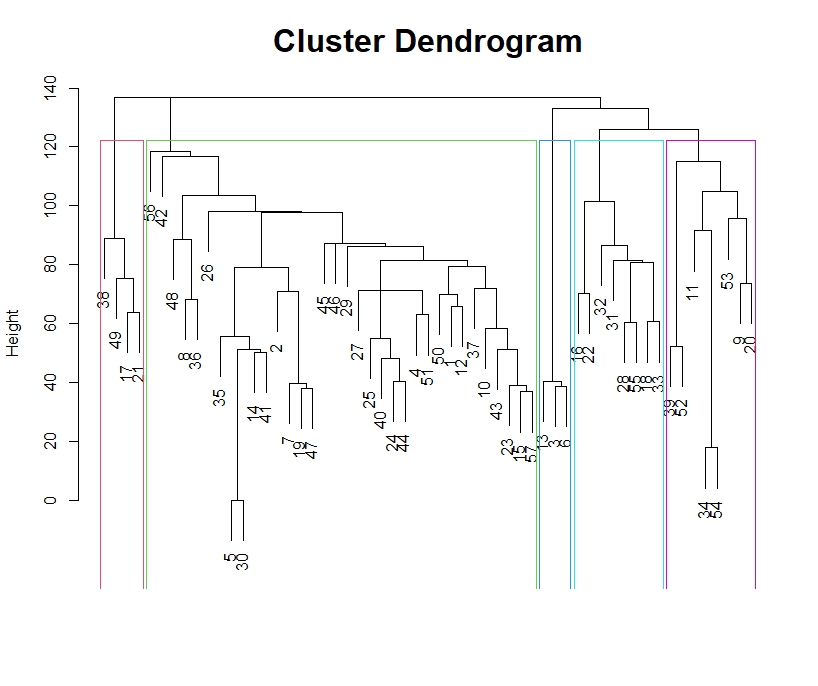}
\caption{Knots chosen by $k$-means on Yinyang data and the Dendrogram for single linkage hierarchical clustering with similarity measured by Voronoi density.}
\label{fig::yingyang1}
\end{figure}
%We compare the performances of different clustering methods on Yinyang data with dimension $10$, $100$, $500$, and $1000$ in Figure \ref{fig::yingyang2}. The studied methods are direct single-linkage hierarchical clustering (SL), spectral clustering (SC), skeleton clustering with average distance density (AD), skeleton clustering with Voronoi density (Voron), skeleton clustering with Face density (Face), and skeleton clustering with Tube density (Tube).

\begin{figure}
\centering
\includegraphics[width=3.5cm]{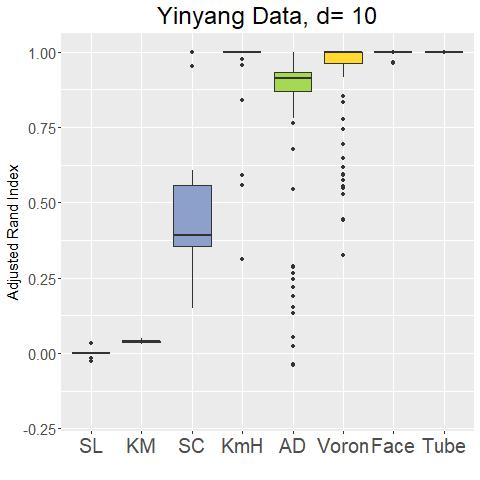}
\includegraphics[width=3.5cm]{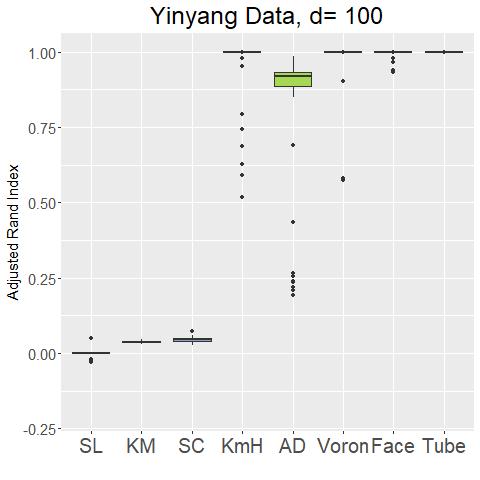}
\includegraphics[width=3.5cm]{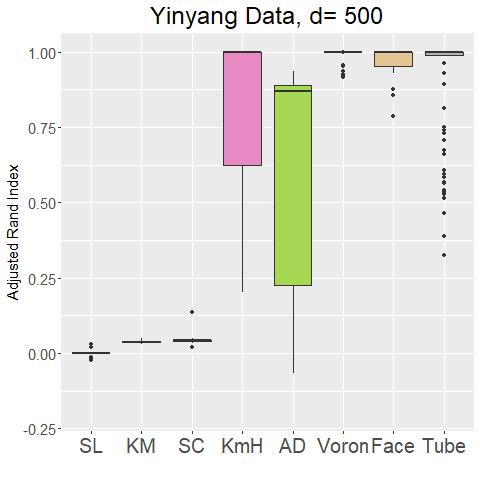}
\includegraphics[width=3.5cm]{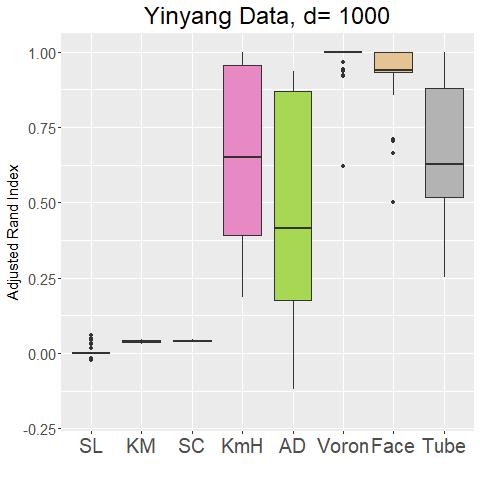}
\caption{Comparison of the final clustering performance in terms of adjusted Rand Index with different clustering methods on Yinyang Data with dimensions $10$, $100$, $500$, and $1000$.}
\label{fig::yingyang2}
\end{figure}

The results are given in Figure~\ref{fig::yingyang2}.
We observe that when dimension increases, traditional methods (SL, KM, SC) fail to give good clustering results while skeleton clustering can generate nearly perfect clustering. 
The KmH approach has better performance than the skeleton clustering using average distance, but skeleton clustering with other proposed density-aided similarity measures has better clustering results.
This illustrates the effectiveness of using the skeleton clustering framework and  highlights the advantage of using the proposed density-aided weights in clustering large-dimensional data.
Across all the data dimensions, the Voronoi density, the simplest measure among the three proposed similarity measures, gives the best performance in the skeleton clustering framework. 
% Average distance density becomes problematic in high-dimensional settings but still gives better performance compared to the classical methods. 

\subsubsection{Mickey Data}	\label{sec::mickey}

\begin{figure}[ht]
\centering
\captionsetup{skip=-0.5em}
\includegraphics[height=4cm]{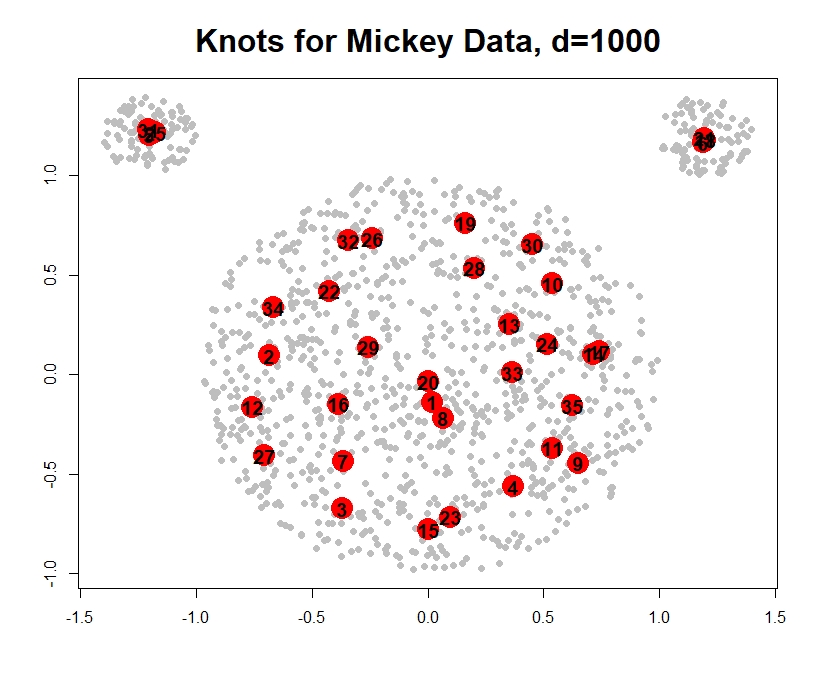}
\includegraphics[height=4cm]{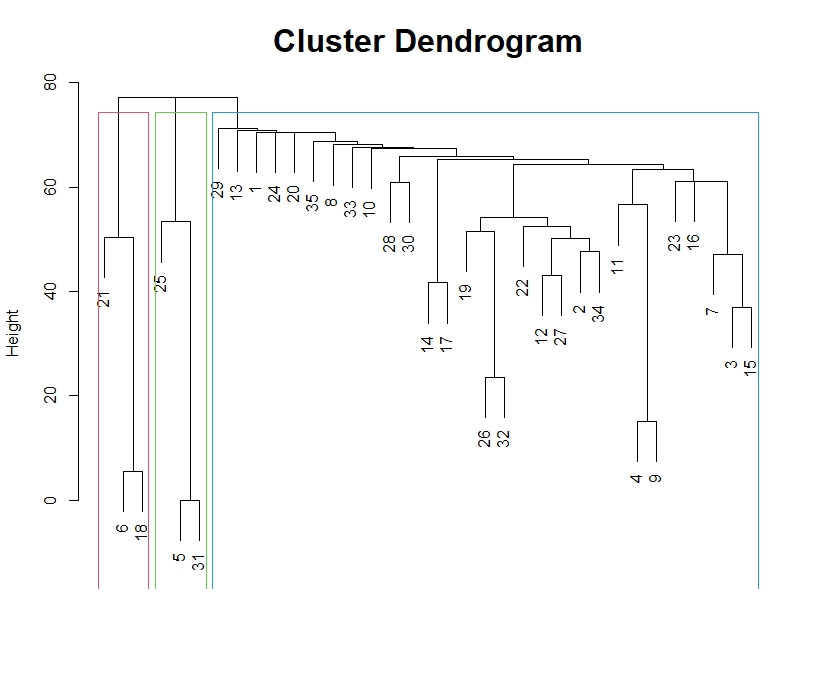}
\caption{An illustration of the analysis of the Mickey data with dimension $100$. 
}
\label{fig::mickey1}
\end{figure}

~~~~The simulated Mickey data is an intrinsically $2$-dimensional data consisting of one large circular region with $1000$ data points and two small circular regions each with $100$ data points. As a result, the structures have unbalanced sizes. The total sample size is $n = 1200$ and we choose the number of knots to be $k= [\sqrt{1200}] = 35$. We include additional variables with random Gaussian noises to make it a high dimensional data ($d=10,100,500,1000$) the same way as in Section \ref{sec::YY}.
The left panel of Figure~\ref{fig::mickey1} shows the scatter plot of the first two dimensions.

%From the Minimum Spanning Tree total length comparison plot in Figure \ref{fig::mickey1} (middle), we see that knots chosen by overfitting K-means do have the noise-reduction effect. 

\begin{figure}[ht]
\centering
\includegraphics[width=3.5cm]{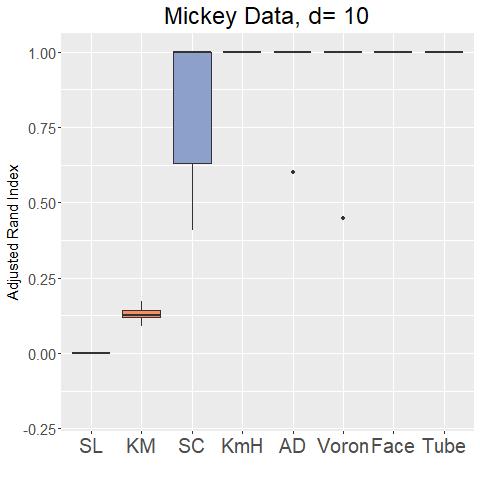}
\includegraphics[width=3.5cm]{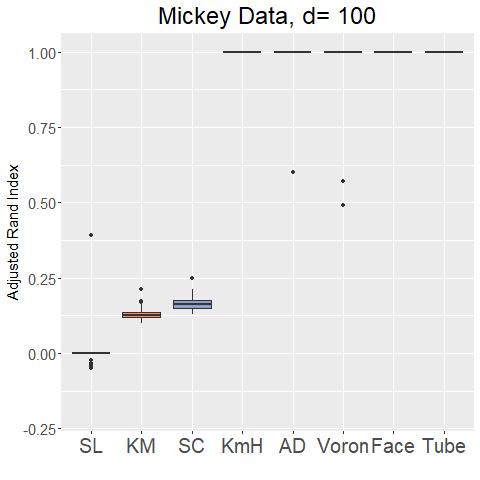}
\includegraphics[width=3.5cm]{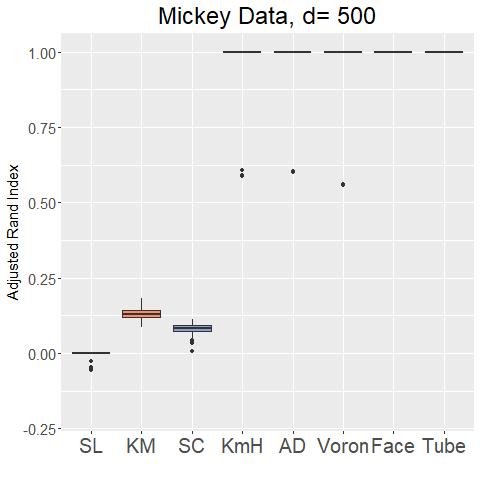}
\includegraphics[width=3.5cm]{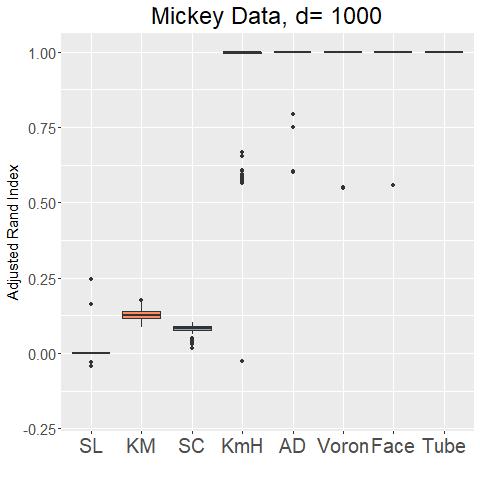}
\caption{Comparison of adjusted Rand index using different similarity measures on Mickey data with dimensions $10$, $100$, $500$, $1000$.}
\label{fig::mickey2}
\end{figure}

We perform the same comparisons as done on the Yinyang data with the  {true number of components $S=3$ provided to all the clustering algorithms,} and the results are displayed in Figure~\ref{fig::mickey2}. All methods perform well when $d$ is small but starting at $d=100$, traditional methods fail to recover the underlying clusters.
On the other hand, all methods in the skeleton clustering framework and the KmH approach work well even when $d=1000$.

%The Rand Indexes of the clustering results generated by different methods are presented in Figure \ref{fig::mickey2}. All the skeleton clustering methods have perfect performance regardless of the type of bone weight used. Spectral clustering has poor performance here because the clusters have unbalanced sizes, and spectral clustering tends to divide up the large circle.

\section{Real Data}	\label{sec::real}
~~~~In this section, we apply skeleton clustering to one real data example: the graft-versus-host disease (GvHD) data \citep{Brinkman2007}. 
 {Additionally, we analyze the Zipcode data \citep{Wernergsl} in Appendix \ref{sec::zip} and the Olive Oil data \citep{TSIMIDOU1987227} in Appendix \ref{sec::olive}.}
% \subsection{GvHD Data}
\label{sec::GvHD}
% ~~~~We adopt the famous flow cytometry data from \citep{Brinkman2007} about the graft-versus-host disease (GvHD). 
%Here we apply skeleton clustering framework to flow cytometry data from \citep{Brinkman2007}. 

GvHD is a significant problem in the field of allogeneic blood and marrow transplantation which occurs when allogeneic hematopoietic stem cell transplant recipients when donor-immune cells in the graft attack the tissues of the recipient.
The data include samples from a patient with GvHD containing $n_1= 9083$ observations and samples from a control patient with $n_2= 6809$ observations. 
Both samples include four biomarker variables, CD4, CD8$\beta$, CD3, and CD8. Previous studies \citep{Lo2008, Baudry2010} have identified the presence of high values in CD3, CD4, CD8$\beta$ cell sub-populations as a significant characteristic in the GvHD positive sample  {and a major objective of our analysis is to rediscovery this region with the proposed skeleton clustering methods}. 
% Thus, we focus on these three variables (CD3, CD4, CD8$\beta$) for plotting but we still use all four biomarker variables in our clustering analysis.
% We apply the proposed skeleton clustering to analyze this dataset. 
In addition, our skeleton clustering procedure shows more information and leads to a novel two-sample test.
%and show that our clustering
%method may lead to a new two-sample test procedure.

%framework with Voronoi density measure for edge weights to the GvHD data.

\begin{figure}
\centering
\includegraphics[height=5cm]{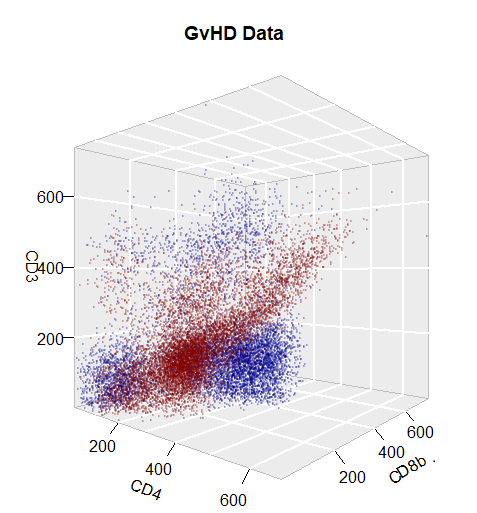}
\includegraphics[height=5cm]{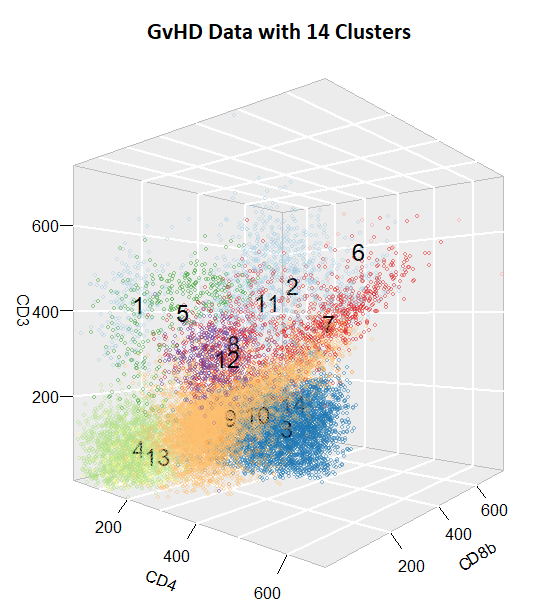}
\caption{ {\bf Left:} 3D scatterplot of the positive sample (red) and the control sample (blue). {\bf Right:} Final clustering result of combined GvHD data.}
\label{fig::GvHD1}
\end{figure}

The two samples are plotted in the left panel of Figure \ref{fig::GvHD1} focusing on the three key variables (CD3, CD4, CD8$\beta$) with blue points from the control sample and the red points from the GvHD positive sample. 
%The three axis are the first three biomarkers of interest. s
We observe that, in addition to the high CD3, CD4, CD8$\beta$ region, the distribution of the positive sample is different from the control sample also in some region with medium to the low CD3, CD4, and CD8$\beta$. Later we will demonstrate that our clustering framework can identify all such differences in distributions. 

To apply the skeleton clustering for a fair comparison of the two samples, we first construct knots from each sample separately.
Specifically, we apply the $k$-means method to find $k_1 = [\sqrt{n_1}]$ knots for the positive sample and find $k_2 = [\sqrt{n_2}]$ knots for the control sample. This ensures that both samples are well-represented by knots. 
We then combine the two samples into one dataset and combine the two sets of knots into one set with $k_1+ k_2$ knots. 
We create edges among the combined knots and apply the Voronoi density (VD) to measure the edge weights. 
To segment the knots, we use the average linkage criterion because there is no clear gap between major components and the analysis in Appendix~\ref{sec::simLinkage} suggests average linkage for this scenario. 
The skeleton clustering result is displayed in the right panel of Figure~\ref{fig::GvHD1}
with the number of final clusters chosen to be $S=14$. This choice follow \cite{Baudry2010} where the authors suggest to fit $9$ components on the positive sample and $3$ to $5$ components on the control sample. We choose $S = 9+5=14$ on the combined data to give a reasonable representation of the structures in the data, and, by empirical exploration, this choice leads to a good clustering result.

For further insights, we examined the weighted proportion of positive observations in each cluster.
A proportionally smaller weight is assigned to each positive observation to accommodate the fact that there are more positive observations ($n_1=9083 > n_2 = 6809$). 
After such normalization, a weighted proportion of $0.5$ means that the positive and control observations are balanced in one region.
A summary of the weighted proportion of clusters is presented in Table \ref{tab::GvHD}. 
We note that clusters 7,9,12, and 13 are majorly composed of positive observations (proportion $>0.75$), and clusters 3 and 6 are majorly composed of observations from the control sample (proportion $<0.25$). 
We also include the p-value for testing if the proportions equal $0.5$.
%, adjusted for False Discovery Rate \citep{fdr}.
Admittedly, because we use the data to  find clusters and use the same data to do the test, the p-values in Table \ref{tab::GvHD} may tend to be small.

%%%Comment out the old table

% \begin{table}
% \footnotesize
% \centering
% \setlength{\tabcolsep}{2pt}
% \begin{tabular}{||c c c c c c c c c c c c c c c ||} 
%  \hline
%  Cluster & 1 & 2 & 3 & 4 & 5 & 6& 7 & 8 & 9 & 10 & 11 & 12 & 13 & 14 \\ 
%  \hline\hline
%  Size & 202 & 948 & 3881& 1859 & 338 &  17 & 812 & 468 &6191 & 251  & 37 & 478 & 402  &  8  \\ 
%  Prop & .458 & .343& .008 & .296 & .341 & .000 & .934 &
%   .690 & .888& .673&  .669 & .794 & .841 & .310\\
%   p-value &.32&  $10^{-19}$& 0 & 8$\times10^{-63}$ & 6$\times10^{-8}$ & $10^{-4}$&
%  3$\times 10^{-102}$ & 3$\times 10^{-13}$ & 0 & 1$\times 10^{-6}$ & .11 &2$\times 10^{-29}$&
% 8 $\times10^{-33}$ & .52\\
%  \hline
% \end{tabular}
% \caption{Table of the sizes of the clusters and the weighted proportion of positive observations within each cluster.
% A proportion $0.5$ indicates that the two sample has equal proportion in the region.
% The $p$-value is the simple proportional test to examine if the two sample has equal proportion in that cluster, adjusted for False Discovery Rate \citep{fdr}.
% {\color{magenta}YC: please change the sign e-x to be $10^{-x}$.}}
% \label{tab::GvHD}
% \end{table}

\begin{table}
\footnotesize
\centering
\setlength{\tabcolsep}{2pt}
\begin{tabular}{||c c c c c c c c ||} 
 \hline
 Cluster & 1 & 2 & 3 & 4 & 5 & 6& 7 \\ 
 \hline
 \hline
 Size & 202 & 948 & 3881& 1859 & 338 &  17 & 812  \\ 
 Prop & .458 & .343& .008 & .296 & .341 & .000 & .934\\
  p-value &.30&  $7\times10^{-20}$& 0 & 3$\times10^{-63}$ & 4$\times10^{-8}$ & $1\times 10^{-4}$&
 6$\times 10^{-103}$ \\
 \hline
  \hline
 Cluster & 8 & 9 & 10 & 11 & 12 & 13 & 14 \\ 
 \hline\hline
 Size  & 468 &6191 & 251  & 37 & 478 & 402  &  8  \\ 
 Prop & .690 & .888& .673&  .669 & .794 & .841 & .310\\
  p-value & 2$\times 10^{-13}$ & 0 & 1$\times 10^{-6}$ & .09 &6$\times 10^{-30}$& 3 $\times10^{-33}$ & .52\\
 \hline
\end{tabular}
\caption{Table of the sizes of the clusters and the weighted proportion of positive observations within each cluster.
A proportion of $0.5$ indicates that the two samples have equal proportions in the region.
The $p$-value is the simple proportional test to examine if the two samples have equal proportions in that cluster.}
%, adjusted for False Discovery Rate.}
\label{tab::GvHD}
\end{table}

\begin{figure}[ht]
\captionsetup{skip=-0.5pt}
\centering
\includegraphics[width= 5cm]{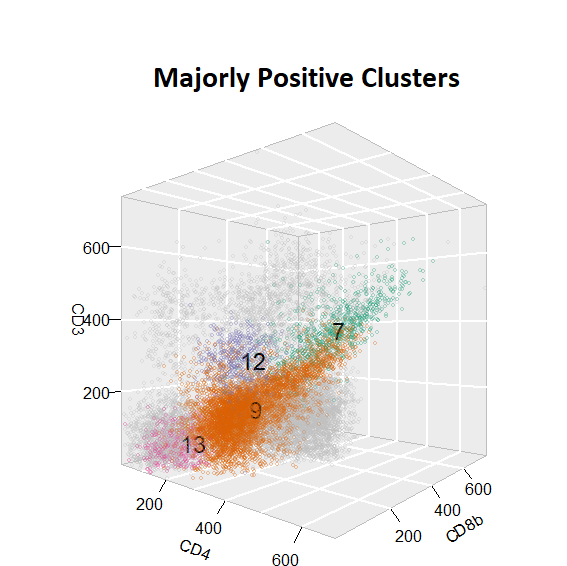}
\includegraphics[width= 5cm]{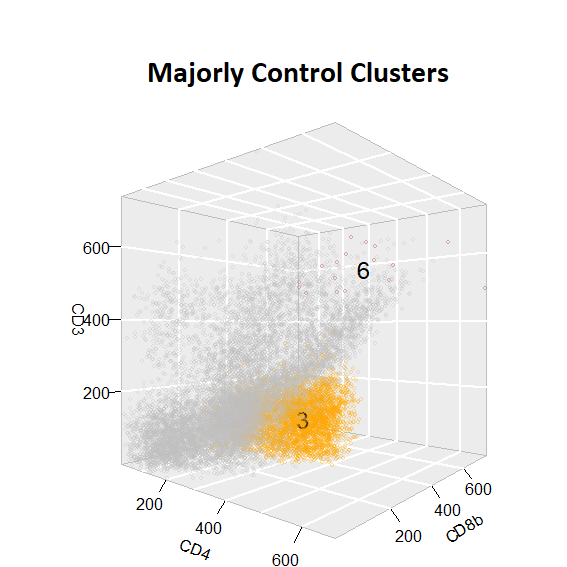}
\caption{Clusters with majorly positive observations and majorly control observations}
\label{fig::GcHD4}
\end{figure}

%Now we trace back the locations of the positive-dominant clusters and control-dominant clusters 
Clusters with majorly positive observations and clusters with majorly control observations are depicted in the two panels in Figure \ref{fig::GcHD4}.  {
Cluster 7 corresponds to the high CD3, CD4, CD8$\beta$ region identified by previous works with nearly all data points belonging to the positive patient. 
Cluster 6 is also scattered in the high CD3, CD4, CD8$\beta$ region but has all the observations coming from the control sample.
However, the small size (only $17$ data points) of Cluster 6 makes it unclear if it is a real structure or due to pure randomness. Overall our method succeeds in identifying the CD3+ CD4+ CD8$\beta$+ area for the GvHD-positive patient like the previous model-based clustering approaches.
Note that the data we are using are two individuals from the original 31 individuals in the GvHD study, which does not account for the inter-individual variability.}

Our clustering approach has some additional findings. Clusters 9, 12, and 13 also have high proportions of positive samples. 
These clusters are in the mid to low CD3, CD4, CD8$\beta$ region.
% , which is different from what was identified in \cite{ Lo2008, Baudry2010}.
%Such clusters can also potentially serve as an indication of GvHD. 
For the control case, in addition to the small Cluster 6, Cluster 3 is a large cluster with nearly all the observations from the control sample. It is located in the high CD8$\beta$ but low CD3 and CD4 region.

 {Model-based clustering approaches \cite{ Lo2008, Baudry2010} have an advantage for managing this cytometry data as they can parametrically describe the behaviors of data samples in different regions. The overlapping between different structures and the overall $4$-dimensional feature space is also applicable with model-based clustering methods. However, the proposed skeleton clustering approach can result in a graphical representation of each cluster that can be visualized for intuitive understanding. We include the skeleton graphs of the GvHD data clusters from the proposed clustering approach in Appendix \ref{sec::GvHDgraph}. Moreover, model-based approaches can still be limited to some regular shapes of the clusters in the ambient space, while applying the proposed clustering method helps identify clusters with complex structures. Cluster 9, for instance, shows a hammer-like structure based on the skeleton representation (see Figure \ref{fig::GvHDgraph}).}

%but with all the $17$ points belonging to the control sample. Therefore patients without GvHD can still have sample points in the identified ``risky'' region, but with lower likelihood. But, based on all the biomarkers in the data, our algorithm can differentiate such extreme points from the control sample from the sample points from positive patient in this region. Moreover, out method run faster than that proposed in \citep{Baudry2010} without estimating the densities, and can generalize to high dimensional without suffering from the curse of dimensionality.

Our results suggest a potential procedure for diagnosing GvHD.  Biomarkers from a new patient can be divided into clusters with respect to the learned segmentation, and doctors can mainly focus on the sample points that fall into regions 3, 7, 9, 12, and 13. If the patient has many points in Clusters 7, 9, 12, and 13, the patient likely has GvHD. 
Note that our current result is only based on two individuals and, with a descriptive purpose, is not accounting for the variability between different individuals and different cases. To use it for practical diagnosis, a more comprehensive analysis based on a larger and more representative sample is required.
%However, with data from more patients, the skeleton clustering framework can be applied to the larger data and learn more accurate clustering structure.

\vspace{-2em}
\section{Conclusion}	\label{sec::conclusion}
~~~~In this work, we introduce the skeleton clustering framework that can handle multivariate and even high-dimensional clustering problems with complex, manifold-based cluster shapes.
Our method adopts the density-based clustering idea to the high dimensional regime.
The key to bypassing the curse of dimensionality is the use
of density surrogates such as Voronoi density, Face density, and Tube density that are less sensitive to the dimension.
We use both theoretical and empirical analysis to
illustrate the effectiveness of the skeleton clustering procedure.
%We also include two extensions, the anomaly detection and combination of tSEN with skeleton clustering,
%in Appendix \ref{sec::extension}.
% high dimensional data in a computation efficient way. 
% We also proposed multiple novel similarity measures between clusters that take advantage of the local geometric structure. And by taking projection, some proposed similarity measures always reduce the estimation to be a $1$-dimensional problem regardless of the high-dimensional setting of the dataset. Such similarity measures can have general usage besides the Skeleton clustering framework. With simulation studies and real data analysis, we showed that skeleton clustering methods generate great clustering results for high-dimensional data. We also discussed possible ways to integrate uncertainty clustering, anomaly detection, and dimension reduction into the skeleton clustering framework. So far we only provided the consistency result for bone weight measures. 

In what follows, we discuss some possible future directions.
First, theoretical results accounting for the randomness of knots should be developed. The randomness of knots can affect the clustering performance because
the location of knots directly impacts the Voronoi cells, which changes the value of the similarity measures and consequently the cluster label assignments.
However, our $k$-means procedure is unlikely to stop at the global minimum, and it is unclear how to derive a theoretical statement based on local minima properly, so we leave this as future work. 
Additionally, the proposed skeleton clustering framework can also be potentially used for tasks such as detecting boundary points between clusters, anomaly and noise detection, and community detection in network data (Appendix \ref{sec::future}).
Overall, given the flexibility of the skeleton clustering framework, other possibilities by incorporating different methods for different steps in the framework can be explored.

% We analyzed the asymptotic behavior of k-means clustering when the number of clusters $k \to \infty$ and derived convergence rates for some proposed similarity measures.

\noindent{\large\bf Acknowledgments}\\
Yen-Chi Chen is supported by NSF DMS-195278, 2112907, 2141808 and NIH U24-AG072122. Zeyu Wei is supported by NSF NSF DMS - 2112907.

The authors report there are no competing interests to declare. 

% \newpage
\begingroup
\setstretch{0.9}
% \footnotesize
\bibliographystyle{abbrvnat}
\bibliography{Skeleton.bib}
\endgroup

\newpage

\appendix
\section*{Appendices}
\addcontentsline{toc}{section}{Appendices}
\renewcommand{\thesubsection}{\Alph{subsection}}

\subsection{Computational Complexity}	\label{sec::complexity}
%~~~~In this section we briefly describe the computational complexity of the skeleton clustering.
%As is known in the literature, the $k$-means clustering has a time complexity $O(ndk\ell)$, where $n$ is the sample size, $d$ is the dimension, $k$ is the total number of knots from the $k$-means, and $\ell$ is the total number of iterations. In our choice, $k=\sqrt{n}$ so the total cost is $O\left(n^{3/2}d\ell\right)$.

~~~~{\bf Knots construction.}
The first step of skeleton clustering is choosing knots, and, in this work, we take overfitting $k$-means as the default method. The $k$-means algorithm of Hartigan and Wong \citep{Hartigan1979} has time complexity $O(ndkI)$, where $n$ is the number of points, $d$ is the dimension of the data, $k$ is the number of clusters for $k$-means, and $I$ is the number of iterations needed for convergence. When using overfitting $k$-means to choose knots, the reference rule is $k = \sqrt{n}$, and hence the complexity is $O(n^{3/2}dI)$. This is a time-consuming step of our clustering framework, and the complexity increases linearly with $d$. Therefore, preprocessing the data with dimension reduction techniques or using subject knowledge to choose knots can be helpful to speed up this process.

{\bf Edges construction.}
For the edge construction step, we approximate the Delaunay Triangulation with $\hat{DT}(\calC)$ by looking at the 2-NN neighborhoods (the Voronoi Density regions in \ref{sec::VD} ). Hence the main computational task for our edge construction step is the 2-nearest knot search. We used the k-d tree algorithm for this purpose, which gives the worst-case complexity of $O(nd k^{(1-1/d)})$. 
%However, the k-d tree algorithm is not that useful for high-dimensional settings, but a brute-force search line search gives the not much worse complexity of $O(ndk)$. 
Notably, the computation complexity at this step is at the worst linear in $d$, which is a much better rate than computing the exact Delaunay Triangulation (exponential dependence on $d$), and our empirical studies have illustrated the effectiveness of such approximation.

{\bf Edge weight construction: VD.}
Next, we consider the computation complexity of the different edge weights measurements. For the VD, its numerator can be computed directly from the $2$-NN search when constructing the edges and hence no additional computation is needed. The denominators are pairwise distances between knots and can be computed with the worst-case complexity of $O(dk^2)$ because the number of nonzero edges is less than $\frac{k(k-1)}{2}$. With $k = \sqrt{n}$, we have the total time complexity of computing the VD to be $O(n d)$.

{\bf Edge weight construction: FD.}
For the Face density, we calculate the projected KDE at the middle point for each pair of neighboring Voronoi cells. The projection of one data point onto one central line can be done by matrix multiplication with complexity $O(d)$. Recall that we only use data points in local Voronoi cells for FD calculation, and the local sample size would be at $n_{loc} = O(\sqrt{n})$ under the conditions in Section \ref{sec::theory} and the reference rule $k = [\sqrt{n}]$. Together it takes $O(d\sqrt{n})$ to calculate the projected data for one edge.
With the projected data, KDE calculation has a time complexity $O(c \log c)$ where $c = \max_{j\neq \ell}\{n_j+n_\ell\} $ for any pair of knot indexes $j,\ell$. Again we have $c = O(n/k) = O(\sqrt{n})$ under the previously mentioned conditions. We need to do KDE for each edge in the skeleton, which gives the overall time complexity of FD weights to $O(k^2 d\sqrt{n}+  k^2 c \log c) = O(n^{3/2} d + n^{3/2} \log n)$.

{\bf Edge weight construction: TD.}
For Tube density, we similarly perform a projected KDE for each edge. Let $\eta$ be the maximum number of points in a tube region $\eta = \max_{j,\ell} |\{X_i: \|\Pi_{j\ell}(X_i)-X_i\|\leq R\}|$, the data projection again takes $O(\eta d)$ complexity. 
Suppose the minimum density is obtained by a grid search with $m$ grid points, the KDE step takes a total of $O(m \eta \log \eta  )$ for one edge.
To compute the whole edge weights matrix with $k = \sqrt{n}$, we have the complexity to be $O(n \eta d + n m \eta \log \eta ) $. Under conditions where the tube regions for TD estimations is also of size $\eta = O(n/k) = O(\sqrt{k})$, we have the overall complexity for VD weights calculation to be $O(k^2 d\sqrt{n}+  k^2 c \log c) = O(n^{3/2} d + m n^{3/2} \log n)$, which is larger than that for FD due to the grid search for minimum density.

{\bf Knots segmentation.}
In this work, we segment the learned weighted skeleton using hierarchical clustering. 
%When the linkage criterion satisfies the reducibility condition \citep{Lance1967, Gordon1987, Murtagh1983}, the time complexity will be $O(k^2) = O(n)$. Note that both the single linkage and the average linkage satisfy the reducibility condition.
With links that can be updated by  Lance-Williams update \citep{Lance1967} and satisfy the reducibility condition \citep{Gordon1987}, hierarchical clustering can be carried out with computation complexity $O(N^2)$, where $N$ is the number of points to start the algorithm with \citep{Murtagh1983}. For our empirical results, we favored single linkage and average linkage, and both satisfy the requirements for an efficient hierarchical clustering algorithm. We perform hierarchical clustering on the $k = \sqrt{n}$ knots, and hence the computation complexity for segmenting the skeleton structure is $O(k^2) = O(n)$.

%\subsubsection{Dip Statistic Edge Measure}
%In \citep{Hartigan1985} the authors proposed the dip statistic as ``the maximum difference between the empirical distribution function and the unimodal distribution function that minimizes that maximum difference''. Such dip statistic tests the unimodality of a distribution against any multimodal distribution. This statistic can also be adopted as a measure of dissimilarity between two knots in skeleton clustering framework. Particularly, similar to the face density and the Tube density, for two knots $c_j, c_\ell$, we project the points in the Voronoi cells $\CC_j, \CC_\ell$ (or the tube region as for Tube density) onto the line passing through the two knots and can get a one-dimensional empirical distribution function. The dip statistic of this empirical distribution function measures the dissimilarity between the two knots/Voronoi cells. However, empirical results are not good.

\subsection{Theory for Face Density}
\label{sec::FDtheory}
~~~~Here we derive the convergence rate of the Face Density estimator. 
%Let $p(x)$ be the density function of the data distribution, $g(x) = \nabla p(x)$ the gradient, and $H(x) = \nabla \nabla p(x)$ the Hessian matrix. 
Recall that $\mu_d$ is the Lebesgue measure on the $d$-dimensional Euclidean space and $F_{j\ell} = \CC_\ell\cap \CC_j$ is the face region between knots $c_j, c_\ell$. Let $\partial F_{j\ell}$ be the boundary of $F_{j\ell}$. We consider the following assumptions:
%We first make the following regularity assumptions:

%{\color{magenta}YC: the $\|\cdot\|_{\max}$ and $\|\cdot\|_{\infty}$ norms are different for matrix. I think $\max$ norm is easier to work with in our case. Also, I rewrite the assumptions since estimating FD requires less assumptions.
%	Besides, there is no need to use a multivariate kernel--the kernel is used in 1D case.}
\begin{itemize}
    \item[\textbf{(D1)}] (Density conditions) 
%    $\sup_{x \in \calX} p(x) < \infty$ and 
The PDF $p$ has compact support $\calX$, is bounded away from zero that $\inf_{x \in \calX} p(x) \geq p_{\min} > 0$, $\sup_{x \in \calX} p(x) \leq p_{\max} <\infty$, and is Lipschitz continuous. 

%    $\sup_{x \in \calX} \{ p(x), ||g(x)||_{\max}, ||H(x)||_{\max}\} < \infty$ and $\inf_{x \in \calX} p(x) \geq f_{\min} > 0$ and $p$ has compact support.
	\item[\textbf{(B2)}] (Bounded face region)  There exist constants $c_0, c_1$ such that the face area 
	$$
	\frac{c_0}{k^{1-\frac{1}{d}}}\leq \min_{(j,\ell)\in E} \mu_{d-1}(F_{j\ell}) \leq \max_{(j,\ell)\in E} \mu_{d-1}(F_{j\ell})\leq \frac{c_1}{k^{1-\frac{1}{d}}}
	$$
	\item[\textbf{(B3)}] (Boundary of face bounded) There exists a constant $c_2$ such that
	$$
	\max_{(j,\ell)\in E} \mu_{d-2}(\partial F_{j\ell})\leq \frac{c_2}{k^{1-\frac{2}{d}}},
	$$
	\item[\textbf{(B4)}] (Intersecting angle condition) There is an angle $\theta_0 < \pi$ such that, for every pair of intersecting face regions $F_{ij}$ and $F_{j\ell}$, the maximal principle angle between the two subspaces $\theta_{ij,j\ell}$ satisfies $\theta_{ij,j\ell} \leq \theta_0$
    \item[\textbf{(K1)}] (Kernel function conditions)  The kernel function $K$ is a positive and symmetric function  
    satisfying
%    such that $K(x) = K(||x||)$ has bounded
%    , continuous third derivatives and 
%    \begin{align*}
%       \int |x| K (x) dx < \infty, 
      $ \int K^2 (x) dx < \infty,\quad \int |x| K (x) dx<\infty, \quad \int x^2 K (x) dx<\infty.$
%       , \int x^2 K^{(\al)} (x) dx < \infty, \int (K^{(\al)}(x))^2 dx < \infty
%    \end{align*}
%    for all $\al = 0,1,2,3$. particularly, for $x \in \calX$,
%    $$\int x^2 K(x) dx = m_2(K) \bm I_d$$ 
%    for some real number $m_2(K)$ with $\bm I_d$ the identity matrix of order $d$.
\end{itemize}

Assumption (D1) is commonly assumed for the density estimation problem, but usually with higher-order smoothness conditions. Notably, for consistency of the FD estimator, we require only the Lipschitz condition since the bias of the sample estimator will be dominated by a geometric difference even if we have a higher-order smoothness (see the discussion after Theorem~\ref{thm::FD} and Appendix \ref{sec::proof} for more detail).
Condition (B2) restricts the shared boundary of two Voronoi cells to scale at the rate of $O(k^{1-\frac{1}{d}})$. While this condition may seem abstract, it is a mild condition. To illustrate this, suppose we have $k = m^d$ points that are on a uniform grid of $[0,1]^d$ for some integer $m$. We form the Voronoi cells of these grid points. 
The $(d-1)$-dimensional volume of the shared boundary of two neighboring Voronoi cells will scale at rate $O(k^{1-\frac{1}{d}})$ as $k\rightarrow\infty$.
(B3) requires the boundaries of the face regions to scale at most at a rate of $O(k^{1-\frac{2}{d}})$, and 
%(B4) requires the angles between the Voronoi cells to be bounded above. 
(B4) requires that we cannot have two nearby faces to be parallel to each other.
Assumptions (B3) and (B4) are needed when bounding the geometric difference between the estimator and the population quantity and are both mild conditions: 
When the knots form a spherical packing of a smooth region,
these conditions hold. 
%when the density is uniform, the centers from k-means will approaching sphere packing structure of 
%when knots are  on a uniform grid of $[0,1]^d$, 
%the conditions will be satisfied.
%Condition (B2) is a stronger version than (B1) in the sense that we require that the smallest volume
%of Voronoi cells is still at a proper order.
Notably, (D1) and (B2) imply (B1) and hence the consistency of FD requires more conditions than the consistency of VD.
The condition (K1) is a common assumption on the kernel function \citep{Wasserman2006, scott2015multivariate}
satisfied by many common kernel functions, including the Gaussian kernel.

%Note that here we require only Lipchitz condition since the bias will be at rate $O(h)$
%even 
%The  additional requirement on the Hessian matrix is to 

\begin{thm}[Face Density] Assume (D1), (K1), and (B2-B4). With $h \to 0$, $k \to \infty$, $h k^{1/d}\to 0$, $\frac{n h}{k^{1-\frac{1}{d}}} \to \infty$, then
for any pair $j\neq\ell$, we have
\begin{align}
\left |\frac{\hat{S}_{j\ell}^{FD}}{S_{j\ell}^{FD}} -1 \right|&=  O\big(hk^{1/d}\big) + O_p\bigg(\sqrt{\frac{k^{1-\frac{1}{d}}}{nh}}\bigg)
%\max_{(j,\ell)\in E}\left |\frac{\hat{S}_{j\ell}^{VD}}{S_{j\ell}^{VD}} -1 \right|&=  O\big(hk^{1/d}\big) + O_p\bigg(\sqrt{\frac{k^{1-\frac{1}{d}}\log k }{nh}}\bigg) 
\end{align}
\label{thm::FD}
\end{thm}
\vspace{-1em}
Theorem~\ref{thm::FD} shows the convergence rate
of estimating the FD. 
Roughly speaking, the rate is similar to a 1-dimensional density estimation problem. 
%and is affected by the ambient dimension only through the number of knots $k$. 
With $d\to \infty$, we have the rate to be $ O\big(h\big) + O_p\bigg(\sqrt{\frac{k}{nh}}\bigg) = O\big(h\big) + O_p\bigg(\sqrt{\frac{1}{n_{loc}h}}\bigg)$, where $n_{loc} = O\big(\frac{n}{k}\big)$ is the local effective sample size. 
Therefore, the effect of the ambient dimension is negligible when $d$ is large, and this is because we are estimating a `projected' density on the central line, which reduces to a 1-dimensional problem. 

Noticeably, the bias term in Theorem~\ref{thm::FD} is of the order $O(h)$.
While this rate is optimal under the Lipschitz smoothness (D1) for density estimation problem, it is slower than the conventional rate $O(h^2)$ when we have a bounded second-order derivative of $p$. 
One may be wondering if higher-order smoothness of $p$ is assumed, can we improve the convergence rate? 
Unfortunately, 
even if $p$ is very smooth, the bias rate will still stay the same at $O(h)$.
This is because there are two sources of bias.
The first one is the usual bias from kernel smoothing, which can be improved to higher order if we have high-order derivatives of $p$.
The other source of bias comes from the different geometric shapes of the Voronoi cells $\CC_j$ and $\CC_\ell$ (for illustration see Figure \ref{fig::faceproof1} in Appendix \ref{sec::proof}).
Consider the characterization of central line as $c_j + t(c_\ell-c_j)$ for $t\in[0,1]$, and the boundary will occur at $t=\frac{1}{2}$. 
Regions projected onto the central line will be different depending on the value of $t$.
Specifically, when $t>\frac{1}{2}$, the projected region is from $\CC_\ell$
whereas when $t<\frac{1}{2}$, the projected region is from $\CC_j$, and those projected regions can have shapes different from the face region.
This difference leads to an additional geometric bias of the order $O(h)$ and cannot be improved by higher-order smoothness of $p$.
 {In a sense, this bias $O(h)$ is similar to the boundary bias in that the density function is continuous but not differentiable. 
%In view of this, one may use a kernel function to correct for
However,
since the non-differentiability is caused by the geometric difference in two nearby Voronoi cells, it is unclear if we can use the conventional
boundary-correction kernels \citep{jones1993simple} to correct this bias. }

%However, points projected onto the central line with $t$ around $\frac{1}{2}$ will have different shapes in general, so the integrated density  will not change smoothly at $t=\frac{1}{2}$, leading to a bias of the order $O(h k^{1/d})$, which dominates and appears in Theorem \ref{thm::FD}. 

From Theorem~\ref{thm::FD},
one can see that the optimal bandwidth scales at rate 
$h \asymp \bigg(\frac{k^{1-3/d}}{2n}\bigg)^{1/3}$.
Recall that our reference rule sets $k = \sqrt{n}$ so that
$n_{loc} = \frac{n}{k} = \sqrt{n}$ is the average number of observations per each knot.
When  $d$ large, $\frac{3}{d}$ is negligible. 
Thus, the optimal bandwidth is given by $h \asymp \big(\frac{k}{n}\big)^{1/3} = n_{loc}^{-1/3}$.
While our empirical rule $ n_{loc}^{-1/5}$
is not optimal in this case,
it still gives a consistent estimator
and our empirical analysis
shows that
such choice leads to reliable clustering results;
see Appendix \ref{sec::bandrate}.

%where $n_{loc} = \frac{n}{k}$ is the average amount of observations per each knot.
%\begin{cor}[Optimal bandwidth for Face Density]
%Under the conditions from last theorem and for a given $k$, the optimal bandwidth to minimize Mean Square Error of the estimation is $h = \bigg(\frac{k^{1-3/d}}{2n}\bigg)^{1/3}
%    %\\
%    %&\left | \frac{\hat{S}_{j\ell}^{FD}}{S_{j\ell}^{FD} }  - 1  \right| = O\big(n^{\frac{2}{3d-6}}\big) 
%    $
%\label{cor::FDrate}
%\end{cor}
%
%
%For the empirical reference rule we choose $k = \sqrt{n}$, and under this choice, 
%for $d$ large so that $\frac{3}{d}$ is negligible, the optimal bandwidth is given by $h = O\big(\big(\frac{k}{n}\big)^{1/3}\big) =  O(n_{loc}^{-1/3})$. 
%On the other hand,
%when the dimension is  $d=3$, the optimal bandwidth is given by $h  \asymp n^{-1/3} = n_{loc}^{-2/3}$,
%which requires 
%a smaller bandwidth. 

%The NR rule $h\asymp n^{-1/5}$ is optimal when $d=15$.
%We empirically tested the performance of bandwidth $h$ ranging  from $O(n_{loc}^{-1/3})$ (when $d=\infty$) to $O(n_{loc}^{-1/6})$ (when $d=3$)
%in Appendix XXX. We found that the clustering performance is stable within this range of rates.
%Therefore, from both theoretical and empirical perspective our reference rule on bandwidth selection is appropriate.
%{\color{magenta}YC: please include graphs to justify this in appendix.}

%{\color{magenta}YC: I change the following paragraph a bit; this is universally true and we don't need
%the optimal case.}
One may notice that a small $k$ in Theorem~\ref{thm::FD}
leads to a better convergence rate, which suggests to use a small $k$.
While this is true from the perspective of estimation,
overall a small $k$ may lead to a poor representation of the data and result in a bad clustering performance.
Empirical results show that we need a sufficiently large number of knots to represent the data in order for the skeleton clustering to perform appropriately. 
Therefore, our reference rule with $k=\sqrt{n}$ is a suitable balance between the trade-off between representation and estimation.
We include an empirical analysis of the effect of $k$ on clustering performance in Appendix \ref{sec::KnotSize}.

\subsection{Theory for Tube Density}
\label{sec::TDtheory}
%We then show that the TD estimator $\widehat{\pDisk}(x,R, v) $ is consistent. 

~~~~In this section we derive the convergence rate of the Tube Density estimator. 
%We start with the setting where the radius $R$ is fixed, and the proof is similar to that for Face density with additional uncertainty from the location of the disk with infimum integrated density.
%Note that we need stronger smoothness conditions than the FD case but we may obtain a faster rate. 
We consider the following assumptions, which are slightly stronger than the corresponding ones in the case of the FD:

\begin{itemize}
    \item[\textbf{(D2)}] (Density conditions) 
    The PDF $p$ has a compact support and is $3$-H{\"o}lder and $\inf_{x \in \calX} p(x) \geq f_{\min} > 0$.
    \item[\textbf{(D3)}] (Disk Density conditions) 
    For any pair $c_j,c_\ell$, the minimum disk density location $t^* = {\sf argmin}_{t\in[0,1]}\pDisk_{j\ell, R}(t)\in(0,1)$
    is unique and the second derivative of the disk density $\pDisk^{(2)}_{j\ell, R}(t^*)\geq c_{\min}>0$.
    \item[\textbf{(K2)}] (Kernel function conditions)  The kernel function $K$ is a positive and symmetric function  
    satisfying
%    such that $K(x) = K(||x||)$ has bounded
%    , continuous third derivatives and 
%    \begin{align*}
%       \int |x| K (x) dx < \infty, 
%       \int K^2 (x) dx < \infty,\quad \int x^2 K (x) dx.
      $ \int x^2 K^{(\al)} (x) dx < \infty, \int (K^{(\al)}(x))^2 dx < \infty,$
%    \end{align*}
    for all $\al = 0,1,2$, where $K^{(\al)}$ denotes the $\al$-th order derivative of $K$.
%    particularly, for $x \in \calX$,
%    $$\int x^2 K(x) dx = m_2(K) \bm I_d$$ 
%    for some real number $m_2(K)$ with $\bm I_d$ the identity matrix of order $d$.
\end{itemize}

(D2) is a stronger version of (D1) that we require additional smoothness condition of $p$.
We need the $3$-H{\"o}lder class (slightly weaker than the requirement of third-order derivatives)
to obtain the rate of estimating the minimum \citep{Chacon2011,Chen2016}. 
%A similar condition appears in 
Also, a stronger condition (K2) on the kernel function is needed to ensure the gradient estimation is consistent.
Fortunately, common kernel functions such as the Gaussian kernel satisfy these conditions.

%{\color{magenta}YC: I think we don't need two theorems. I will only include one theorem
%and leave the other as a remark.}
%

\begin{thm}[Tube Density Consistency]  
%Let $\calC$ the given set of knots. 
Assume (D2), (D3), and (K2). Let $h \to 0$, $k \to \infty$, $R \to 0$, $n h^3\to \infty$,$n h R^{d-1} \to \infty$.
Suppose that for every pair $c_j, c_\ell$, $\inf_{t \in [0,1]} \pDisk_{j\ell,R}(t)$ and  $\inf_{t \in [0,1]} \widehat{\pDisk}_{j\ell,R}(t)$ do not occur at the boundary $t = 0,1$. Then for any pair $j\neq\ell$ that shares an edge, we have
\begin{align}
\pDisk_{j\ell,R}(t) &= O(R^{d-1}),\\
%    \max_{(j,\ell) \in E}
\bigg\vert 
    \frac{\hat{S}_{j\ell}^{TD}}{S_{j\ell}^{TD}}-1\bigg\vert &= O(h^2)+ O_p \bigg( \sqrt{\frac{1 }{nh R^{d-1}}} \bigg)  +O_p \bigg( \frac{1 }{nh^3} \bigg) 
\end{align}
\label{thm::TD1}
\end{thm}
\vspace{-1em}
%\begin{remark}
%The requirement for minimum $t$ to be in $(0,1)$ is saying in a sense that the density on the line segment between $c_j$ and $c_\ell$ is not changing monotonically.
%\end{remark}
%We take ratio between the estimator and the population Tube density because the population TD $\pDisk_{j\ell} = \pDisk_{j\ell,R} = O(R^{d-1})$  depends on $R$ as we are integrating a region that is changing with $R$. 
Theorem~\ref{thm::TD1} shows that the TD estimator converges to
the population TD with a rate consisting of three components. 
We allow $R\rightarrow0$ as $n\rightarrow \infty$ but this result also applies
to scenarios where $R$ is fixed. 
The first component $O(h^2)$ is the usual smoothing bias. 
The second component $O_p \bigg( \sqrt{\frac{1}{nhR^{d-1}}} \bigg)$ is similar to the stochastic variation part from usual KDE but with an additional dependence on $R^{d-1}$. This is due to the fact that, when $R\rightarrow0$, we are using fewer and fewer observations to perform smoothing, and $nR^{d-1}$ serves as the effective sample size. 
%The $\log k$ is again from the maximal deviation across all pairs. 
The third component $O_p \bigg(\frac{1}{nh^3}\bigg)$
is due to the error of estimating the location of the minimum. It is a squared term because the density behaves like a quadratic function around its minimum
due to (D3). 

While the convergence rate of TD requires stronger conditions (D2) and (K2) compared to the conditions (D1) and (K1) when estimating the FD,
the TD estimator has a smaller bias than the FD estimator (comparing Theorem \ref{thm::FD} and~\ref{thm::TD1}).
This is because the TD is evaluated on a ``regular shape'', which leads to a smoother quantity being estimated. 
%The second improvement is in the stochastic variation part that the $O_P$ term of estimating the TD does not involve any polynomial order of $k$ (in contrast to VD and FD). This is because when $R$ is fixed, the quantity $S^{TD}$ is a fixed quantity that does not shrink with respect to the growth of $k$.

For the stochastic variation part, the second term in Theorem~\ref{thm::TD1} gives $O_p \bigg( \sqrt{\frac{1 }{nh R^{d-1}}} \bigg)$ while the second term in Theorem~\ref{thm::FD} gives $ O_p\bigg(\sqrt{\frac{k^{1-\frac{1}{d}} }{nh}}\bigg) $. Note that empirically we choose $R$ to be the average of the root mean squared distances of each Voronoi cell (Section \ref{sec::TD}), which is of order $O(k^{-1/d})$ with cell sizes to have the same rates. 
Hence $k^{1-1/d}$ and $\frac{1}{R^{d-1}}$ are at the same rate and the stochastic variation part is comparable for TD and FD estimators. 
However, for TD we have another source of variation coming from the uncertainty of the location of minimum, which can cause TD to have a larger variation than the FD estimator.

Based on the above reasoning, our choice of $R$ leads to $\frac{1}{R^{d-1}}\asymp k^{1-1/d}$, which implies the rate
$O(h^2) + O_p\left(\sqrt{\frac{k^{1-1/d}}{nh}}\right) + O_p\left(\frac{1}{nh^3}\right)$.
Under our reference rule $k = \sqrt{n}$
the optimal bandwidth is $h\asymp n^{-\frac{1}{10}(1+\frac{1}{d})}$.
Recall that the local sample size is about $n_{ loc} = n/k = \sqrt{n}$ and hence the optimal bandwidth is $h\asymp n_{ loc}^{-\frac{1}{5}(1+\frac{1}{d})}$. When $d\rightarrow\infty$, this leads to $h\asymp n_{ loc}^{-1/5}$,
which is the same rate on sample size as given by the Silverman's rule of thumb.

 {
\begin{remark}
\small
Similar uniform bounds of the Face and Tube density can be derived with an extra $\log k$ factor in the rates through the concentration bound for kernel density estimator \citep{gine2002rates}. Also, similar concentration bounds on the Adjusted Rand Indexes can be achieved for partition based on the Face and Tube density.
\end{remark}}

\subsection{Proofs}	\label{sec::proof}
\subsubsection{Proofs for Voronoi Density Results}

We restate the assumption:
\begin{itemize}
\item[\textbf{(B1)}] There exists a constant $c_0$ such that the minimal knot size $\min_{(j,\ell)\in E} \mathbb{P}(A_{j\ell})\geq \frac{c_0}{k}$
and $\min_{(j,\ell)\in E}\|c_j-c_\ell\|\geq \frac{c_0}{k^{1/d}}$, where $A_{j\ell}$ is the 2-NN region of knots $c_j,c_\ell$ as defined in Equation \ref{eq::define2NN}.
\end{itemize}

%\begin{manualtheorem}{\ref{thm::VD}}[Voronoi Density]
%Assume (B1).
%Then for any pair $j\neq\ell$,
%the similarity measure based on the Voronoi density satisfies
%\begin{align*}
%\left |\frac{\hat{S}_{j\ell}^{VD}}{S_{j\ell}^{VD}} -1 \right|&= O_p\left(\sqrt{\frac{k}{n}}\right),\\
%\end{align*}
%when $n\rightarrow\infty, k\rightarrow\infty, \frac{n}{k \log k} \to \infty$.
%\label{thm::VD}
%\end{manualtheorem}

\begin{proof}[ of Theorem~\ref{thm::VD}]

For given knots $c_j, c_\ell$, the distance $||c_j - c_\ell||$ is also given. 
We denote the numerator of $S_{j\ell}^{VD}$ as
\begin{align*}
    p_{j\ell} = \PP(A_{j\ell}) &= \EE I(X_i: d(X_i, c_m) > max\{d(X_i, c_j), d(X_i, c_\ell), \forall m \neq j,l\} )
\end{align*}
and note that the numerator of $\hat{S}_{j\ell}^{VD}$ is 
\begin{align*}
   \hat{P}_n(A_{j\ell}) &= \frac{1}{n} \sum_{i=1}^{n} I(X_i: d(X_i, c_m) > max\{d(X_i, c_j), d(X_i, c_\ell), \forall m \neq j,l\} ),
\end{align*}
which
is a sum of binary variables and has variance $\sigma_{j\ell}^2 = \frac{p_{j\ell}(1-p_{j\ell})}{n}$. 
By Chebyshev's inequality,
\begin{align*}
%\begin{split}
%    &\PP\big(\big\vert\hat{P}_n(A_{j\ell}) - p_{j\ell}\big\vert \geq t \sig_{j\ell} \big) \leq \frac{1}{t^2}\\
    &\big\vert\hat{P}_n(A_{j\ell}) - p_{j\ell}\big\vert= O_p(\sig_{j\ell}^{1/2})= O_p\bigg(\bigg[\frac{p_{j\ell}(1-p_{j\ell})}{n}\bigg]^{1/2}\bigg)
%\end{split}
\end{align*}
Note that the region $A_{j\ell}$ is changing with respect to $k$.
%and hence we focus the ratio between the estimator and the population quantity. 
%Then have
The ratio is then
\begin{align*}
%\begin{split}
    \left |\frac{\hat{S}_{j\ell}^{VD}}{S_{j\ell}^{VD}} -1 \right| &= \left |\frac{\hat{P}_n(A_{j\ell}) }{\PP(A_{j\ell})} -1 \right| = \frac{1}{p_{j\ell} }O_p\bigg(\bigg[\frac{p_{j\ell}(1-p_{j\ell})}{n}\bigg]^{1/2}\bigg)\\
    &= O_p\bigg(\bigg[\frac{(1-p_{j\ell})}{np_{j\ell}}\bigg]^{1/2}\bigg)
    =O_p\bigg(\bigg[\frac{(1-c_0/k)}{n c_0/k}\bigg]^{1/2}\bigg) = O_p\bigg(\bigg(\frac{k}{n}\bigg)^{1/2}\bigg)
\label{eq::VDrate}
%\end{split}
\end{align*}
by assumption (B1) that $\min_{(j,\ell)\in E} \mathbb{P}(A_{j\ell})\geq \frac{c_0}{k}$,
which completes the proof for Equation \ref{eq::VoronConv}.
%Let $n_{j \ell} = n \PP(A_{j\ell})$ be the local effective sample size and by assumption (B1) we have that $n_{j \ell} = O_p(n/k)$, and the rate in (\ref{eq::VDrate}) is the common concentration rate based on the local effective sample size.

%The maximal ratio as in (\ref{eq::VDmaxrate}) follows
%from the fact that $\frac{\hat{S}_{j\ell}^{VD}}{S_{j\ell}^{VD}} -1$
%can be written as IID sum of sub-Gaussian random variables
%and there are $k(k-1)/2$ distinctive entries in the bone weight matrices.
%So the result follows from the usual Gaussian concentration method. 
%note that there are $k(k-1)/2$ distinctive entries in the bone weight matrices.
%Because 
%\begin{align*}
%P\left(\max_{(j,\ell)\in E}\left |\frac{\hat{S}_{j\ell}^{VD}}{S_{j\ell}^{VD}} -1 \right|>t\right)&
%\leq \sum_{(j,\ell)\in E}P\left(\left |\frac{\hat{S}_{j\ell}^{VD}}{S_{j\ell}^{VD}} -1 \right|>t\right)
%\end{align*}
%
%and hence
%\begin{align*}
%    \max_{(j,\ell)\in R}\left |\frac{\hat{S}_{j\ell}^{VD}}{S_{j\ell}^{VD}} -1 \right|
%    \leq \sum_{(j,\ell)\in R}\left |\frac{\hat{S}_{j\ell}^{VD}}{S_{j\ell}^{VD}} -1 \right|  = O_p\bigg( \sqrt{\frac{k\log k}{n}}\bigg)
%\end{align*}

 {To get the uniform bound, we first start with the concentration bound. Note that $\big(I(X_i \in A_{j\ell}) - p_{j\ell}\big)$ has zero mean and $\left|I(X_i \in A_{j\ell}) - p_{j\ell}\right|\leq 1$. Hence by Bernstein's inequalities, we have 
\begin{align*}
    \PP\left\{\left| \frac{\hat{P}_n(A_{j\ell}) }{p_{j\ell}}  -1 \right|> \eps \right\}
    &=\PP\left\{\big\vert\hat{P}_n(A_{j\ell}) - p_{j\ell}\big\vert > \eps p_{j\ell} \right\} \\
    &= \PP\left\{ \left|\frac{1}{n} \sum_{i=1}^n I(X_i \in A_{j\ell}) - p_{j\ell} \right| > \eps p_{j\ell} \right\} \\
    &= 2\PP\left\{ \sum_{i=1}^n \left(I(X_i \in A_{j\ell}) - p_{j\ell}\right) > n \eps p_{j\ell} \right\}\\
    &\leq  2\exp\left\{-\frac{ \frac{1}{2} \eps^2p_{j\ell}^2 n^2}{ \sum_{i=1}^n \EE\left[\left(I(X_i \in A_{j\ell}) - p_{j\ell}\right)^2 \right] + \frac{1}{3}\eps p_{j\ell} n} \right\}\\
    &=  2\exp\left\{-\frac{ \frac{1}{2} \eps^2p_{j\ell}^2 n^2}{ np_{j\ell}(1-p_{j\ell})  + \frac{1}{3}\eps p_{j\ell} n} \right\}\\
    &=2\exp\left\{-\frac{ \frac{1}{2} \eps^2p_{j\ell}^2 n}{ p_{j\ell}(1-p_{j\ell})  + \frac{1}{3}\eps p_{j\ell} } \right\}
\end{align*}
Note that plugging in the $p_{j\ell} = \Omega\left(\frac{1}{k}\right)$ rate to above concentration bound we can recover the $O_p\bigg(\sqrt{\frac{k}{n}}\bigg)$ rate in Equation \ref{eq::VoronConv}.
Then by union bound we have
\begin{align*}
\PP\left\{ \max_{(j,\ell) \in \calS} |\hat{S}_{j\ell} /S_{j\ell}  -1| >  \eps \right\}
    &\leq \PP\left\{ \max_{j,\ell} |\hat{S}_{j\ell} /S_{j\ell}  -1| > \eps \right\}\\
    &\leq  \sum_{j,\ell} \PP\left\{ |\hat{S}_{j\ell} /S_{j\ell}  -1|> \eps \right\}\\
    &\leq \frac{k(k-1)}{2} \max_{j,\ell}\PP\left\{\left| \frac{\hat{P}_n(A_{j\ell}) }{p_{j\ell}}  -1 \right|> \eps \right\}\\
    &\leq k(k-1) \max_{j,\ell} \left\{ \exp\left(-\frac{ \frac{1}{2} \eps^2p_{j\ell}^2 n}{ p_{j\ell}(1-p_{j\ell})  + \frac{1}{3}\eps p_{j\ell} } \right) \right\}\\
    &\leq  k(k-1)  \exp\left(-\frac{ \frac{1}{2} \eps^2p_{min} n}{ (1-p_{min})  + \frac{1}{3}\eps } \right) 
%\end{split}
\end{align*}
where $p_{min} = \min_{j\ell}p_{j\ell}$. Therefore we can derive the uniform error bound that 
\begin{align*}
\max_{j,\ell}\left |\frac{\hat{S}_{j\ell}^{VD}}{S_{j\ell}^{VD}} -1 \right|&= O_p\left(\sqrt{\frac{k}{n}} \log k\right),
\end{align*}
when $n\rightarrow\infty, k\rightarrow\infty, \frac{n}{k } \to \infty$.}

\end{proof}

For comprehensiveness, we provide the definition of the adjusted Rand Index below.
For two partitions $X = \set{X_1,\dots, X_r}$ and $Y = \set{Y_1,\dots, Y_s}$, let $n_{ij} = \left| X_i \cap Y_j \right|$, we have the contingency table
$$
\begin{array}{c |c c c c | c} 
& Y_1 & Y_2 & \cdots & Y_s & Sums\\
X_1 & n_{11} & n_{12} & \cdots & n_{1s} & a_1\\
X_2 & n_{21} & n_{22} & \cdots & n_{2s} & a_2\\
\vdots & \vdots& \vdots & \ddots & \vdots & \vdots\\
X_r & n_{r1} & n_{r2} & \cdots & n_{rs} & a_r\\
 \hline
Sums & b_1 &b_2 &\cdots & b_s&
\end{array}
$$
And the Adjusted Rand Index (ARI) adjusting for permutation chance is
\begin{align*}
    ARI = \frac{\sum_{ij} {n_{ij} \choose 2} - \left[ \sum_i {a_i \choose 2} \sum_j {b_j \choose 2} \right]\bigg/ {n \choose 2} }{ \frac{1}{2} \left[ \sum_i {a_i \choose 2} + \sum_j {b_j \choose 2} \right] - \left[ \sum_i {a_i \choose 2} \sum_j {b_j \choose 2} \right]\bigg/ {n \choose 2}}
\end{align*}

{
% The adjusted Rand index is a correction of the Rand index that measures the similarity between two classifications of the same objects by the proportions of agreements between the two partitions. The correction is obtained by subtracting from the Rand index its expected value.

% ~~~~We again take the skeleton graph $\calS$ as given with knots $\calC = \{c_1, \dots, c_k\}$ with the corresponding Voronoi cells $\CC = \{\CC_1, \dots, \CC_k\}$. The connected edges are also provided. We let $S_{ij}$ be the population similarity measure between knots $c_i$ and $c_j$ that are connected in the skeleton graph. 
% We define the ``true'' clusters as the connected components of the skeleton graph with edges having true similarity over a known threshold $\tau > 0$. We here show that cutting the skeleton graph based on estimated edge similarities at the same threshold $\tau$ retrieves the true clustering.

% More precisely, we suppose that the true clustering can be achieved by partition $(\tau(1 - \eps), \tau(1+ \eps) ) $ for small $\eps > 0$, and in practice this is always the case that cutting the edges within a range of thresholds can result in the same partitions. We denote the true partition as a collection of sets of knots that $\calL^* = \{\calL_\ell^*\}_{\ell = 1, \dots, L}$ and $\calL_\ell^* $ collects all the knot indices belonging to the partition $\ell$. Similarly we denote the partition based on estimated edge similarities by $\hat \calL$. Let $ARI(\calL^*, \hat \calL)$ denotes the Adjusted Rand Index of the estimated partition.
\begin{proof} of Theorem~\ref{thm::ARIVoron} (Performance guarantee for Voronoi density)
We note that, assuming (P1),
\begin{align*}
%\begin{split}
        \PP\left\{ARI(\calL^*, \hat \calL) < 1\right\} &\leq \PP\left\{ \text{there exists at least one wrongly cut edge}\right\} \\
    &= \PP\left\{ \max_{(j,\ell) \in \calS} |\hat{S}_{j\ell} /S_{j\ell}  -1| >  \eps \right\}\\
    &\leq k(k-1)  \exp\left(-\frac{ \frac{1}{2} \eps^2p_{min} n}{ (1-p_{min})  + \frac{1}{3}\eps } \right) 
%\end{split}
\end{align*}
\end{proof}

by the uniform bound derived above.
}

\subsubsection{Proofs for Face Density Consistency}
\label{sec::FDproof}
~~~~Let $p(x)$ be the density function of the data distribution, let $\mu_d$ be the Lebesgue measure on the $d$-dimensional Euclidean space, let $F_{j\ell} = \bar{\CC}_\ell\cap \bar{\CC}_j$ denote the face between knots $c_j, c_\ell$, and let $\partial F_{j\ell}$ be the boundary of $F_{j\ell}$. We consider the following assumptions:
%$g(x) = \nabla p(x)$ the gradient and $H(x) = \nabla \nabla p(x)$ the Hessian matrix. 
%We first make the following regularity assumptions:
%
%{\color{magenta}YC: the $\|\cdot\|_{\max}$ and $\|\cdot\|_{\infty}$ norms are different for matrix. I think $\max$ norm is easier to work with in our case. Also, I rewrite the assumptions since estimating FD requires less assumptions.
%	Besides, there is no need to use a multivariate kernel--the kernel is used in 1D case.}
Again, we recall the assumptions:

\begin{itemize}
    \item[\textbf{(D1)}] (Density conditions) 
%    $\sup_{x \in \calX} p(x) < \infty$ and 
The PDF $p$ has compact support $\calX$, is bounded away from zero that $\inf_{x \in \calX} p(x) \geq p_{\min} > 0$, $\sup_{x \in \calX} p(x) \leq p_{\max} <\infty$, and is Lipschitz continuous. 

%    $\sup_{x \in \calX} \{ p(x), ||g(x)||_{\max}, ||H(x)||_{\max}\} < \infty$ and $\inf_{x \in \calX} p(x) \geq f_{\min} > 0$ and $p$ has compact support.
	\item[\textbf{(B2)}] There exist constants $c_0, c_1$ such that the face area 
	$$
	\frac{c_0}{k^{1-\frac{1}{d}}}\leq \min_{(j,\ell)\in E} \mu_{d-1}(F_{j\ell}) \leq \max_{(j,\ell)\in E} \mu_{d-1}(F_{j\ell})\leq \frac{c_1}{k^{1-\frac{1}{d}}}
	$$
	\item[\textbf{(B3)}] There exists a constant $c_2$ such that
	$
	\max_{(j,\ell)\in E} \mu_{d-2}(\partial F_{j\ell})\leq \frac{c_2}{k^{1-\frac{2}{d}}},
	$
	\item[\textbf{(B4)}] There is an angle $\theta_0 < \pi$ such that, for every pair of intersecting face regions $F_{ij}$ and $F_{j\ell}$, the maximal principle angle between the two subspaces $\theta_{ij,j\ell}$ satisfies $\theta_{ij,j\ell} \leq \theta_0$
    \item[\textbf{(K1)}] (Kernel function conditions)  The kernel function $K$ is a positive and symmetric function  
    satisfying
%    such that $K(x) = K(||x||)$ has bounded
%    , continuous third derivatives and 
%    \begin{align*}
%       \int |x| K (x) dx < \infty, 
      $ \int K^2 (x) dx < \infty,\quad \int |x| K (x) dx<\infty, \quad \int x^2 K (x) dx<\infty.$
%       , \int x^2 K^{(\al)} (x) dx < \infty, \int (K^{(\al)}(x))^2 dx < \infty
%    \end{align*}
%    for all $\al = 0,1,2,3$. particularly, for $x \in \calX$,
%    $$\int x^2 K(x) dx = m_2(K) \bm I_d$$ 
%    for some real number $m_2(K)$ with $\bm I_d$ the identity matrix of order $d$.
\end{itemize}

%\begin{manualtheorem}{\ref{thm::FD}}[Face Density] 
%Assume (D1), (K1), and (B2-B4). With $h \to 0$, $k \to \infty$, $h k^{1/d}\to 0$, $\frac{n h}{k^{1-\frac{1}{d}}\log k} \to \infty$, then we have
%\begin{align*}
%\left |\frac{\hat{S}_{j\ell}^{FD}}{S_{j\ell}^{FD}} -1 \right|&=  O\big(hk^{1/d}\big) + O_p\bigg(\sqrt{\frac{k^{1-\frac{1}{d}}}{nh}}\bigg) 
%\max_{(j,\ell)\in E}\left |\frac{\hat{S}_{j\ell}^{VD}}{S_{j\ell}^{VD}} -1 \right|&=  O\big(hk^{1/d}\big) + O_p\bigg(\sqrt{\frac{k^{1-\frac{1}{d}}\log k }{nh}}\bigg) 
%\end{align*}
%Assume (D1), (K1), and (B2-B4). With $h \to 0$, $k \to \infty$, $h k^{1/d}\to 0$, $\frac{n h}{k^{1-\frac{1}{d}}\log k} \to \infty$, then
%for any pair $j\neq\ell$, we have
%\begin{align*}
%\left |\frac{\hat{S}_{j\ell}^{FD}}{S_{j\ell}^{FD}} -1 \right|&=  O\big(hk^{1/d}\big) + O_p\bigg(\sqrt{\frac{k^{1-\frac{1}{d}}}{nh}}\bigg).
%\end{align*}
%\end{manualtheorem}

\begin{proof}[ of Theorem~\ref{thm::FD}]

Our analysis starts with the usual bias-variance decomposition that 
$$
\hat S_{j\ell}^{FD} -  S_{j\ell}^{FD} = \underbrace{\hat S_{j\ell}^{FD} - \EE(\hat S_{j\ell}^{FD})}_\text{\bf stochastic variation} +\underbrace{\EE(\hat S_{j\ell}^{FD} )-  S_{j\ell}^{FD}}_{\bf bias}.
$$
We analyze the two term separately. 
Before we start our proof, we first recall some useful notations.

Recall that 
the face region between two knots $c_j, c_\ell $ is $F_{j\ell} \equiv \overline{\CC}_j \cap \overline{\CC}_l$ and $c_* = c_j + \frac{1}{2}(c_\ell - c_j) = \frac{1}{2}(c_\ell+c_j)$
and $\LL_{j\ell} = \{c_j - a(c_\ell - c_j) : a \in [0,1]\}$ is the central line passing through $c_j$ and $c_\ell$, and for a value $a \in [0,1]$. 
The face $F_{j\ell} = \big\{x \in \overline{\CC}_j \cup \overline{\CC}_l : \Pi_{j\ell}(x) = c_*  \big\}$, where $\Pi_{j\ell}$ denotes the projection onto $\LL_{j\ell}$.
The quantity $\mu_s(dx)$ denotes the integration with respect to $s$-dimensional volume.
We now reparametrize any point in $\LL_{j\ell}$ using a unit distance $t$.
Let $T_{j\ell,t} = \big\{x \in \calX :  \Pi_{j\ell}(x) = c_* + t\frac{c_\ell - c_j}{ ||c_\ell - c_j||} \big\}$ be the subspace orthogonal to $\LL_{j\ell}$ at the point $c_* + t\frac{c_\ell - c_j}{ ||c_\ell - c_j||}$. $t$ is $1$-dimensional distance to $c_*$ along the line passing through $c_j$ and $c_\ell$. Let
\begin{align*}
    q_{j\ell}(t) = \int_{(\overline\CC_j \cup \overline\CC_\ell) \cap T_{j\ell, t}} p(x) \mu_{d-1}(dx)
\end{align*}

With these quantities,
$S_{j\ell}^{FD} = q_{j\ell}(0) $ and that $q_{j\ell}(t)$ is a $1$-dimensional quantity. 
Our estimator is
$$
\hat S_{j\ell}^{FD} = \frac{1}{nh} \sum_{i = 1}^n K \bigg( \frac{\Pi_{j\ell} (X_i) - c_* }{h} \bigg) I(X_i \in \overline \CC_j \cup \overline\CC_\ell).
$$

\textbf{Bias}: We study the bias part first. A direct computation shows that
\begin{align}
%\begin{split}
    \EE[\hat{S}_{j\ell}^{FD}] &= \EE \bigg( \frac{1}{nh} \sum_{i = 1}^n K \bigg( \frac{\Pi_{j\ell} (X_i) - c_* }{h} \bigg) I(X_i \in \overline \CC_j \cup \overline\CC_\ell)  \bigg)\\
    &= \frac{1}{h} \int_{x \in \calX } K \bigg( \frac{\Pi_{j\ell} (x) - c_* }{h} \bigg) I(x \in \overline\CC_j \cup \overline\CC_\ell)  p(x) \mu_d(dx)\\
    &= \frac{1}{h} \int_{\LL_{j\ell}} K \bigg( \frac{ c_* 
    + t\frac{c_\ell - c_j }{||c_\ell - c_j ||}- c_* }{h} \bigg)  \bigg(\int_{(\overline\CC_j \cup \overline\CC_\ell) \cap T_{j\ell, t}} p(y) \mu_{d-1}(d y) \bigg) d \bigg(c_j + t\frac{c_\ell - c_j }{||c_\ell - c_j ||}\bigg)\\
    &= \frac{1}{h} \int_{\RR} K \bigg( \frac{\big\Vert t\frac{c_\ell - c_j }{||c_\ell - c_j ||} \big\Vert}{h} \bigg)  q_{j\ell}(t)  d t \\
    &= \frac{1}{h} \int_{\RR} K \bigg( \frac{t}{h} \bigg)  q_{j\ell}(t)  d t\\
    &=  \int_{\RR} K(u) q_{j\ell} (  hu ) du,
%\end{split}
\label{eq::FDbias1}
\end{align}
where for the third equality, we split the integration with respect to $c_j + t\frac{c_\ell - c_j }{||c_\ell - c_j ||} \in \LL_{j\ell}$ and the integration with respect to the subspace orthogonal to $\LL_{j\ell}$ at $c_j + t\frac{c_\ell - c_j }{||c_\ell - c_j ||}$. 
This is possible because all the points in $T_{j\ell,t}$ have the same projection onto $\LL_{j\ell}$. For the fourth equality, we used 
the symmetry of the kernel function.
the property of the kernel function that $K(x) = K(\|x\|)$. 
For the last equality, we used the change of variable that $u = \frac{t}{h}  $ and got the simplified form.

The expansion of 
$$
q_{j\ell}(t) = \int_{(\overline\CC_j \cup \overline\CC_\ell) \cap T_{j\ell, t}} p(y) \mu_{d-1}(d y)
$$
is more involved when $t\approx0$.
Let 
\begin{align*}
W_{j\ell}(t) &= (\overline\CC_j \cup \overline\CC_\ell) \cap T_{j\ell, t}\\
&=\begin{cases}
\overline\CC_j\cap T_{j\ell, t},\quad t<0,\\
\overline\CC_\ell\cap T_{j\ell, t},\quad t>0,\\
(\overline\CC_j \cup \overline\CC_\ell) \cap T_{j\ell, 0} = F_{j\ell},\quad t=0
\end{cases}
\end{align*}
be the region that leads to $q_{j\ell}(t)$.
For a face $F_{j\ell}$ and a real number $t\in\RR$, we denote 
$$
F_{j\ell}\oplus t = \left\{x+t\frac{c_\ell - c_j }{||c_\ell - c_j ||}  : x\in F_{j\ell}\right\}.
$$
By the above notation,
we can decompose
$$
W_{j\ell}(t) = [F_{j\ell}\oplus t] \cup \Delta_{j,\ell}(t),
$$ 
where $\Delta_{j,\ell}(t)$ is the additional region when 
moving away from $t=0$;
see Figure \ref{fig::faceproof1} for an example.

\begin{figure}
\captionsetup{skip=1pt}
\centering
%    \begin{subfigure}[t]{0.28\textwidth}
%        \centering
%        \includegraphics[width=\linewidth]{Yinyang_MST.JPG} 
%        \caption{}
%    \end{subfigure}
%    \begin{subfigure}[t]{0.34\textwidth}
%        \centering
        \includegraphics[width=5cm]{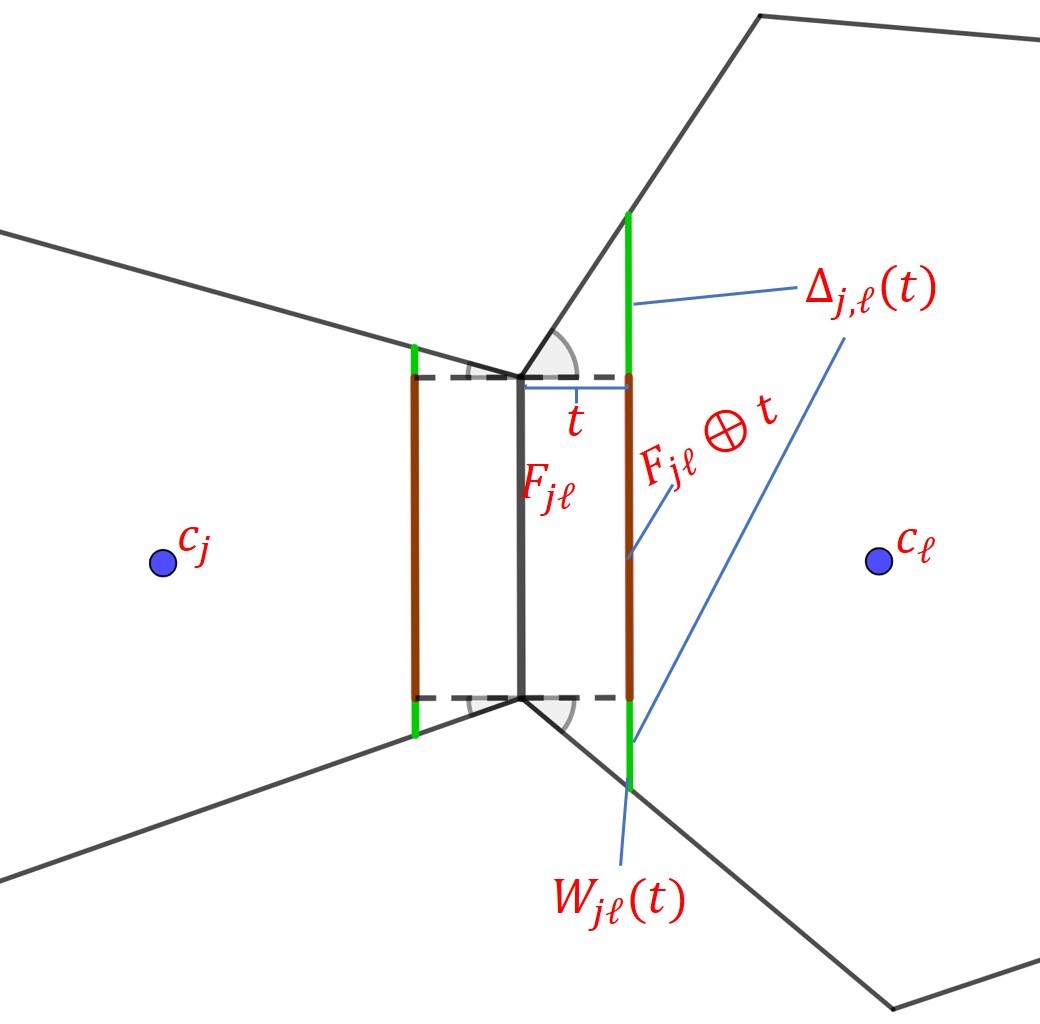}
%        \caption{}
%    \end{subfigure}
\caption{Decomposition of $W_{j\ell}(t)$. The dark red segment is $F_{j\ell}\oplus t$, which has the same shape with $F_{j\ell}$. The green segments consist $\Delta_{j,\ell}(t)$, the part leading to geometric bias.}
\label{fig::faceproof1}
\end{figure}

Thus, the difference 
\begin{align*}
q_{j\ell}(hu) - q_{j\ell}(0) &= \int_{W_{j\ell}(hu)} p(y) \mu_{d-1}(d y) -  \int_{W_{j\ell}(0)} p(y) \mu_{d-1}(d y)\\
& =  \underbrace{\int_{F_{j\ell}\oplus hu} p(y) \mu_{d-1}(d y) - \int_{F_{j\ell}} p(y) \mu_{d-1}(d y)}_{(I)} + \underbrace{\int_{\Delta_{j\ell}(hu)} p(y) \mu_{d-1}(d y)}_{(II)}.
\end{align*}

(I) is the usual bias caused by the change of density.
Note that
the Lipchitz condition in (D1) implies that  there is a constant $C_g$ such that $|p(x_1)-p(x_2)| \leq C_g |x_1 - x_2|$. 
Since every point can be matched nicely between $F_{j\ell}\oplus hu$ and $F_{j\ell}$,
it can be bounded by 
$$
|(I)|\leq \mu_{d-1}(F_{j\ell})C_g h |u|.
$$

\begin{figure}
\captionsetup{skip=1pt}
\centering
%    \begin{subfigure}[t]{0.28\textwidth}
%        \centering
%        \includegraphics[width=\linewidth]{Yinyang_MST.JPG} 
%        \caption{}
%    \end{subfigure}
%    \begin{subfigure}[t]{0.34\textwidth}
%        \centering
        \includegraphics[width=7cm]{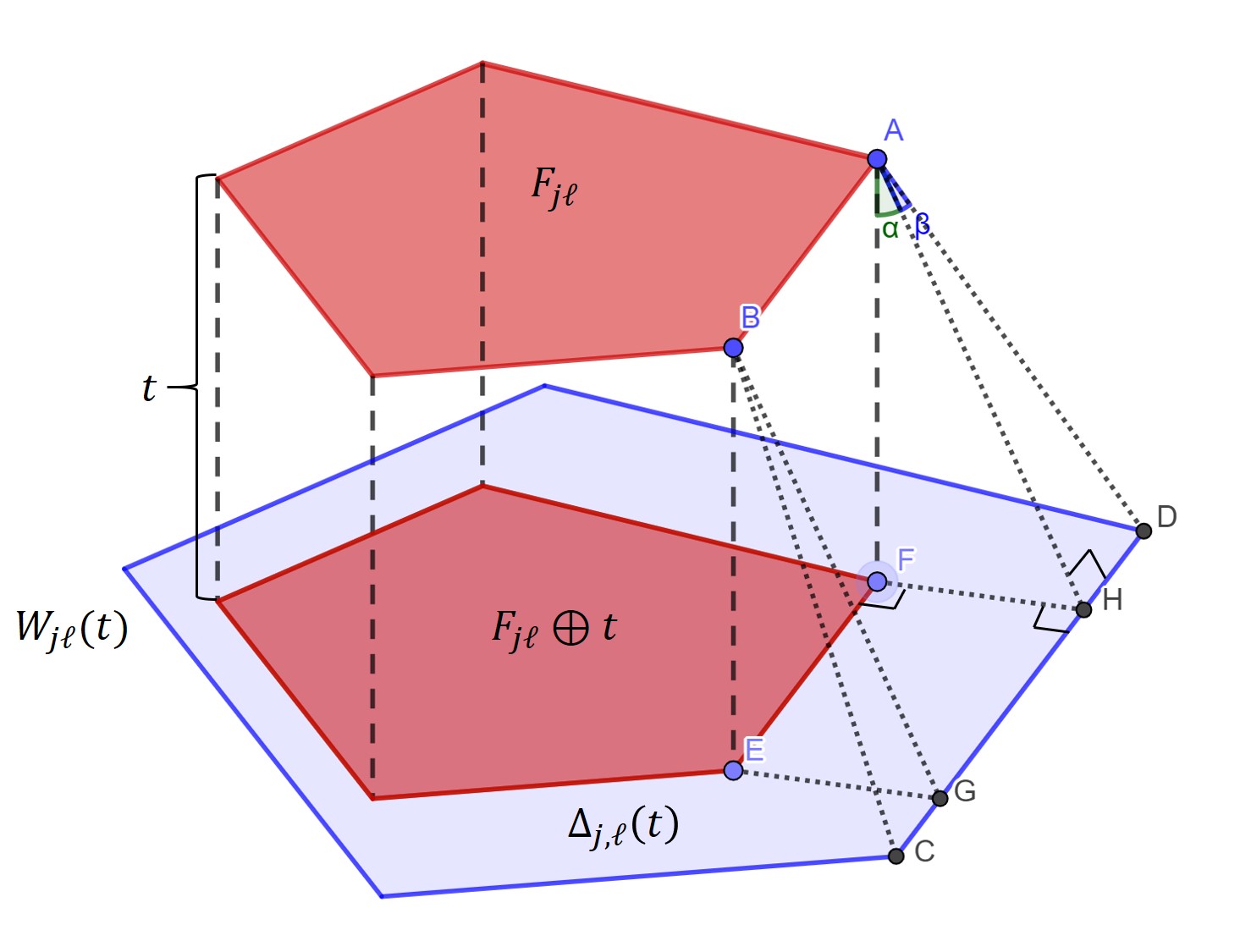}
%        \caption{}
%    \end{subfigure}
\caption{Decomposition of $W_{j\ell}(t)$. The red regions are $F_{j\ell}$ and the projected $F_{j\ell}\oplus t$, while the blue band region denotes $\Delta_{j,\ell}(t)$. All the $\alpha$ angles such as $\angle FAH$ and all the $\beta$ angles such as $\angle HAD$ are bounded by $\theta_0$ from assumption (B4).}
\label{fig::faceproof2}
\end{figure}

(II) is the bias due to the change of volume, so we call it a geometric bias.
With an upper bound of the density, 
(II) can be bounded by $(II)\leq \mu_{d-1}(\Delta_{j,\ell}(hu))\cdot p_{max}.$
Thus, we only need to bound the volume $\mu_{d-1}(\Delta_{j,\ell}(hu))$.

$\Delta_{j,\ell}(t)$ is illustrated by the blue region in Figure  \ref{fig::faceproof2}. The width of the band region like $FH $ will all be bounded by $t\tan(\theta_0) = O(t)$, and as $t\rightarrow 0$ the surface area (circumference) will be bounded by $O\big( \mu_{d-2}(\partial F_{j\ell} \big))$.
%Note that we can decompose the region into the increment from the boundary $\partial F_{j\ell}$ and the `corner' part of the boundary $\partial F_{j\ell}$. 
%The contribution from the boundary is from regions like the the rectangular area $EFGH$. 
%The corner region consists of area of the triangles $DFH$ and $CEG$. 

%which also implies that the triangle region $DFH$ and $CEG$ will be of the order $O(h^2).$
Thus, the volume of the blue region $\mu_{d-1}(\Delta_{j,\ell}(t)) \leq O\big(\mu_{d-2}(\partial F_{j\ell})t \big)$,
which leads to the bound
$$
(II)\leq   O\big( h|u| \cdot \mu_{d-2}(\partial F_{j\ell}) \big)\cdot p_{max} .
$$

Putting altogether,
we have
\begin{align}
     \big|q_{j\ell} ( hu ) -  q_{j\ell} ( 0) \big|
     \leq 
     \mu_{d-1}(F_{j\ell})C_g h |u| +  p_{max}  h|u|\cdot O\big(\mu_{d-2}(\partial F_{j\ell})\tan(\theta_0)\big) 
%     p_{max} \bigg\{\sum_{p=1}^{d-1} O(|hu|^{p}) \mu_{d-p-1}(\partial^p F_{j\ell})\bigg\}
\label{eq::FDdecomp}
\end{align}

This, together with equation \eqref{eq::FDbias1}, implies that 
\begin{align*}
\begin{split}
|\EE[\hat{S}_{j\ell}^{FD}] - \underbrace{q_{j\ell}(0)}_{=S_{j\ell}^{FD}}|&= 
\left|\int_{\RR} K(u) [q_{j\ell} (  hu )- q_{j\ell}(0)] du\right|\\
&\leq \int_{\RR} K(u) |q_{j\ell} (  hu )- q_{j\ell}(0)| du\\
&\leq h \bigg[\int_{\RR} |u| K(u)du \bigg] \times  \bigg[ \mu_{d-1}(F_{j\ell})C_g + p_{max} O\big(\mu_{d-2}(\partial F_{j\ell}) \big) \bigg] \\
%& \qquad+ O(h^2) \cdot \int_{\RR} u^2 K(u)du \\
%& \ \ \ \ + p_{max} \sum_{p=1}^{d-1} O(|h|^{p}) \mu_{d-p-1}(\partial^p F_{j\ell}) \int_{\RR} |u|^p K(u)du\\
& \overset{(B2-3)}{=}O\left(h \cdot\left[\frac{1}{k^{1-1/d}}\right] \right) + O\left(h \cdot\left[\frac{1}{k^{1-2/d}}\right] \right)
%h\int |u|K(u)du\cdot\mu_{d-2}(\partial F_{j\ell})\tan(\theta_0) p_{max} + O(h^2)
%\sum_{p=1}^{d-1} O\left(h^p \cdot\left[\frac{1}{k^{1-(p+1)/d}}\right] \right)
\end{split}
\end{align*}
As a result,
\begin{align}
    \begin{split}
         | \EE[\hat{S}_{j\ell}^{FD}] - S_{j\ell}^{FD}  |  = O\bigg(\frac{h}{k^{1-1/d}}\bigg) +  O\bigg(\frac{h}{k^{1-2/d}}\bigg)
    \end{split}
\label{eq::FDbias2}
\end{align}

Moreover, note that 
\begin{align}
    \frac{h}{k^{1-1/d}}\times\frac{k^{1-2/d}}{h} = \frac{1}{k^{1/d}} \to 0
\label{eq::FDratecompare1}
\end{align}
since $k\to \infty$. Therefore the bias given by the geometric difference (II) dominates the bias given by the change in density (I). 
Even if we assume a higher-order derivative, the bias in (II) will still dominate the component in (I).

%Note that if we require higher order smoothness of the density function that $p$ is of class $C^{r}$ for integer $r>1$, then we would have the bias term given by the change in density to be of rate $O\big(\frac{h^r}{k^{1-1/d}}\big)$ and note that $h\to 0$ so
%\begin{align}
%    \frac{h^r}{k^{1-1/d}}\times\frac{k^{1-2/d}}{h} = \frac{h^{r-1}}{k^{1/d}} \to 0
%\end{align}
%again would be dominated by the bias given by the geometric difference.

Therefore, the overall bias can be expressed as
reduces to
\begin{align}
    \begin{split}
         | \EE[\hat{S}_{j\ell}^{FD}] - S_{j\ell}^{FD}  |  = O\bigg(\frac{h}{k^{1-2/d}}\bigg)
    \end{split}
\label{eq::FDbias3}
\end{align}

\textbf{Stochastic variation}: For the stochastic variation part, we have
\begin{align}
    \begin{split}
        Var(\hat{S}_{j\ell}^{FD}) &= Var \bigg( \frac{1}{nh} \sum_{i = 1}^n K \bigg( \frac{\Pi_{j\ell} (X_i) - c_*)}{h} \bigg) I(X_i \in \overline \CC_j \cup \overline\CC_\ell)  \bigg)\\
        &\leq \frac{1}{n h^2} \EE\bigg[ K^2 \bigg( \frac{\Pi_{j\ell} (X_i) - c_*}{h} \bigg) I(X_i \in \overline \CC_j \cup \overline\CC_\ell)  \bigg]\\
        &\leq \frac{1}{n h} \int K^2(u) \bigg(q_{j\ell}(0) +\mu_{d-1}(F_{j\ell})C_g + p_{max}  h |u| \mu_{d-2}(\partial F_{j\ell}) \tan(\theta_0) \bigg) du\\
        &\leq \frac{1}{n h} \int K^2(u) \bigg(q_{j\ell}(0) +O\bigg(\frac{h}{k^{1-1/d}}\bigg) +  O\bigg(\frac{h}{k^{1-2/d}}\bigg)\bigg) du\\
%        &= \frac{1}{n h} \bigg\{ q_{j\ell}(0) \bigg[\int K^2(u) d u \bigg] + \mu_{d-1}(F_{j\ell})C_H h^2 \bigg[ \int K^2(u) u^2 d u\bigg]\\
%        & \ \ \ \ + p_{max}  h  \mu_{d-2}(\partial F_{j\ell}) \tan(\theta_0) \bigg[ \int K^2(u) |u| d u \bigg] \bigg\} \\
%        &= \frac{1}{nh}\bigg\{ O(q_{j\ell}(0)) + O\bigg(\frac{h^2}{k^{1-1/d}}\bigg) +  O\bigg(\frac{h}{k^{1-2/d}}\bigg) \bigg\}
    \end{split}
\label{eq::FDvariation}
\end{align}
by the same decomposition in (\ref{eq::FDbias1}) and the bound in (\ref{eq::FDdecomp}) and the assumptions  (K1). 
Note that similar to(\ref{eq::FDratecompare1}), the second term in (\ref{eq::FDvariation}) is at a slower rate than the third term,
so we can simplify it as
\begin{align}
Var(\hat{S}_{j\ell}^{FD}) = O\bigg( \frac{q_{j\ell}(0)}{nh} \bigg)+ O\bigg(\frac{1}{nk^{1-2/d}}\bigg).
\label{eq::FDvariation2}
\end{align}

Combining (\ref{eq::FDbias2}) and (\ref{eq::FDvariation}),
we conclude that
for $\forall j,\ell$,
\begin{align}
\begin{split}
    | \hat{S}_{j\ell}^{FD}  - S_{j\ell}^{FD}   | &=    O\bigg(\frac{h}{k^{1-2/d}}\bigg) +O_p\bigg(\sqrt{\frac{q_{j\ell}(0)}{nh}}\bigg) +  O_p\bigg(\sqrt{\frac{1}{nk^{1-2/d}}}\bigg)
\end{split}
\end{align}
Note that the volume of face region $F_{j\ell}$ decreases when $k$ increases.
By assumption (D1) and (B2), we have
\begin{align}
    q_{j\ell}(0) = S_{j\ell}^{FD} \geq p_{\min} \min_{(j,\ell)\in E} \mu_{d-1}(F_{j\ell}) = p_{\min} \frac{c_0}{k^{1-\frac{1}{d}}}.
\end{align}
For the theorem, we again take the ratio between the estimated and the true face density to accommodate the fact that the true face density is decreasing with the number of knots, and we have that
This implies that
\begin{align}
    \left | \frac{\hat{S}_{j\ell}^{FD}}{S_{j\ell}^{FD} }  - 1  \right| =  O\big(hk^{1/d}\big) + O_p\bigg(\sqrt{\frac{k^{1-\frac{1}{d}}}{nh}}\bigg) + O_p\bigg(\sqrt{\frac{k}{n}}\bigg)
\label{eq::FDrate1}
\end{align}

When $h  k^{1/d} \to 0$,
\begin{align}
    \frac{k^{1-\frac{1}{d}}}{nh} \times \frac{n}{k} = \frac{1}{h k^{1/d}} \to \infty,
\end{align}
so the second term dominates the third term in (\ref{eq::FDrate1}) and
the rate reduces to
\begin{align}
    \left | \frac{\hat{S}_{j\ell}^{FD}}{S_{j\ell}^{FD} }  - 1  \right| = O\big(hk^{1/d}\big) + O_p\bigg(\sqrt{\frac{k^{1-\frac{1}{d}}}{nh}}\bigg) ,
\label{eq::FDrate2}
\end{align}
which completes the proof.
%The stochastic variation part in (\ref{eq::FDrate2}) is similar to the usual variation for KDE with an extra term depending on $k$ accounting for the change in the face area.

%The uniform bound across all pairs can be derived based on
%a similar maximal deviation approach.
%Since it is a standard procedure, we omit the proof.
%
%There are $\frac{k(k-1)}{2}$ combinations of different pairs of centers, and hence
%\begin{align}
%\begin{split}
%\max_{(j,\ell)\in E}\left |\frac{\hat{S}_{j\ell}^{VD}}{S_{j\ell}^{VD}} -1 \right| 
%&\leq \sum_{(j,\ell)\in E}\left |\frac{\hat{S}_{j\ell}^{VD}}{S_{j\ell}^{VD}} -1 \right|\\ 
%&= O\big(hk^{1/d}\big) + O_p\bigg(\sqrt{\frac{k^{1-\frac{1}{d}}\log k }{nh}}\bigg) 
%\end{split}
%\end{align}

\end{proof}

\subsubsection{Proofs for Tube Density Consistency}

We consider the following assumptions, which are slightly stronger than those in the case of the FD:

\begin{itemize}
    \item[\textbf{(D2)}] (Density conditions) 
    The PDF $p$ has compact support, is in the $3$-H{\"o}lder class, and $\inf_{x \in \calX} p(x) \geq f_{\min} > 0$.
    \item[\textbf{(D3)}] (Disk Density conditions) 
    For any pair $c_j,c_\ell$, the minimum disk density location $t^* = {\sf argmin}_{t\in[0,1]}\pDisk_{j\ell, R}(t)\in(0,1)$
    is unique and satisfies $\pDisk^{(2)}_{j\ell, R}(t^*)\geq c_{\min}>0$.
    \item[\textbf{(K2)}] (Kernel function conditions)  The kernel function $K$ is a positive and symmetric function  
    satisfying
%    such that $K(x) = K(||x||)$ has bounded
%    , continuous third derivatives and 
%    \begin{align*}
%       \int |x| K (x) dx < \infty, 
%       \int K^2 (x) dx < \infty,\quad \int x^2 K (x) dx.
      $ \int x^2 K^{(\al)} (x) dx < \infty, \int (K^{(\al)}(x))^2 dx < \infty,$
%    \end{align*}
    for all $\al = 0,1,2$, where $K^{(\al)}$ denotes the $\al$-th order derivative of $K$.
%    particularly, for $x \in \calX$,
%    $$\int x^2 K(x) dx = m_2(K) \bm I_d$$ 
%    for some real number $m_2(K)$ with $\bm I_d$ the identity matrix of order $d$.
\end{itemize}

%{\color{magenta}YC: I think we should just write one single proof of decreasing $R$.}

%\begin{manualtheorem}{\ref{thm::TD1}}[Tube Density Consistency]  
%Assume (D2) and (K2), and $h \to 0$, $k \to \infty$, $R \to 0$, $\frac{n h^3}{\log k} \to \infty$,$\frac{n h R^{d-1}}{\log k} \to \infty$.
%Suppose that for every pair $c_j, c_\ell$, $\inf_{t \in [0,1]} \pDisk_{j\ell,R}(t)$ and  $\inf_{t \in [0,1]} \widehat{\pDisk}_{j\ell,R}(t)$ do not occur at the boundary $t = 0,1$. Then for any pair $j\neq\ell$, we have
%\begin{align*}
%\pDisk_{j\ell,R}(t) &= O(R^{d-1}).
%\end{align*}
%\end{manualtheorem}
%
\begin{proof}[ of Theorem~\ref{thm::TD1}]

Let $t^* = {\sf argmin}_t \pDisk_{j\ell,R}(t)$ and $\hat t^* = {\sf argmin}_t \hat\pDisk_{j\ell,R}(t)$.
Then the tube densities
\begin{align*}
S_{j\ell}^{TD} &= \inf_{t \in [0,1]} \pDisk_{j\ell,R}(t) = \pDisk_{j\ell,R}(t^*),\\
\hat S_{j\ell}^{TD} &= \inf_{t \in [0,1]} \hat\pDisk_{j\ell,R}(t) = \hat \pDisk_{j\ell,R}(\hat t^*).
\end{align*}
Since the ratio difference 
$$
\frac{\hat{S}_{j\ell}^{TD}}{S_{j\ell}^{TD}}-1 = \frac{1}{S_{j\ell}^{TD}}\left(\hat S_{j\ell}^{TD} - S_{j\ell}^{TD}\right),
$$
we will focus on the difference $\hat S_{j\ell}^{TD} - S_{j\ell}^{TD}$.

The difference admits the following decomposition:
\begin{align*}
\hat S_{j\ell}^{TD} - S_{j\ell}^{TD} & = \hat \pDisk_{j\ell,R}(\hat t^*) - \pDisk_{j\ell,R}(t^*)\\
& = \underbrace{\hat \pDisk_{j\ell,R}(\hat t^*) - \hat \pDisk_{j\ell,R}( t^*)}_{(I)}+
\underbrace{\hat \pDisk_{j\ell,R}( t^*) - \EE(\hat \pDisk_{j\ell,R}( t^*))}_{(II)}\\
&\qquad\qquad+
\underbrace{\EE(\hat \pDisk_{j\ell,R}( t^*)) - \pDisk_{j\ell,R}(t^*)}_{(III)}.
\end{align*}
It is easier to start with term (III) and then term (II) and then term (I).

%We begin by analyzing the estimation of density on a fixed disk (given $c_j, c_\ell$ and fixed $t$). 
%\begin{align*}
%\begin{split}
%     | \widehat{\pDisk}_{j\ell, R}(t)  - \pDisk_{j\ell, R}(t)  | \leq & | \widehat{\pDisk}_{j\ell, R}(t)  - \EE[\widehat{\pDisk}_{j\ell, R}(t)]  | \\
%     &+ | \EE[\widehat{\pDisk}_{j\ell, R}(t)] - \pDisk_{j\ell, R}(t)  |
%\end{split}
%\end{align*}

%Those terms can be analyzed by Kernel Density Estimation techniques. First for notation let 
%The first term is the stochastic variation and the second term is the bias.
Recall that 
\begin{align*}
\begin{split}
    q_{v,R}(y) = \int_{Disk(y, R, v)} p(x) dx,
\end{split}
\end{align*}
and hence $\pDisk_{j\ell, R}(t) = q_{c_\ell - c_j, R}(c_j - t(c_\ell - c_j))$.
%We start with the analysis of bias first.

\textbf{(III): Bias.} Note that the kernel weights $w(x) = K \big( \frac{\Pi_{j\ell} (x) - c_j - t(c_\ell - c_j)}{h} \big)$ is the same for all $ x \in Disk(c_j - t(c_\ell - c_j), R, c_\ell - c_j)$. Let $\LL_{j\ell} = \{c_j - t(c_\ell - c_j) : t \in \RR\}$ be the line passing through $c_j$ and $c_\ell$. Then
\begin{align*}
\begin{split}
    \EE[\widehat{\pDisk}_{j\ell, R}(t)] &= \EE \bigg( \frac{1}{nh} \sum_{i = 1}^n K \bigg( \frac{\Pi_{j\ell} (X_i) - c_j - t(c_\ell - c_j)}{h} \bigg) I\big(||X_i - \Pi_{j\ell}(X_i) ||\leq R\big)  \bigg)\\
    &= \frac{1}{h} \int_{x \in \calX } K \bigg( \frac{\Pi_{j\ell} (x) - c_j - t(c_\ell - c_j)}{h} \bigg) I(||x - \Pi_{j\ell}(x) || \leq R)  p(x) \mu_{d}(d x)\\
    &= \frac{1}{h} \int_{\LL_{j\ell}} K \bigg( \frac{z - c_j - t(c_\ell - c_j)}{h} \bigg)  \bigg(\int_{Disk(z, R, c_\ell - c_j)} p(y) \mu_{d-1}(d y) \bigg)  d z\\
    &= \frac{1}{h} \int_{\LL_{j\ell}} K \bigg( \frac{z - c_j - t(c_\ell - c_j)}{h} \bigg)  q_{c_\ell - c_j, R}(z)   d z\\
    &= \frac{||c_j - c_\ell||}{h} \int_{\LL_{j\ell}} K \bigg( \frac{(s-t) ||c_j - c_\ell||}{h} \bigg)  q_{c_\ell - c_j, R}(c_j - s(c_\ell - c_j))  d s    
\end{split}
\end{align*}
where for the third equality we split the integration with respect to $z \in \LL_{j\ell}$ and the integration with respect to $y \in Disk(z, R, c_\ell - c_j)$, and for the last equality we set $z = c_j - s(c_\ell - c_j)$ and utilized the symmetry of the kernel function $K$.

Then by another change of variable that $u = \frac{(s-t)||c_\ell - c_j||}{h}$ and Taylor expansion, we have 
\begin{align*}
    \begin{split}
        \EE[\widehat{\pDisk}_{j\ell, R}(t)] &= \int K(u) q_{c_\ell - c_j, R} \bigg( c_j - t(c_\ell - c_j) - hu \frac{c_\ell - c_j}{ ||c_j - c_\ell||} \bigg) du\\
        &=  \int K(u) \bigg(q_{c_\ell - c_j, R}( c_j - t(c_\ell - c_j))  + hu \cdot g_1 + \frac{1}{2}h^2 u^2 \cdot g_2+ O(h^2) \bigg) du\\
    \end{split}
\end{align*}
where
\begin{align*}
    \begin{split}
        g_1 &= \bigg(\frac{c_\ell - c_j}{ ||c_j - c_\ell||}\bigg)^T \cdot \nabla q_{c_\ell - c_j, R}( c_j - t(c_\ell - c_j)) \\
        g_2 &= \bigg(\frac{c_\ell - c_j}{ ||c_j - c_\ell||}\bigg)^T \cdot \nabla \nabla q_{c_\ell - c_j, R}( c_j - t(c_\ell - c_j)) \bigg(\frac{c_\ell - c_j}{ ||c_j - c_\ell||}\bigg)
    \end{split}
\end{align*}

When $R\rightarrow0$,
assumption (D2) implies that
there is a constant $C_{d-1}$ that 
\begin{equation}
    2 p_{\min} C_{d-1} R^{d-1}\leq \pDisk_{j\ell, R}(t) \leq 2 p_{\max} C_{d-1} R^{d-1} = O(R^{d-1})
    \label{eq::D2new}
\end{equation}
where $0 < p_{\min} \leq \inf_{x \in \calX} p(x)$, $\sup_{x \in \calX} p(x) \leq p_{\max}  < \infty$. 
Since the disk density is shrinking at rate $O(R^{d-1})$,
one can easily verify that the gradient and Hessian of the disk density function are also at rate $O(R^{d-1})$.
Namely,
$$
g_1 = O(R^{d-1}),\qquad g_2 = O(R^{d-1}).
$$

By assumption (\textbf{D2}) we have $g_1$ and $g_2$ to be bounded and therefore
Thus,
\begin{align*}
    \begin{split}
        \EE[\widehat{\pDisk}_{j\ell, R}(t)] &=  q_{c_\ell - c_j, R}( c_j - t(c_\ell - c_j)) \int K(u) d u + h \bigg[\int u K(u)  du \bigg] \cdot g_1 \\
        &+ \frac{1}{2}h^2 \bigg[\int u^2 K(u) du \bigg] \cdot g_2 + O(h^2R^{d-1})  \\
        &= q_{c_\ell - c_j, R}( c_j - t(c_\ell - c_j)) + O(h^2R^{d-1})\\
        &= \pDisk_{j\ell, R}(t) + O(h^2R^{d-1}),
    \end{split}
\end{align*}
where for the second equality we used, by assumption (K)
\begin{align*}
    \begin{split}
         \int K(u) d u = 1, \int u K(u) d u = 0,\int u^2 K(u) d u  < \infty
    \end{split}
\end{align*}
so we conclude that  $| \EE[\widehat{\pDisk}_{j\ell, R}(t)] - \pDisk_{j\ell, R}(t)  |  = O(h^2R^{d-1})$

\textbf{(II): Stochastic variation.}
\begin{align*}
%    \begin{split}
        Var(\widehat{\pDisk}_{j\ell, R}(t)) &= Var \bigg( \frac{1}{nh} \sum_{i = 1}^n K \bigg( \frac{\Pi_{j\ell} (X_i) - c_j - t(c_\ell - c_j)}{h} \bigg) I(||X_i - \Pi_{j\ell}(X_i) \leq R)  \bigg)\\
        &\leq \frac{1}{n h^2} \EE\bigg[ K^2 \bigg( \frac{\Pi_{j\ell} (X_i) - c_j - t(c_\ell - c_j)}{h} \bigg) I(||X_i - \Pi_{j\ell}(X_i) \leq R)  \bigg]\\
        &= \frac{1}{n h} \int K^2(u) \bigg(q_{c_\ell - c_j, R}( c_j - t(c_\ell - c_j))  + hu \cdot g_1 + O(h^2) \bigg) du\\
        &= O\bigg(\frac{1}{nh}\bigg)
%    \end{split}
\end{align*}
by the same analysis procedure as for Face Density and the assumptions (D1), (K1). 

Now, by assumption (D2), the face density $q_{c_\ell - c_j, R}( c_j - t(c_\ell - c_j)) = O(R^{d-1})$, which leads to 
$$
Var(\widehat{\pDisk}_{j\ell, R}(t))  = O\left(\frac{R^{d-1}}{nh}\right).
$$

Therefore, 
\begin{align*}
    | \widehat{\pDisk}_{j\ell, R}(t)  - \EE[\widehat{\pDisk}_{j\ell, R}(t)]  | = O_p \bigg(\sqrt{\frac{R^{d-1}}{nh}}\bigg)
\end{align*}
and
\begin{equation}
    | \widehat{\pDisk}_{j\ell, R}(t)  - \pDisk_{j\ell, R}(t)  | = O(h^2R^{d-1}) + O_p \bigg( \sqrt{\frac{R^{d-1}}{nh}} \bigg).
    \label{eq::TD::rate1}
\end{equation}

\textbf{(I): Change in position.}
Finally, we bound the term
$$
(I) = \hat\pDisk_{j\ell, R}(\hat{t}^*) - \hat\pDisk_{j\ell, R}({t}^*).
$$
Note that the minimizer $\hat{t}^*$ satisfies the gradient condition 
$$
\hat\pDisk'_{j\ell, R}(\hat{t}^*) = 0.
$$
By a simple Taylor expansion at $\hat t^*$,
we obtain
\begin{align*}
(I)& =- (\hat\pDisk_{j\ell, R}({t}^*)- \hat\pDisk_{j\ell, R}(\hat{t}^*) )\\
& = -(t^* - \hat t^*) \underbrace{\hat\pDisk'_{j\ell, R}(\hat{t}^*)}_{=0} - \frac{1}{2}(t^* - \hat t^*)^2 \hat\pDisk''_{j\ell, R}(\hat{t}^*) + O(|t^* - \hat t^*|^3)\\
& = O(|t^* - \hat t^*|^2).
\end{align*}
Thus, we only need to derive the rate of $t^* - \hat t^*$.

Now by the fact that $t^*$ solves the population gradient condition $\pDisk'_{j\ell, R}({t}^*) = 0,$
we have 
\begin{align*}
\hat\pDisk'_{j\ell, R}({t}^*) - \pDisk'_{j\ell, R}({t}^*) 
& = \hat\pDisk'_{j\ell, R}({t}^*) - \hat\pDisk'_{j\ell, R}(\hat{t}^*) \\
& = \hat\pDisk''_{j\ell, R}({t}^*) (t^* - \hat t^*) +  O(|t^* - \hat t^*|^2).
\end{align*}
Because $ \hat\pDisk''_{j\ell, R}({t}^*)  \overset{P}{\rightarrow}  \pDisk''_{j\ell, R}({t}^*) $
from the analysis of term (II) and (III), we conclude that 
$$
\hat t^* - t^* = O(\hat\pDisk'_{j\ell, R}({t}^*) - \pDisk'_{j\ell, R}({t}^*) ) = O(h^2R^{d-1}) + O_P\left(\sqrt{\frac{R^{d-1}}{nh^3}}\right).
$$
Note that the above rate analysis follows from the same analysis as term (II) and (III)
except that we are using gradient rather than the density. 

As a result, we conclude that 
$$
(I) = O(|t^* - \hat t^*|^2) = O(h^4R^{2d-2}) + O_P\left({\frac{R^{d-1}}{nh^3}}\right).
$$

Combining together, we have
\begin{align*}
    \begin{split}
        |\hat{S}_{j\ell}^{TD} - S_{j\ell}^{TD} | & = (I)+(II)+(III)\\
        &= O(h^4R^{2d-2}) + O_p \bigg(\frac{R^{d-1}}{nh^3}\bigg) + O(h^2R^{d-1}) + O_p \bigg( \sqrt{\frac{R^{d-1}}{nh}} \bigg)\\
        &= O(h^2R^{d-1})  + O_p \bigg( \sqrt{\frac{R^{d-1}}{nh}} \bigg) + O_p \bigg(\frac{R^{d-1}}{nh^3}\bigg).
    \end{split}
\end{align*}
Using the fact that $S_{j\ell}^{TD} \geq 2 p_{\min} C_{d-1} R^{d-1}$ from equation \eqref{eq::D2new},
we conclude that 
$$
\left|\frac{\hat{S}_{j\ell}^{TD}}{{S}_{j\ell}^{TD}}-1\right| = O(h^2)  + O_p \bigg( \sqrt{\frac{1}{nhR^{d-1}}} \bigg) + O_p \bigg(\frac{1}{nh^3}\bigg),
$$
which completes the proof.

%In terms of concentration inequality, let $\eps > 0$, then there exists small $h$ such that $| \EE[\widehat{\pDisk}_{j\ell, R}(t)] - \pDisk_{j\ell, R}(t)  |   \leq \eps/3$ and there exists $a_1, a_2, a_3, a_4$ sucht that 
%\begin{align}
%\begin{split}
%    \PP(|\hat{S}_{j\ell}^{TD} - S_{j\ell}^{TD} | > \eps) \leq a_1 e^{- a_2 \cdot nh \eps^2/9} + a_3 e^{- a_4 \cdot n^2 h^6 \eps^2/9}
%\end{split}
%\end{align}
%
%
%Note that there are $\frac{k(k-1)}{2}$ combinations of different pairs of centers, and hence
%\begin{align}
%\begin{split}
%\max_{(j,\ell) \in E} |\hat{S}_{j\ell}^{TD} - S_{j\ell}^{TD} | &\leq \sum_{j,l} |\hat{S}_{j\ell}^{TD} - S_{j\ell}^{TD} |  \\
%&= O(h^2)  + O_p \bigg( \sqrt{\frac{\log k}{nh}} \bigg)+ O_p \bigg(\frac{\log k}{nh^3} \bigg)
%\end{split}
%\end{align}

\end{proof}

{
\subsection{Choice of Linkage} \label{sec::simLinkage}
~~~~In this section, we use different simulations to investigate the effect of different linkage criteria under our skeleton clustering framework.
We start with the same Yinyang data to illustrate how different linkages cope with well-separated clusters in Appendix \ref{sim::yinyangLinkage}. 
Next, we  add noisy observations to the Yinyang data and make the comparison again in Appendix \ref{sim::noisyYinyang}.
%assess how the proposed skeleton clustering framework with different choices of linkage perform with the presence noises.
Moreover, we repeat this comparison using different simulation scenarios when there are overlapping clusters; 
the comparisons in Appendix \ref{sim::mixMickey}, \ref{sim::noisyMixMickey}, \ref{sim::mixStar}, and \ref{sim::noisymixStar}.

%we illustrate that our clustering framework can deal with overlapping clusters with appropriate linkage method in Section \ref{sim::mixMickey}, 
%and finally we study a data scenario with both noisy data and overlapping clusters in Section \ref{sim::noisyMixMickey} to provide further guidance for linkage choice. 
%Although denoising the data before applying skeleton clustering approaches (Remark \ref{remark::denoise}) can improve clustering performance, we do not carry out this pre-processing step for simulations in this section to test the robustness of our framework to background noise. 

%For all the experiments in this section, we repeat simulate the data $100$ times, construct the skeleton graph, and calculate the edge weights given by Voronoi density estimates as in section \ref{sec::VoronDensity}. 
Except for the linkage criterion, all other procedures are the same with the following settings:
we use $k$-means clustering with $k =\sqrt{n}$ to find knots 
and use
the Voronoi density as the density-aided similarity measure.
We vary the total number of final clusters from $1$ to $40$ and compare
the adjusted Rand Index (ARI) to the actual cluster label. 
The entire procedure is repeated $100$ times for the comprehensive comparison of various linkage methods from the \texttt{hclust} function in R.
% We include more comprehensive comparisons of other linkage methods in Appendix \ref{sec::simLinkage}. Overall, single linkage and average linkage have superior performance with respect to different simulation scenarios.
%	we, with $100$ repetitions, randomly simulate the data, construct the skeleton graph, and calculate the edge weights given by Voronoi density estimates as in section \ref{sec::VoronDensity}.
%	 Different linkage methods are then used to partition the skeleton into a some numbers of final clusters. 
The medium performances of the resulting clusterings are summarized in Table \ref{tab::linkageComparison}. For datasets without noisy points, we only present the medium ARI at the true number of clusters, while for data with noisy points we show the best medium ARI across different $S$ and record the corresponding $S$ in the bracket. The best linkages for each data scenario are in bold. 

\begin{table}[ht]
\scriptsize
\centering
\setlength{\tabcolsep}{3pt}
\begin{tabular}{|c |c| c| c |c |c |c| c| c|} 
 \hline
  & average & centroid& complete & mcquitty &median& minimax & single & Ward\\ 
 \hline
 Yinyang,d=10 & {\bf 1.000} & 0.119 & -0.017 & {\bf1.000}&  0.111 & 0.027 & {\bf  1.000}  & {\bf 1.000}  \\ 
 Yinyang,d=100 &  {\bf  1.000}&   0.098 & -0.008 &  {\bf 1.000} &  0.097 &  0.055 &  {\bf 1.000} &   {\bf 1.000}  \\ 
 Yinyang,d=500 &  0.560 &   0.074  & -0.028 &   0.587   & 0.054 &   0.062  &  {\bf 1.000}   & 0.526 \\ 
 Yinyang,d=10000 &0.533  &0.107& -0.029 & 0.555&  0.021  &0.106&  {\bf 1.000}  &0.456 \\ 
 \hline
 MixMickey,d=10 & {\bf 0.731} & -0.005&   0.017 &  0.380 &  0.007 &  0.010&  -0.004 &  0.194  \\ 
 MixMickey,d=100 & {\bf 0.740}&  -0.005 &  0.005&   0.341&   0.010 &  0.043 & -0.001&   0.129  \\ 
 MixMickey,d=500&  {\bf 0.710}&  -0.003 &  0.003&   0.356 &  0.013&  -0.003&  -0.004  & 0.180 \\ 
 MixMickey,d=10000&  {\bf 0.692}&  -0.006 & -0.014 &  0.297 &  0.011&  -0.045 & -0.006 &  0.217 \\ 
  \hline
 MixStar,d=10
&\textbf{0.763}  
&0.0001
&0.00532  
&0.510    
&0.001
&0.0488   
&0.0001
& 0.424 \\ 
 MixStar,d=100
&\textbf{0.763}    
&0.0001
&0.007  
&0.540    
&0.001
&0.0503   
&0.0001
&0.415 \\ 
 MixStar,d=500
&\textbf{0.762 }   
&0.0001
&0.004
&0.537    
&0.001
&0.039   
&0.0001
&0.444\\ 
 MixStar,d=1000 
&\textbf{0.721}  
&0.0001
&0.005  
&0.533    
&0.001
&0.050   
&0.0001
&0.418 \\ 
 \hline
 NoisyYinyang,d=10& 0.875(S=4)
& 0.182(4)
& 0.102(35)
& 0.397(3)
&  0.180(13)
&  0.132(28)
&  {\bf 0.968(16)}
&   0.535(4)\\ 
 NoisyYinyang,d=100  & 0.875(S=3)
& 0.182(6)
& 0.103(35)
& 0.798(2)
& 0.242(20)
& 0.135(23)
& {\bf 0.999(14)}
&  0.695(4)\\ 
 NoisyYinyang,d=500& 0.875(S=3)
&0.121(10)
&0.107(28)
& 0.783(3)
& 0.209(20)
&0.143(21)
&{\bf 0.999(11)}
&0.539(4)\\ 
 NoisyYinyang,d=1000 &0.875(S=3)
&0.176(7)
&0.111(27)
&0.875(3)
&0.193(28)
&0.149(19)
&{\bf 0.998(10)}
& 0.372(5)\\ 
 \hline
 NoisyMixMickey,d=10 
&{\bf 0.686(S=5)}
&0.119(34)
& 0.093(29)
&0.413(6)
&0.077(39)
&0.157(15)
&0.501(31)
&0.235(5)\\ 
 NoisyMixMickey,d=100
&{\bf 0.700(S=5)}
&0.141(37)
&0.094(29)
&0.358(6)
&0.095(39)
&0.158(16)
&0.506(31)
&0.221(6)
\\ 
 NoisyMixMickey,d=500 
& {\bf 0.697(S=5)}
& 0.095(37)
& 0.091(30)
& 0.359(7)
& 0.098(39)
& 0.155(17)
& 0.502(31)
& 0.232(6)
\\ 
 NoisyMixMickey,d=1000
& {\bf 0.692(S=5)}
&0.122(36)
&0.091(29)
&0.386(6)
&0.104(39)
&0.153(17)
&0.497(31)
&0.241(5)
\\ 
 \hline
 NoisyMixStar,d=10
&{\bf 0.783(S=10)}
&0.109(40)
&0.221(30)
&0.613(11)
&0.140(40)
&0.330(17)
&0.623(31)
&0.476(4)\\ 
 NoisyMixStar,d=100
&{\bf 0.779(S=9)}
&0.129(40)
&0.220(28)
&0.627(10)
&0.171(40)
&0.334(18)
&0.667(30)
&0.487(4)\\ 
 NoisyMixStar,d=500
&{\bf 0.788(S=8)}
&0.115(40)
&0.220(29)
&0.604(9)
&0.158(40)
&0.328(16)
&0.651(30)
&0.498(4)\\ 
 NoisyMixStar,d=1000 
&{\bf 0.791(S=9)}
&0.113(40)
&0.219(29)
&0.599(9)
&0.150(40)
&0.333(15)
&0.621(30)
&0.476(4)\\ 
 \hline
\end{tabular}
\caption{Comparison of the linkage methods across different simulated datasets. All reported values are mediums of $100$ random simulations. For datasets without noisy points, the performance at the true number of clusters is reported ($S=5$ for Yinyang, $S=3$ for Mix Mickey and Mix Star). For datasets with noisy points, we report the best performance across different numbers of clusters and include the number of clusters at which the max is achieved in the bracket.}
\label{tab::linkageComparison}
\end{table}

% \begin{table}[ht]
% % \footnotesize
% \centering
% % \setlength{\tabcolsep}{2pt}
% \begin{tabular}{||c ||c| c| c |c |c|} 
%  \hline
%   & Yinyang & Noisy Yinyang& MixMickey & Noisy MixMickey & Noisy MixStar \\ 
%  \hline
%  average &   & &  & &\\ 
%   \hline
%  single &  & &  & &\\ 
%   \hline
%  complete &  & &  & &\\ 
%   \hline
%  minimax &  & &  & &\\ 
%   \hline
%  centroid  &  & &  & &\\ 
%   \hline
%  mcquitty&  & &  & &\\ 
%   \hline
%  median&  & &  & &\\ 
%   \hline
%  Ward&  & &  & &\\ 
%  \hline
% \end{tabular}
% \caption{Table of the linkage comparisons}
% \label{tab::linkageComparison}
% \end{table}

From Table \ref{tab::linkageComparison}, either average linkage or single linkage
achieve the best and most reliable performance. 
Thus, we recommend using one of them as the linkage criterion. 
We include a more detailed analysis of each dataset in the following subsections and we plot the 5th percentile, medium, and 95th percentile
of the adjusted Rand index for single linkage, average linkage, and complete linkage. Plots comparing all the linkages on the different datasets are deferred to Appendix \ref{sec::simLinkageAll}.

\subsubsection{Yinyang Data} \label{sim::yinyangLinkage}
~~~~We begin by comparing the different linkage methods on the Yinyang datasets with different numbers of noisy dimensions (same data as in Section \ref{sec::YY}). The results are shown in Figure \ref{fig::linkageYinyang}. For each dimension ($d=10,100,500,1000$), the medium adjusted Rand index of the $100$ runs is plotted with the solid line, and the $5$ percentile to $95$ percentile range is depicted with a lighter color band. The true number of clusters $S=5$ is shown as the red dotted vertical line. 

\begin{figure}[ht]
\captionsetup{skip=1pt}
\centering
        \includegraphics[width=0.8\linewidth]{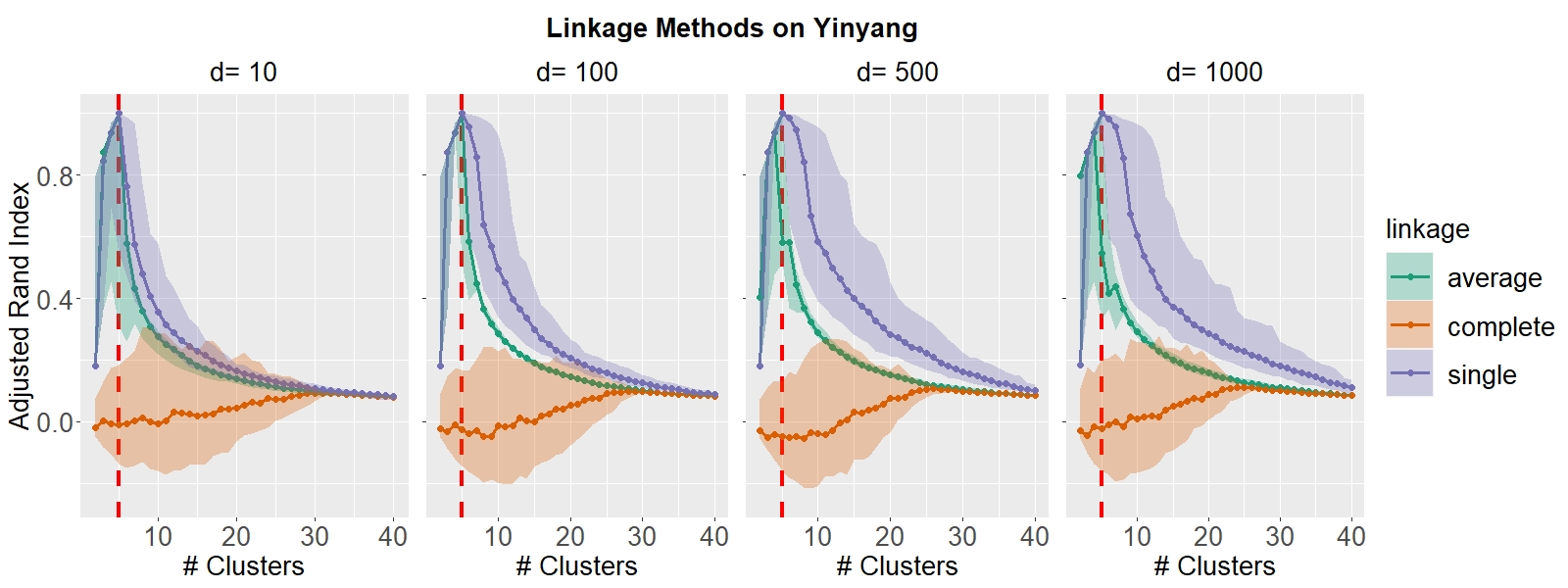} 
\caption{Clustering results with different linkage methods across different numbers of final clusters on Yinyang data. The line is for medium and the band is from 5th percentile to 95th percentile. The vertical red dashed line indicates the true number of $5$ clusters.}
\label{fig::linkageYinyang}
\end{figure}

We observe that single linkage and average linkage have similar performance for lower dimensions $d = 10$ and $d = 100$, with medium performance achieving nearly perfect clustering at the true number of clusters. However, the clustering results returned by single linkage are more stable, having a narrower band while the band of average linkage is much wider. For cases with higher dimensions $d=500,1000$, we observe single linkage still stably achieves nearly perfect clustering at $k=5$, which corroborates our results in Section \ref{sec::YY}, but average linkage fails to get such good clustering performance when dimensions get higher. Therefore, single linkage has superior performance on the Yinyang data, arguably because the true manifold of the data has well-separated clusters that single linkage is suitable for separation.

% \begin{figure}[ht]
% \captionsetup{skip=1pt}
% \centering
%     \begin{subfigure}[t]{0.4\textwidth}
%         \centering
%         \includegraphics[width=\linewidth]{figures/linkageYinyang2_d10.jpg} 
%     \end{subfigure}
%     \begin{subfigure}[t]{0.4\textwidth}
%         \centering
%         \includegraphics[width=\linewidth]{figures/linkageYinyang2_d100.jpg}
%     \end{subfigure}\\
    
%     \begin{subfigure}[t]{0.4\textwidth}
%         \centering
% %        \includegraphics[width=\linewidth]{Yinyang_noise1000_dendro_ave.png}
%         \includegraphics[width=\linewidth]{figures/linkageYinyang2_d500.jpg}
%     \end{subfigure}
%     \begin{subfigure}[t]{0.4\textwidth}
%         \centering
% %        \includegraphics[width=\linewidth]{Yinyang_noise1000_ave.png} 
%         \includegraphics[width=\linewidth]{figures/linkageYinyang2_d1000.jpg}
%     \end{subfigure}    
% \caption{Clustering results with different linkage methods across different numbers of final clusters on Yinyang data. Line for medium and band from 5th percentile to 95th percentile. The vertical red dashed line indicates the true number of $5$ clusters.}
% \label{fig::linkageYinyang}
% \end{figure}

\subsubsection{Noisy Yinyang Data} \label{sim::noisyYinyang}
~~~~
%We assess how the skeleton clustering with different linkage methods can perform on data with additive noises. 
To create additional noise,
we added $640$ (20\% of the number of signals) noisy points to the Yinyang dataset, sampled uniformly from $[-3,3]\times [-3,3] $ in the first two dimensions, with random Gaussian variables in the other dimensions the same way we generated Yinyang data. The adjusted Rand indexes are calculated only for the true signal data points and the results are shown in Figure \ref{fig::linkageNoiseYinyang}.

% \begin{figure}[ht]
% \captionsetup{skip=1pt}
% \centering
%     \begin{subfigure}[t]{0.4\textwidth}
%         \centering
%         \includegraphics[width=\linewidth]{figures/linkageNoiseYinyang2_d10.jpg} 
%     \end{subfigure}
%     \begin{subfigure}[t]{0.4\textwidth}
%         \centering
%         \includegraphics[width=\linewidth]{figures/linkageNoiseYinyang2_d100.jpg}
%     \end{subfigure}\\
    
%     \begin{subfigure}[t]{0.4\textwidth}
%         \centering
% %        \includegraphics[width=\linewidth]{Yinyang_noise1000_dendro_ave.png}
%         \includegraphics[width=\linewidth]{figures/linkageNoiseYinyang2_d500.jpg}
%     \end{subfigure}
%     \begin{subfigure}[t]{0.4\textwidth}
%         \centering
% %        \includegraphics[width=\linewidth]{Yinyang_noise1000_ave.png} 
%         \includegraphics[width=\linewidth]{figures/linkageNoiseYinyang2_d1000.jpg}
%     \end{subfigure}    
% \caption{Clustering results with different linkage methods across different numbers of final clusters on Yinyang data with noisy points. }
% \label{fig::linkageNoiseYinyang}
% \end{figure}

\begin{figure}[ht]
\captionsetup{skip=1pt}
\centering
        \includegraphics[width=0.8\linewidth]{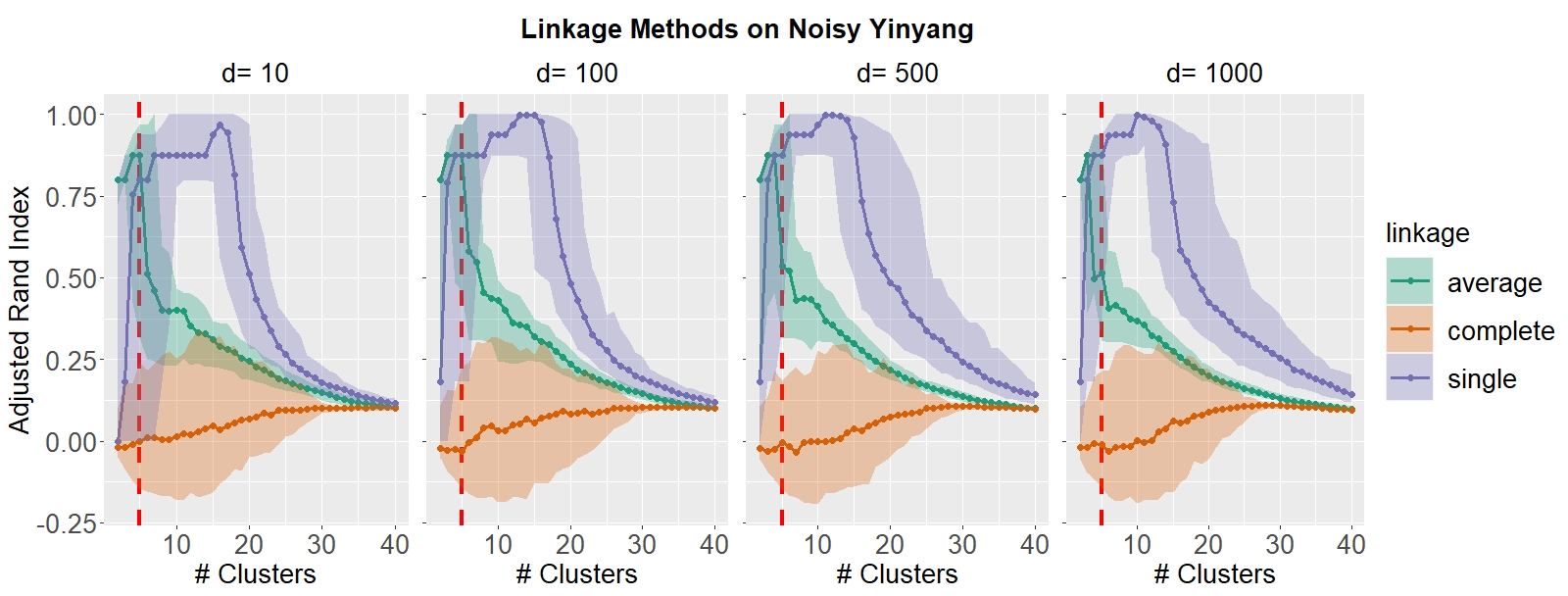} 
\caption{Clustering results with different linkage methods across different numbers of final clusters on Yinyang data with noisy points. The vertical red dashed line indicates the true number of $5$ clusters.}
\label{fig::linkageNoiseYinyang}
\end{figure}

Average linkage can achieve slightly better performance than single linkage around the true number of clusters $S = 5$ for lower dimensions ($d = 10, 100$), but fails to achieve satisfactory clustering performance when dimensionality gets higher ($d=500,1000$). The performance of single linkage improves with $S$ being slightly larger than the actual number $5$ and can yield nearly perfect clusters with $S$ being around 15 to 20. 
	A further investigation reveals that large $S$ will group noisy points into separate clusters and hence improves the clustering performance; see Figure \ref{fig::noiseYinyangVis}. This suggests that our framework may be used for anomaly detection.

\begin{figure}[ht]
\captionsetup{skip=1pt}
\centering
    \begin{subfigure}[t]{0.22\textwidth}
        \centering
        \includegraphics[width=\linewidth]{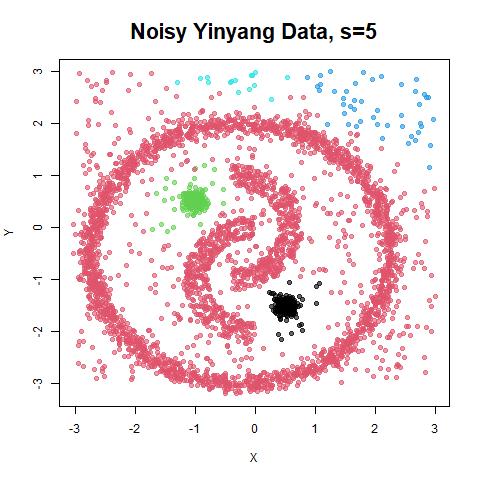} 
    \end{subfigure}
    \begin{subfigure}[t]{0.22\textwidth}
        \centering
        \includegraphics[width=\linewidth]{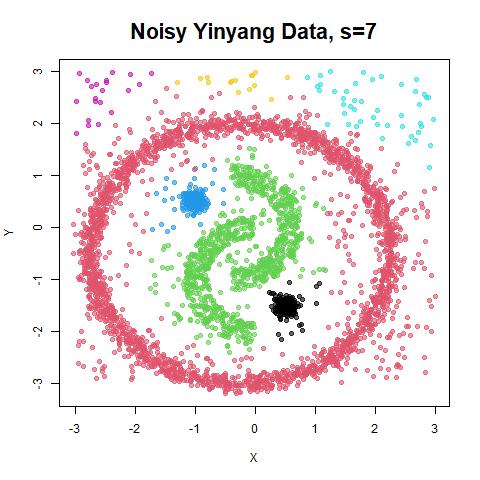}
    \end{subfigure}
    \begin{subfigure}[t]{0.22\textwidth}
        \centering
        \includegraphics[width=\linewidth]{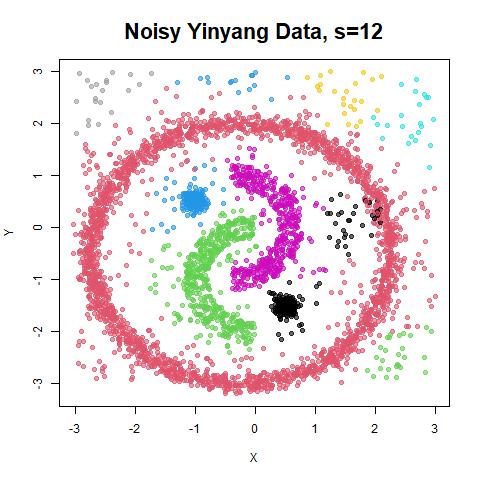}
    \end{subfigure}
    \begin{subfigure}[t]{0.22\textwidth}
        \centering
        \includegraphics[width=\linewidth]{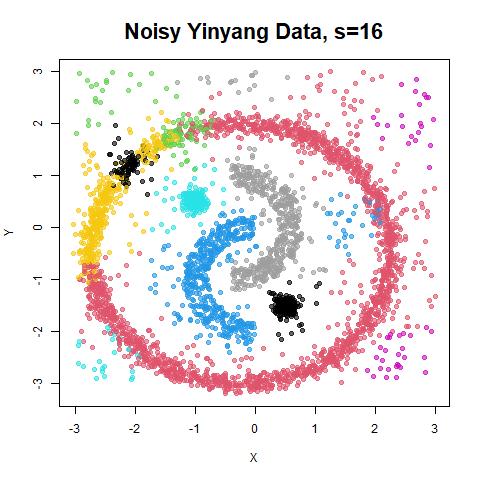}
    \end{subfigure}    
\caption{The clustering results with single linkage in skeleton clustering with different numbers of final clusters $S$ for Noisy Yinyang data, $d=1000$.}
\label{fig::noiseYinyangVis}
\end{figure}

\vspace{-1em}
\subsubsection{Mix Mickey Data} \label{sim::mixMickey}

~~~~The well-separated structures in the Yingyang data may provide advantages to the single linkage. 
To investigate the effect of linkage criteria on the overlapping clusters, we consider 
a three-Gaussian mixture model in 2D case that we call the Mix Mickey data.
\begin{figure}
    \centering
       \includegraphics[width=1.8in]{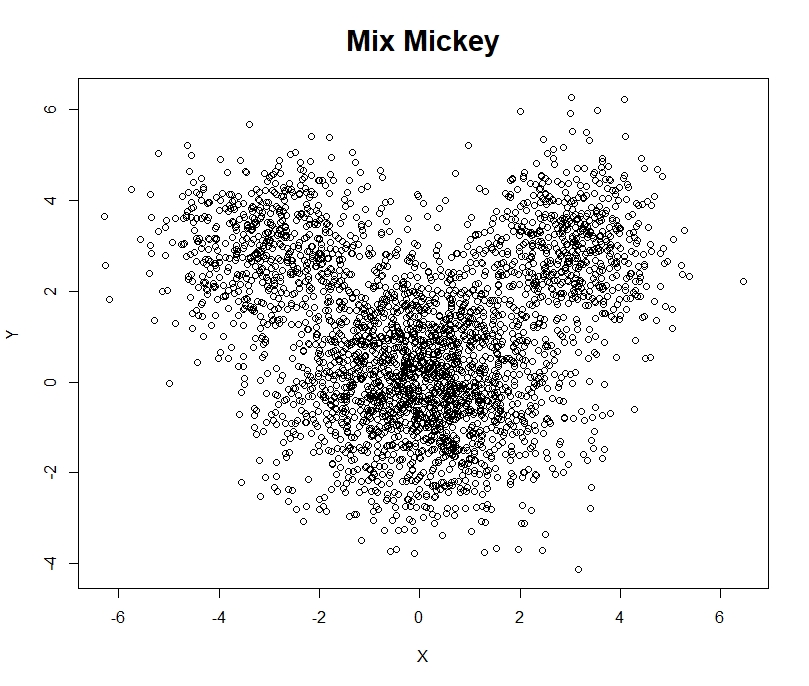}
      \captionsetup{skip=-0.5em}
    \caption{First two dimensions of Mix Mickey data.}
    \label{fig::OC3}
\end{figure}
%In such cases, employing other linkage methods, such as average linkage, can give more reliable clustering results within our skeleton clustering framework.
%In this experiment, we consider a three-Gaussian mixture model in 2D case that we call it the Mix Mickey data.
% the Mix Mickey data consisted of three $2$-dimensional multivariate Gaussian clusters. 
The large cluster is centered at $(0,0)$ with the covariance matrix being a diagonal matrix of $2$ and has $2000$ points. The two smaller clusters are centered at $(3,3)$ and $(-3,3)$ respectively, and both have a covariance matrix being a diagonal matrix of $1$, and each has $600$ points. Random Gaussian variables are added to make the data $d=10,100,500,1000$ dimensions via the same way we generate the Yinyang data.
Figure~\ref{fig::OC3} presents a scatter plot of the first two dimensions; the three clusters have a substantial amount of overlap
so it is difficult for clustering methods to separate them into three distinct clusters.
The results under the same linkages analysis pipeline are shown in Figure \ref{fig::linkageMixMickey}.

\begin{remark}
\small
GMM can be favored in this data example but is unstable and cannot work with too many noisy dimensions. We present some comparisons including GMM in Appendix \ref{sec::withGMM}.
\end{remark}

% \begin{figure}[ht]
% \captionsetup{skip=1pt}
% \centering
%     \begin{subfigure}[t]{0.4\textwidth}
%         \centering
%         \includegraphics[width=\linewidth]{figures/linkageMixMickey2_d10.jpg} 
%     \end{subfigure}
%     \begin{subfigure}[t]{0.4\textwidth}
%         \centering
%         \includegraphics[width=\linewidth]{figures/linkageMixMickey2_d100.jpg}
%     \end{subfigure}\\
    
%     \begin{subfigure}[t]{0.4\textwidth}
%         \centering
%         \includegraphics[width=\linewidth]{figures/linkageMixMickey2_d500.jpg}
%     \end{subfigure}
%     \begin{subfigure}[t]{0.4\textwidth}
%         \centering
%         \includegraphics[width=\linewidth]{figures/linkageMixMickey2_d1000.jpg}
%     \end{subfigure}    
% \caption{Clustering results with different linkage methods across different numbers of final clusters on Mix Mickey data. The vertical red dashed line indicates the true number of $3$ clusters.}
% \label{fig::linkageMixMickey}
% \end{figure}

\begin{figure}[ht]
\captionsetup{skip=1pt}
\centering
        \includegraphics[width=0.8\linewidth]{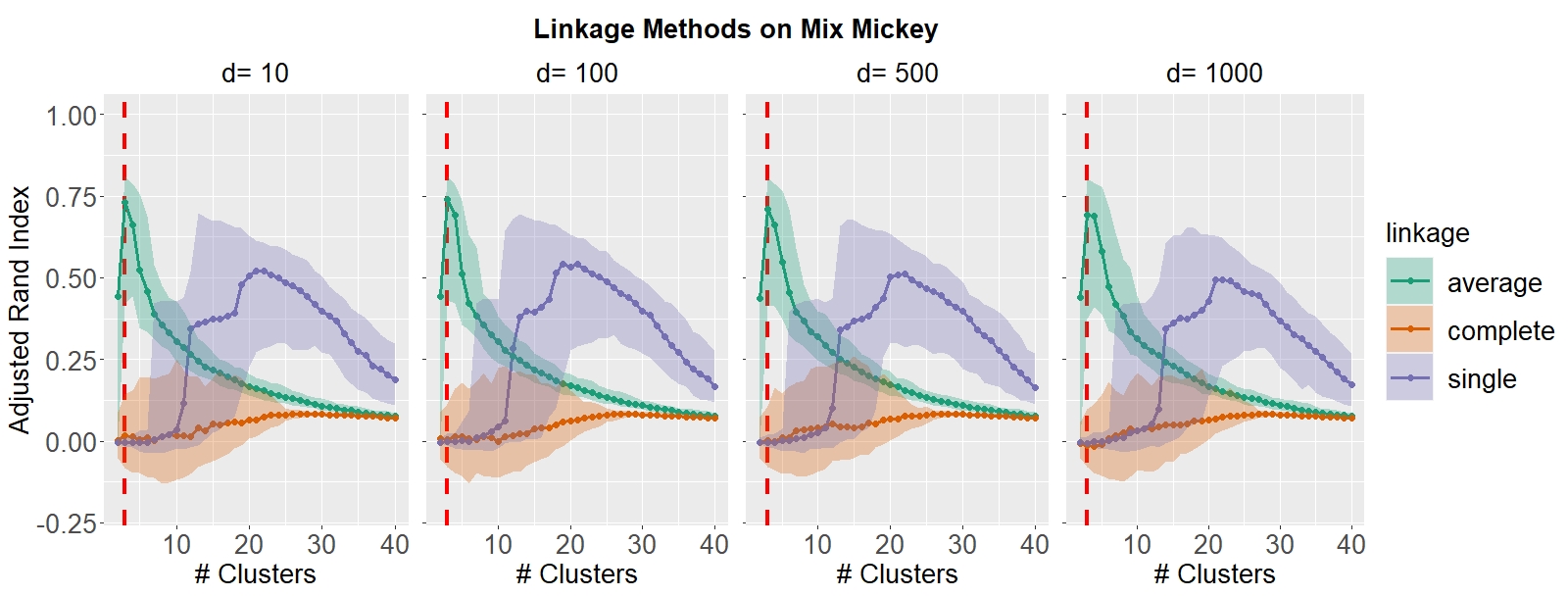} 
\caption{Clustering results with different linkage methods across different numbers of final clusters on Mix Mickey data. The vertical red dashed line indicates the true number of $3$ clusters.}
\label{fig::linkageMixMickey}
\end{figure}

We observe that average linkage gives good performance at $S = 3$ (the true number of clusters) and single linkage
fails to give a satisfying performance under this scenario, giving non-informative clusters at low $S$ (only extracting small clusters) and too fragmented clusters at high $S$.
The average linkage is a criterion that tends to create spherical clusters with similar sizes and hence is better suited for this simulated data.
Thus, our experiment shows that, for data containing overlapping clusters with roughly spherical shapes, the average linkage criterion 
in the knots segmentation step is preferred.
%, and our skeleton clustering framework is still capable of dealing with such scenarios.

\subsubsection{Noisy Mix Mickey Data} \label{sim::noisyMixMickey}

~~~~In this section, we experiment with a scenario with both overlapping clusters and noisy observations. We added $640$ (20\% of the number of signals) noisy points to the Mix Mickey dataset, sampled uniformly from $[-6,6]\times [-5,6] $ in the first two dimensions, with random Gaussian noises in the other dimensions the same way as in Mix Mickey data. The adjusted Rand indices are measured only on the true signal data points with the results shown in Figure \ref{fig::linkageNoiseMixMickey}.

% \begin{figure}[ht]
% \captionsetup{skip=1pt}
% \centering
%     \begin{subfigure}[t]{0.4\textwidth}
%         \centering
%         \includegraphics[width=\linewidth]{figures/linkageNoiseMixMickey2_d10.jpg} 
%     \end{subfigure}
%     \begin{subfigure}[t]{0.4\textwidth}
%         \centering
%         \includegraphics[width=\linewidth]{figures/linkageNoiseMixMickey2_d100.jpg}
%     \end{subfigure}\\
    
%     \begin{subfigure}[t]{0.4\textwidth}
%         \centering
% %        \includegraphics[width=\linewidth]{Yinyang_noise1000_dendro_ave.png}
%         \includegraphics[width=\linewidth]{figures/linkageNoiseMixMickey2_d500.jpg}
%     \end{subfigure}
%     \begin{subfigure}[t]{0.4\textwidth}
%         \centering
% %        \includegraphics[width=\linewidth]{Yinyang_noise1000_ave.png} 
%         \includegraphics[width=\linewidth]{figures/linkageNoiseMixMickey2_d1000.jpg}
%     \end{subfigure}    
% \caption{Clustering results with different linkage methods across different numbers of final clusters on Mix Mickey data with Noise. }
% \label{fig::linkageNoiseMixMickey}
% \end{figure}
\begin{figure}[ht]
\captionsetup{skip=1pt}
\centering
        \includegraphics[width=0.8\linewidth]{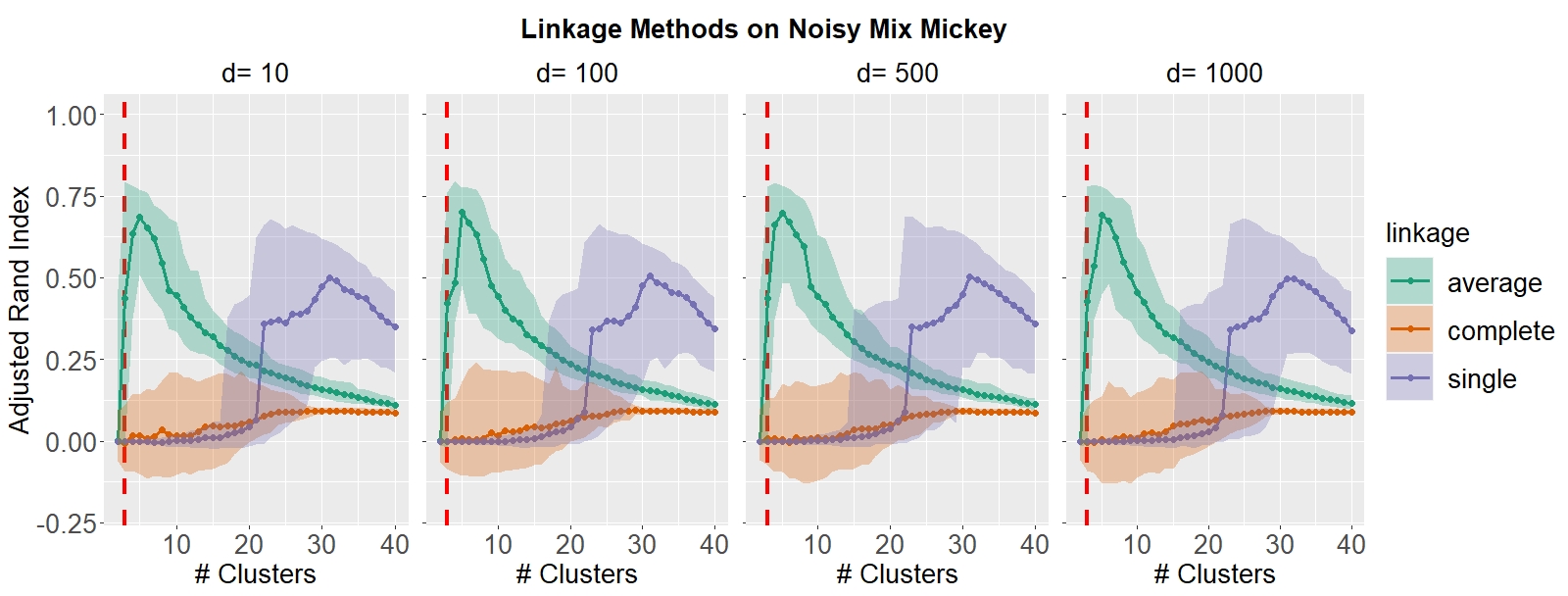} 
\caption{Clustering results with different linkage methods across different numbers of final clusters on Mix Mickey data with Noise. The vertical red dashed line indicates the true number of $3$ clusters.}
\label{fig::linkageNoiseMixMickey}
\end{figure}

Average linkage still gives good performance and is superior to the single linkage, which fails to give reasonable clustering performance under a decent number of clusters. Notably, average linkage achieves the best performance with the $S$ being slightly higher than $3$ due to the introduction of noisy data points.

\subsubsection{Mix Star Data} \label{sim::mixStar}

~~~~We present here the Mix Star dataset, another 3-GMM data but with a more elongated shape  as illustrated in Figure \ref{fig::star}.
%~~~~ In this experiment, we consider a three-Gaussian mixture model in 2D case that we call it the Mix Star data.
The three clusters are all generated as 2D Gaussian with $5$ and $0.3$ on the diagonal of the covariance matrix with respective centers at $(4,0)$, $(-4,0)$, and $(0,-4)$, and then are rotated to get a star-like shape. Each cluster has $1000$ sample points, and random Gaussian variables with standard deviation $0.1$ are added to make the data $d=10,100,500,1000$ dimensions. There is still a decent overlap among clusters, but each cluster is more distinct compared to Mix Mickey. 
\begin{figure}[ht]
\centering
\includegraphics[width=3.5cm]{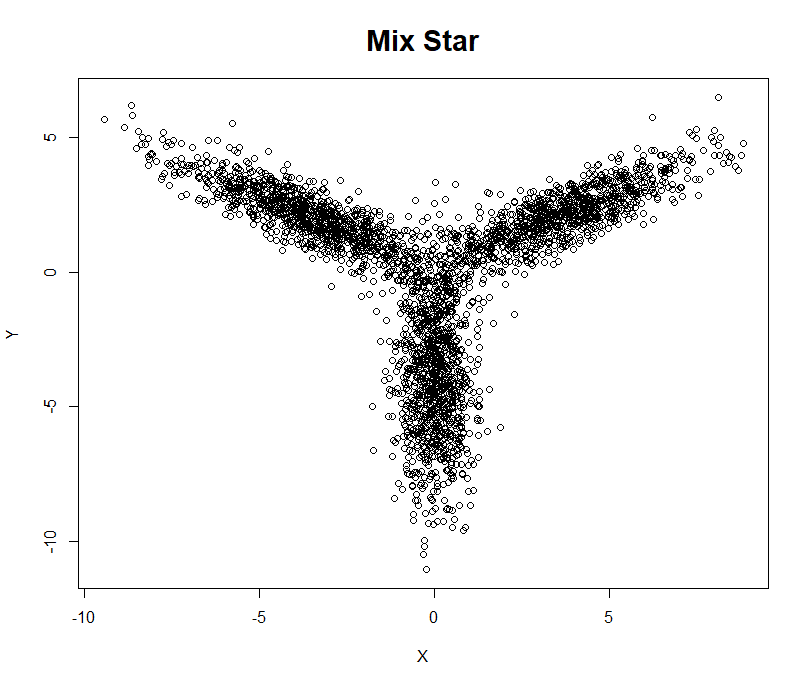}
\caption{First two dimensions of the Mix Star data. }
\label{fig::star}
\end{figure}
We apply the same analysis pipeline as the Yinyang and Mix Mickey data and compare different linkage criteria. 
Figure \ref{fig::linkageMixStarAll2}  displays the median clustering performance. Again, we see that average linkage has the best performance.

\begin{figure}[ht]
\captionsetup{skip=1pt}
\centering
        \includegraphics[width=0.8\linewidth]{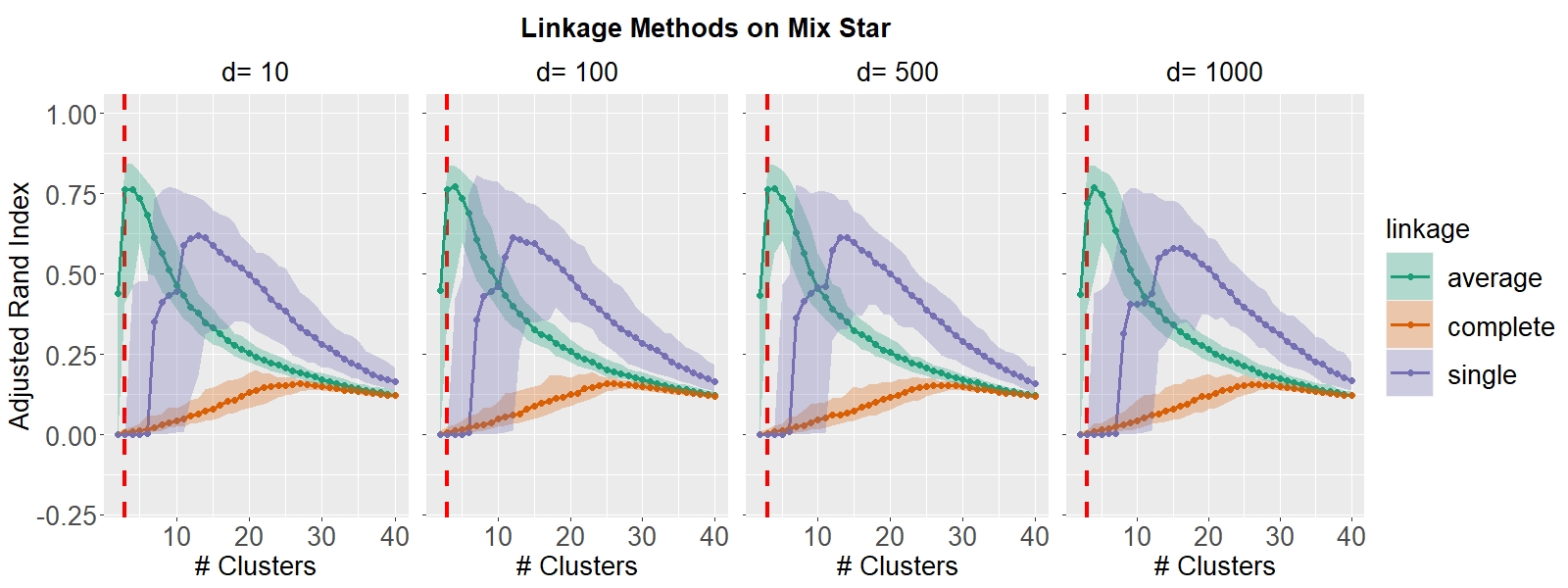} 
\caption{Clustering results with different linkage methods across different numbers of final clusters on Mix Star data. The vertical red dashed line indicates the true number of $3$ clusters.}
\label{fig::linkageMixStarAll2}
\end{figure}

%An exploratory analysis by applying different linkage methods to the skeleton constructed with Voronoi density on $d=1000$ Mix Star data is shown in Figure \ref{fig::star}. 
%We see that average linkage performs the best.

% \begin{figure}[ht]
% \centering
% \includegraphics[width=3.5cm]{figures/Mixstar.png}
% \includegraphics[width=3.5cm]{figures/Mixstar_ave.png}
% \includegraphics[width=3.5cm]{figures/Mixstar_single.png}
% \includegraphics[width=3.5cm]{figures/Mixstar_complete.png}
% \includegraphics[width=3.5cm]{figures/Mixstar_wpgma.png}
% \includegraphics[width=3.5cm]{figures/Mixstar_median.png}
% \includegraphics[width=3.5cm]{figures/Mixstar_ward.png}
% \includegraphics[width=3.5cm]{figures/Mixstar_minimax.png}
% \caption{Comparing linkage criteria in segmentation on the Mix Star data, d = 1000. }
% \label{fig::starAll}
% \end{figure}

% {\color{magenta}YC: not sure if we need the 2D scatetrplot for the mix star; it shows a similar result as Figure~\ref{fig::MixMickey2} so I remove it.}

\subsubsection{Noisy Mix Star}
\label{sim::noisymixStar}

~~~~To investigate the effect of added noises,
we make the data similar to the Noisy Mix Mickey by adding $600$ (20\% of the number of signals) noisy points to the Mix Star dataset, sampled uniformly from $[-10,10]\times [-10,5] $ in the first two dimensions, with random Gaussian noises in the other dimensions generated the same way. The results of linkage comparison results are shown in Figure \ref{fig::linkageNoiseMixStar2}. Average linkage still gives the best clustering results in this scenario.

\begin{figure}
\captionsetup{skip=1pt}
\centering
        \includegraphics[width=0.8\linewidth]{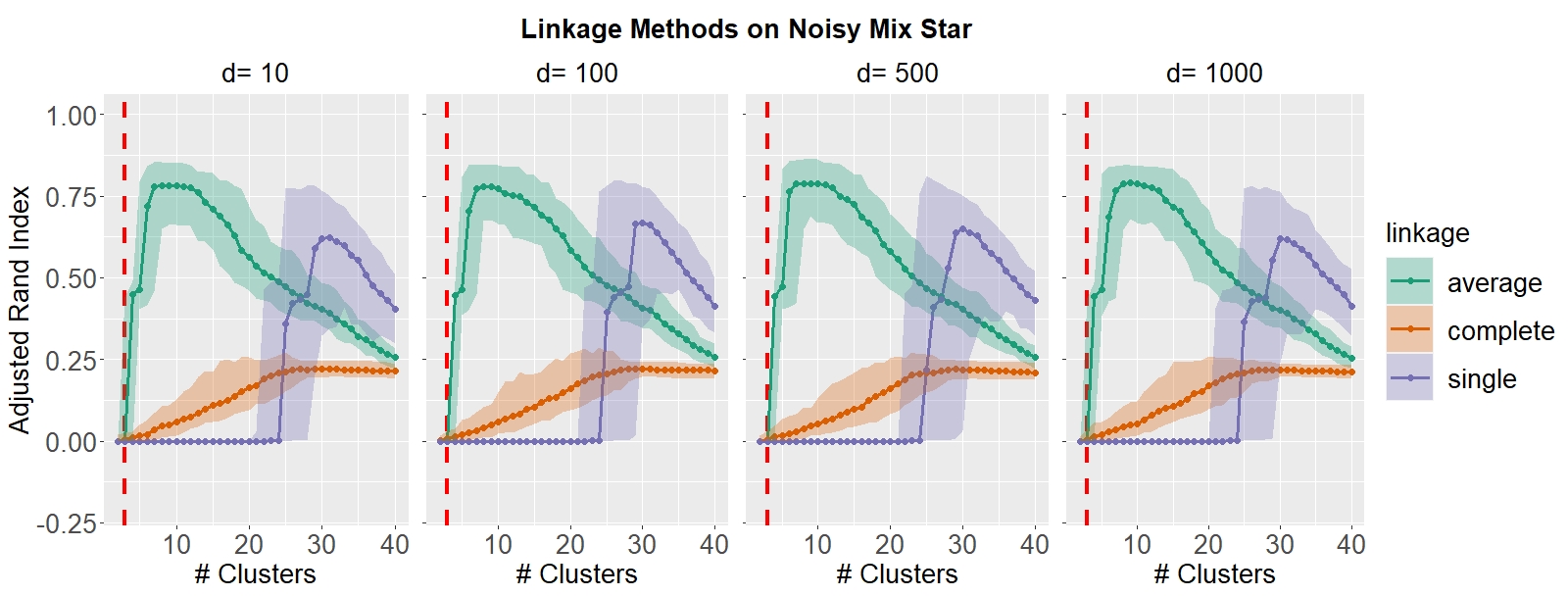} 
\caption{Clustering results with different linkage methods across different numbers of final clusters on Mix Star data with Noise. }
\label{fig::linkageNoiseMixStar2}
\end{figure}

% \begin{figure}[ht]
% \captionsetup{skip=1pt}
% \centering
%     \begin{subfigure}[t]{0.4\textwidth}
%         \centering
%         \includegraphics[width=\linewidth]{figures/linkageNoiseMixStar2_d10.jpg} 
%     \end{subfigure}
%     \begin{subfigure}[t]{0.4\textwidth}
%         \centering
%         \includegraphics[width=\linewidth]{figures/linkageNoiseMixStar2_d100.jpg}
%     \end{subfigure}\\
    
%     \begin{subfigure}[t]{0.4\textwidth}
%         \centering
% %        \includegraphics[width=\linewidth]{Yinyang_noise1000_dendro_ave.png}
%         \includegraphics[width=\linewidth]{figures/linkageNoiseMixStar2_d500.jpg}
%     \end{subfigure}
%     \begin{subfigure}[t]{0.4\textwidth}
%         \centering
% %        \includegraphics[width=\linewidth]{Yinyang_noise1000_ave.png} 
%         \includegraphics[width=\linewidth]{figures/linkageNoiseMixStar2_d1000.jpg}
%     \end{subfigure}    
% \caption{Clustering results with different linkage methods across different numbers of final clusters on Mix Star data with Noise. }
% \label{fig::linkageNoiseMixStar}
% \end{figure}

% \begin{figure}[ht]
% \captionsetup{skip=1pt}
% \centering
%         \includegraphics[width=\linewidth]{figures/linkageNoiseMixStarAll2.jpeg} 
% \caption{Clustering results with different linkage methods across different numbers of final clusters on Mix Star data with Noise. }
% \label{fig::linkageNoiseMixStar}
% \end{figure}

In summary, as illustrated by all the simulations in this section, our skeleton clustering framework is able to handle noisy data points by tuning the number of final clusters and can cope with overlapping clusters by choosing an appropriate linkage criterion for skeleton segmentation. Broadly speaking, the appropriate choice of linkage method depends on the intrinsic geometric structure of the data and may require subject matter knowledge or exploratory analysis. Specifically, if the intrinsic clusters are well-separated, single linkage is preferred as it gives clear cuts for disjoint components. But if the clusters are believed to have some degree of overlapping with each cluster approximately spherically shaped, average linkage criterion can lead to better performance.

%------------------------------------------------------------------------------------

\subsubsection{All Linkage Comparisons} \label{sec::simLinkageAll}

%\subsubsection{Yinyang} \label{sim::yinyangLinkageAll}

~~~~Figures~\ref{fig::linkageYinyangAll} and \ref{fig::linkageNoiseYinyangAll}
display the median clustering performances of all linkage methods under different numbers of clusters using  Yinyang and noisy Yinyang data. 
We see that average linkage and single linkage dominate all other methods, while single linkage is superior in those two cases.

%~~~~The additional results for different linkage methods on Yinyang data are shown in Figure \ref{fig::linkageYinyangAll}. 

% \begin{figure}[ht]
% \captionsetup{skip=1pt}
% \centering
%     \begin{subfigure}[t]{0.4\textwidth}
%         \centering
%         \includegraphics[width=\linewidth]{figures/linkageYinyang_d10.jpg} 
%     \end{subfigure}
%     \begin{subfigure}[t]{0.4\textwidth}
%         \centering
%         \includegraphics[width=\linewidth]{figures/linkageYinyang_d100.jpg}
%     \end{subfigure}\\
    
%     \begin{subfigure}[t]{0.4\textwidth}
%         \centering
% %        \includegraphics[width=\linewidth]{Yinyang_noise1000_dendro_ave.png}
%         \includegraphics[width=\linewidth]{figures/linkageYinyang_d500.jpg}
%     \end{subfigure}
%     \begin{subfigure}[t]{0.4\textwidth}
%         \centering
% %        \includegraphics[width=\linewidth]{Yinyang_noise1000_ave.png} 
%         \includegraphics[width=\linewidth]{figures/linkageYinyang_d1000.jpg}
%     \end{subfigure}    
% \caption{Clustering results with different linkage methods across different numbers of final clusters on Yinyang Data. }
% \label{fig::linkageYinyangAll}
% \end{figure}

\begin{figure}[ht]
\captionsetup{skip=1pt}
\centering
        \includegraphics[width=\linewidth]{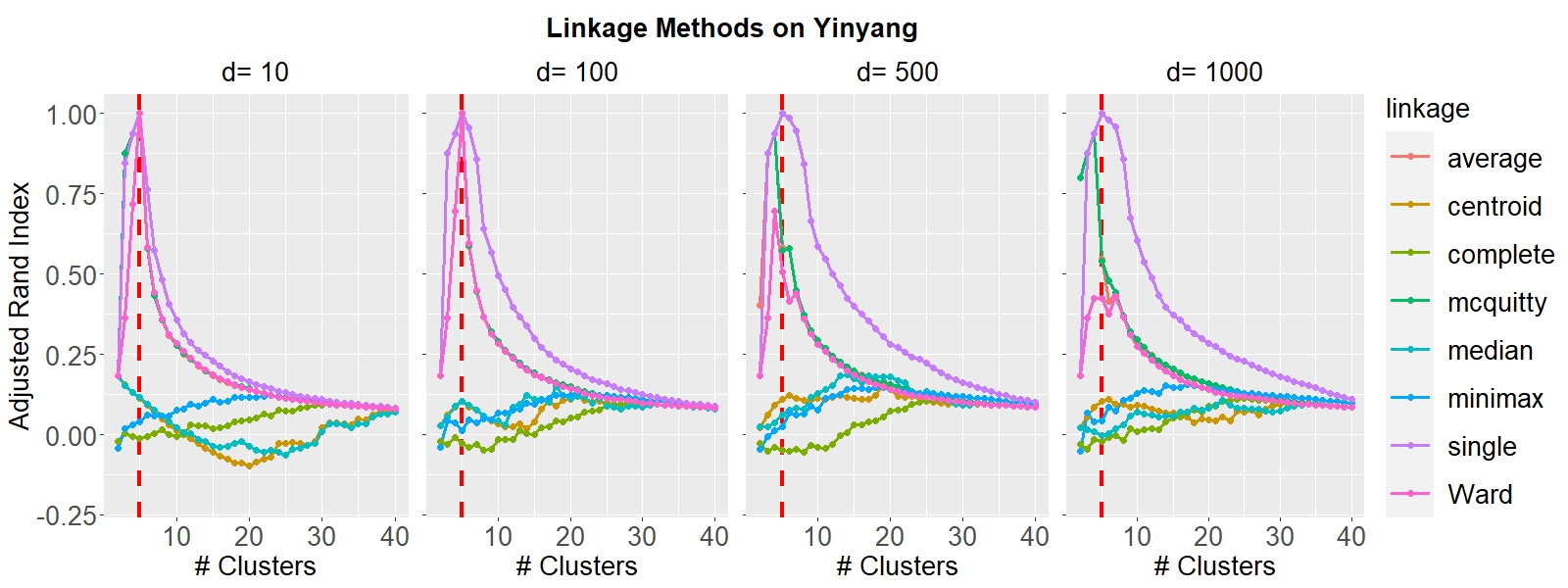}
\caption{Clustering results with different linkage methods across different numbers of final clusters on Yinyang Data. }
\label{fig::linkageYinyangAll}
\end{figure}

\begin{figure}[ht]
\captionsetup{skip=1pt}
\centering
        \includegraphics[width=\linewidth]{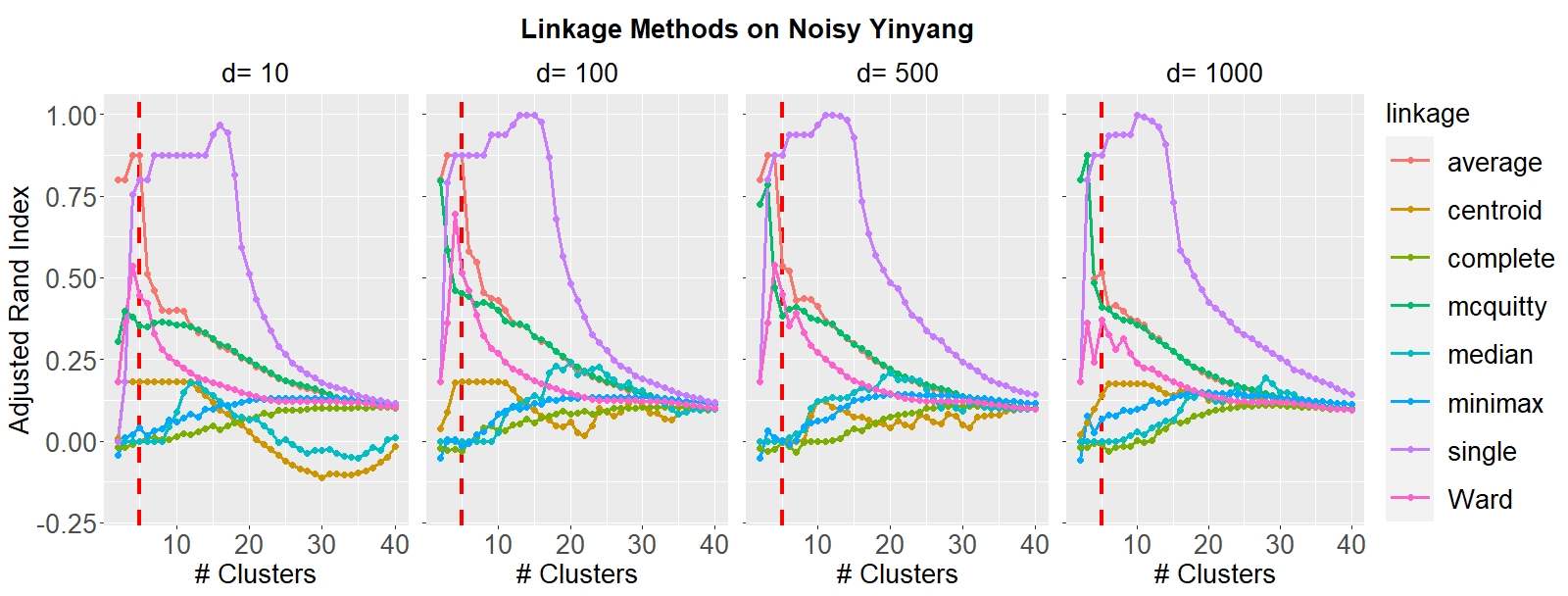}
\caption{Clustering results with different linkage methods across different numbers of final clusters on Noisy Yinyang Data. }
\label{fig::linkageNoiseYinyangAll}
\end{figure}

%\subsubsection{Noisy Yinyang} \label{sim::noisyYinyangAll}
%~~~~The additional results for different linkage methods on Yinyang data with noisy points are shown in Figure \ref{fig::linkageNoiseYinyangAll}. 

% \begin{figure}[ht]
% \captionsetup{skip=1pt}
% \centering
%     \begin{subfigure}[t]{0.4\textwidth}
%         \centering
%         \includegraphics[width=\linewidth]{figures/linkageNoiseYinyang_d10.jpg} 
%     \end{subfigure}
%     \begin{subfigure}[t]{0.4\textwidth}
%         \centering
%         \includegraphics[width=\linewidth]{figures/linkageNoiseYinyang_d100.jpg}
%     \end{subfigure}\\
    
%     \begin{subfigure}[t]{0.4\textwidth}
%         \centering
% %        \includegraphics[width=\linewidth]{Yinyang_noise1000_dendro_ave.png}
%         \includegraphics[width=\linewidth]{figures/linkageNoiseYinyang_d500.jpg}
%     \end{subfigure}
%     \begin{subfigure}[t]{0.4\textwidth}
%         \centering
% %        \includegraphics[width=\linewidth]{Yinyang_noise1000_ave.png} 
%         \includegraphics[width=\linewidth]{figures/linkageNoiseYinyang_d1000.jpg}
%     \end{subfigure}    
% \caption{Clustering results with different linkage methods across different numbers of final clusters on Yinyang data with Noise. }
% \label{fig::linkageNoiseYinyangAll}
% \end{figure}

Figures~\ref{fig::linkageMixMickeyAll} and \ref{fig::linkageNoiseMixMickeyAll}
present the median clustering performance under different numbers of clusters for the Mix Mickey and noisy Mix Mickey data
(same setup in Section \ref{sec::simLinkage}). 
Similar to the case of Yinyang data, we observe that average linkage and single linkage dominate all other methods, while average linkage is superior among the two.

To further investigate what the clusters will be like in high dimensions,
we present 2D scatterplot of clustering results under $S=3$ (final number of clusters is 3)
of the first two coordinates in Figure~\ref{fig::MixMickey2}. 
We use the data with $d=1000$ and color the clusters using red, green, and blue.
Clearly, average linkage successfully recovers the actual clusters while other methods fail to recover. 
Note that single linkage does not perform well because clusters are overlapping with each other.

%\subsubsection{Mix Mickey} \label{sim::mixMickeyAll}
%~~~~The additional results for different linkage methods on Mix Mickey data are shown in Figure \ref{fig::linkageMixMickeyAll}. 
% \begin{figure}[ht]
% \captionsetup{skip=1pt}
% \centering
%     \begin{subfigure}[t]{0.4\textwidth}
%         \centering
%         \includegraphics[width=\linewidth]{figures/linkageMixMickey_d10.jpg} 
%     \end{subfigure}
%     \begin{subfigure}[t]{0.4\textwidth}
%         \centering
%         \includegraphics[width=\linewidth]{figures/linkageMixMickey_d100.jpg}
%     \end{subfigure}\\
%     \begin{subfigure}[t]{0.4\textwidth}
%         \centering
%         \includegraphics[width=\linewidth]{figures/linkageMixMickey_d500.jpg}
%     \end{subfigure}
%     \begin{subfigure}[t]{0.4\textwidth}
%         \centering
%         \includegraphics[width=\linewidth]{figures/linkageMixMickey_d1000.jpg}
%     \end{subfigure}    
% \caption{Clustering results with different linkage methods across different numbers of final clusters on Mix Mickey data. }
% \label{fig::linkageMixMickeyAll}
% \end{figure}
\begin{figure}
\captionsetup{skip=1pt}
\centering
        \includegraphics[width=\linewidth]{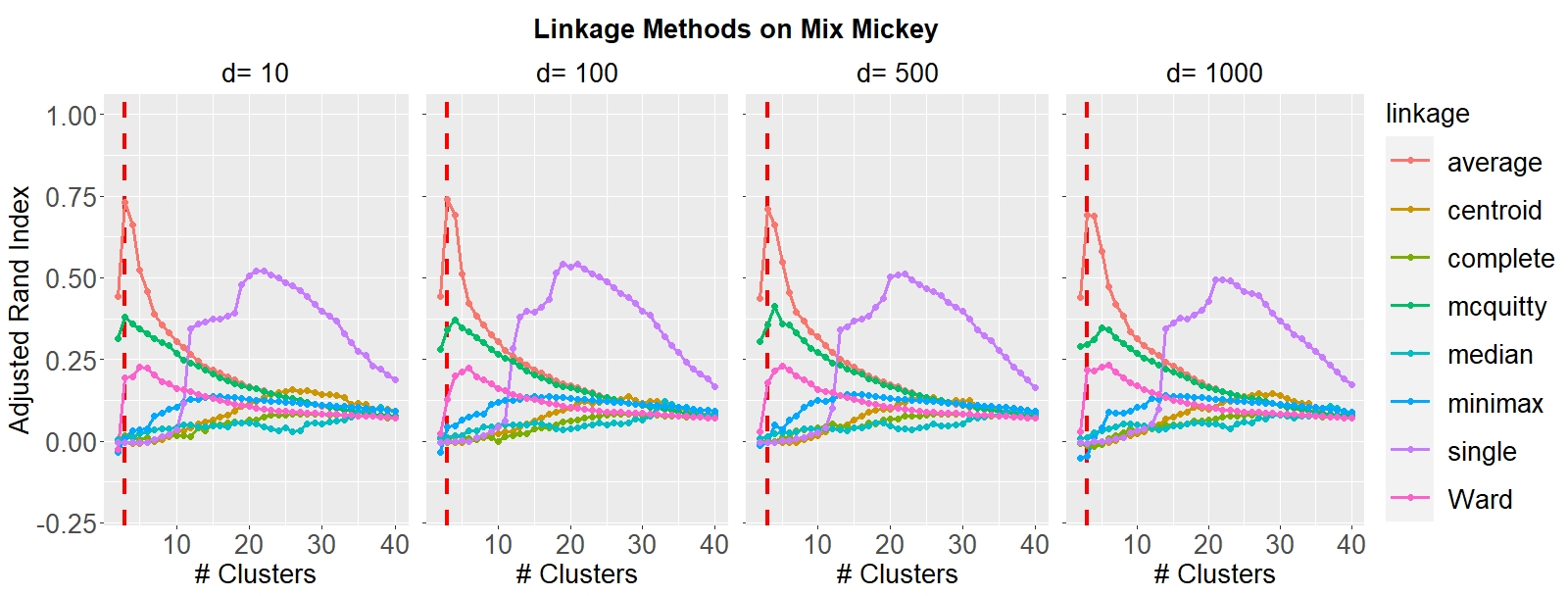} 
\caption{Clustering results with different linkage methods across different numbers of final clusters on Mix Mickey data. }
\label{fig::linkageMixMickeyAll}
\end{figure}

\begin{figure}
\captionsetup{skip=1pt}
\centering
        \includegraphics[width=\linewidth]{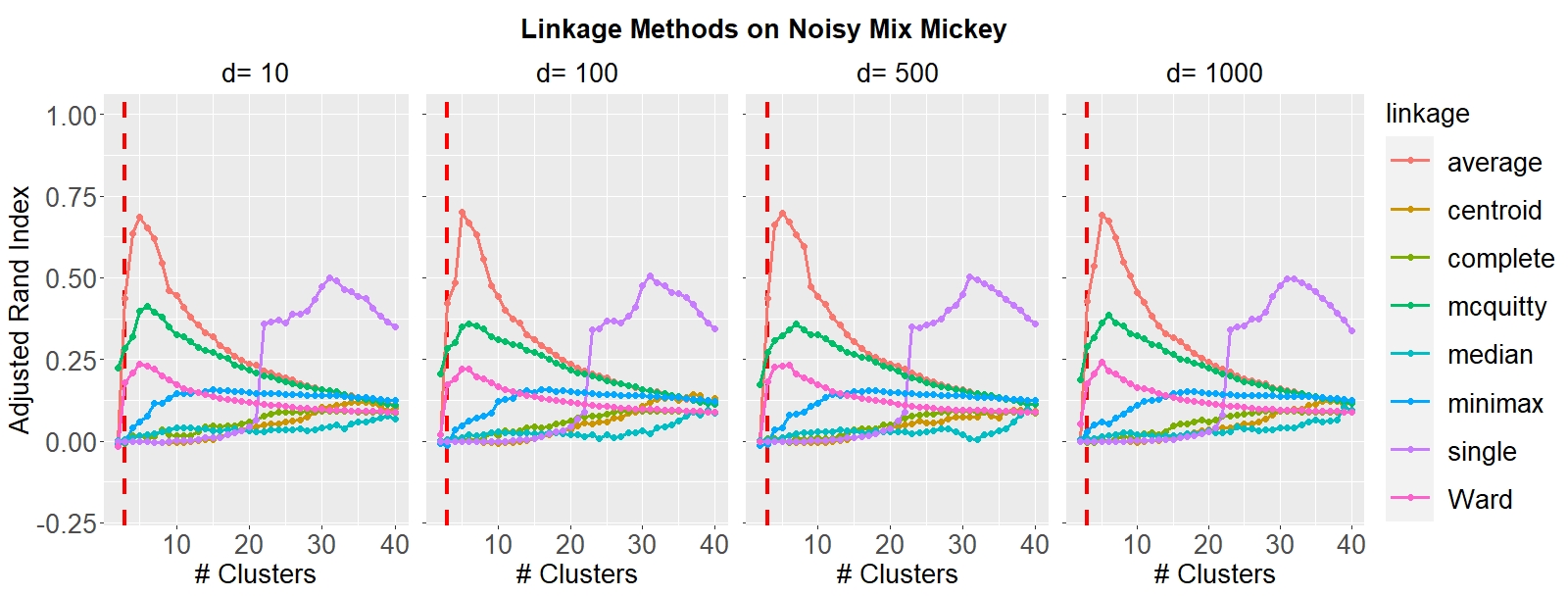} 
\caption{Clustering results with different linkage methods across different numbers of final clusters on Mix Mickey data with Noise. }
\label{fig::linkageNoiseMixMickeyAll}
\end{figure}

%\begin{figure}[ht]
%    \centering
%       \includegraphics[width=2.5in]{figures/NoisyMixMickey.jpeg}
%    \caption{First two dimensions of Noisy Mix Mickey data.}
%    \label{fig::NoiseMixMickey}
%\end{figure}

%Additionally, we visualization the $3$ clusters given by different linkage methods in Figure \ref{fig::MixMickey2}

\begin{figure}
\centering
\includegraphics[width=3.5cm]{figures/MixMickey.jpeg}
\includegraphics[width=3.5cm]{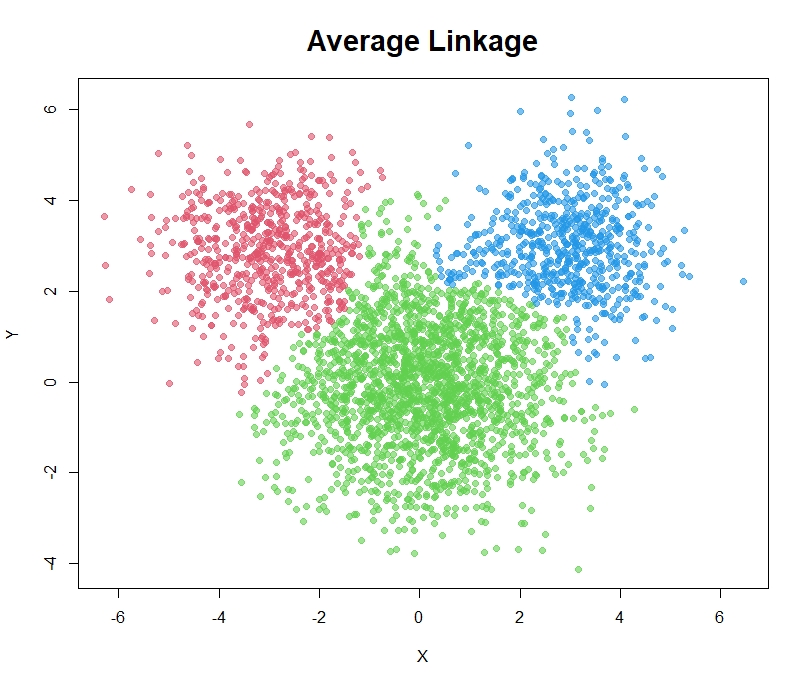}
\includegraphics[width=3.5cm]{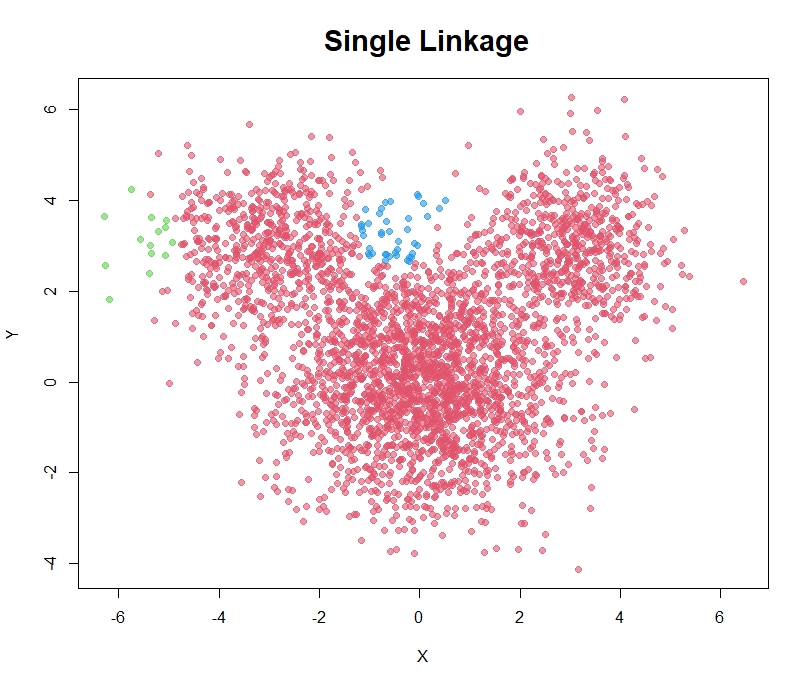}
\includegraphics[width=3.5cm]{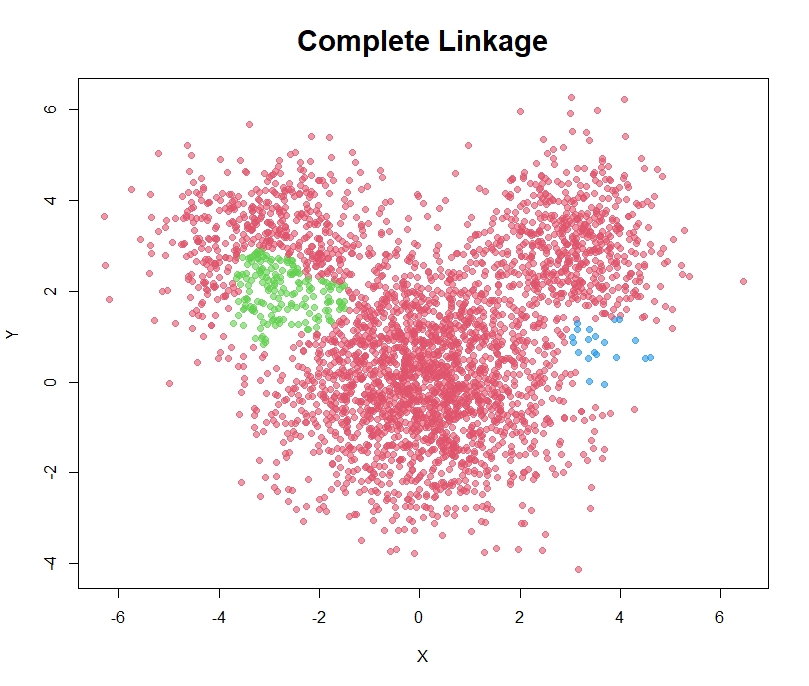}
\includegraphics[width=3.5cm]{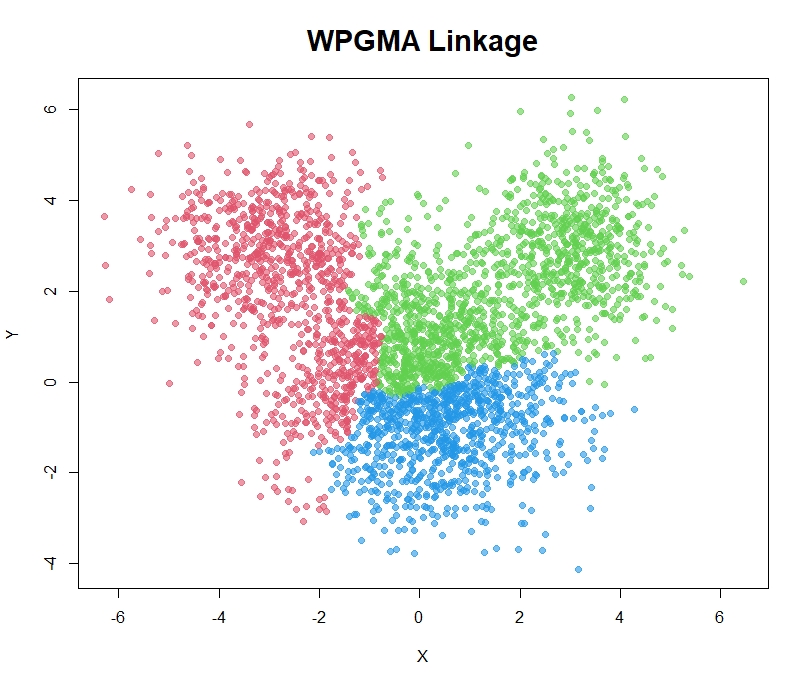}
\includegraphics[width=3.5cm]{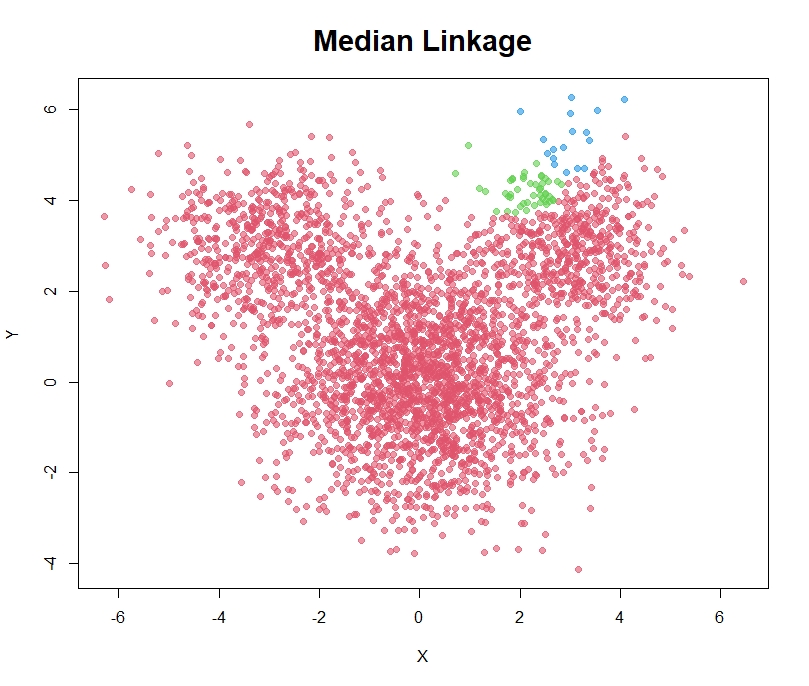}
\includegraphics[width=3.5cm]{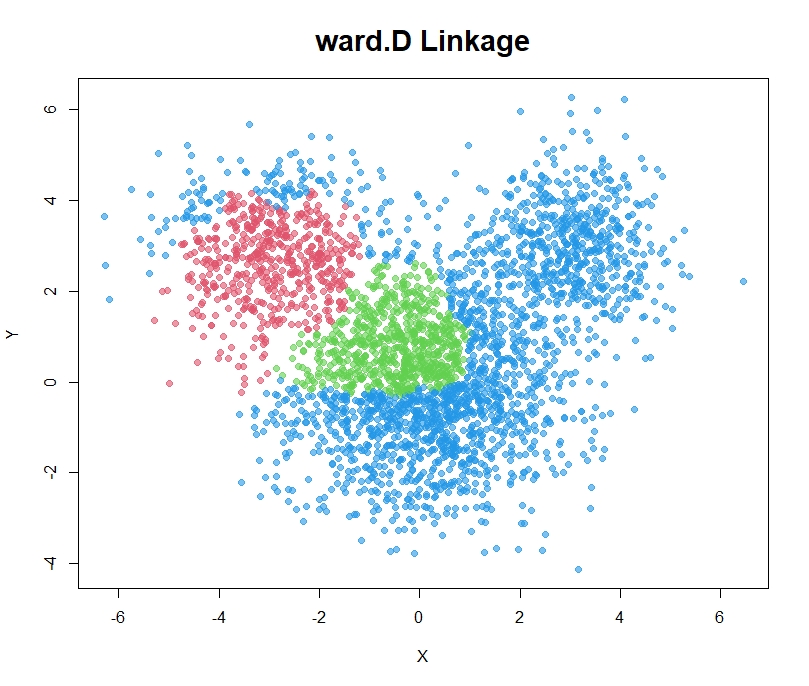}
\includegraphics[width=3.5cm]{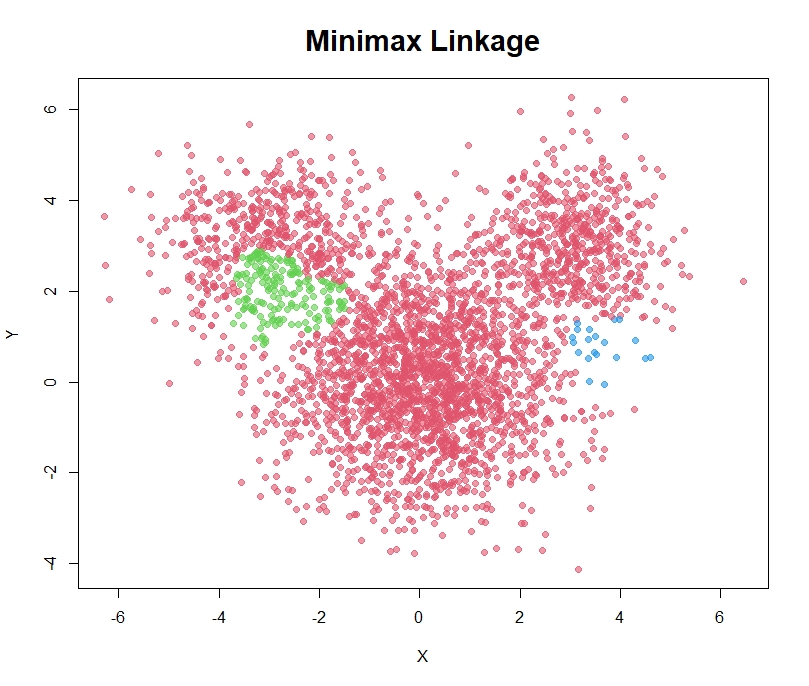}
\caption{
Comparing linkage criteria in segmentation on the Mix Mickey data, $d = 1000$.}
\label{fig::MixMickey2}
\end{figure}

%\subsubsection{Noise Mix Mickey} \label{sim::noisyMixMickeyAll}
%~~~~The first 2 dimensions of Mix Mickey data with noisy points is visualized in Figure \ref{fig::NoiseMixMickey}. The additional results for different linkage methods on Noisy Mix Mickey data are shown in Figure \ref{fig::linkageNoiseMixMickeyAll}. 

% \begin{figure}[ht]
% \captionsetup{skip=1pt}
% \centering
%     \begin{subfigure}[t]{0.4\textwidth}
%         \centering
%         \includegraphics[width=\linewidth]{figures/linkageNoiseMixMickey_d10.jpg} 
%     \end{subfigure}
%     \begin{subfigure}[t]{0.4\textwidth}
%         \centering
%         \includegraphics[width=\linewidth]{figures/linkageNoiseMixMickey_d100.jpg}
%     \end{subfigure}\\
    
%     \begin{subfigure}[t]{0.4\textwidth}
%         \centering
%         \includegraphics[width=\linewidth]{figures/linkageNoiseMixMickey_d500.jpg}
%     \end{subfigure}
%     \begin{subfigure}[t]{0.4\textwidth}
%         \centering
%         \includegraphics[width=\linewidth]{figures/linkageNoiseMixMickey_d1000.jpg}
%     \end{subfigure}    
% \caption{Clustering results with different linkage methods across different numbers of final clusters on Mix Mickey data with Noise. }
% \label{fig::linkageNoiseMixMickeyAll}
% \end{figure}

%\begin{figure}[ht]
%\captionsetup{skip=1pt}
%\centering
%        \includegraphics[width=\linewidth]{figures/linkageNoiseMixMickeyAll.jpeg} 
%\caption{Clustering results with different linkage methods across different numbers of final clusters on Mix Mickey data with Noise. }
%\label{fig::linkageNoiseMixMickeyAll}
%\end{figure}

Figures~\ref{fig::linkageMixStarAll} and \ref{fig::linkageNoiseMixStarAll}
present the median clustering performance under different numbers of clusters for the Mix Star and noisy Mix Star data. 
We observe that average linkage and single linkage dominate all other methods.

\begin{figure}
\captionsetup{skip=1pt}
\centering
        \includegraphics[width=\linewidth]{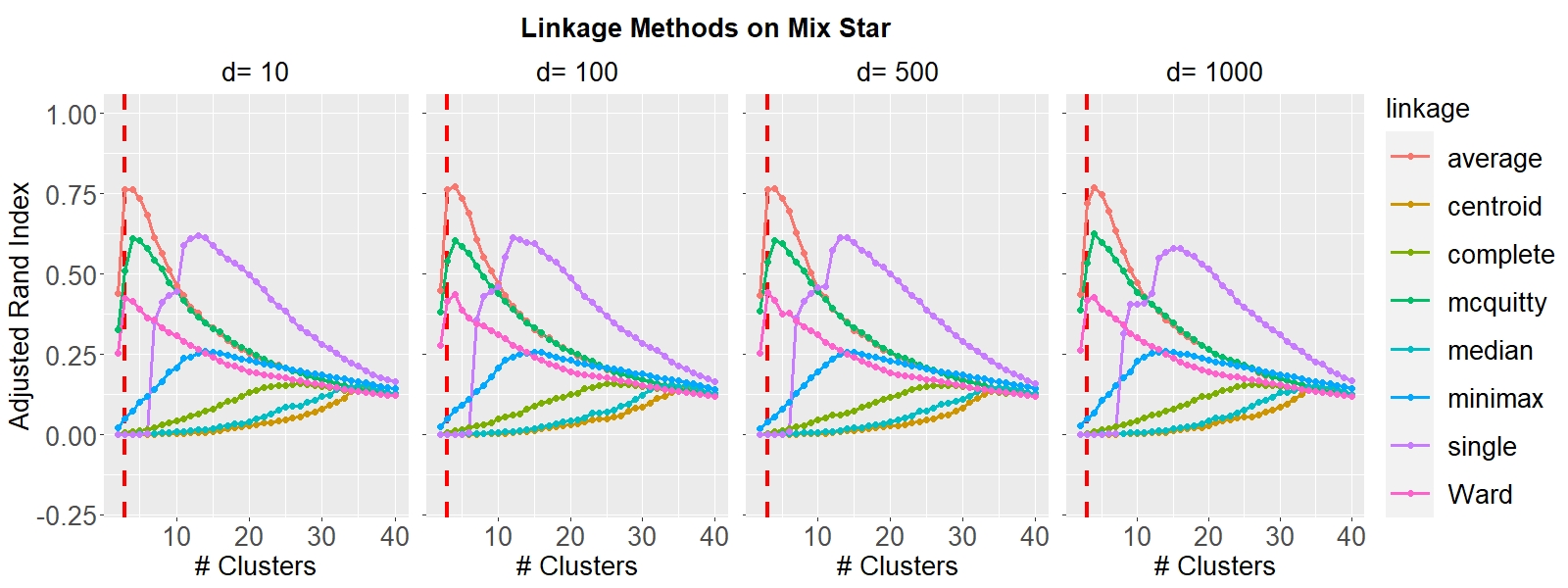} 
\caption{Clustering results with different linkage methods across different numbers of final clusters on Mix Star data. }
\label{fig::linkageMixStarAll}
\end{figure}

\begin{figure}
\captionsetup{skip=1pt}
\centering
        \includegraphics[width=\linewidth]{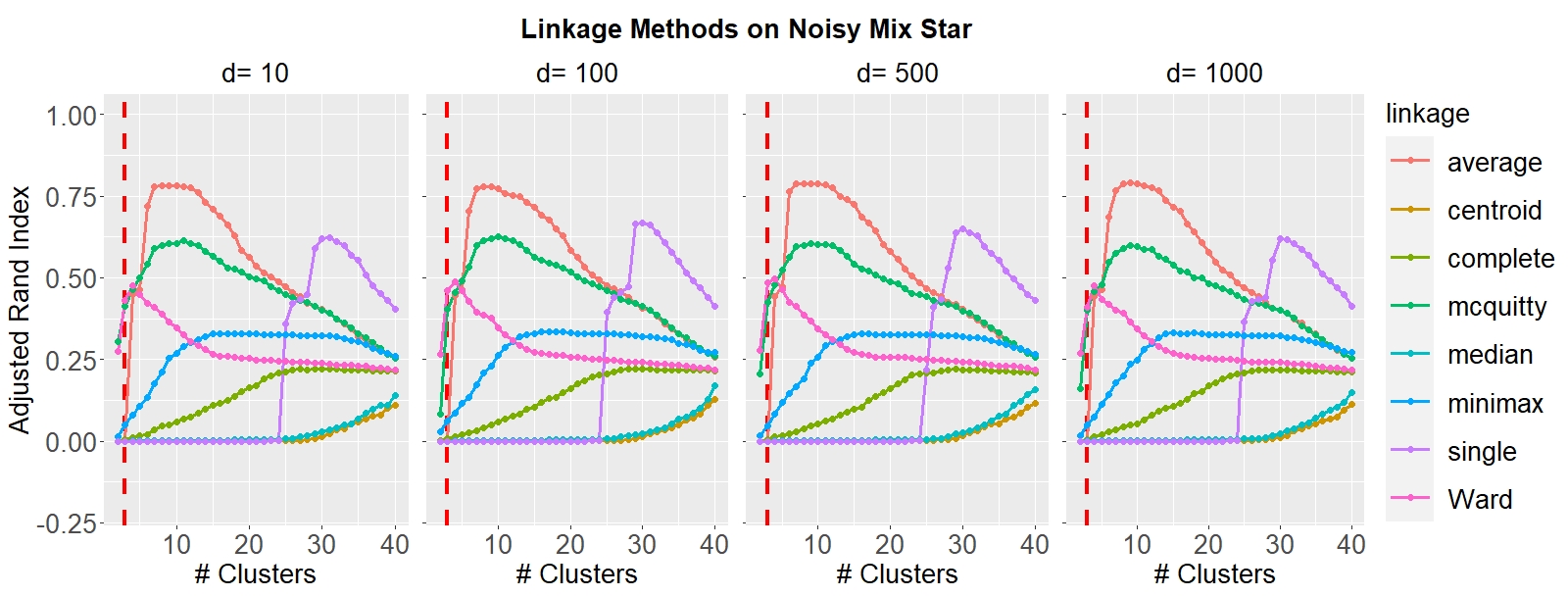} 
\caption{Clustering results with different linkage methods across different numbers of final clusters on Mix Star data with Noise. }
\label{fig::linkageNoiseMixStarAll}
\end{figure}

} %end coloring

\subsection{Additional Data Analysis}	\label{sec::additional}

\subsubsection{Performance with Different Number of Knots}
\label{sec::KnotSize}

\begin{figure}[ht]
\centering
\includegraphics[width=6cm]{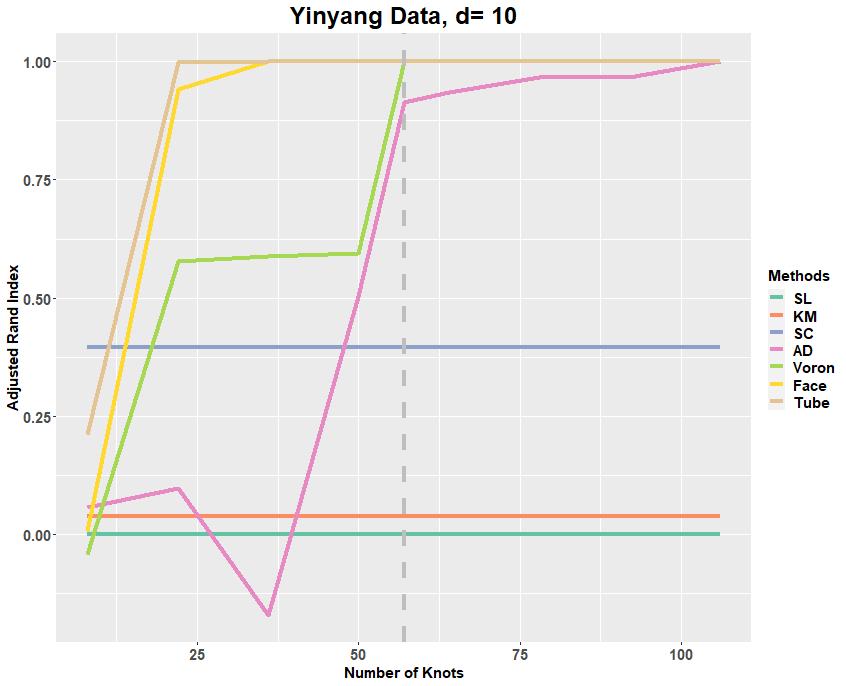}
\includegraphics[width=6cm]{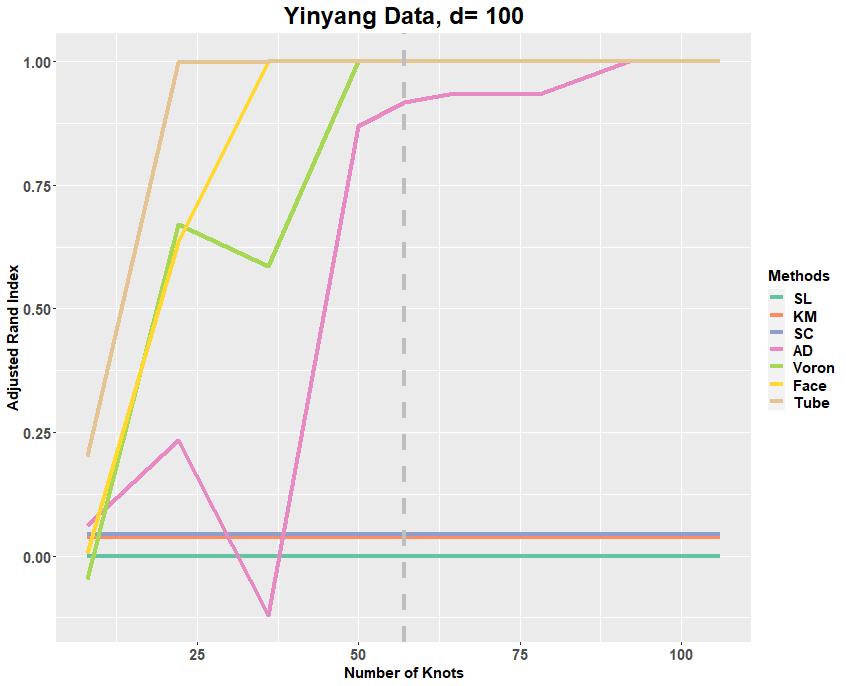}\\
\includegraphics[width=6cm]{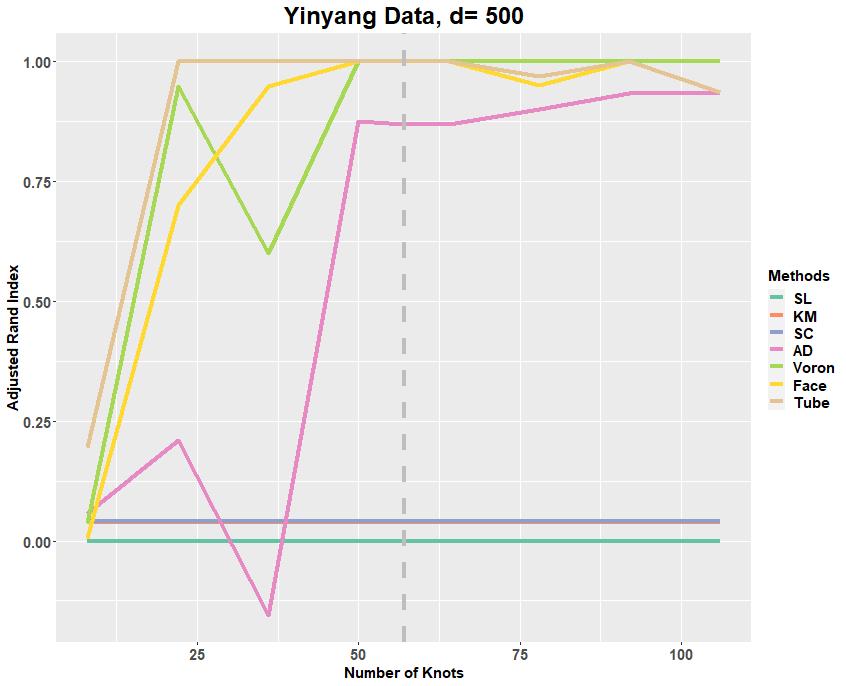}
\includegraphics[width=6cm]{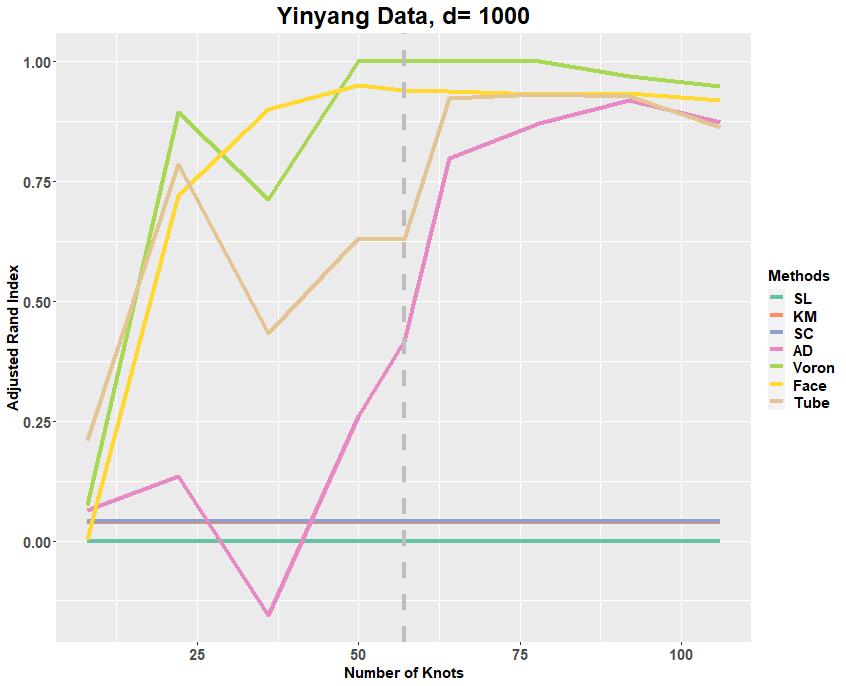}
\caption{Adjusted Rand indexes of different clustering methods against different numbers of knots on $100$ simulated Yinyang data.}
\label{fig::kvaryrands}
\end{figure}

~~~~We analyze how the number of knots would affect the performance of the skeleton clustering. We empirically test the effect of the number of knots, $k$, on the final clustering performance on Yinyang data with dimensions $10,100,500$ and $1000$. For each dimension, we simulated the Yinyang data $100$ times, and for each simulated data we carried out the default skeleton clustering procedure with single linkage and different $k$ (other steps the same as in Section \ref{sec::YY}). 
Figure \ref{fig::kvaryrands} displays the median adjusted Rand index given by each method across different $k$, where the reference rule with $k=57$ is marked by the vertical dash line.
%The skeleton clustering works well in all scenarios compared to other method.
We see that as long as $k$ is sufficiently large, 
skeleton clustering works well. 

% In Figure \ref{fig::KnotSize2}, we show 
% the knot-size diagrams under different numbers of knots.
% This is based on the Yingyang data with $d=200$. 
%how the numbers of points in the resulting Voronoi cells change with different number of knots. We applied $k$-means clustering with different $k$ to the simulated Yinyang dataset with dimension to be $200$. 
% Note that our reference rule $k=\sqrt{n}$ takes $57$ knots with the knot-size diagram shown in Figure \ref{fig::ClusSize}. 
%We note that the cell sizes become balanced when $k$ is around our reference rule (when $k \geq 50$) and stay stable with $k$ goes larger. 
%Therefore our reference rule setting  $k = [\sqrt{n}]$ leads to Voronoi cells with balanced sizes.

% \begin{figure}[ht]
% \centering
% \includegraphics[width=3.5cm]{figures/Yinyang_ClusDistr_k8.jpg}
% \includegraphics[width=3.5cm]{figures/Yinyang_ClusDistr_k22.jpg}
% \includegraphics[width=3.5cm]{figures/Yinyang_ClusDistr_k36.jpg}
% \includegraphics[width=3.5cm]{figures/Yinyang_ClusDistr_k50.jpg}
% \includegraphics[width=3.5cm]{figures/Yinyang_ClusDistr_k64.jpg}
% \includegraphics[width=3.5cm]{figures/Yinyang_ClusDistr_k78.jpg}
% \includegraphics[width=3.5cm]{figures/Yinyang_ClusDistr_k92.jpg}
% \includegraphics[width=3.5cm]{figures/Yinyang_ClusDistr_k106.jpg}
% \caption{The knot-size diagram with different $k$ on Yinyang data with dimension $200$.}
% \label{fig::KnotSize2}
% \end{figure}

%The median adjusted rand indexes of each method with the specific number of knots on the $100$ simulated data with the given dimension are plotted in Figure \ref{fig::kvaryrands}. The reference rule with $k=57$ is marked by the vertical dash line. We observe that our reference rule is empirically sufficient to give a good performance for VD, FD, and TD.

\subsubsection{Self-Organizing Map} 	\label{sec::SOM}
~~~~The Self-Organizing Map (SOM)
is another popular prototype clustering method
and can be used as an alternative to $k$-means clustering
in finding knots.
Thus, here we conduct a simple experiment to
examine the performance of using SOM to find knots.
We examine the performance using Yingyang data with $d=10$ to $d= 1000$.
The identical procedure as in Section~\ref{sec::YY} is applied except that the knots are now detected by the SOM rather than overfitting $k$-means.
The total number of grid points in the SOM is the total number of knots we obtain and, to be comparable to $k$-means with $k=\sqrt{n}$ knots, we used $\lceil n^{1/4} \rceil$ breaks for each dimension of the SOM grid, giving a total of $\lceil n^{1/4} \rceil^2$ initial grid points. However, the SOM may return knots with tiny sample sizes, on which the density-aided similarity measures cannot be calculated. Therefore, we remove knots with less than $3$ data points and use the remaining ones for skeleton construction. 

\begin{figure}
\centering
\includegraphics[height=1.5in]{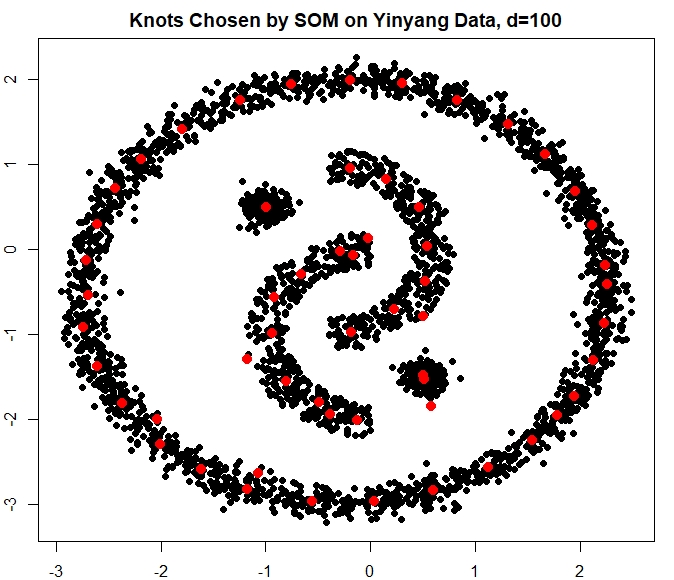}
\includegraphics[height=1.5in]{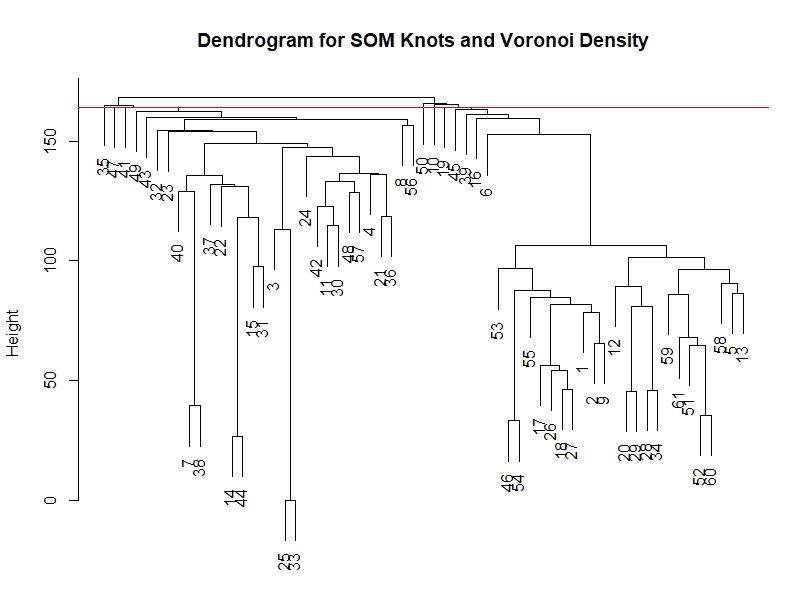}\\
%\caption{Knots chosen by SOM on Yinyang Data}
%\end{figure}
%
%The Rand index using SOM for knots construction and using different bone weights are illustrated in Figure 17. 
%
%\begin{figure}
%\centering
\includegraphics[height=3.5cm]{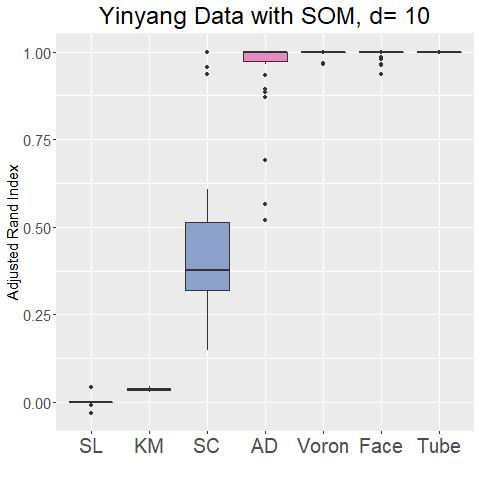}
\includegraphics[height=3.5cm]{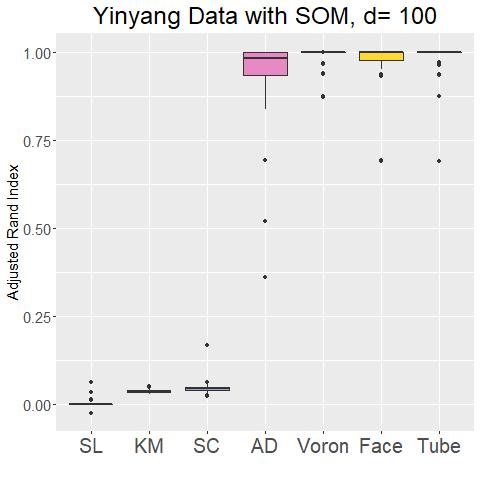}
\includegraphics[height=3.5cm]{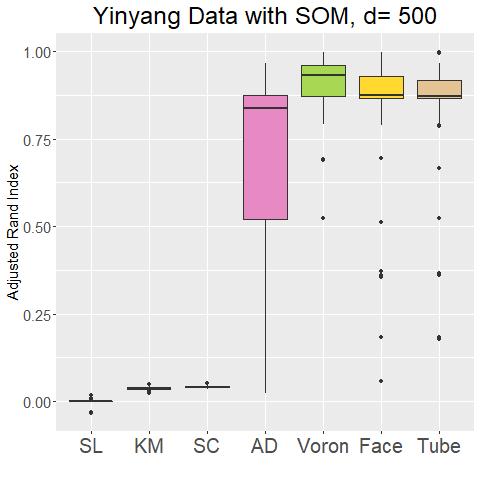}
\includegraphics[height=3.5cm]{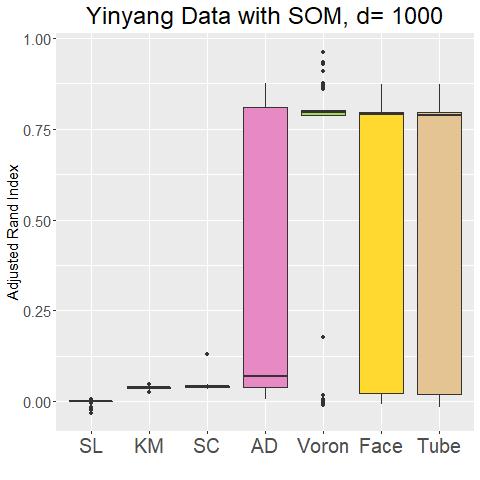}
\caption{Adjusted Rand indexes using SOM for knots selection on Yinyang data.}
\label{fig::SOM}
\end{figure}

Figure~\ref{fig::SOM} summarizes the result.
The top left panel shows the knots from the SOM (after post-processing), which are located around the main data structures and are representative of the original data as well. The dendrogram shows the cluster structure of the SOM knots using Voronoi density on one $100$-dimensional Yinyang data.
In the bottom row, we display the adjusted Rand indices from the clustering methods. 
Compared to the results of Figure~\ref{fig::yingyang1}, the adjusted Rand indices given by the skeleton clustering with SOM knots are similarly good when the dimension is not so high ($d=10$ and $100$). But when the data dimension becomes high ($d=500,1000$), knots constructed by SOM lead to worse clustering results. Therefore, overfitting $k$-means is favored in this work.
Another limitation of SOM is that we need to perform some post-processing to remove tiny knots;
in the case of k-means, we do not need such a procedure.

\subsubsection{Bandwidth Selection Yinyang Data}	\label{sec::bandwidth}

~~~~The estimations of the FD and the TD involve the use of the projected kernel density estimation, for which the type of kernel and the bandwidth need to be specified. 
Similar to the usual KDE, the kernel function does not affect the final performance much, so by default
we use the Gaussian kernel in all of our empirical studies.
It is worth noting that using the uniform kernel can save some computation since it has compact support, but empirically we find using the Gaussian kernel leads to better final clustering results.
In what follows, we focus on the bandwidth selection. 

%For kernel type, both Gaussian kernel and uniform (rectangular) kernel are tested. Note that for a uniform kernel, the Face density and Tube density reduces to counting the points around the face (disk) region. Empirically, both Gaussian kernel and uniform kernel gives good clustering performance for the Face density, but for Tube density, uniform kernel gives significantly worse results than using Gaussian kernel. Therefore, Gaussian kernel is used by default in this work, but if only Face density is of interest, then using a uniform kernel can save some computation. 

\begin{figure}
\centering
    \begin{subfigure}[t]{\textwidth}
        \centering
        \includegraphics[width=0.7\linewidth]{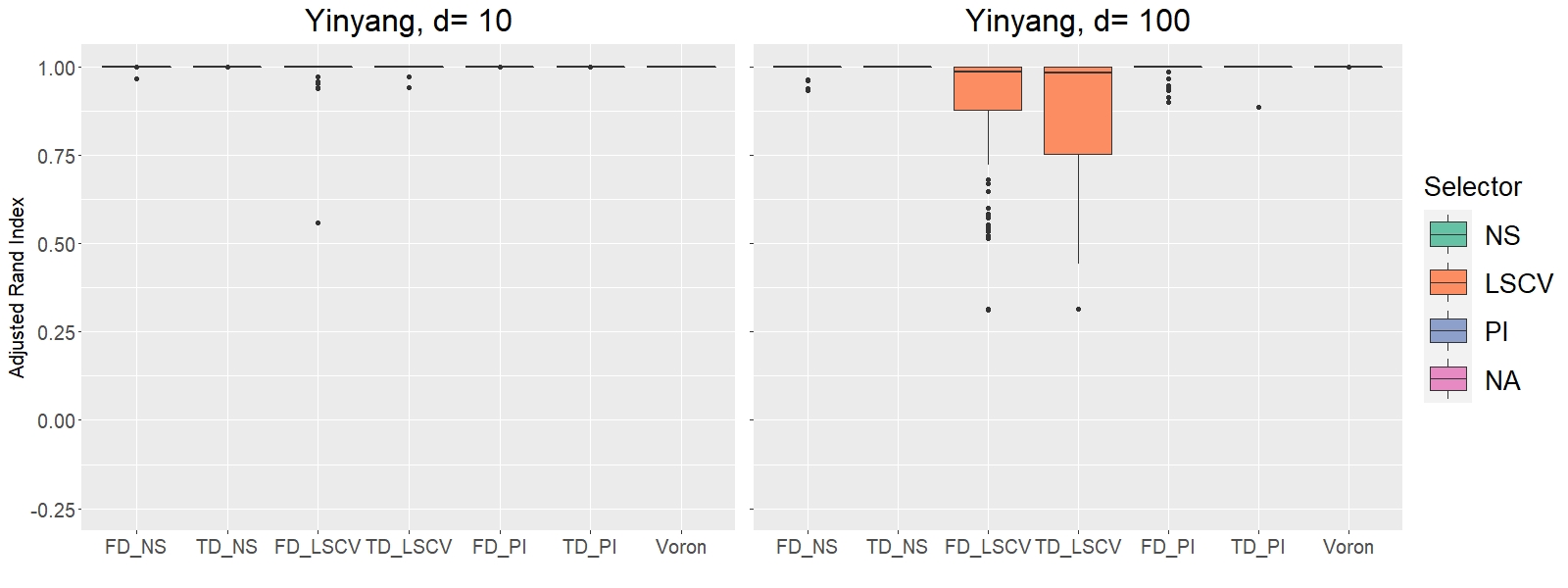} 
    \end{subfigure} \\
        \begin{subfigure}[t]{\textwidth}
        \centering
        \includegraphics[width=0.7\linewidth]{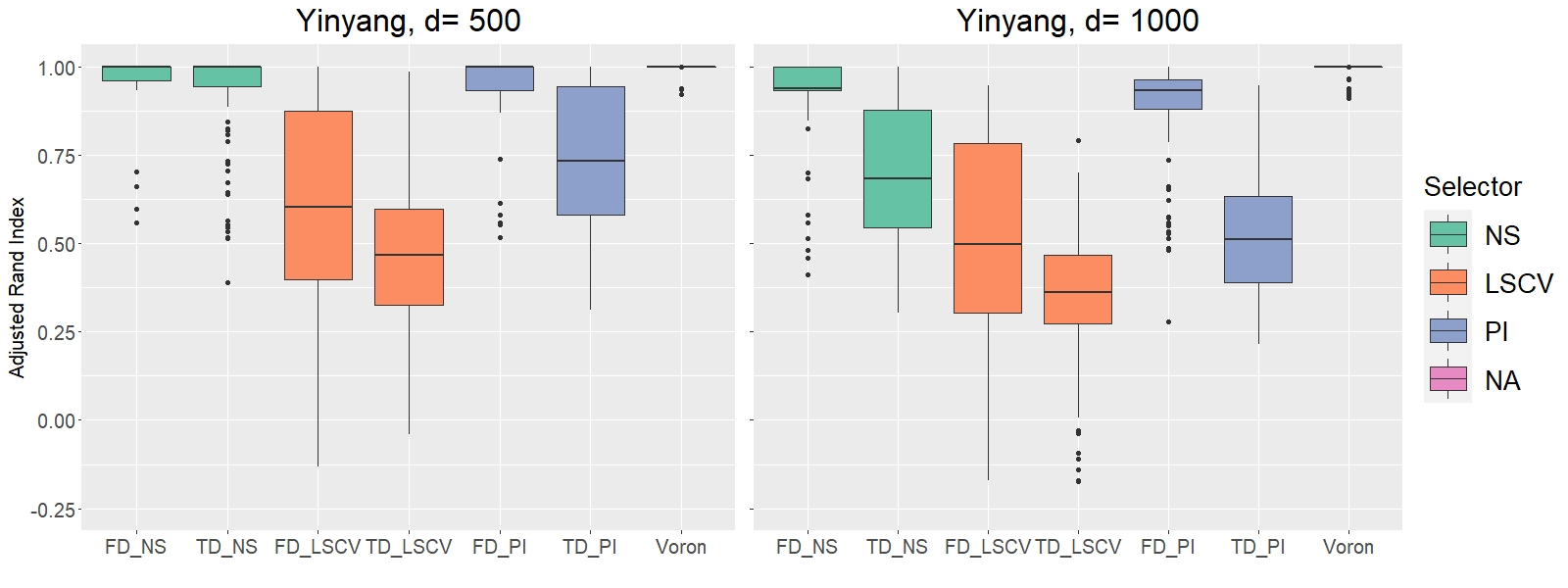} 
    \end{subfigure} \\
\caption{Performance of skeleton clustering on Yinyang data $d=10,100,500,1000$ with Face and Tube density by different bandwidth selectors. Voronoi density result is included for comparison.}
\label{sec::comp}
\end{figure}

It is known that the bandwidth is a pivotal parameter that can significantly affect the estimation result of a kernel density estimator. 
 {In Figure~\ref{sec::comp},
we conduct a simulation using the Yinyang data with different dimensions of noisy Gaussian variables (see Section~\ref{sec::YY} for more details)} 
and compare the performance of three common bandwidth selectors:
the normal scale bandwidth (NS) \citep{Chacon2011}, the least-squared cross-validation (LSCV) \citep{hlscv1, hlscv2},
and the plug-in approach (PI) \citep{wand1994multivariate}.
Each edge is allowed to have its own  bandwidth. 
 {Voronoi density performance results are also included for comparison.}
We found that the NS performs reliably well while the others may have unstable performance. A similar comparison of the bandwidth selectors on another dataset is presented in Appendix \ref{sec::bandselectMixMickey} and the NS also performs relatively better than the other bandwidth selectors..
As a result, we recommend using the NS as the default bandwidth selector.
Additionally, since the density estimations are all 1-dimensional, in practice it is possible to examine the estimated density to assess the degree of oversmoothing or undersmoothing and manually adjust the bandwidth. 

In addition to different bandwidth selectors, we also study how the bandwidth should depend on the sample size for clustering purpose. In $1$-dimensional data, the normal scale bandwidth agrees with Silverman's rule of thumb \citep{silverman1986density} giving the bandwidth as $h = \frac{4}{3}^{1/5} \hat{\sig} n_{loc}^{-1/5}$, where $\hat{\sig}$ is the standard deviation of the sample used in the edge weight calculation, and $n_{loc}$ the number of sample points used. Empirically we tested the clustering performance with FD and TD calculated under bandwidth with rates on $n_{loc}$ from $-1/3$ to $-1/10$ (see Appendix \ref{sec::bandrate}). We found that the clustering performance with FD and TD generally stays stable with varying bandwidth rates, although a larger bandwidth (slower rate than $O\big(n_{loc}^{-1/5}\big)$)
%a larger than $O(n_{loc}^{-1/5})$ rate bandwidth 
may give better clustering results with TD when the dimension of the data is high.

{

\subsubsection{Bandwidth Selection with Mix Mickey}	\label{sec::bandselectMixMickey}

\begin{figure}[ht]
\centering
    \begin{subfigure}[t]{\textwidth}
        \centering
        \includegraphics[width=0.8\linewidth]{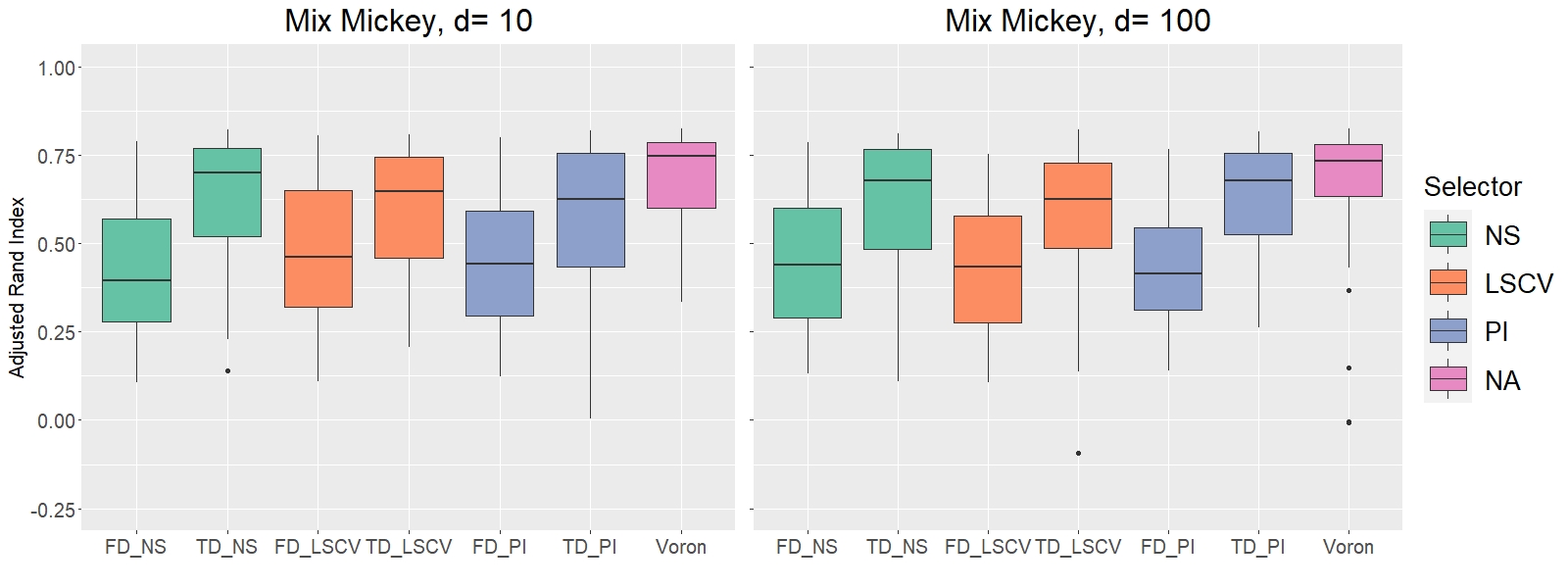} 
    \end{subfigure} \\
        \begin{subfigure}[t]{\textwidth}
        \centering
        \includegraphics[width=0.8\linewidth]{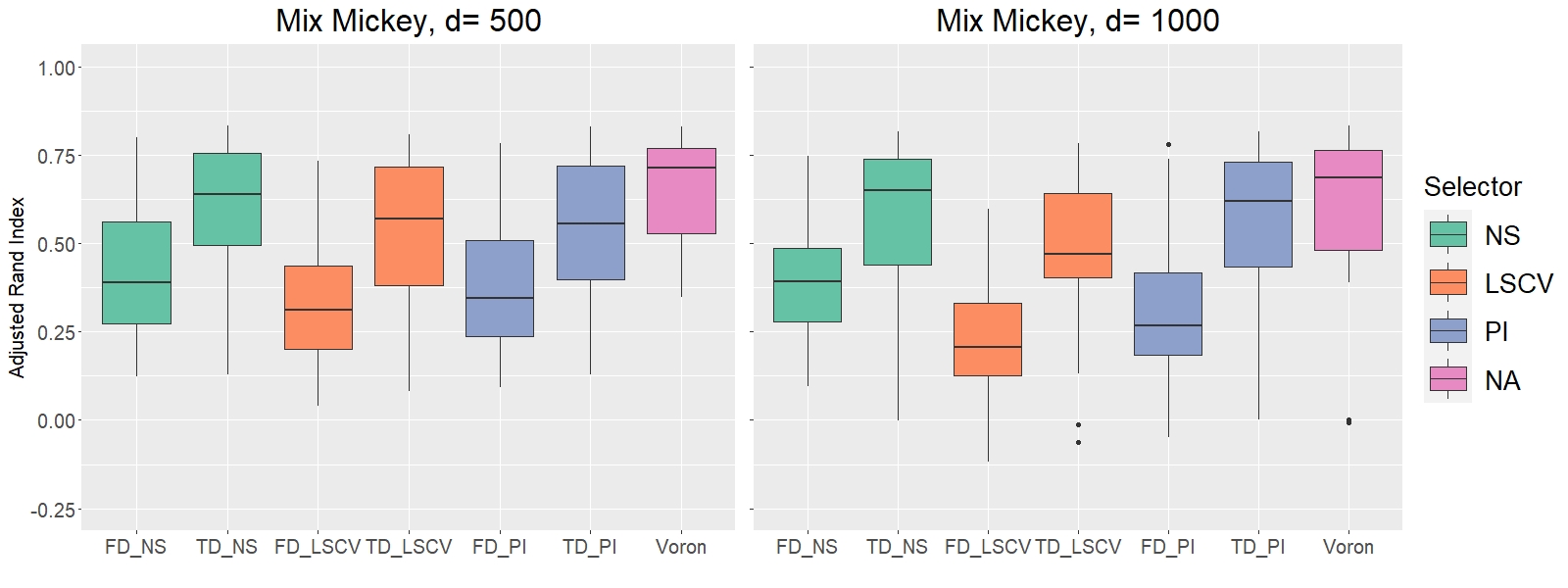} 
    \end{subfigure} \\
\caption{Performance of skeleton clustering on Mix Mickey data $d=10,100,500,1000$ with Face and Tube density by different bandwidth selectors. Voronoi density result is included for comparison.}
\label{fig::MixMickeyBandSelect}
\end{figure}
~~~~We present additional results comparing different bandwidth selectors on the Mix Mickey dataset generated the same way as in Section \ref{sim::mixMickey}. We use average linkage for all the included skeleton clustering approaches. The results are presented in Figure \ref{fig::MixMickeyBandSelect}.
The selectors have similar performances on this Mix Mickey dataset, but NS again seems to perform better with larger dimensions, which corroborates our default choice of using NS for bandwidth.

\subsubsection{Performance under Different Bandwidth Rate}
\label{sec::bandrate}

~~~~In this section we present empirical results on how changing the bandwidth rate affects the performance of clustering.
We consider the Yinyang data in Section \ref{sec::YY} with $d = 10, 100, 500, 1000$. 
We compare 
the Face and Tube density where the bandwidth is selected by Silverman's rule of thumb with different  rates, ranging from $n_{loc}^{-1/3}$ to $n_{loc}^{-1/10}$.
Note that the original  Silverman's rule of thumb will be at rate $n_{loc}^{-1/5}$.
We repeat the experiment $100$ times and record the adjusted Rand index in Figure \ref{fig::hraterands}.

\begin{figure}[ht]
\centering
\includegraphics[width=6cm]{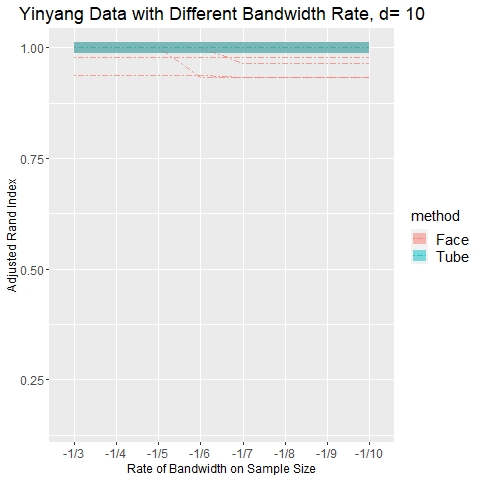}
\includegraphics[width=6cm]{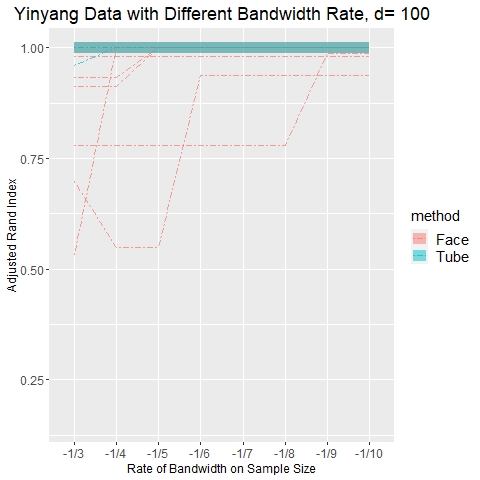}\\
\includegraphics[width=6cm]{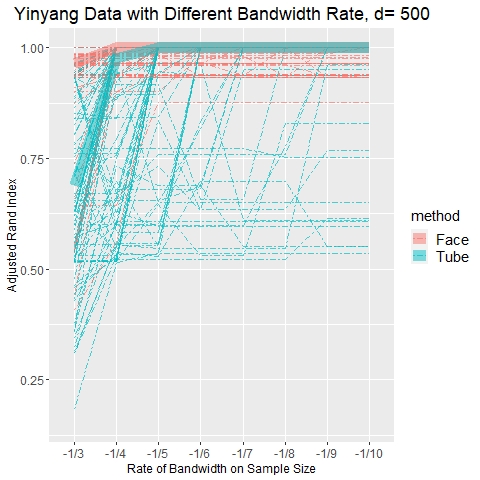}
\includegraphics[width=6cm]{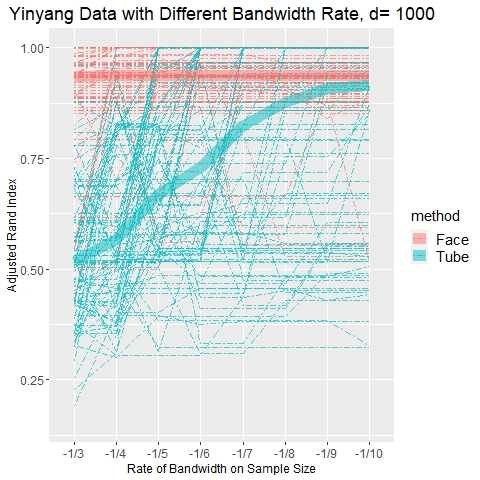}
\caption{Adjusted Rand indexes of skeleton clustering with Face and Tube density under different bandwidth rates on $100$ simulated Yinyang datasets. The thick lines indicate the median adjusted Rand index of a given method.}
\label{fig::hraterands}
\end{figure}

When the dimension is low (top panels), all bandwidth within this range works well. 
When the dimension is large (bottom panels), 
a slower rate (larger bandwidth) seems to be showing better performance for the TD.
Interestingly, the face density yields a robust result across different rates of bandwidth.
Note that for the TD, the theory (Theorem~\ref{thm::TD1}) suggests the choice at rate $h\asymp n_{loc}^{-1/5}$
is optimal for estimation in large $d$, the same rate may not lead to the optimal clustering performance.
Figure \ref{fig::hraterands} bottom-right panel suggests that the choice $h\asymp n_{loc}^{-1/10}$
may have a better clustering performance in this case.

\subsubsection{Adaptive Radius for Tube Density}	\label{sec::adaptiveTube}

\begin{figure}[ht]
\centering
\includegraphics[width=3.5cm]{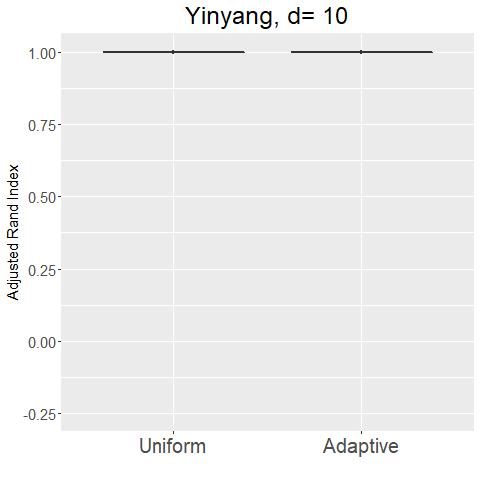}
\includegraphics[width=3.5cm]{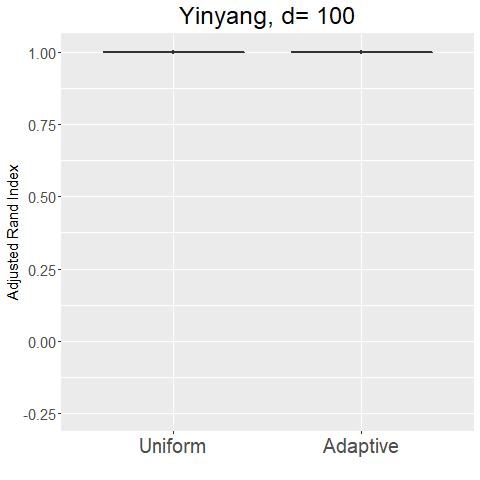}
\includegraphics[width=3.5cm]{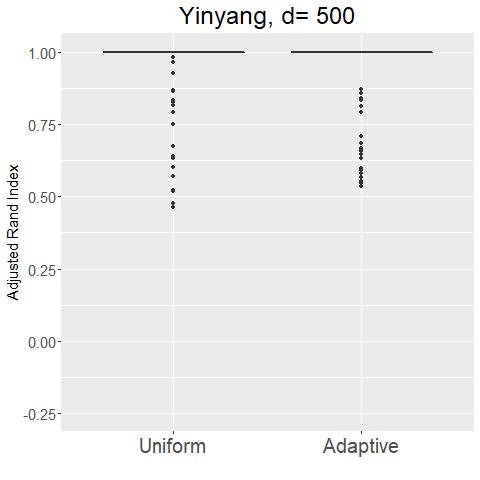}
\includegraphics[width=3.5cm]{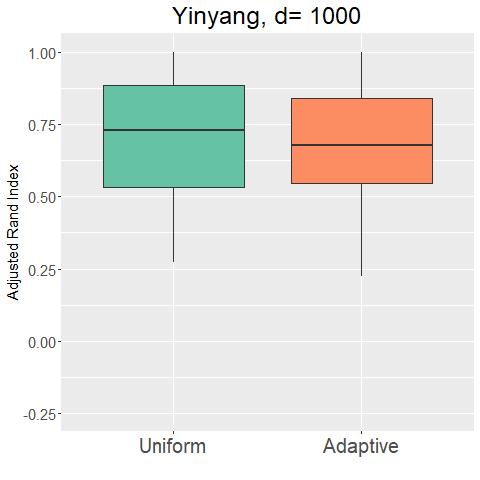}
\caption{Comparison of radius choices on Yinyang data with dimensions $10$, $100$, $500$, $1000$.}
\label{fig::adaptiveFrustum}
\end{figure}

~~~~We compare the clustering performance of Tube density when using fixed radius and that when using adaptive radius as described in Section \ref{sec::TD}. The data is the same Yinyang data in Section \ref{sec::YY} and the results are presented in Figure \ref{fig::adaptiveFrustum}. The two approaches (adaptive and fixed radius) have  a similar performance.

\begin{figure}[ht]
\centering
\includegraphics[width=3.5cm]{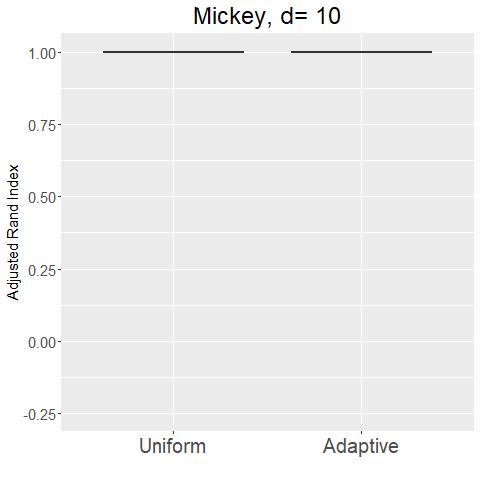}
\includegraphics[width=3.5cm]{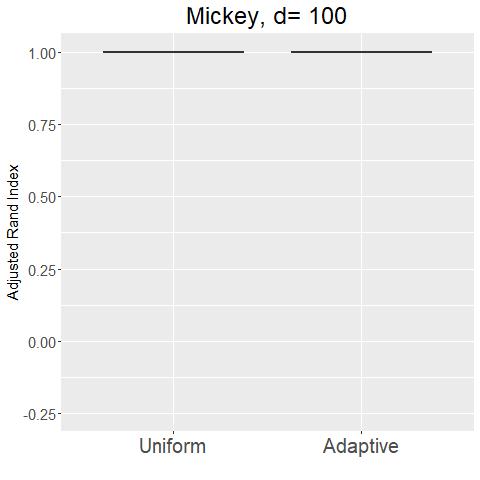}
\includegraphics[width=3.5cm]{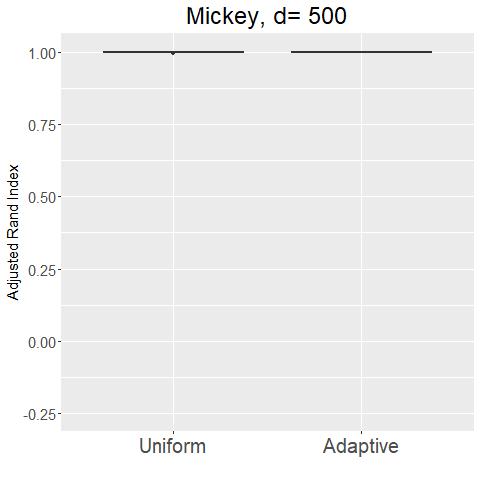}
\includegraphics[width=3.5cm]{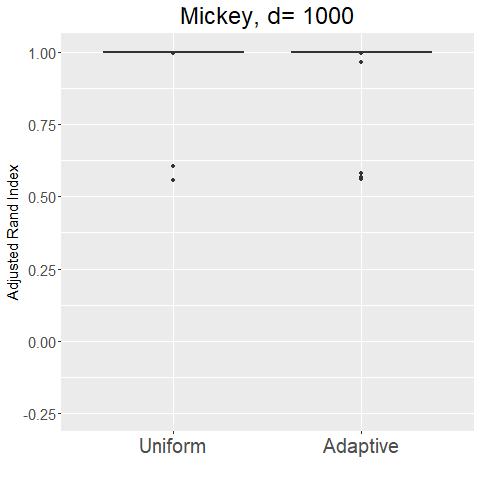}
\caption{Comparison of radius choices on Mickey data with dimensions $10$, $100$, $500$, $1000$.}
\label{fig::adaptiveFrustumMickey}
\end{figure}

~~~~For comprehensiveness, we also compare the clustering performance of Tube density when using fixed radius and that when using adaptive radius on the Mickey data same as in Section \ref{sec::mickey}, and the results are presented in Figure \ref{fig::adaptiveFrustumMickey}. The two approaches also have a similar performance. 
In Figure \ref{fig::radiusDispersion}, we plot the distribution of the adaptive radius on one Yinyang data and one Mickey data. We note that there are some variations across different Voronoi cells, but the variation is not large, and this can be a consequence of constructing the knots using the k-Means method so that the within-cluster variations are minimized. 
\begin{figure}[ht]
\centering
\includegraphics[width=6cm]{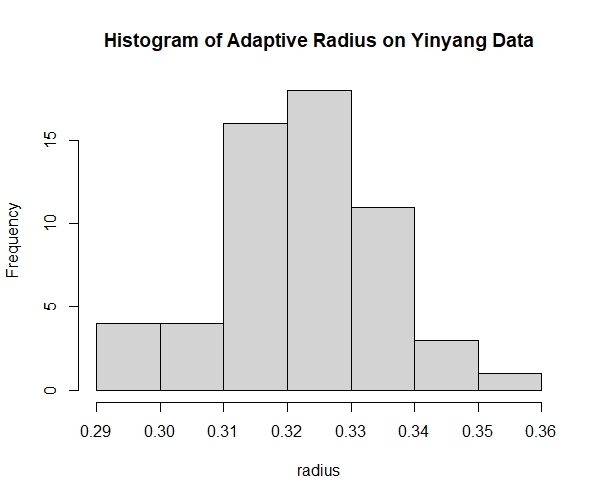}
\includegraphics[width=6cm]{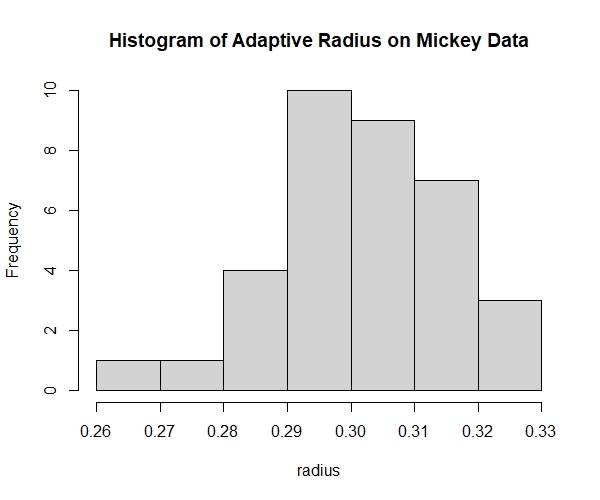}
\caption{Dispersion of radius on one Yinyang data and one Mickey data.}
\label{fig::radiusDispersion}
\end{figure}

\subsubsection{Higher Standard Deviations for Noisy Dimensions}	\label{sec::highSigma}

~~~~We investigate how changing the noise level of the added noisy dimensions of our simulation examples changes the clustering performance. Here we simulate Yinyang data with different standard deviations of the added dimensions. We apply the same analysis procedure as in Section \ref{sec::YY} is applied. The adjusted Rand indexes of the clustering methods on $100$ simulated datasets with under setting are presented in Figure \ref{fig::highSigma}. 

\begin{figure}[ht]
\captionsetup{skip=1pt}
\centering
    \begin{subfigure}[t]{0.22\textwidth}
        \centering
        \textbf{$\sig$ = 0.1}
    \end{subfigure}\\
    \begin{subfigure}[t]{0.22\textwidth}
        \centering
        \includegraphics[width=\linewidth]{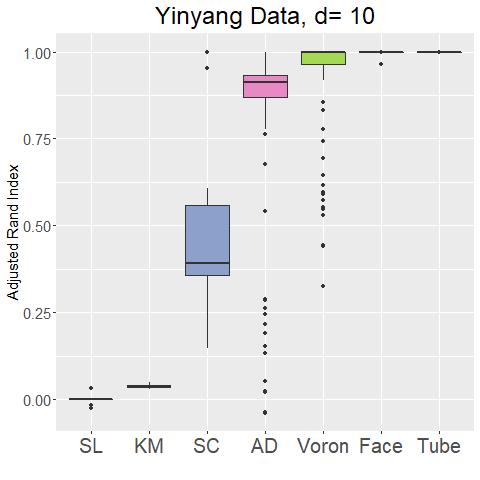} 
    \end{subfigure}
    \begin{subfigure}[t]{0.22\textwidth}
        \centering
        \includegraphics[width=\linewidth]{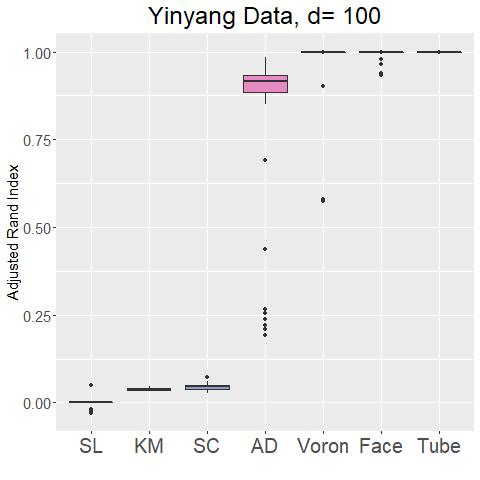}
    \end{subfigure}
    \begin{subfigure}[t]{0.22\textwidth}
        \centering
        \includegraphics[width=\linewidth]{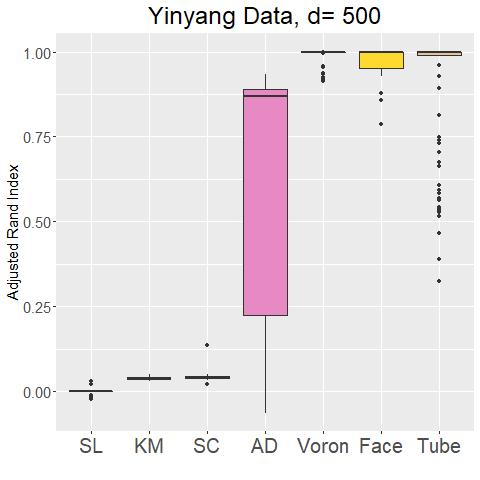}
    \end{subfigure}
    \begin{subfigure}[t]{0.22\textwidth}
        \centering
        \includegraphics[width=\linewidth]{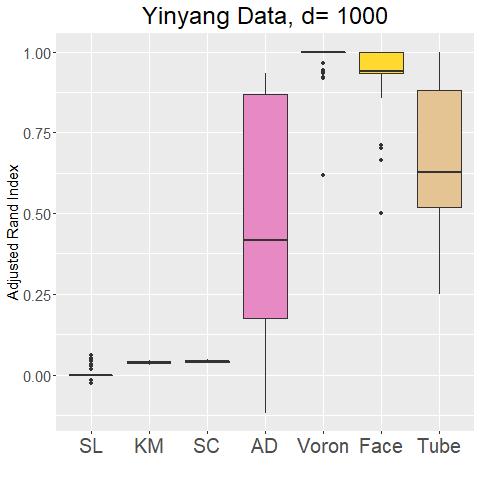}
    \end{subfigure}\\
    \begin{subfigure}[t]{0.22\textwidth}
        \centering
        \textbf{$\sig$ = 0.2}
    \end{subfigure}\\
    \begin{subfigure}[t]{0.22\textwidth}
        \centering
        \includegraphics[width=\linewidth]{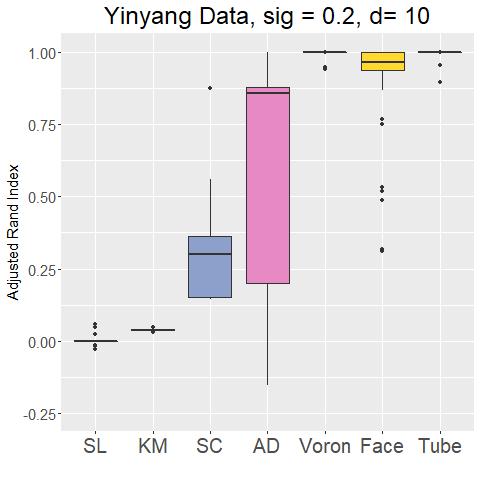} 
    \end{subfigure}
    \begin{subfigure}[t]{0.22\textwidth}
        \centering
        \includegraphics[width=\linewidth]{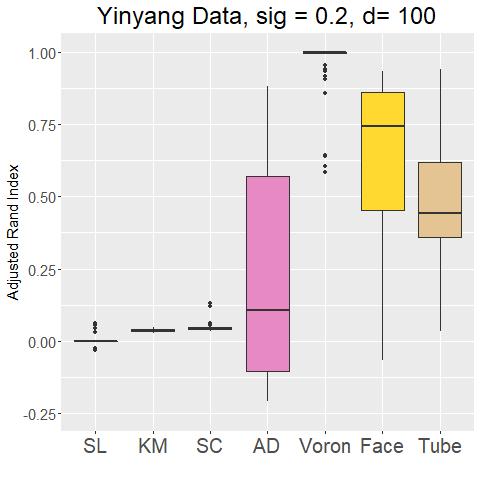}
    \end{subfigure}
    \begin{subfigure}[t]{0.22\textwidth}
        \centering
        \includegraphics[width=\linewidth]{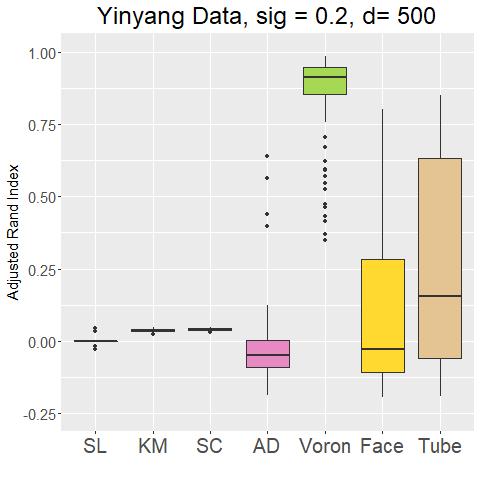}
    \end{subfigure}
    \begin{subfigure}[t]{0.22\textwidth}
        \centering
        \includegraphics[width=\linewidth]{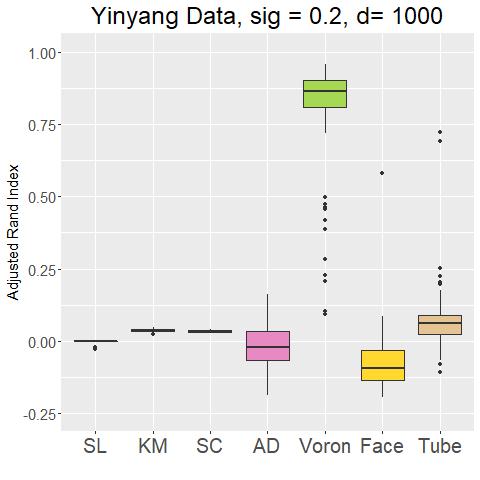}
    \end{subfigure}\\
        \begin{subfigure}[t]{0.22\textwidth}
        \centering
        \textbf{$\sig$ = 0.3}
    \end{subfigure}\\
        \begin{subfigure}[t]{0.22\textwidth}
        \centering
        \includegraphics[width=\linewidth]{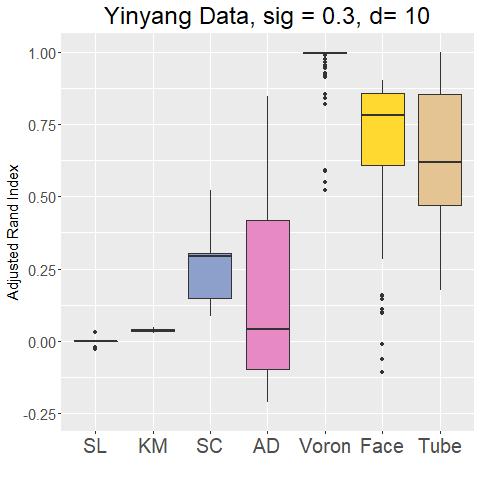} 
    \end{subfigure}
    \begin{subfigure}[t]{0.22\textwidth}
        \centering
        \includegraphics[width=\linewidth]{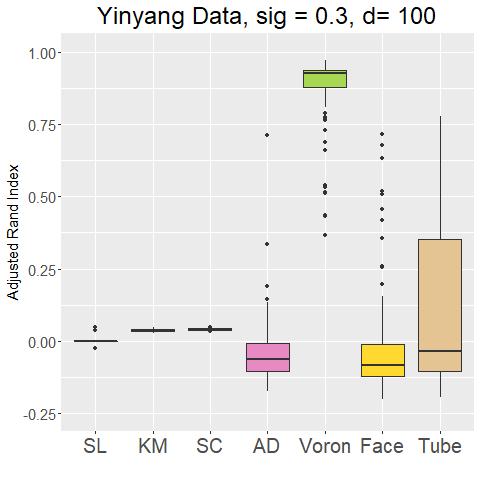}
    \end{subfigure}
    \begin{subfigure}[t]{0.22\textwidth}
        \centering
        \includegraphics[width=\linewidth]{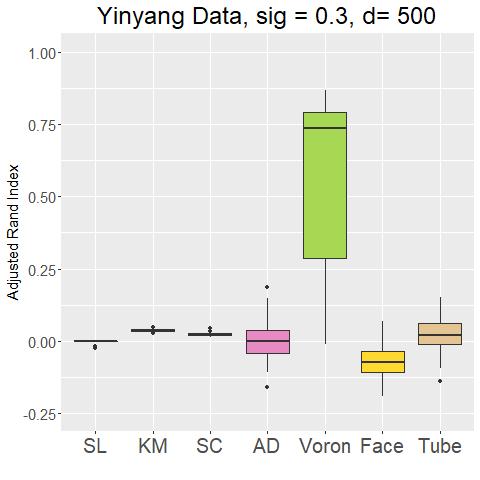}
    \end{subfigure}
    \begin{subfigure}[t]{0.22\textwidth}
        \centering
        \includegraphics[width=\linewidth]{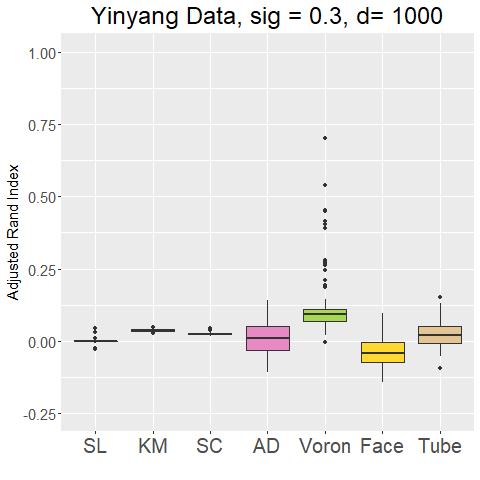}
    \end{subfigure}\\
            \begin{subfigure}[t]{0.22\textwidth}
        \centering
        \textbf{$\sig$ = 0.4}
    \end{subfigure}\\
        \begin{subfigure}[t]{0.22\textwidth}
        \centering
        \includegraphics[width=\linewidth]{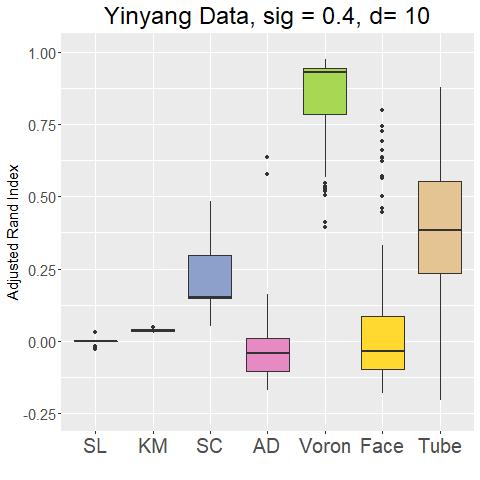} 
    \end{subfigure}
    \begin{subfigure}[t]{0.22\textwidth}
        \centering
        \includegraphics[width=\linewidth]{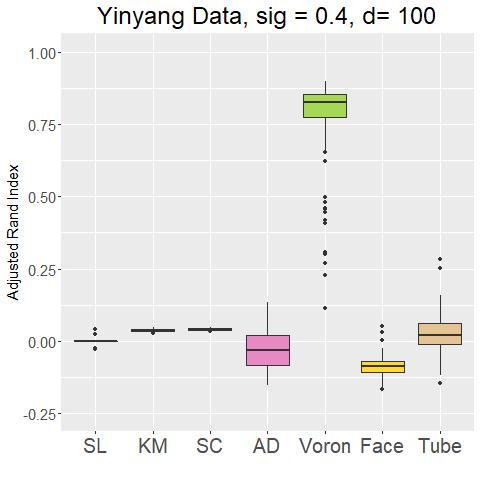}
    \end{subfigure}
    \begin{subfigure}[t]{0.22\textwidth}
        \centering
        \includegraphics[width=\linewidth]{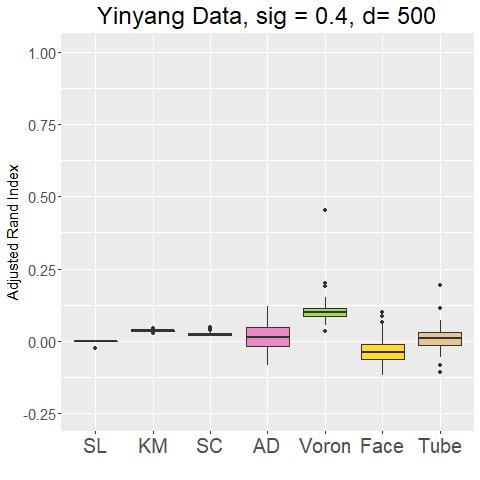}
    \end{subfigure}
    \begin{subfigure}[t]{0.22\textwidth}
        \centering
        \includegraphics[width=\linewidth]{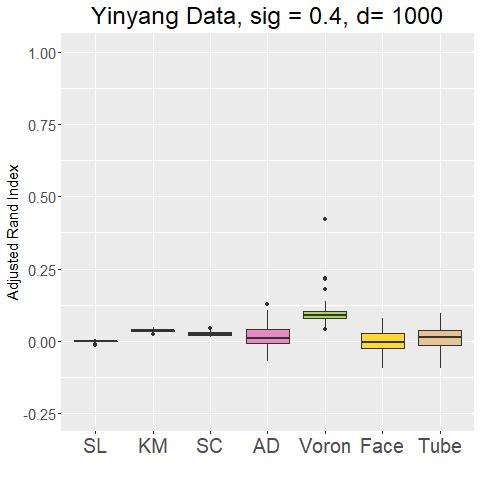}
    \end{subfigure}\\
\caption{Adjusted Rand index performance of clustering methods on Yinyang data with different standard deviation for added dimensions.}
\label{fig::highSigma}
\end{figure}

We observe that increasing the standard deviation of the noisy dimensions  (noise level)
has a stronger impact than adding more noisy variables.
%can be more challenging than adding more noisy variables. 
For example, increasing $\sig = 0.1 \to 0.2$ scales the standard deviation by a factor of $2$ (scales the variance $4$ times), but the clustering performance with $\sig = 0.2, d =100$ is worse than that with $\sig = 0.1, d=500$. However, we still observe that the skeleton clustering with Voronoi density similarity measure can give good clustering performance even under the setting with $\sig = 0.4$ and $d=100$.
It can be observed that the VD performance improves under $\sigma=0.1$ (first row) v.s. $\sigma=0.2$ (second row). While this difference is not large, it is unexpected as in other cases the performance is not better compared to smaller noises. We think that this might be due to Monte Carlo errors.

\subsubsection{Mix Mickey with GMM}	\label{sec::withGMM}

~~~~We compare the performance of Gaussian Mixture Models (GMMs) to our methods using the Mix Mickey data same as in Section \ref{sim::mixMickey}. Unfortunately, the GMM method from \texttt{clusterR} package in R cannot work with dimension $500$ and $1000$ case because of too many noisy dimensions, so we only compare the case of dimension $10$  and $100$. For the skeleton clustering, we use average linkage for the segmentation step the same as in Section \ref{sim::mixMickey}. 
	Because this data is generated from 3-GMM and we fit the GMM with $3$ components, the GMM naturally has the best performance.
%Since this dataset is truly composed of $3$ Gaussian components and we tell the clustering algorithms the true number of clusters, this is a favorable setting for GMM, and GMM indeed gives the best clustering result. 
	However, our proposed approaches achieve  performance comparable to that given by the GMM 
	and are capable of handling high dimensional data ($d=500, 1000$).
%	but can work with larger dimensional datasets and are suitable for clusters with complex structures.

\begin{figure}[ht]
\centering
\includegraphics[width=0.3\textwidth]{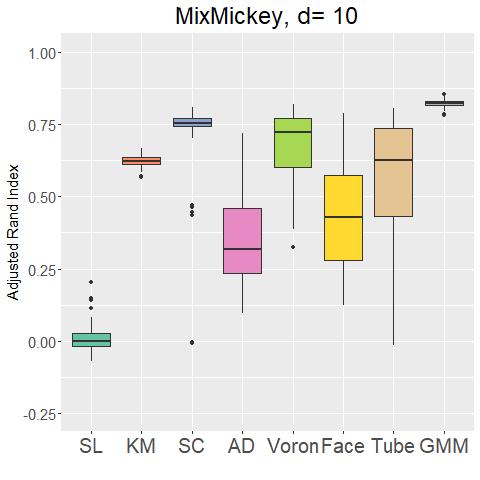}
\includegraphics[width=0.3\textwidth]{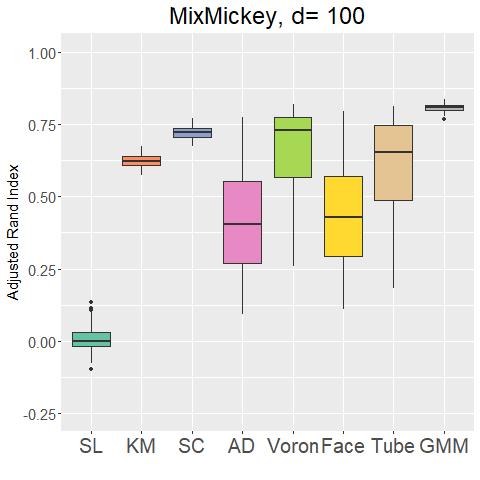}
\caption{Comparison of clustering methods on Mix Mickey data $d=10,100$ with GMM included.}
\label{fig::withGMM}
\end{figure}

\subsubsection{Comparison with Combining Mixture Components for Clustering}
\label{sec::mcomb}
~~~~In this section, we present the empirical comparison with the combining mixture components approach (mComb) from \cite{Baudry2010} with the implementation from the {\bf mclust} package in R \footnote{\url{https://mclust-org.github.io/mclust/}}.
The mComb algorithm builds upon the Gaussian finite mixture models fitted via the EM algorithm and proceeds with a hierarchy of combined clusterings following an entropy-based criterion.

To make a comparison between this model-based merging clusters approach with our proposed skeleton framework, we apply the mComb algorithm to the Yinyang data (Section \ref{sec::YY}) and the Mickey data (Section \ref{sec::mickey}).
However, with a large dimension of noisy variables, the initial stage of Gaussian mixture modeling is likely to identify the data points as one large component without distinguishing the different components, and the overall algorithm is unstable.
Therefore, we only present the empirical results for merging Gaussian mixtures on datasets with dimension $d = 10$.

We carry out the combining mixture components approach in two ways. 
In one way, we provide the true number of final clusters ($5$ for Yinyang data and $3$ for Mickey data) to the algorithm (\textbf{mCombOracle}). 
In another way, we use the optimal number of clusters by combining mixture components based on the entropy method as proposed in \cite{Baudry2010} (\textbf{mCombOptim}). 
We repeat the experiments for $100$ random instances for the two different simulation datasets and the results are presented in Figure \ref{fig::mComb}, where are included the result given by the spectral clustering (SC) and the skeleton clustering with Voronoi density (Voron) for comparison.

\begin{figure}
\centering
\includegraphics[width=0.4\textwidth]{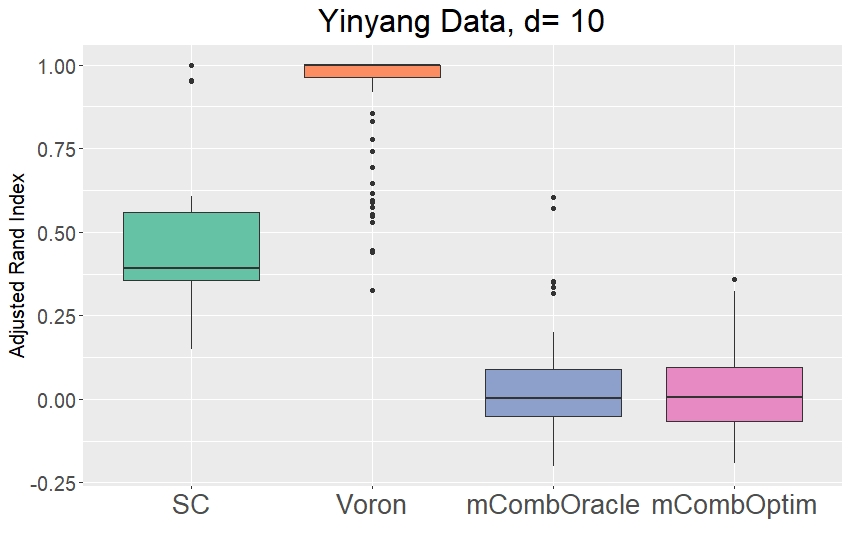}
\includegraphics[width=0.4\textwidth]{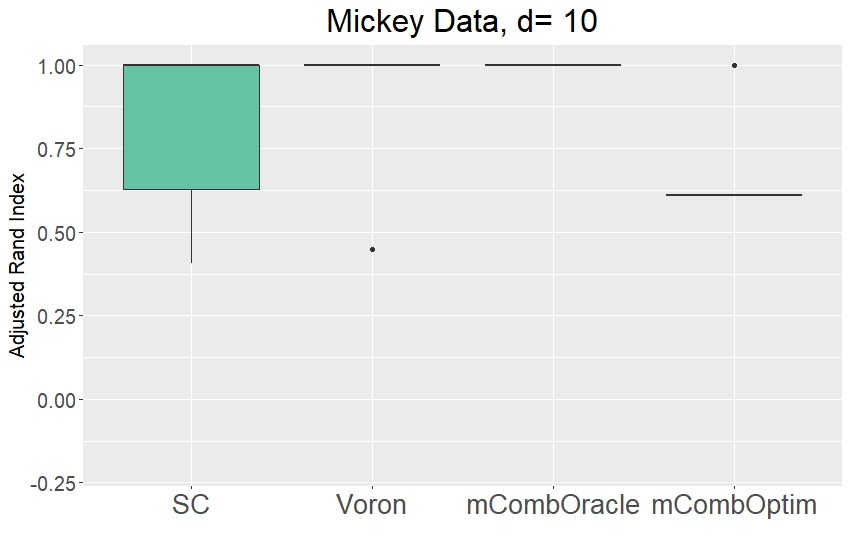}
\caption{Results from combining mixture components on Yinyang and Mickey data $d=10$. }
\label{fig::mComb}
\end{figure}

On the Yinyang data, the optimal number of clusters chosen by the mComb algorithm varies from $2$ to $8$, while only $12\%$ of the runs have the optimal number of clusters to be the true number of clusters $5$.
The performance given by the combining mixture components approach as assessed by the adjusted Rand Index is not satisfactory compared to the classical clustering methods and the skeleton clustering approaches. 
Merging Gaussian components is not sufficient to handle the complex structures of the Yinyang data.

For the Mickey data, $94\%$ of the mCombOptim runs identify $2$ to be the optimal number of clusters, while only $6\%$ of the times the optimal number of clusters is determined to be $3$, the true number of clusters.
Therefore, combining mixture components with an auto-chosen number of clusters does not give good performance.
However, the combining mixture components approach with the number of components given does lead to perfect clustering results on the Mickey data where the clusters are spherically shaped.

\subsubsection{Graphical Representation of GvHD Data Clusters}	\label{sec::GvHDgraph}

\begin{figure}
\captionsetup{skip=1pt}
\centering
    \begin{subfigure}[t]{0.19\textwidth}
        \centering
        \includegraphics[width=\linewidth]{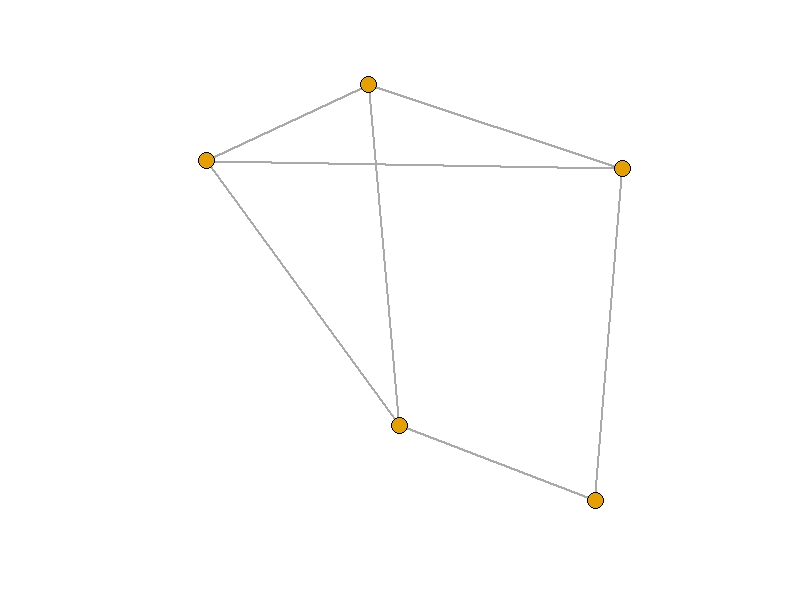} 
        \caption{Cluster 1}
    \end{subfigure}
    \begin{subfigure}[t]{0.19\textwidth}
        \centering
        \includegraphics[width=\linewidth]{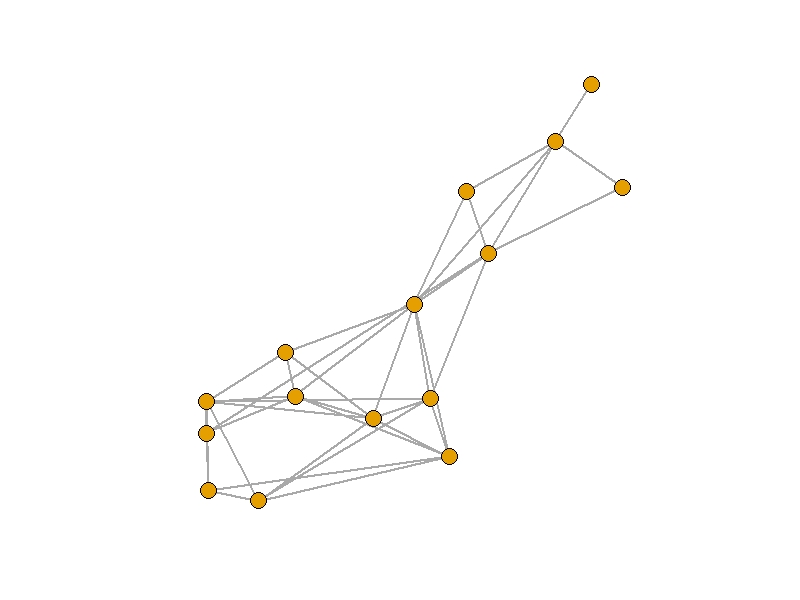}
        \caption{Cluster 2}
    \end{subfigure}
    \begin{subfigure}[t]{0.19\textwidth}
        \centering
        \includegraphics[width=\linewidth]{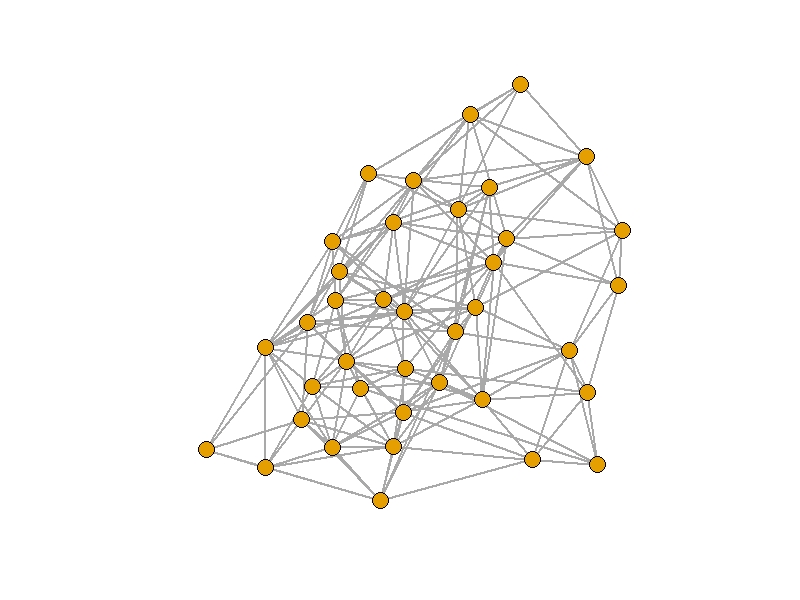}
        \caption{Cluster 3}
    \end{subfigure}
    \begin{subfigure}[t]{0.19\textwidth}
        \centering
        \includegraphics[width=\linewidth]{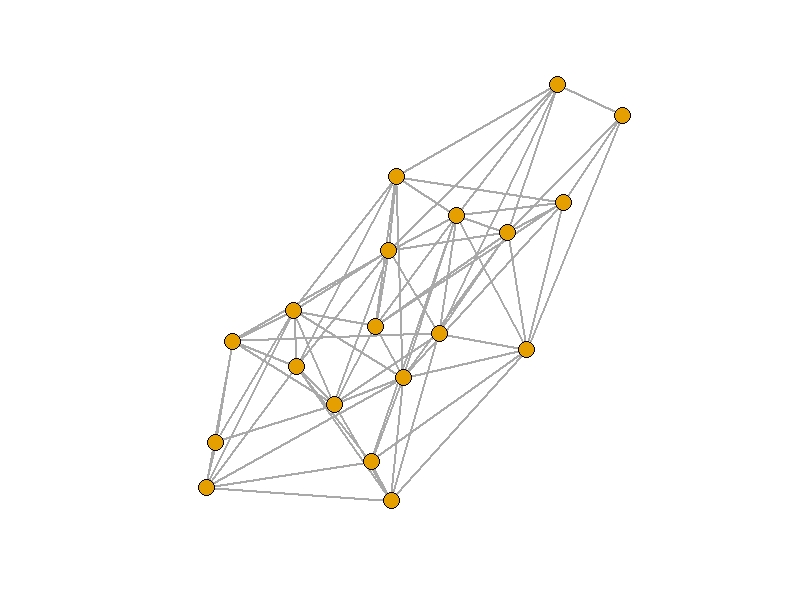} 
        \caption{Cluster 4}
    \end{subfigure}
    \begin{subfigure}[t]{0.19\textwidth}
        \centering
        \includegraphics[width=\linewidth]{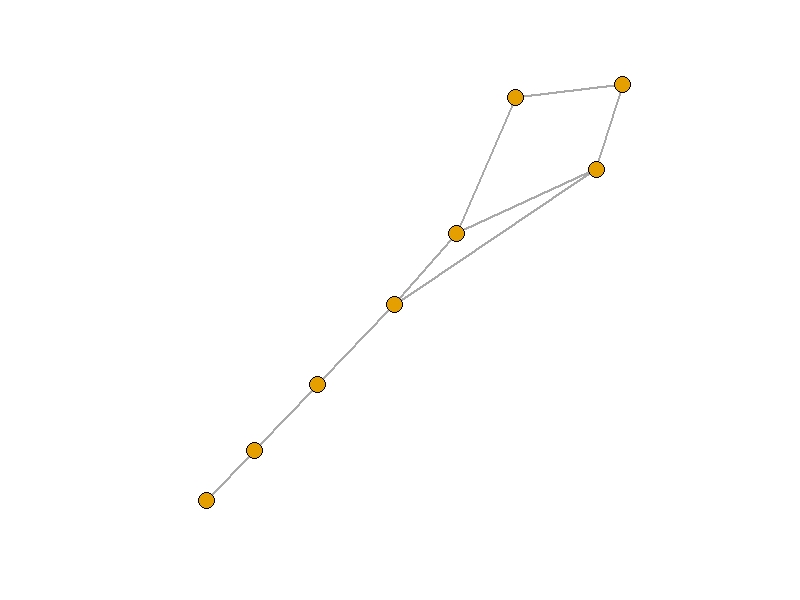} 
        \caption{Cluster 5}
    \end{subfigure} \\
    \begin{subfigure}[t]{0.19\textwidth}
        \centering
        \includegraphics[width=\linewidth]{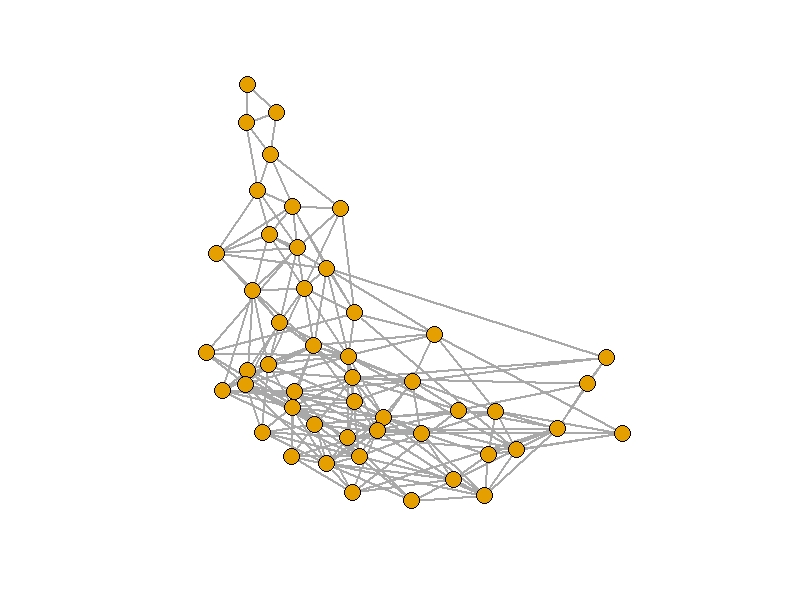} 
        \caption{Cluster 7}
    \end{subfigure}
    \begin{subfigure}[t]{0.19\textwidth}
        \centering
        \includegraphics[width=\linewidth]{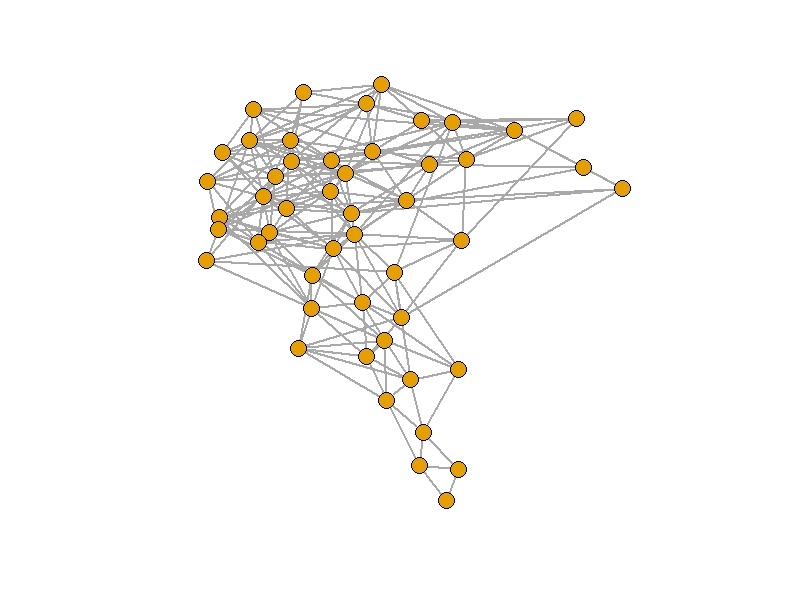}
        \caption{Cluster 8}
    \end{subfigure}
    \begin{subfigure}[t]{0.19\textwidth}
        \centering
        \includegraphics[width=\linewidth]{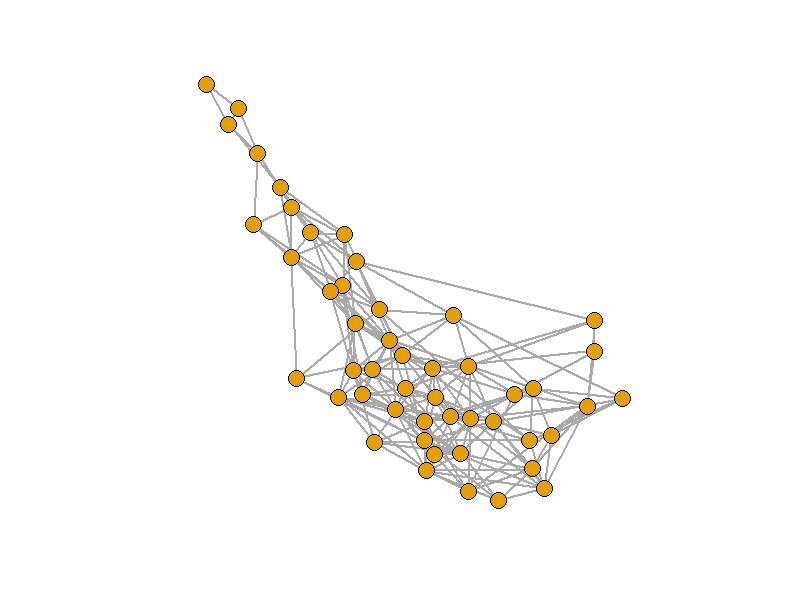}
        \caption{Cluster 9}
    \end{subfigure}
    \begin{subfigure}[t]{0.19\textwidth}
        \centering
        \includegraphics[width=\linewidth]{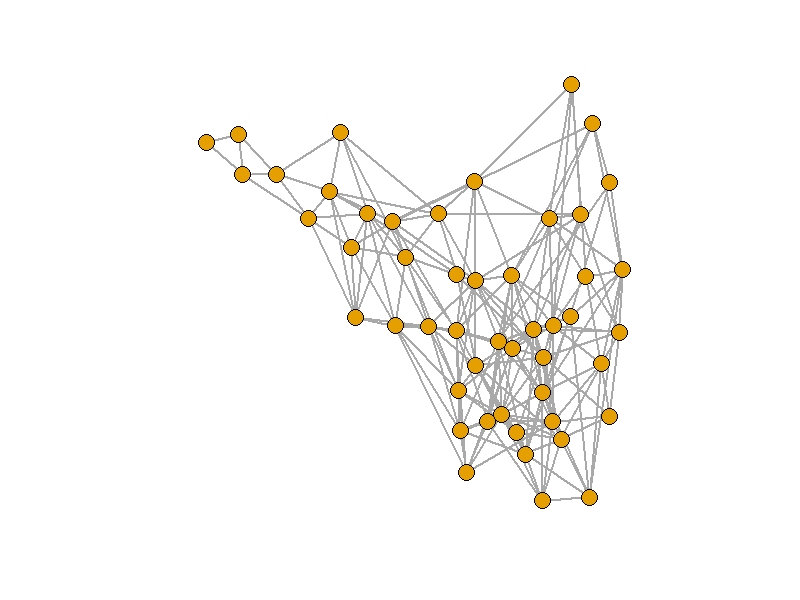} 
        \caption{Cluster 10}
    \end{subfigure}
    \begin{subfigure}[t]{0.19\textwidth}
        \centering
        \includegraphics[width=\linewidth]{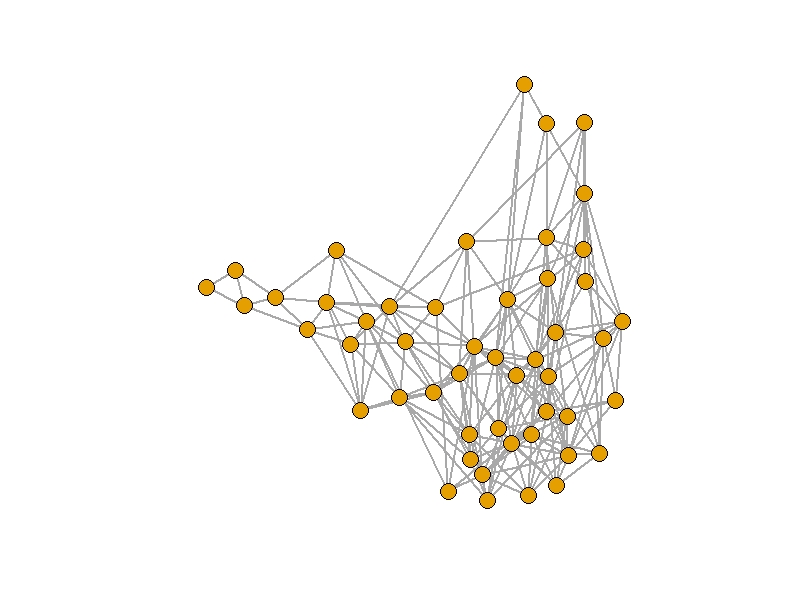} 
        \caption{Cluster 11}
    \end{subfigure} \\
       \begin{subfigure}[t]{0.19\textwidth}
        \centering
        \includegraphics[width=\linewidth]{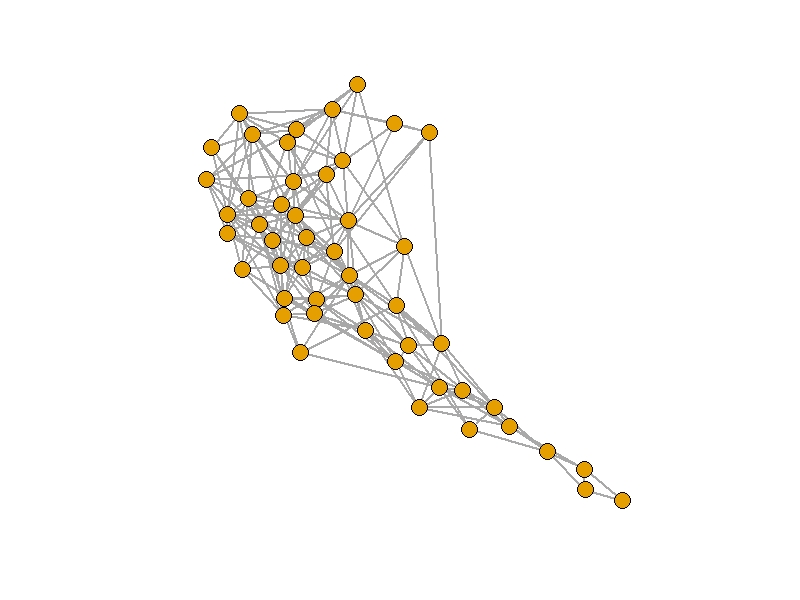} 
        \caption{Cluster 12}
    \end{subfigure}
    \begin{subfigure}[t]{0.19\textwidth}
        \centering
        \includegraphics[width=\linewidth]{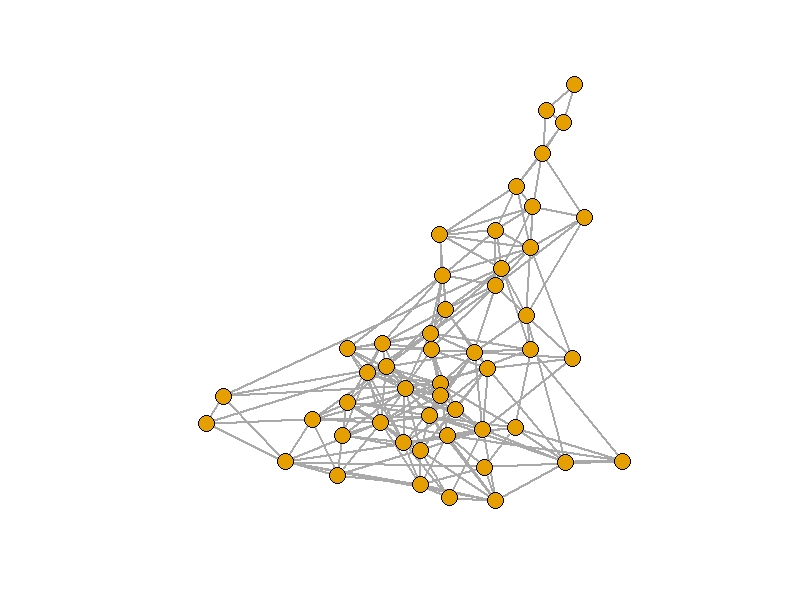}
        \caption{Cluster 13}
    \end{subfigure}
    \begin{subfigure}[t]{0.19\textwidth}
        \centering
        \includegraphics[width=\linewidth]{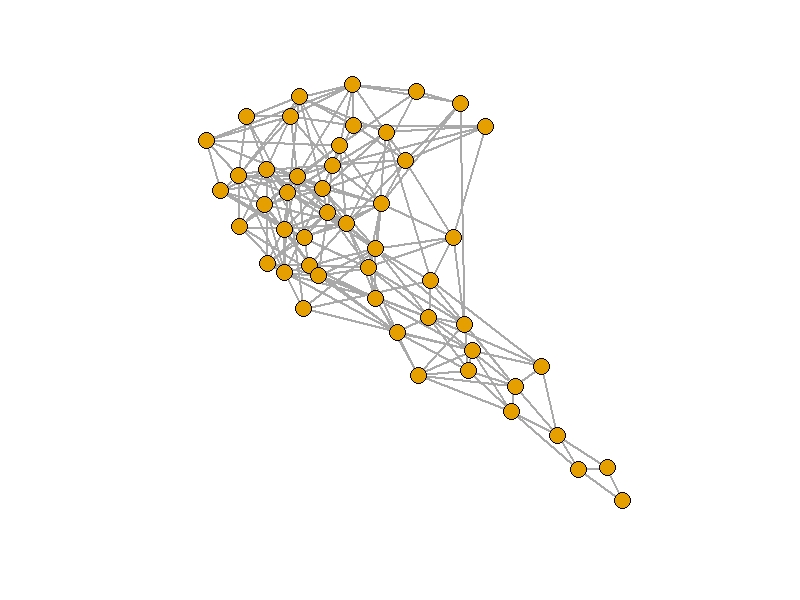}
        \caption{Cluster 14}
    \end{subfigure}
\caption{Skeleton structures of the clusters identified for the GvHD dataset in Section \ref{sec::GvHD}}
\label{fig::GvHDgraph}
\end{figure}

~~~~We visualize the skeleton structure of the clusters identified on the GvHD dataset in Section \ref{sec::GvHD}.
	These graph representations are generated by the \texttt{igraph} package in R. 
	Cluster $6$ only has $1$ knot with $17$ corresponding data points and is hence omitted in Figure \ref{fig::GvHDgraph}. 
	We observe that most clusters display a hammer-like structure, which is non-spherical and not favorable for some classical clustering methods.
	Only Cluster $3$ has a spherical shape in this data.
} %end coloring

\subsection{Additional Simulated Data Examples} \label{sec::simDataAdd}

\subsubsection{Manifold Mixture Data} \label{sim::ManifoldMixture}

\begin{figure}
\centering
\includegraphics[height=4.5cm]{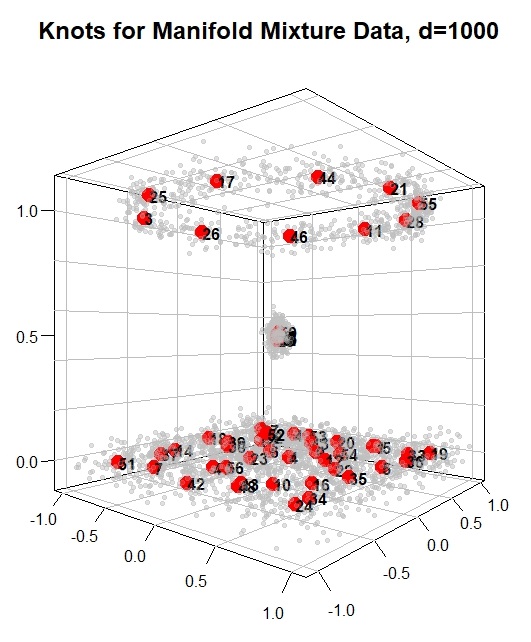}
\includegraphics[height=4.5cm]{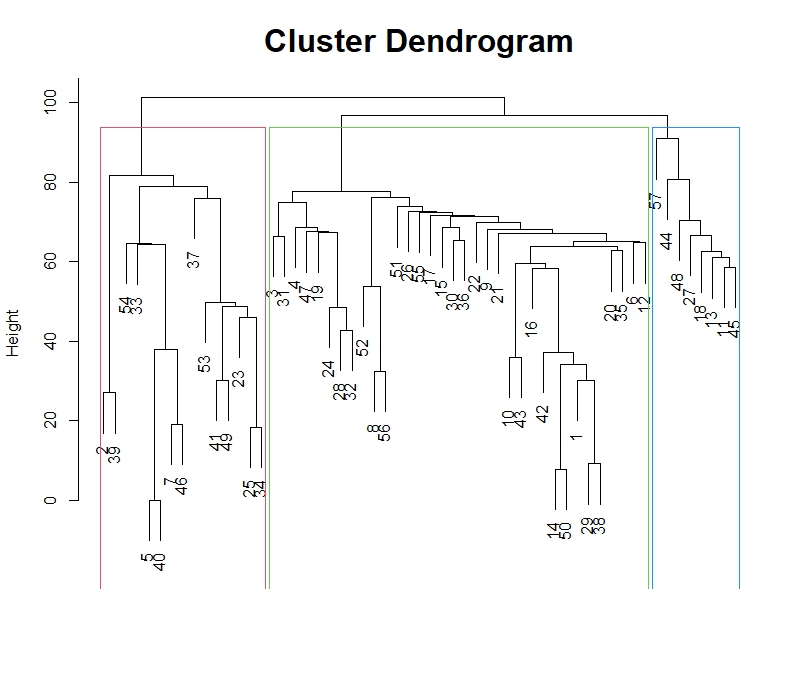}
\caption{Results on Manifold Mixture data with dimension $100$.}
\label{fig::MM1}
\end{figure}

~~~~In the Yinyang data and the Mix Mickey data experiments,
the underlying components are all two-dimensional structures. 
Here we consider the data composed of structures of different intrinsic dimensions called the manifold mixture data.
The simulated manifold mixture data, as illustrated in the left panel of Figure~\ref{fig::MM1}, consists of a $2$-dimensional plane with $2000$ data points, a $3$-dimensional Gaussian cluster with $400$ data points, and an essentially 1-dimensional ring shape with $800$ data points. There are a total of $3200$ observations and we choose $k = [\sqrt{3200}] = 57$ knots.
Similar to the other two simulations, we include Gaussian noise variables to make the data high-dimensional ($d=10,100,500,1000$) and make comparisons between the same set of clustering methods. The true number of components $S=3$ is provided to all the clustering algorithms.

%We test on this simulated dataset the ability of Skeleton Clustering to handle structures of different dimensions. For $d>3$ dimensions, random Gaussian noises are added to each observation. The data has $3200$ observations and $\sqrt{3200} \approx 57$. From the Minimum Spanning Tree total length comparison plot in Figure \ref{fig::MM1} (middle), we see that knots chosen by overfitting K-means do have the noise-reduction effect. 

\begin{figure}[ht]
\centering
\includegraphics[width=3.5cm]{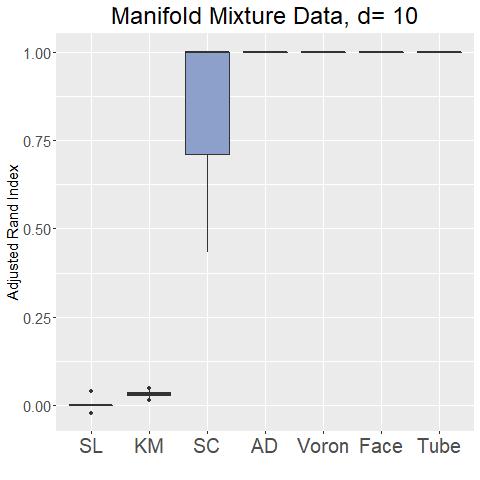}
\includegraphics[width=3.5cm]{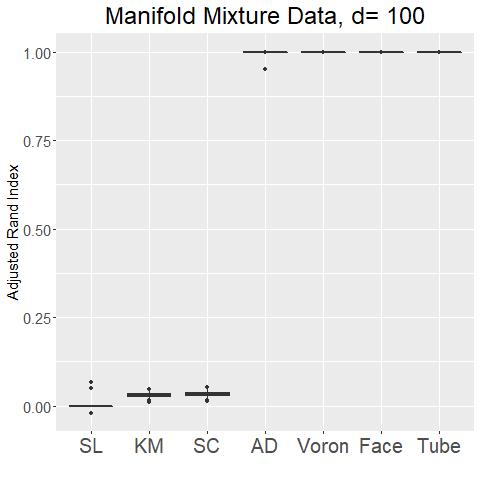}
\includegraphics[width=3.5cm]{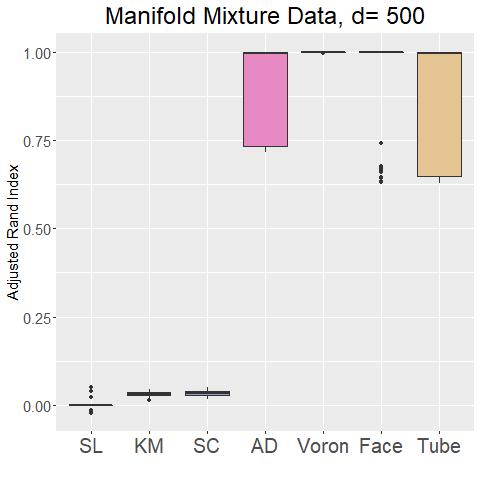}
\includegraphics[width=3.5cm]{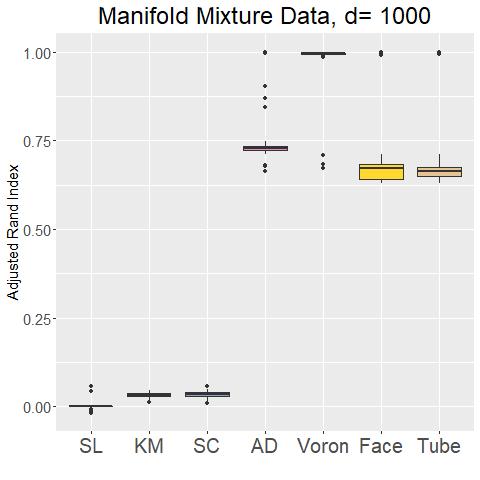}
\caption{Comparison of adjusted Rand index using different similarity measures on Manifold Mixture data with dimensions $10$, $100$, $500$, $1000$.}
\label{fig::MM2}
\end{figure}

Figure~\ref{fig::MM2} summarizes the performance of each method.
%The Rand Indexes of the clustering results generated by different methods are presented in Figure \ref{fig::MM2}.
Traditional methods (SL, KM, and SC) do not perform well when $d>10$ while all methods of skeleton clustering perform very well when $d\leq 500$.
Notably, the skeleton clustering with VD still has a perfect performance even
when $d=1000$, whereas skeleton clustering based on other similarity measures gives satisfying results.

\subsubsection{Ring Data} \label{sec::ring}

%For another simulated data, we set the two leading features to have a circular shape with some data at the center. All higher dimensions are Gaussian noises. This simulated data has $1200$ observations and $\sqrt{1200} \approx 35$, which is indicated as the purple vertical line in the bottom-right plot in Figure 10. We see that choosing the number of knots to be $\sqrt{n}$ works for Ring data.

\begin{figure}
\centering
\includegraphics[width=3.5cm]{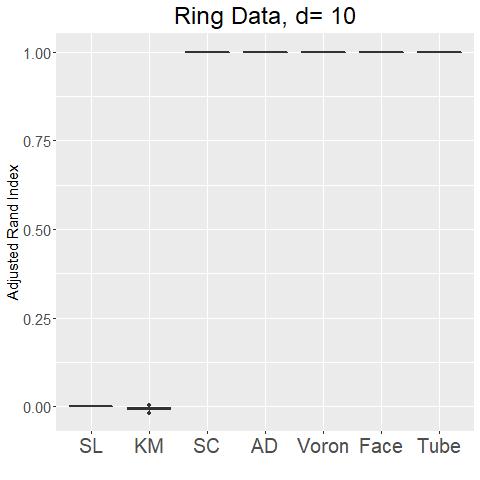}
\includegraphics[width=3.5cm]{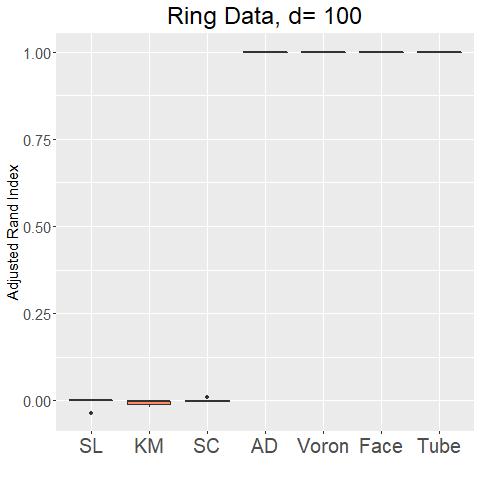}
\includegraphics[width=3.5cm]{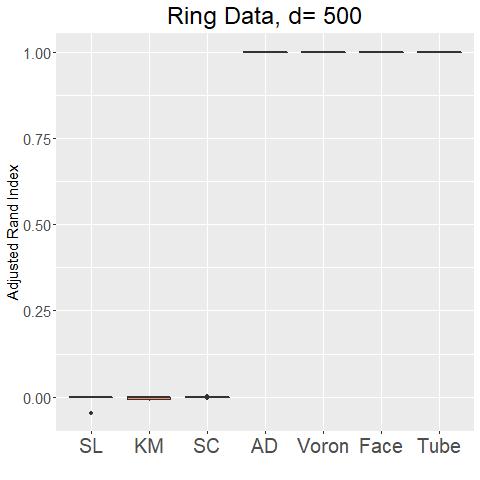}
\includegraphics[width=3.5cm]{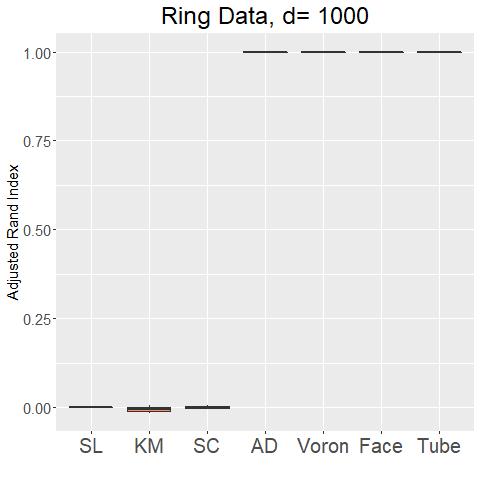}
\caption{Comparison of the rand index using different similarity measures on Ring data with dimensions $10$, $100$, $500$, $1000$. Medium of 100 repetitions.}
\label{fig::ring2}
\end{figure}

~~~~The ring data is constructed by a mixture distribution such that with a probability of $\frac{1}{6}$
we sample from the ring structure and with a probability of $\frac{5}{6}$ 
we sample from the central part.
The ring structure is generated by
a uniform distribution over the ring $\{(x_1,x_2):x_1^2+x_2^2 = 1\}$
and is corrupted with an additive Gaussian noise $N(0, 0.2^2\mathbf{I}_2)$.
The central part is simply a Gaussian $N(0, 0.2^2\mathbf{I}_2)$.
We generate a total of $n=1200$ points from the above mixture and add the high dimensional noise with the same procedure as in Section~\ref{sec::HD}. The same skeleton clustering approaches are applied as well as the classical approaches, with the final number of clusters chosen to be $2$.
The result is displayed in Figure~\ref{fig::ring2}.
Again, the density-based skeleton clustering methods work
well even when the dimension is large.

\begin{figure}
\centering
\includegraphics[height=5cm]{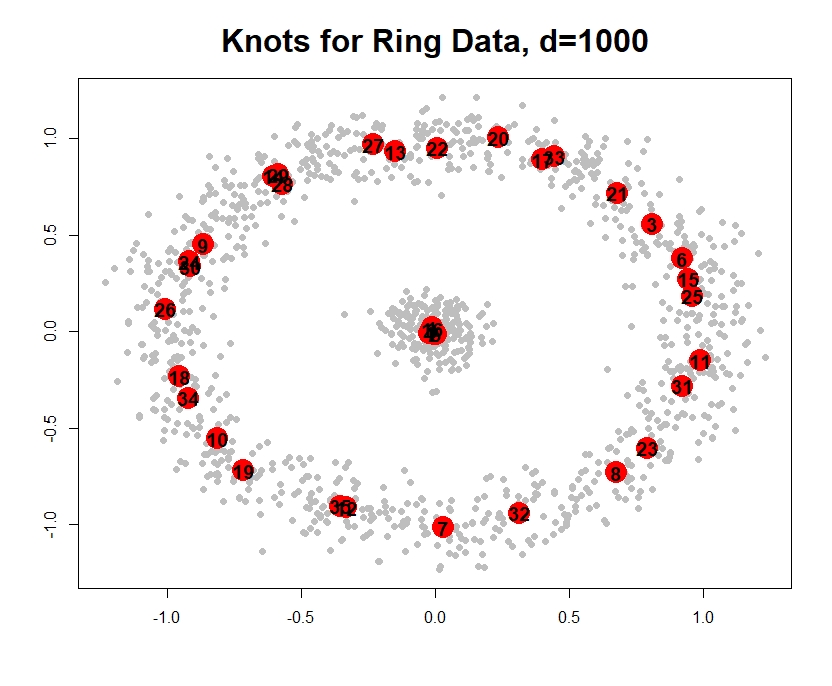}
\includegraphics[height=5cm]{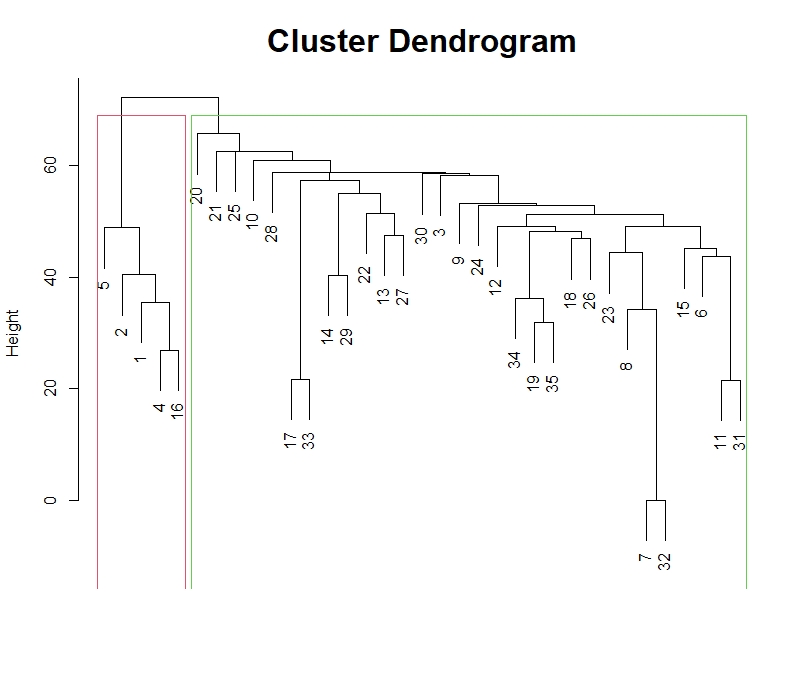}
\caption{Results on Ring data with dimension $1000$. 
}
\label{fig::ring1}
\end{figure}

\subsection{Additional Real Data Examples} \label{sec::realdataAdd}

\subsubsection{Zipcode Data}	\label{sec::zip}
~~~~This dataset consists of $n=2000$ $16\times16$ images of handwritten Hindu-Arabic numerals from \citep{Wernergsl}.
We use the overfitting $k$-means to find $k=45$ knots.
Similar to the procedure in Section~\ref{sec::HD},
we consider four similarity measures to obtain the edge weight: VD, FD, TD, and AD. We use single linkage for the four skeleton clustering approaches and compare them
to three traditional methods: direct single linkage hierarchical clustering (SL), direct $k$-means clustering (KM), and spectral clustering  (SC).

\begin{figure}
\captionsetup{skip=1pt}
    \centering
%    \begin{subfigure}[t]{0.35\textwidth}
%        \centering
        \includegraphics[width=2.3in]{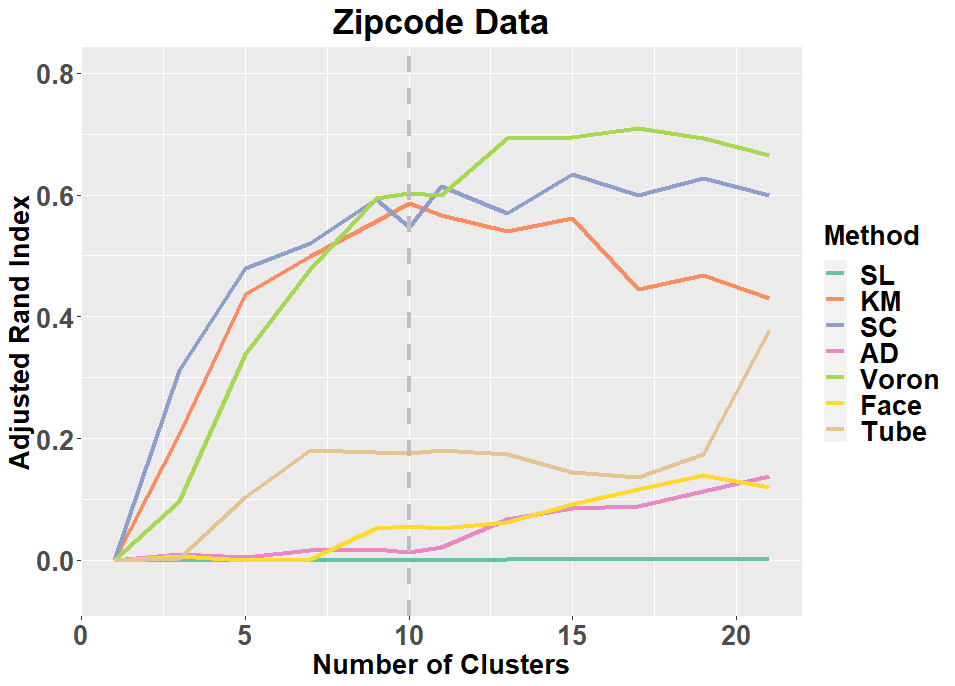}
%    \end{subfigure}
%    \begin{subfigure}[t]{0.35\textwidth}
%        \centering
       \includegraphics[width=2.3in]{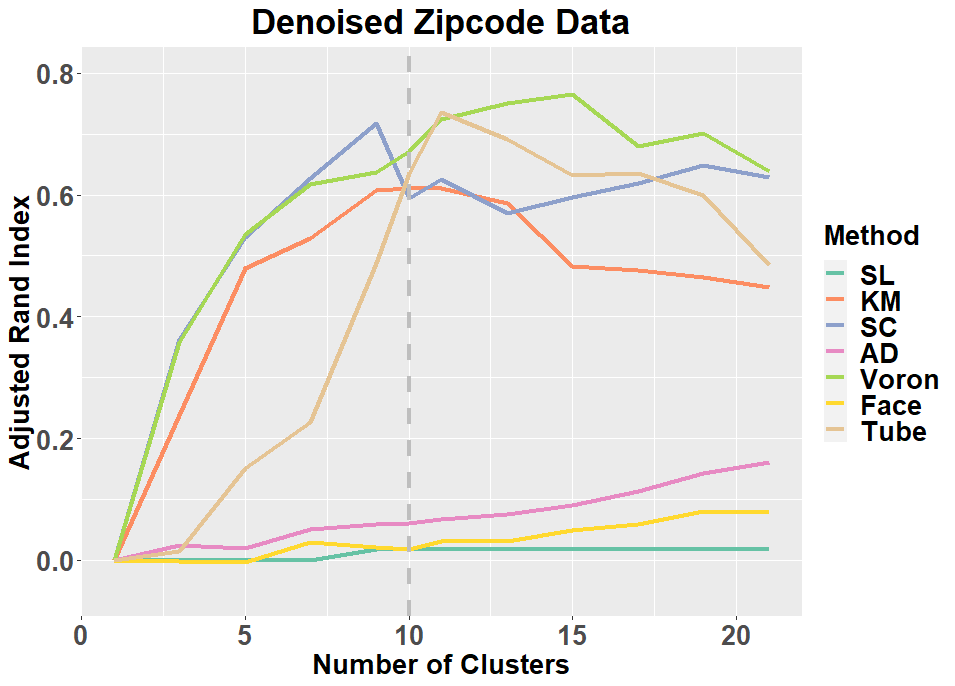}
%    \end{subfigure}
    \caption{Comparison of different similarity measures on all Zipcode Data.}
    \label{fig::zip1}
\end{figure}

The result is shown in the left panel of Figure~\ref{fig::zip1} with the adjusted Rand index plotted against different numbers of total clusters $S$. 
The gray vertical line indicates $S=10$, which is the actual number of digits.
%Clearly, the skeleton clustering with VD, FD, or TD all work well compared to the traditional methods (SL, SC) and the average distance (AD).
The skeleton clustering with VD (Voron) gives the best clustering result in terms of adjusted Rand index at the true $10$ clusters and gives good clustering results when the number of clusters is specified to be larger than the truth. However, we note that spectral clustering (SC) and naive $k$-means clustering (KM) give comparably good results with a small number of clusters.
%The fact that the AD does not work well again highlight the effectiveness of including density information in the edge weight construction. 

The right panel of Figure~\ref{fig::zip1} is the ``denoised'' version of the digits.
We estimate the density of each observation by $[\sqrt{n}]$-nearest-neighbor density estimator and remove the observations with the lowest $10\%$ density. 
%This leads to observations that are in a higher density region.
We see that all clustering results are slightly improved, but such improvement may come from the decreased total sample size after denoising. Notably, the skeleton clustering with Tube density (Tube) generates significantly better clustering results after denoising the data, giving adjusted Rand indexes comparable to skeleton clustering with Voronoi density. This shows skeleton clustering with Tube density can be sensitive to noises in real data but still has the potential to give insightful clustering results.
%This again shows that the `density' could provide useful information in clustering so methods that utilizes this feature would tend to have a better performance (such the VD, FD, and TD). 

%We use overfitting K-means to choose $k = [\sqrt{n}]$ knots, where $n$ is the sample size. Different similarity measures are compared on these two datasets. Specifically, for Face and Tube density, we use the estimator defined in equation 6 and equation 14 with Gaussian kernel for density estimation, and the bandwidth is selected by the Rule of Thumb. Single linkage hierarchical clustering is used to segment the skeleton structure, and the $1$ nearest knot method is used to assign the cluster label for individual data points. We also applied the clustering methods to the denoised Zipcode data. To denoise the data, we find the $H$-Nearest Neighbor distance of each data point with $H = [\sqrt{n}]$ and removed data points with the largest $10\%$ $H$-NN distance.

%The performance of different clustering methods on the  Zipcode data is presented in Figure \ref{fig::zip1} with different colors and the gray vertical line indicates the true number of digits used. 

%Here we see that Voronoi and Face density give the best performance with Rand Index $0.60$ and $0.62$ at the true $10$ clusters. However, denoising the dataset does not provide much improvement on the clustering result as measured by Rand Index. We have additional empirical results on Zipcode data in Appendix C.2.

%\begin{comment}
%
%\textcolor{red}{tangent distance, Discriminant adaptive nearest-neighbor metric}
%
%\end{comment}

\subsubsection{Olive Oil Data}	\label{sec::olive}

~~~~We consider another real dataset: the Olive Oil data \citep{TSIMIDOU1987227}, a popular dataset for cluster analysis.
This data set represents $d=8$ chemical measurements on different specimens of olive oil produced in $9$ different regions in Italy (northern Apulia, southern Apulia, Calabria, Sicily, inland Sardinia, and coast Sardinia, eastern and western Liguria, Umbria). There are a total of $n = 572$ observations in the dataset.
%\begin{figure}
%\centering
%\includegraphics[width=4.3cm]{Olive_MST.JPG}
%\includegraphics[width=5cm]{Olive_clusSize.jpeg}
%\caption{Results on Olive Oil data}
%\end{figure}
%Measuring the noisy level of the knots by MST total length, we again see that the knots chosen by overfitting K-means are better than knots subsampled from the original data and have some degree of dimension reduction and noise reduction effect. The default choice number of knots $\sqrt{572} \approx 24$ is depicted as the vertical line in Figure 18 (right), and we see it is around the elbow of the cluster sizes. 

\begin{figure}
\centering
\includegraphics[width=6cm]{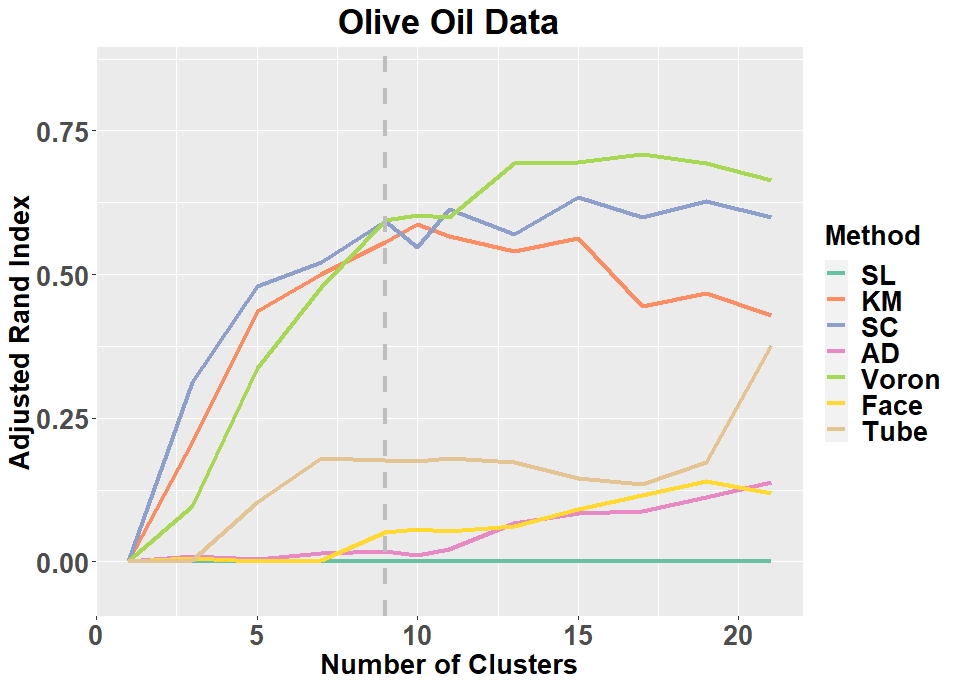}
\includegraphics[width=6cm]{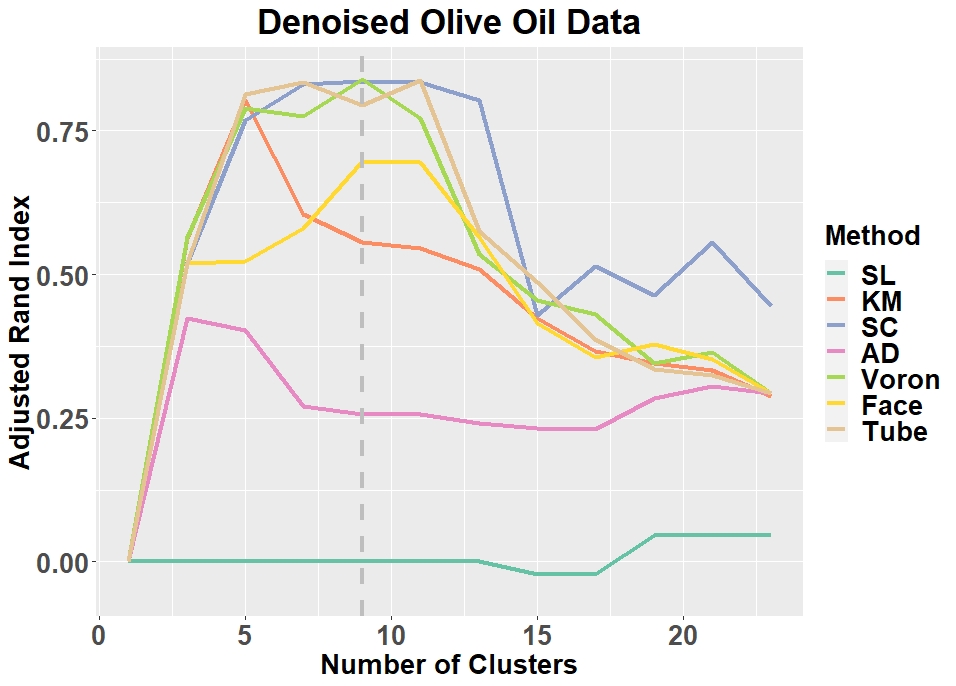}
\caption{The clustering performance under different numbers of final clusters of the Olive oil data.}
\label{fig::Olive}
\end{figure}

The same comparison procedure as in Section \ref{sec::zip} is employed, finding $k=24$ knots by $k$-means and using single linkage for the skeleton clustering approaches. The performance of different similarity measures is presented in Figure \ref{fig::Olive}. Different color denotes different similarity measures and the gray vertical line indicates the actual number of clusters $9$. 
Overall, the skeleton clustering with Voronoi density and Tube density works well;
spectral clustering also performs well in this case.
The fact that average distance fails to capture clusters in the data
highlights the importance of using a density-aided similarity in this case.
Note that we also include the clustering performance on the `denoised' data, in which we remove
the 10\% observation with the lowest $\sqrt{n}$-Nearest-Neighbor density estimate.

\subsection{Future Work}
\label{sec::future}
We discuss some future directions below:
\begin{itemize}
%Moreover, the newly proposed similarity measures can have more general applications beyond the scope of this paper. Particularly, Voronoi density can be computed based on only the pairwise distances, which makes skeleton clustering suitable for various clustering tasks. Overall skeleton clustering framework is flexible, and potentially new methods for different steps can be combined to provide new insights for different data sets even when the data are high-dimensional and large in scale.

\item {\bf Accounting for the randomness of knots.}
For our current theoretical analysis, we assume that the knots are given and non-random to simplify the problem.
But in practice, knots are computed from the sample data with inherent uncertainty. 
 {
The randomness of knots can affect the clustering performance because
the location of knots directly impacts the Voronoi cells, which changes the value of the similarity measures and consequently the cluster label assignments.
In particular, observations on the boundary of clusters will be more sensitive to any perturbations
in the location of knots. 
% We will investigate the effect of this randomness in the future.
%Admittedly, changes in the location of the knots can have complicated implications for the later procedures: changing the Voronoi cells partitioning the data points, introducing extra randomness in the estimations of the edge similarity measures, modifying the segmentation of the skeleton graph, and potentially leading to different clustering results of the data points. 
%However, some degree of variation in the knots can be tolerated in our framework. 
%The final clustering is essentially based on the nearest-knot partitions of the segmented skeleton components, where each disjoint component is a graphical representation of the corresponding cluster. The theoretical results presented in Section \ref{sec::theory} show that our proposed density-aided similarity measures are estimated consistently and can preserve the segmentation by the population edge similarities. In this sense, some changes in the knots will not affect the final clustering performance much as long as the knots are representative of the cluster structures and the population-based similarities between the knots can distinguish the true structures.
}
%Future effort can still be put to account for the randomness of the knots. 
Currently, there are two technical challenges when dealing with random knots. First, the randomness of knots may be correlated with the randomness of estimated edge weight, so the calculation of rates is much more complicated.
Second, while there are established theories for $k$-means algorithm \citep{Graf2000,Graf2002,Hartigan1979}, these results only apply to the global minimum of the objective function.
In reality, we are unlikely to obtain the global minimum, but instead, our inference is based on a local minimum.
It is unclear how to properly derive a theoretical statement based on local minima, so we leave this as future work.
%The analysis on a local minimum is much more complicated
%since 

\item {\bf Skeleton clustering with similarity matrix.}
The idea of skeleton clustering may be generalized to data where we only observe the similarity/distance matrices
such as network data. 
Knots can be restricted to indices in the data and we choose them by minimizing some network-based or diffusion-related criteria. 
While Face and Tube density can be difficult to adopt, the Voronoi density is still applicable since we only need the information about pairs of observations.
This might provide a new approach for community detection in network data \citep{zhao2017survey, abbe2017community}.

\item{\bf Detecting boundary points between clusters.}
Our skeleton clustering method can be applied to detect points on the boundary between two clusters.
The idea is simple: in the final cluster assignment, instead of assigning only one label to an observation, we assign $h$ labels to an observation based on the cluster labels of $h$-nearest knots. 
The homogeneity of the label assignments can be used as a quantity to detect if a point is on the boundary or in the interior of a cluster and may serve as an uncertainty quantification of clustering.
%Our preliminary results in Appendix \ref{sec::boundary} 
%seem to be promising. However, a more detailed analysis is needed to understand the effectiveness of this idea and 
We will pursue this in the future.

 {\item{\bf Anomaly and noise detection.} 
As illustrated in Appendix \ref{sim::noisyYinyang}, \ref{sim::noisyMixMickey}, and \ref{sim::noisymixStar},
the single linkage criterion in our Skeleton clustering framework may
detect noisy observations in the data.
This suggests the possibility of using our approach for noises or anomalies 
similar to the  DBSCAN \citep{Campello2015, EsterKriegel1996}.
We will explore this direction in the future. 
% For future research explicit noise detection techniques can be combined into our proposed skeleton clustering framework.
}

%\item{\bf Clustering after dimension reduction.}
%When the dimension of the data is ultra-high, it might be infeasible to perform clustering directly on the data.
%A common practice is to perform a dimension reduction procedure first and then apply a clustering algorithm.
%However, our preliminary finding shows a conflicting result.
%Sometimes, this idea works well but sometimes it leads to a worse result;
%see Appendix~\ref{sec::DR} for more details.
%In the future, we will investigate this topic and find out
%when this idea will work and when it will not work. 

%For future work, the theoretical property of the skeleton clustering framework still need to be studied. In this work we only present the consistency for the similarity measures, but lack theoretical guarantee on the final clustering result. For future the theoretical understanding of the population version of the similarity measures can be developed, and the overall theoretical guarantee of the skeleton clustering framework can be studied. One promising line of reasoning is to prove the cluster tree consistency accounting for not only the randomness of the similarity measures but also the choice of knots. 
\end{itemize}

\end{document}